\documentclass[runningheads]{llncs}

\usepackage{eccv}

\usepackage{eccvabbrv}

\usepackage{graphicx}
\usepackage{booktabs}

\usepackage[accsupp]{axessibility}  % Improves PDF readability for those with disabilities.

\usepackage{hyperref}

\usepackage{orcidlink}

\usepackage{xcolor}
\usepackage{xspace}
\usepackage{colortbl}
\usepackage{multirow}
\usepackage{soul}
\usepackage{multirow}
\usepackage[percent]{overpic}

\definecolor{lightred}{rgb}{1.0, 0.4, 0.4}
\definecolor{lightorange}{rgb}{1.0, 0.6, 0.2}
\definecolor{lightyellow}{rgb}{1.0, 1.0, 0.4}
\newcommand{\best}{\cellcolor{lightred}}
\newcommand{\sbest}{\cellcolor{lightorange}}
\newcommand{\tbest}{\cellcolor{lightyellow}}
\newcommand{\hlred}[1]{{\sethlcolor{lightred}\hl{#1}}}
\newcommand{\hlorange}[1]{{\sethlcolor{lightorange}\hl{#1}}}
\newcommand{\hlyellow}[1]{{\sethlcolor{lightyellow}\hl{#1}}}

\begin{document}

% ---------------------------------------------------------------
% TODO REVIEW: Replace with your title
\title{GS2Mesh: Surface Reconstruction from Gaussian Splatting via Novel Stereo Views} 

% TODO REVIEW: If the paper title is too long for the running head, you can set
% an abbreviated paper title here. If not, comment out.
\titlerunning{GS2Mesh}

% TODO FINAL: Replace with your author list. 
% Include the authors' OCRID for the camera-ready version, if at all possible.
\author{Yaniv Wolf\inst{*} \and
Amit Bracha\inst{*} \and
Ron Kimmel}

% TODO FINAL: Replace with an abbreviated list of authors.
\authorrunning{Y.~Wolf et al.}
% First names are abbreviated in the running head.
% If there are more than two authors, 'et al.' is used.

% TODO FINAL: Replace with your institution list.
\institute{Technion - Israel Institute of Technology, Haifa, Israel\\
\email{\{yaniv.wolf, amit.bracha, ron\}@cs.technion.ac.il}\\
\url{https://gs2mesh.github.io}}

\maketitle
\def\thefootnote{*}\footnotetext{Indicates equal contribution}\def\thefootnote{\arabic{footnote}}

\begin{figure}[]
    \centering
    \begin{tabular}{cccc}
        \begin{overpic}[width=0.24\textwidth]{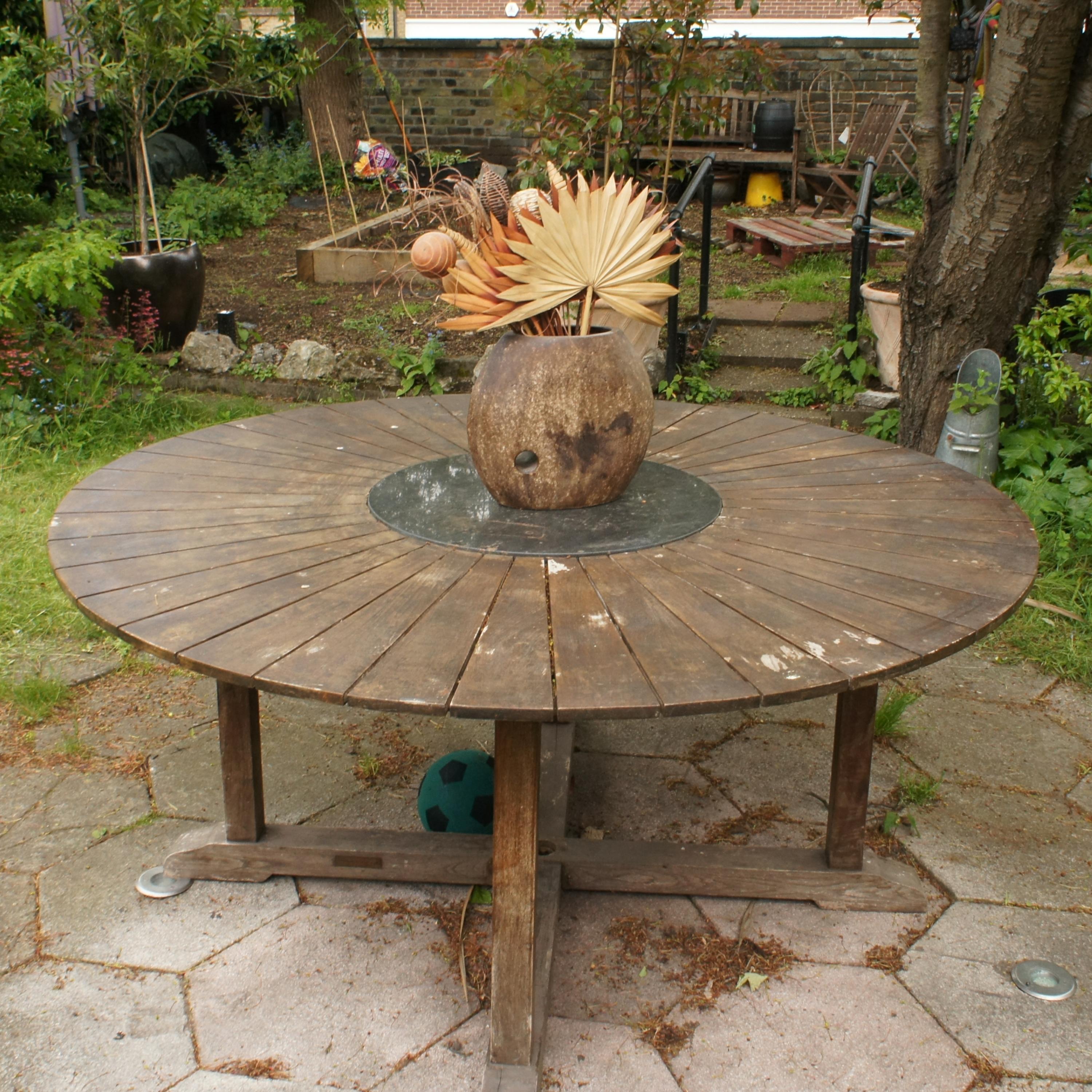}
            \put(5,5){\color{white}\scriptsize\textbf{Image}}
        \end{overpic} & 
        \begin{overpic}[width=0.24\textwidth]{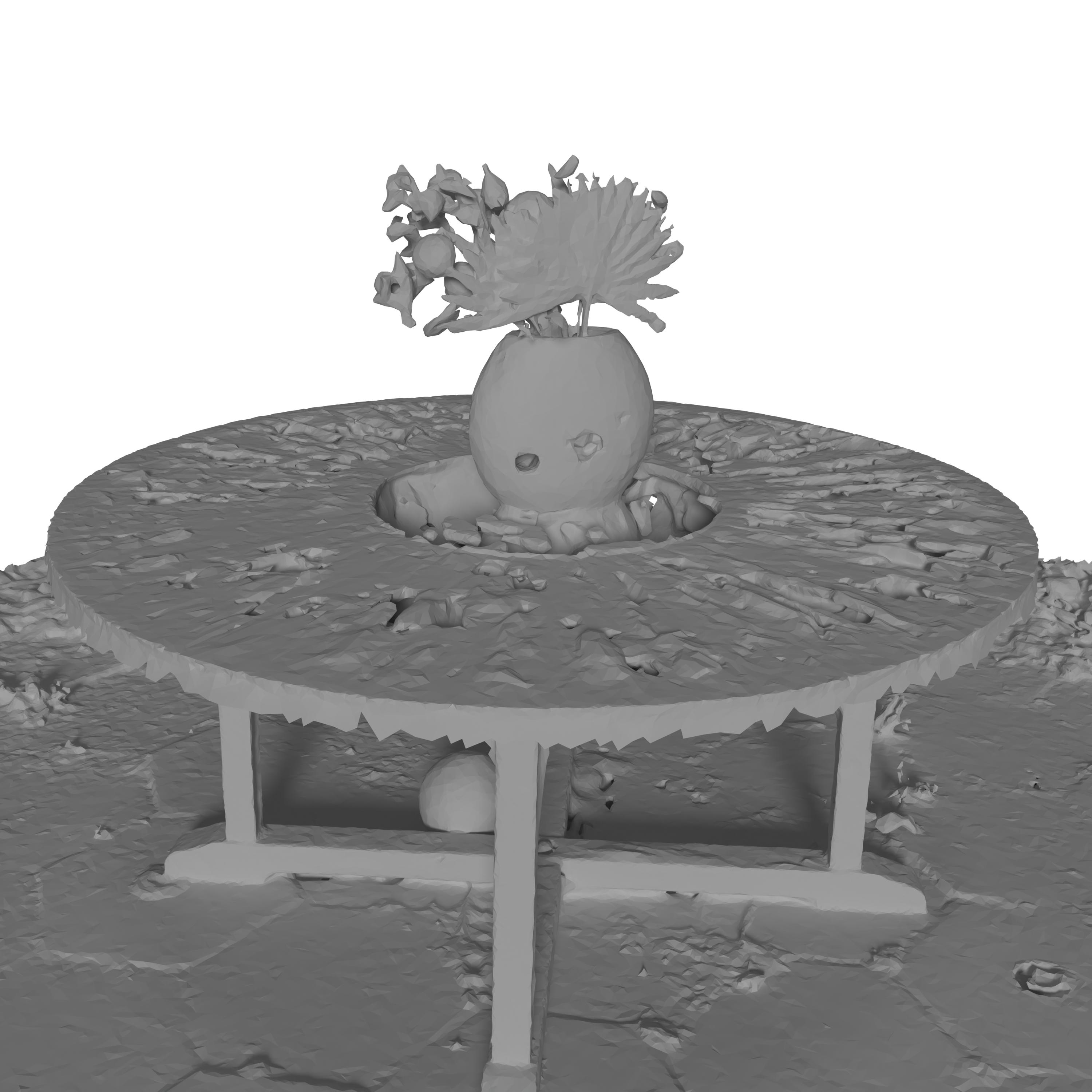}
            \put(5,5){\color{white}\scriptsize\textbf{SuGaR (2H)}}
        \end{overpic} &
        \begin{overpic}[width=0.24\textwidth]{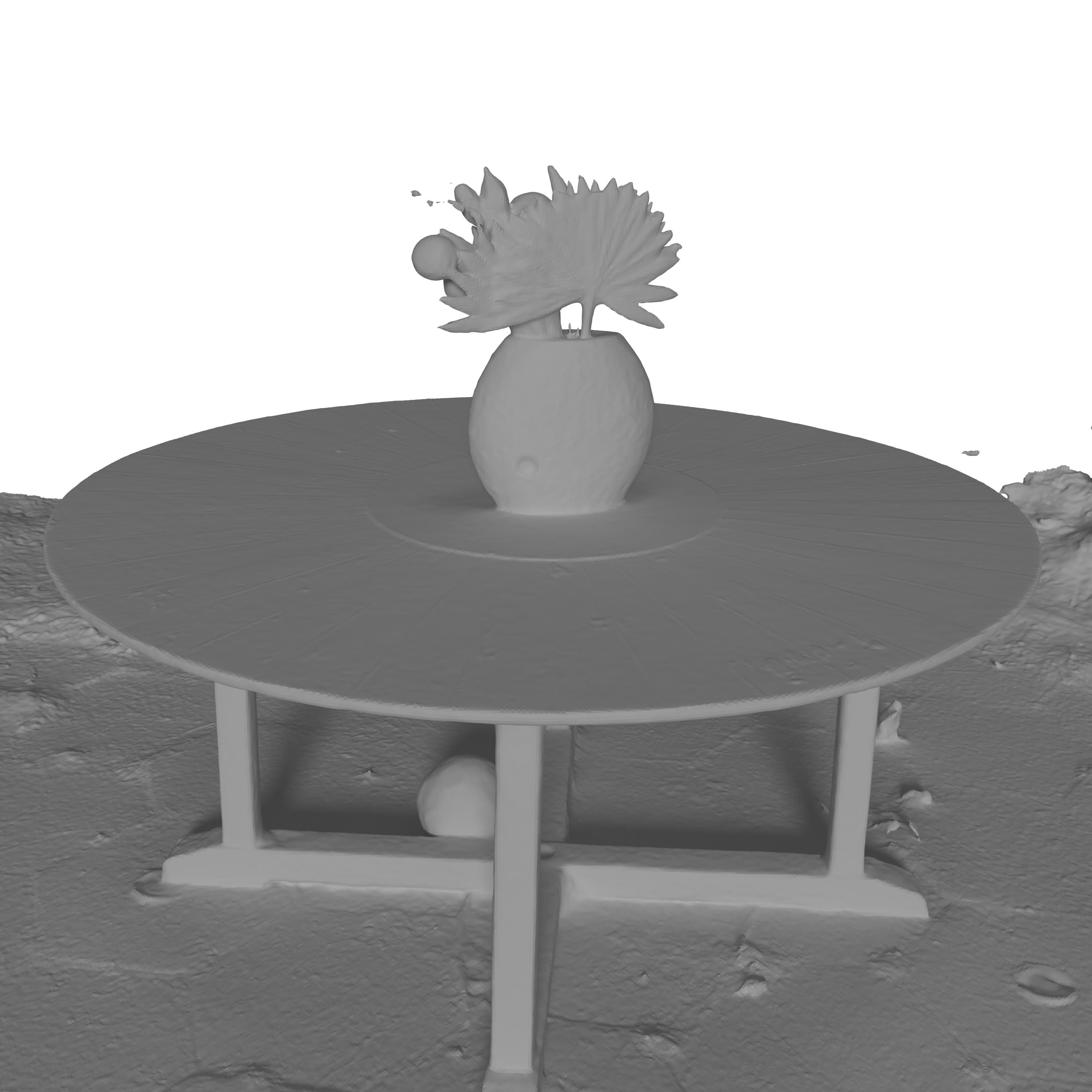}
            \put(5,5){\color{white}\scriptsize\textbf{BakedSDF (>24H)}}
        \end{overpic} & 
        \begin{overpic}[width=0.24\textwidth]{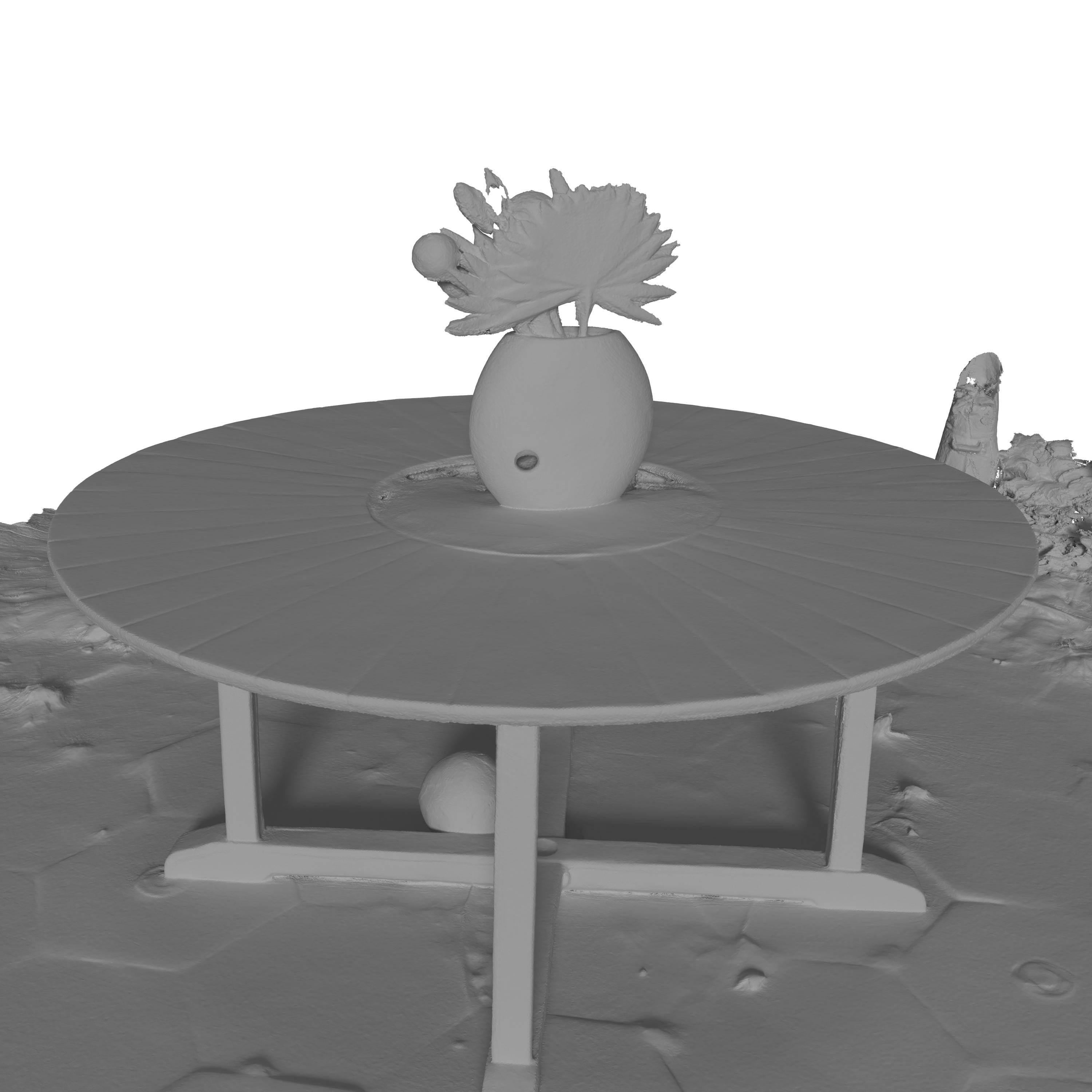}
            \put(5,5){\color{white}\scriptsize\textbf{Ours (1H)}}
        \end{overpic} \\
    \end{tabular}
    \caption{Qualitative results on Mip-NeRF360\cite{Mip-NeRF} dataset garden scene.}
    \label{fig:Mip2}
\end{figure}
% %__________________________________

\begin{abstract}
Recently, 3D Gaussian Splatting (3DGS) has emerged as an efficient approach for accurately representing scenes. 
However, despite its superior novel view synthesis capabilities, extracting the geometry of the scene directly from the Gaussian properties remains a challenge, as those are optimized based on a photometric loss. 
While some concurrent models have tried adding geometric constraints during the Gaussian optimization process, they still produce noisy, unrealistic surfaces.

We propose a novel approach for bridging the gap between the noisy 3DGS representation and the smooth 3D mesh representation, by injecting real-world knowledge into the depth extraction process.
Instead of extracting the geometry of the scene directly from the Gaussian properties, we instead extract the geometry through a pre-trained stereo-matching model.
We render stereo-aligned pairs of images corresponding to the original training poses, feed the pairs into a stereo model to get a depth profile, and finally fuse all of the profiles together to get a single mesh.

The resulting reconstruction is smoother, more accurate and shows more intricate details compared to other methods for surface reconstruction from Gaussian Splatting, while only requiring a small overhead on top of the fairly short 3DGS optimization process.

We performed extensive testing of the proposed method on in-the-wild scenes, obtained using a smartphone, showcasing its superior reconstruction abilities. 
Additionally, we tested the method on the Tanks and Temples and DTU benchmarks, achieving state-of-the-art results.

\end{abstract}
%_________________________________________________

\begin{figure}[]
\centering
\includegraphics[width=\textwidth]{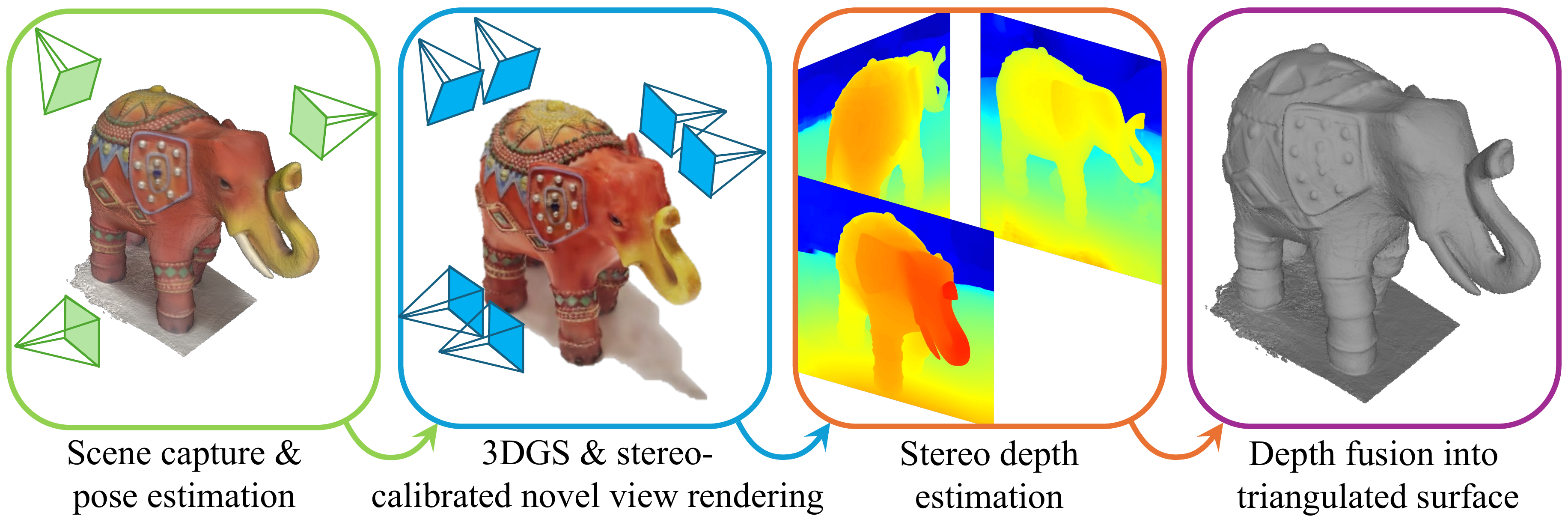}
\caption{
The proposed pipeline for surface reconstruction. 
    First, we represent the scene by applying a 3DGS model. 
    We then use the 3DGS model to render stereo-aligned pairs of images corresponding to the original views.
    For each pair, using a shape from stereo algorithm, we reconstruct an RGB-D structure, which is then integrated from all views using TSDF \cite{TSDF} into a triangulated mesh of the scene.}
\label{fig:Pipeline}
\end{figure}
%_____________________________
\section{Introduction}
The Gaussian Splatting Model for radiance field rendering (3DGS) \cite{GS} has recently marked a significant leap forward in the realm of novel view synthesis, surpassing previous neural rendering methods in both speed and accuracy.
By optimizing the distribution, size, color, and  opacity of a cloud of Gaussian elements, and projecting, or splatting them onto virtual cameras, 3DGS is able to generate realistic images of complex scenes from novel viewing directions in real-time.
However, direct reconstruction of surfaces from 3DGS involves significant challenges.
The main problem is that the locations of the Gaussian elements in 3D space do not form a geometrically consistent surface, as those are optimized for best matching the input images when projected back onto their image planes.
Consequently, reconstructing surfaces based on the centers of the Gaussians yields noisy and inaccurate results. Current state-of-the-art methods attempt to regularize the 3DGS optimization process by adding additional geometric constraints \cite{SuGaR}, flattening the Gaussian elements, \cite{2DGS}, or extracting the geometry using opacity fields \cite{GOF}, but they still rely on the Guassian locations and form noisy, unrealistic surfaces. 

We propose an alternative approach for extracting depth from the optimized Gaussian point cloud, which does not rely on the noisy locations of the Gaussians. 
Instead, we take advantage of a powerful geometric regularizer, trained on real-world data - a pre-trained stereo matching model. Stereo matching models solve a correspondence problem on stereo-aligned pairs of images, from which accurate depth can be extracted. 
Our main observation is that through 3DGS rendering, we can artificially create stereo-aligned pairs of images corresponding to the original views, feed these pairs into a pre-trained stereo model, and fuse the resulting depths using the Truncated Signed Distance Function (TSDF) algorithm \cite{TSDF}. 
The result is a smooth, geometrically consistent surface, that is extracted from the noisy 3DGS cloud using real-world regularization.

The proposed method reduces surface reconstruction time dramatically, taking only a small overhead on top of the 3DGS capturing of the scene, which is significantly faster compared to neural surface reconstruction methods. 
For instance, reconstruction of an in-the-wild scene taken by a standard smartphone camera requires less than five minutes of additional computation time after a 3DGS scene capture. 
Additionally, since we reconstruct the surface based on the 3DGS capture, it is straightforward to bind the mesh to the original model, as mentioned in \cite{GaMeS, SuGaR}, for mesh-based manipulation of the Gaussian elements. 
Moreover, since our mesh is more accurate, it does not require any additional refinements \cite{SuGaR}.

We tested the proposed method on the Tanks and Temples (TnT) benchmark \cite{tnt} as well as the DTU \cite{DTU} benchmark, two commonly used 3D reconstruction datasets, and achieved state-of-the-art results.
Additionally, we extensively tested our method on in-the-wild scenes captured with a smartphone, showing qualitative results of the proposed method's reconstruction abilities. To summarize, our main contribution,
\begin{itemize}
    \item We propose a novel method for fast and accurate in-the-wild surface reconstruction, by using a pre-trained stereo matching model as a geometric prior for extracting depth from a 3DGS model.
\end{itemize}

%_______________________________________
\section{Related Efforts}
\subsection{Multi-View Stereo and Stereo Matching}
Multi-View Stereo (MVS) is a fundamental geometry reconstruction method, where depth maps are extracted for each reference image based on correspondences with neighboring images.
In the field of deep MVS methods, the pioneering work of MVSNet \cite{mvsnet} introduced an end-to-end framework for MVS learning, which can be divided into three parts: 2D feature extraction, homography, and 3D cost volume with 3D convolutions.
Latter methods presented an improvement to this scheme, by improving the 3D cost volume \cite{zhang2020visibility,ma2022multiview}, improving the architecture for 2D feature extraction \cite{wang2021patchmatchnet}, using a vision transformer (ViT) architecture for feature extraction \cite{mvsformer}, and improving 3D convolutions for more efficient computations using a coarse-to-fine method \cite{gu2020cascade, yang2020cost}.
To fuse the extracted depth maps into one point cloud or mesh there are two main methods: Fusibile \cite{Gipuma}, which has recently been generalized by \cite{yan2020dense}, and TSDF \cite{TSDF}. 
MVS methods deeply rely on accurate camera poses for the calculation of epipolar lines, and on in-the-wild scenes they struggle to achieve high accuracy as in the controlled environment, since small errors in pose estimation result in a noisy reconstruction, as we show in our ablation study.

Deep stereo matching methods \cite{kendall2017end, chang2018pyramid,zhang2019ga,shen2021cfnet, poggi2021synergies, laga2020survey} are related to deep MVS methods, however, since it is guaranteed that matching pixels between two images must lie in the same row, the cost volume layer works on the disparity instead of the depth.
Recent state-of-the-art stereo matching models, such as RAFT \cite{RAFT}, IGEV \cite{xu2023iterative}, and DLNR \cite{DLNR}, use iterative refinement using GRU or LSTM layers. 
Unlike MVS methods, stereo methods require only two images which are typically closer to each other compared to MVS, and share the same image plane, resulting in less occluded regions which are visible only in one of the views.

%_______________________________________
\subsection{Neural Rendering for Novel View Synthesis and Surface Reconstruction}
 Novel view synthesis methods are trained on a set of images from a scene, and aim to render views of a scene from any given pose.
 The pioneering work of Neural Radiance Fields (NeRF) \cite{Nerf} presented a major leap forward in accuracy by incorporating importance sampling and positional encoding to enhance rendering quality.
However, the use of relatively large Multi-Layer Perceptrons (MLP) to capture the scene resulted in long training times. 
Later, Mip-NeRF \cite{Mip-NeRF} improved the quality of the rendered view with a different sampling method, although training and rendering times remained long. 
InstantNGP \cite{mueller2022instant} tackled the extended training times of previous efforts, by incorporating a hash grid and an occupancy grid with a small MLP.

 Neural surface reconstruction methods \cite{Siren,Neus,bakedsdf,IDR}, in addition to accurately rendering novel views, are also capable of reconstructing the surface of the scene.
 IDR \cite{IDR} trained an SDF represented by an MLP for both color and geometry reconstruction. 
 Neus \cite{Neus} reduced the geometric error by utilizing weighted volume rendering, and HF-Neus \cite{HF-NeuS} enabled coarse-to-fine refinement for high-frequency detail reconstruction by decomposing the implicit SDF into a base function and a displacement function. 
 RegSDF \cite{RegSDF} used a point cloud obtained from shape-from-motion (SfM) as regularization, in addition to regularizing the curvature of the zero-level of the SDF function. 
 NeuralWarp \cite{Neural_warping} suggested refining the geometry by regularizing image consistency between different views through warping based on implicit geometry. 
 Neuralangelo \cite{Neuralangelo}, using a 3D hash encoded grid, enabled detailed reconstruction and achieved state-of-the-art results on leading benchmarks. 
 However, reconstruction time of these methods can reach up to several days per scene. 
 In the context of novel stereo views, a recent model \cite{tosi2023nerf} has managed to successfully perform unsupervised training of a stereo model using rendered stereo-aligned triplets from a neural scene reconstruction method, showcasing the possibility of novel view synthesis as a data factory.

\subsection{Gaussian Splatting for Novel View Synthesis and Surface Reconstruction}
Recently, a major leap forward was presented by 3DGS \cite{GS}, a faster and more accurate method for scene capturing. 3DGS represents the scene as a point cloud of 3D Gaussians, where each Gaussian has the properties of opacity, rotation, scale, location, and spherical harmonics. 
The scaling of the Gaussians is anisotropic, which allows them to represent thin structures in the scene.
The Gaussians are initialized using an SfM algorithm \cite{schoenberger2016sfm, schoenberger2016mvs}, which extracts the camera poses and provides an initial guess for the locations of the Gaussians. The capturing time is short compared to other methods based on MLPs, and it is capable of real-time rendering.
Concurrent works on GS \cite{wu2024recent,chen2024survey,fu2023colmapfree,cheng2024gaussianpro,yu2023mipspalt} have improved on the vanilla 3DGS in various ways, such as by reducing optimization time, increasing accuracy, reducing aliasing and removing the need for COLMAP \cite{schoenberger2016mvs,schoenberger2016sfm} poses.

As discussed earlier, the Gaussian locations in the vanilla 3DGS do not form a geometrically consistent surface. 
Recent methods try to manipulate the Gaussian elements to extract more accurate surfaces \cite{SuGaR, 2DGS, Neusg}. 
SuGaR \cite{SuGaR}, the pioneering method in surface reconstruction from Gaussian Splatting, added a regularization term for post-process optimization based on the opacity levels of the Gaussians, forcing the Gaussian element cloud to align with the surface. 
2DGS \cite{2DGS} flattens the Gaussians into 2D elements, and GOF \cite{GOF} extracts the surface by creating an opacity field from the Gaussians.
However, since these methods utilize the location and opacity of the Gaussian elements, they reconstruct the surface with noisy undulations.

%_______________________________________________
\section{Method}

We propose a novel pipeline for surface reconstruction from 3DGS, as illustrated in \cref{fig:Pipeline}. In this section, we will explain in detail each step of the pipeline. 
We note that additionally, we can mask out specific objects by projecting segmentation masks from Segment Anything Model (SAM) \cite{SAM} between consecutive images using depth maps. 
Additional information on masking is available in the supplementary material.

%_______________________________________________
\subsection{Scene Capture and Pose Estimation}
We start with a video or images of a static scene as input. 
Following the vanilla 3DGS, we employ COLMAP \cite{schoenberger2016sfm, schoenberger2016mvs} for SFM to identify points of interest and deduce camera matrices from the provided images. 

\subsection{3DGS and Stereo-Aligned Novel View Rendering}
The elements extracted from the previous stage are then fed into the 3DGS model to accurately represent the scene. 
For completeness, we will give a short formulation of the 3DGS process;
In 3DGS, 3D Gaussian elements are defined in space by
    $G(\boldsymbol{x}) = \exp(-\frac{1}{2}(\boldsymbol{x} - \boldsymbol{x}_p)^\top\boldsymbol{\Sigma}^{-1}(\boldsymbol{x} - \boldsymbol{x}_p))$, where 
$\boldsymbol{x}_p$ is the center of the Gaussian, and $\boldsymbol{\Sigma}$ is its covariance matrix. 
During optimization, $\boldsymbol{\Sigma}$ is factorized into the rotation $\boldsymbol{R}$ and scale $\boldsymbol{S}$ matrices: $\boldsymbol{\Sigma} = \boldsymbol{RSS}^\top\boldsymbol{R}^\top$. 
When rendering, the Gaussians are projected onto the image plane: $\boldsymbol{\Sigma}' = \boldsymbol{JW\Sigma W}^\top \boldsymbol{J}^\top$, where $\boldsymbol{W}$ is the view transformation, and $\boldsymbol{J}$ is the Jacobian of the affine projective transformation onto the image plane. 
By removing the last row and column of $ \boldsymbol{\Sigma}'$, we remain with 2D Gaussians in the image plane. To calculate the color of a pixel in the image plane, 3DGS employs alpha blending which applies weights to the opacity from front to back, $C = \sum_{i \in N}\alpha_i c_i\prod_{j=1}^{i-1}(1-\alpha_j)$, where $\alpha_i$ is the product of the the $i^{th}$ 2D Gaussian with its opacity parameter, and $c_i$ is the the directional appearance component. For more details, see the original 3DGS paper \cite{GS}.

During the 3DGS process the Gaussians are optimized based on the photometric loss between the given source images and their corresponding rendered images. 
This creates a representation of the scene that allows rendering novel views which were not present in the original training data.
It is important to note that the vanilla 3DGS relies on sufficient coverage of the scene, and in areas lacking sufficient coverage, noisy artifacts might appear, as seen in \cref{fig:limitations}. 
Additionally, since the 3DGS is optimized based on the training images, staying close to a training image will result in a cleaner render.

Therefore, when generating novel stereo views of the scene, we input a sufficient amount of images that cover the region of interest. 
Additionally, we stay as close as possible to the original training poses, by choosing the left image of the stereo pair to be at the same pose $R_L, T_L$ as a training image. 
Following this choice, the right pose with a horizontal baseline of $b$ is formulated as follows: $R_R = R_L$, $T_R = T_L+(R_L\times [b,0,0])$. 
This ensures that the resulting left-right cameras are stereo-calibrated.

%_______________________________________________

\subsection{Stereo Depth Estimation}
With the rendered stereo-aligned image pairs, we can essentially turn a scene captured from a single camera into a scene captured from a pair of stereo-calibrated cameras, using the novel view synthesis capabilities of 3DGS. 
We then apply a stereo matching algorithm to form depth profiles from every stereo pair. 
We have tested several stereo matching algorithms in the experimental section, and achieved the best qualitative and quantitative results with DLNR \cite{DLNR}, a state-of-the-art neural stereo matching model, with the pre-trained Middlebury \cite{middlebury} weights. 
To further enhance the resulting reconstructions, we apply several masks to the output of the stereo model.
The first mask is an occlusion mask, which is calculated by applying a threshold on the difference between the left-to-right and right-to-left disparities of the same pair of images. 
This masks out parts of the scene that were only visible in one of the cameras, and therefore the stereo model's output in these areas is unreliable. 
We justify the use of this mask by the fact that the occluded areas will be filled in from adjacent stereo views. 
An example of an occlusion mask can be seen in \cref{fig:dtu_example}, and we added an experiment in the supplementary material which demonstrates the effectiveness of the occlusion mask.

The second mask is applied based on the depth of the stereo output.
The relationship between stereo matching errors can be described as $\epsilon(Z) \approx \dfrac{\epsilon(d)}{f_x \cdot B}Z^2$ \cite{bracha2021depth}, where $\epsilon(d)$ represents the disparity output error, $Z$ is the ground-truth depth, $\epsilon(Z)$ is the error of the depth estimate, $f_x$ denotes the camera's horizontal focal length, and $B$ is the baseline. 
Conversely, the disparity between matching pixels in two images of an object that is positioned at a short distance from the cameras can exceed the maximum disparity limit produced by stereo matching algorithms.
Thus, estimating the depth of an object that  is too close to the camera can result in an error due to the limitation of the matching algorithms, and estimating the depth of an object that is distant results in a quadratic error. 
Therefore, we consider depth in the range $4B  \leq  Z \,\leq \, 20 B$. This approach enhances the overall accuracy and reliability of the depth estimation process, ensuring more consistent geometric reconstructions. 
With the above considerations taken into account, we now have two contradicting factors when setting the horizontal baseline of the stereo pair; 
On the one hand, a larger baseline allows for a wider ``sweet spot'' for the stereo model. 
On the other hand, the 3DGS limits how far we can stray from the original training images without producing noisy renders. 
In the experimental section, we tested different baselines and found that a horizontal baseline of $7\%$ of the scene radius, or $3.5\%$ of the scene diameter, which allows for a ``sweet spot'' in the range of $14\%$ to $70\%$ of the scene diameter, provides the best results.

\begin{figure}[]
    \centering
    \includegraphics[width=\textwidth]{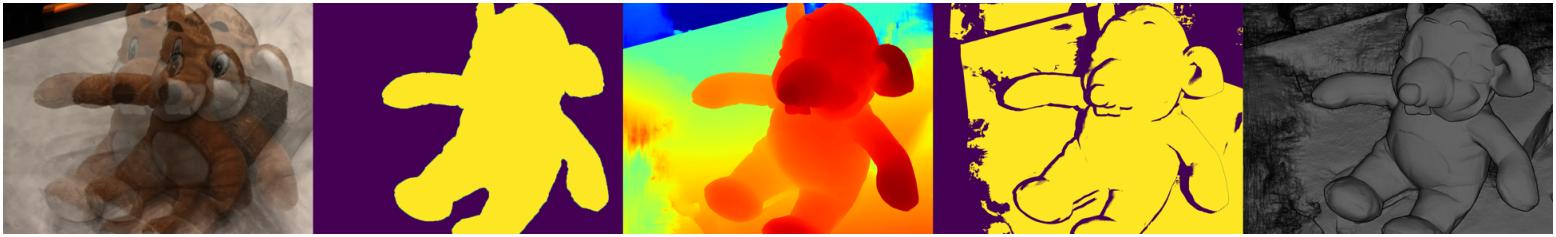}
    \caption{
    Example of our method's output on DTU \cite{DTU} scan105. From left to right: The rendered left and right images, segmentation mask, left-right disparity, occlusion mask, and shading - depth gradient.}
    \label{fig:dtu_example}
\end{figure}

%____________________________________
\subsection{Depth Fusion into Triangulated Surface}
To further enhance geometric consistency and smooth out any noise and errors which might have originated from the individual depth profiles, we aggregate all of the extracted depths using the Truncated Signed Distance Function (TSDF) algorithm \cite{TSDF}, followed by the Marching-Cubes meshing algorithm \cite{marching_cubes}.

%_______________________________________________
\section{Experiments and Results}
We present experiments which demonstrate that our method is able to accurately reconstruct surfaces in a more geometrically consistent way than other 3DGS-based or MVS approaches, as well as achieve comparable performance to neural reconstruction methods while taking significantly less time to run.
For quantitative results, we tested our method on the Tanks and Temples \cite{tnt} and DTU \cite{DTU}datasets, and compared our results to various neural and 3DGS-based reconstruction methods. 
We also compared between different versions of our model to justify our design choices.
Additionally, we show qualitative reconstruction results from
Mip-NeRF360 \cite{Mip-NeRF}, demonstrating that our method achieves comparable visual quality to neural reconstruction methods, and on in-the-wild videos taken from smartphones, we show our superiority in terms of geometric consistency and smoothness when compared to SuGaR \cite{SuGaR}. 
Finally, we perform an ablation study on the MobileBrick \cite{li2023mobilebrick} dataset, which validates the contribution of novel-view image generation and stereo,  by replacing two different points in our pipeline with a deep MVS model.
We note that in the MobileBrick dataset the camera poses are manually refined, and are shown to be more accurate than COLMAP \cite{schoenberger2016mvs,schoenberger2016sfm} poses for reconstruction \cite{li2023mobilebrick}. 
The comparison we present thus favors MVS models in that respect.

\begin{figure}[]
    \centering
    \begin{tabular}{cccc}
        \begin{overpic}[width=0.22\textwidth]{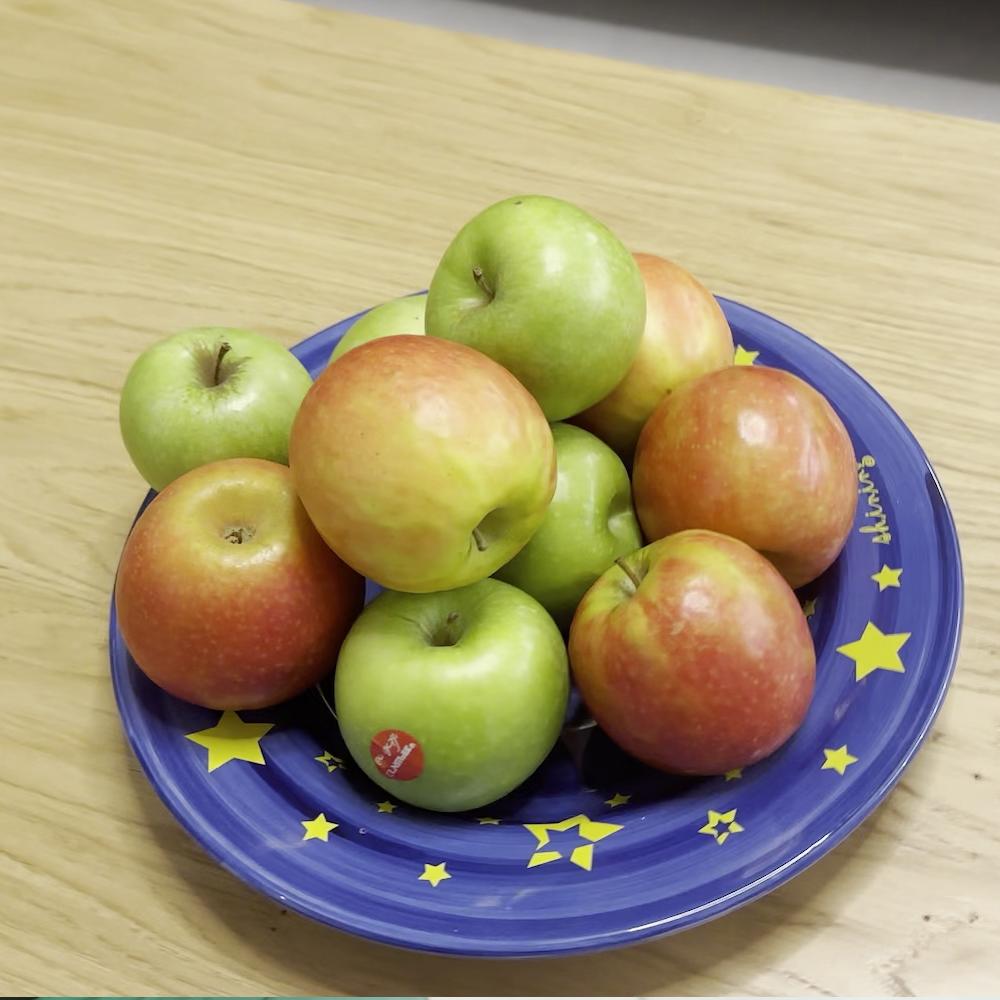}
            \put(5,5){\color{white}\textbf{Image}}
        \end{overpic} & 
        \includegraphics[width=0.22\textwidth]{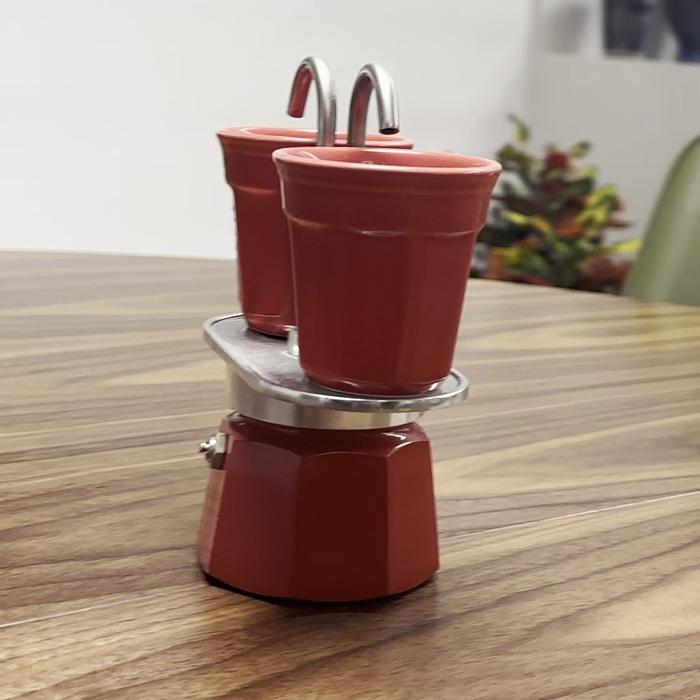} &
        \includegraphics[width=0.22\textwidth]{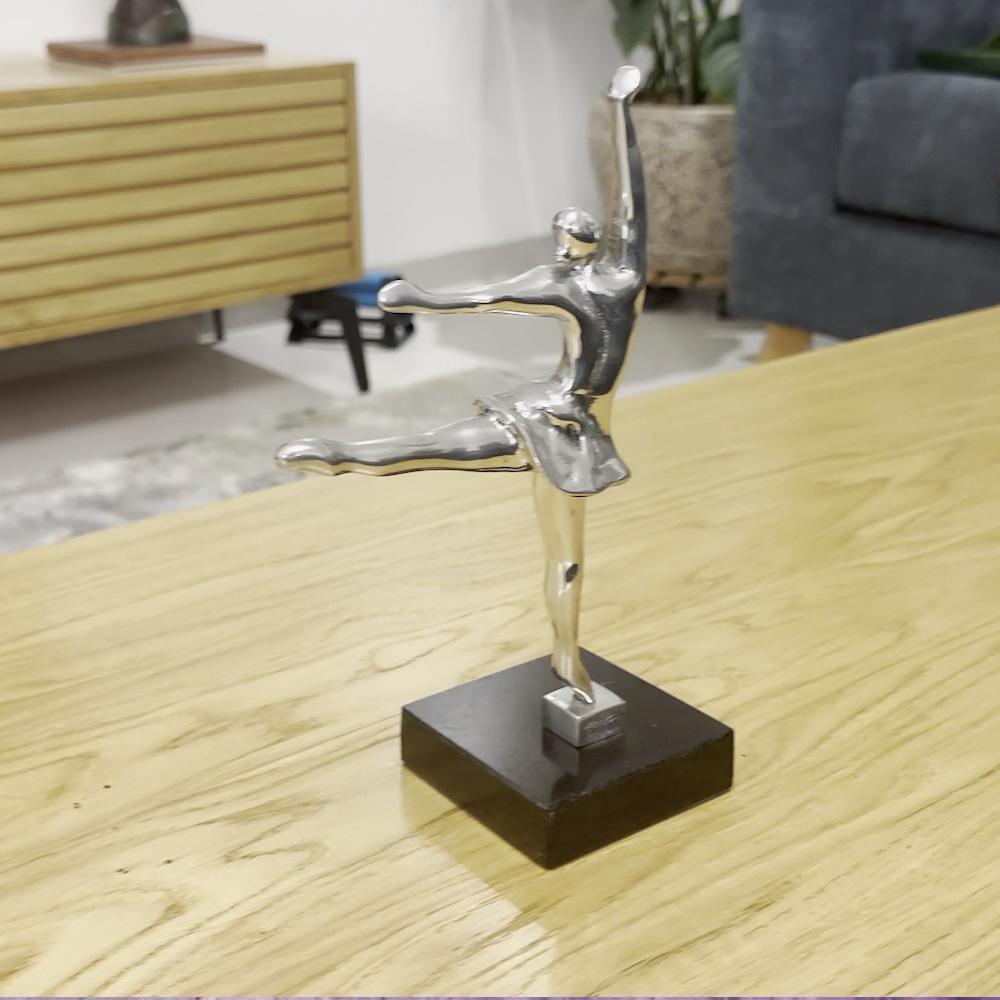} & 
        \includegraphics[width=0.22\textwidth]{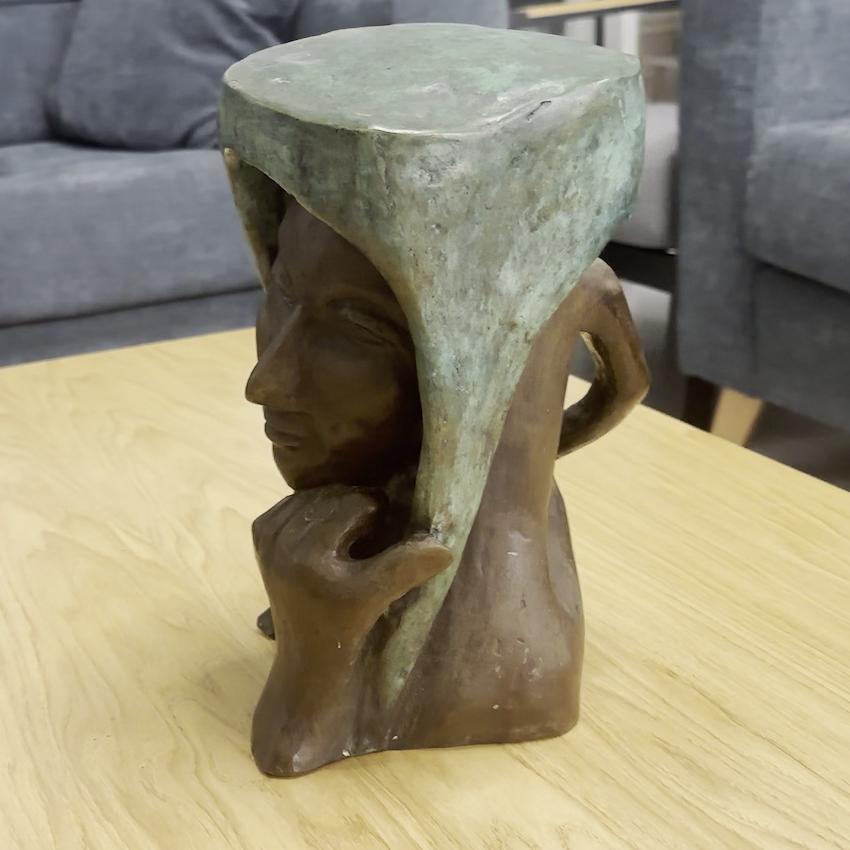} \\
        \begin{overpic}[width=0.22\textwidth]{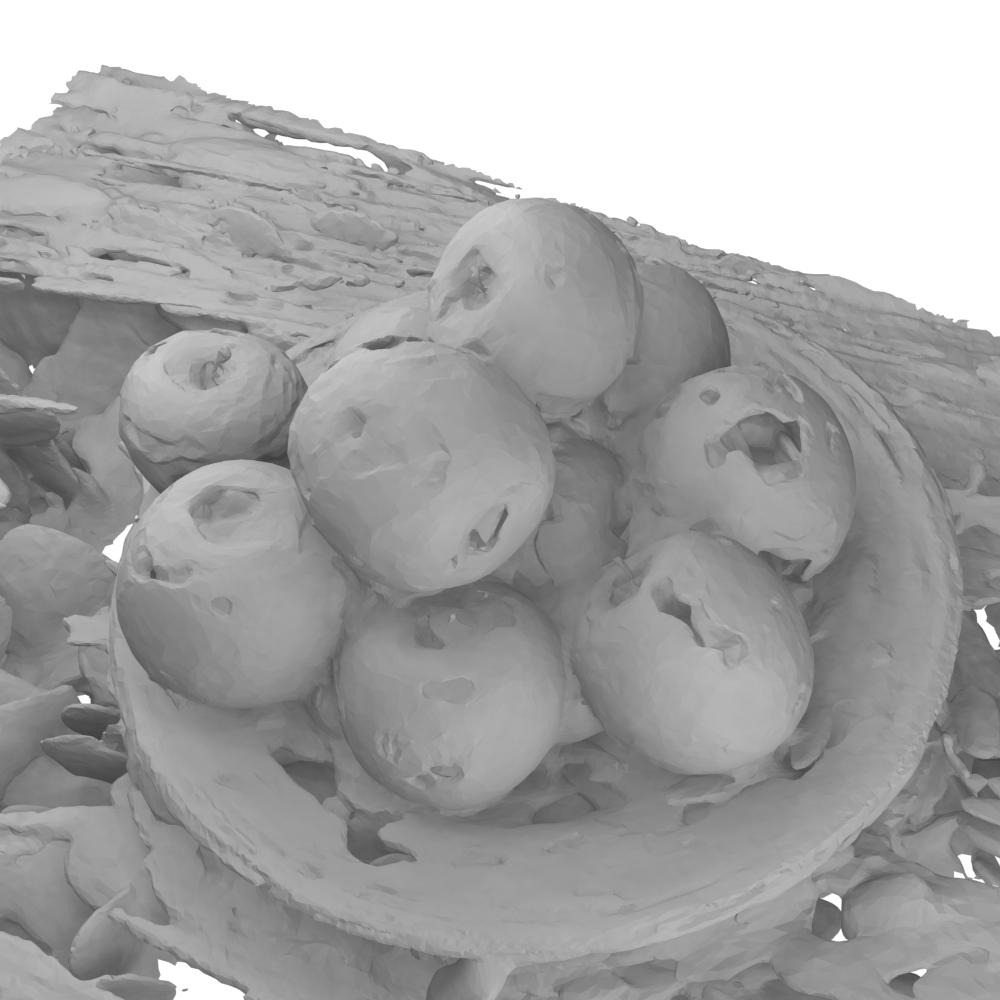}
            \put(5,5){\color{white}\textbf{SuGaR}}
        \end{overpic} & 
        \includegraphics[width=0.22\textwidth]{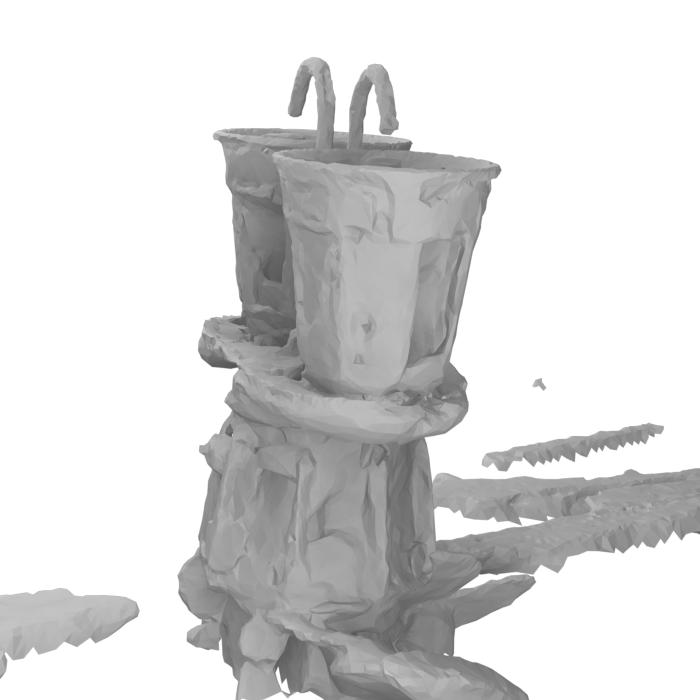} &
        \includegraphics[width=0.22\textwidth]{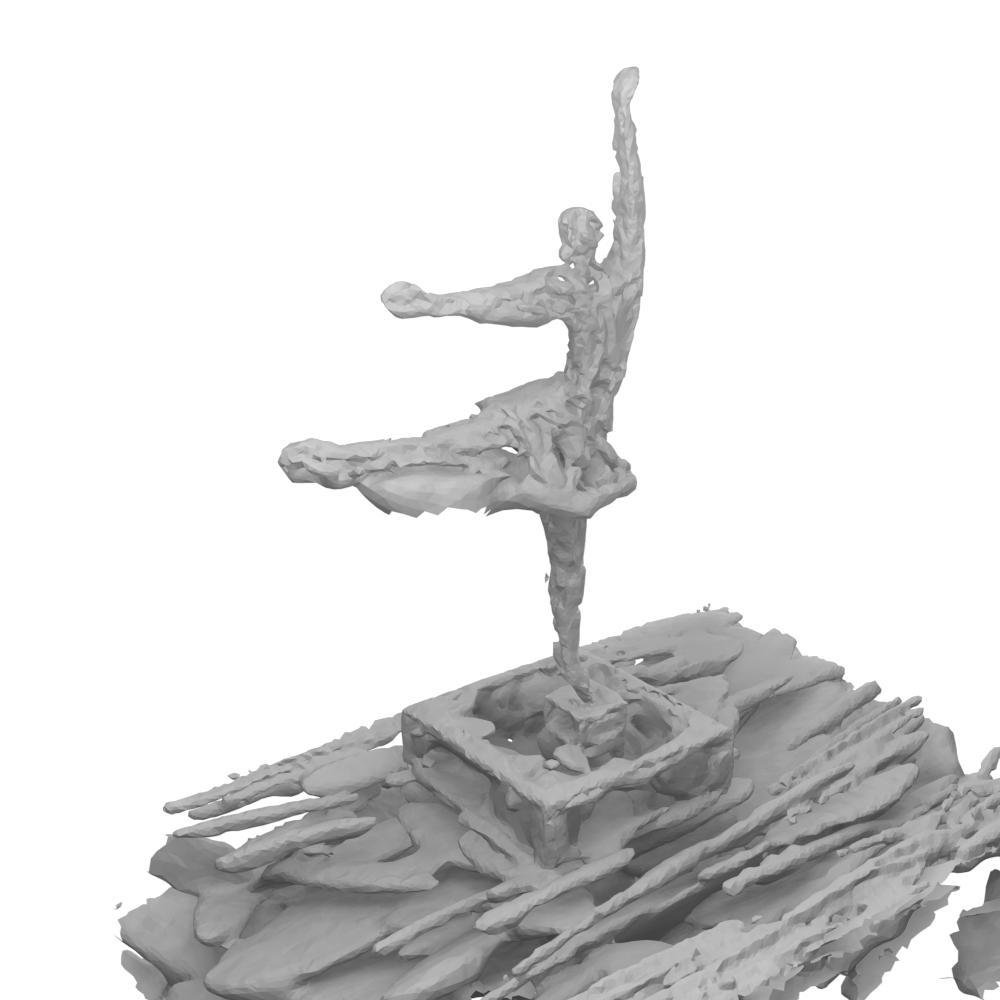} & 
        \includegraphics[width=0.22\textwidth]{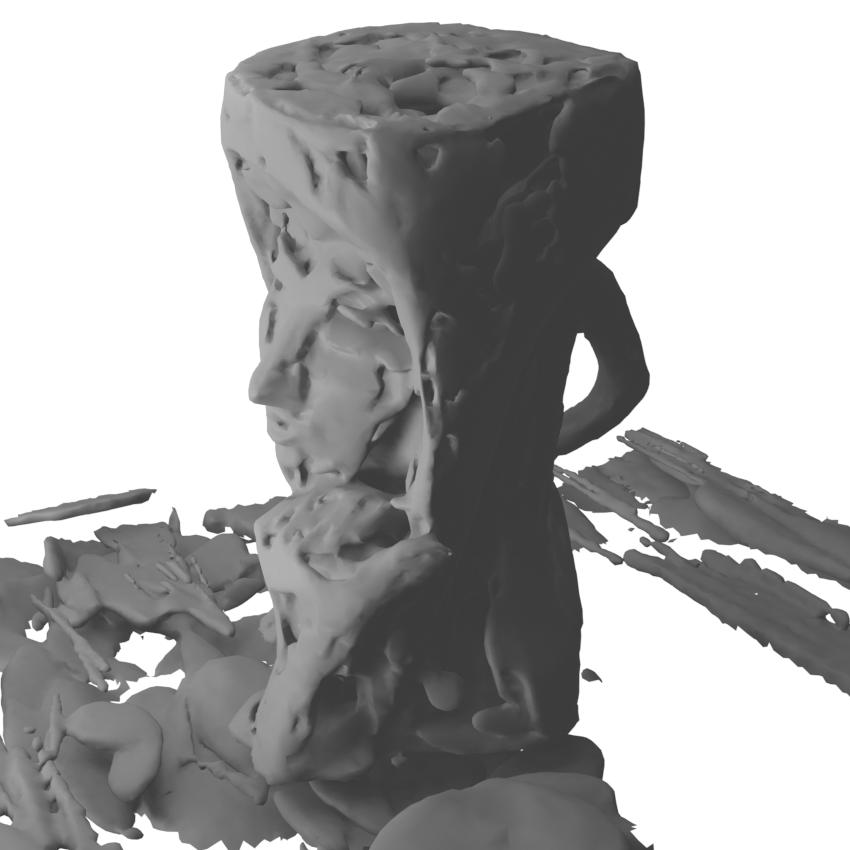} \\
        \begin{overpic}[width=0.22\textwidth]{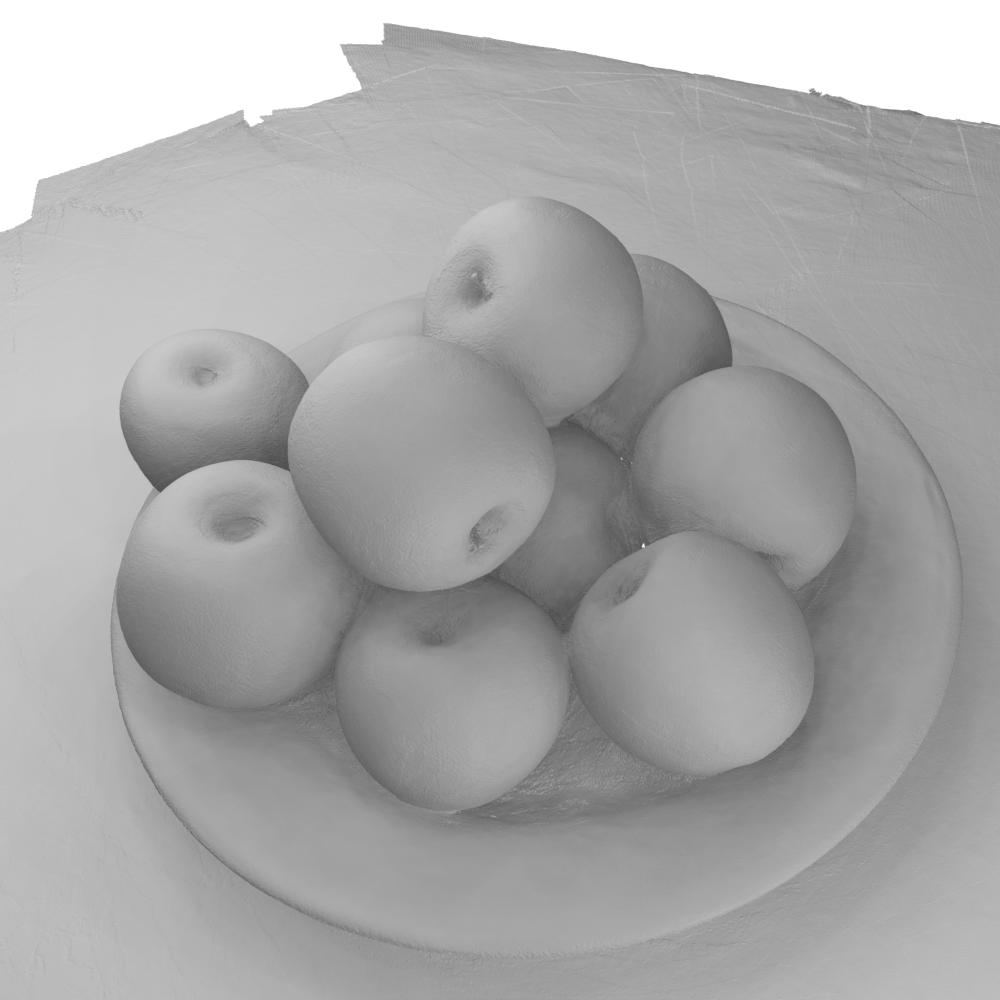}
            \put(5,5){\color{white}\textbf{Ours}}
        \end{overpic} & 
        \includegraphics[width=0.22\textwidth]{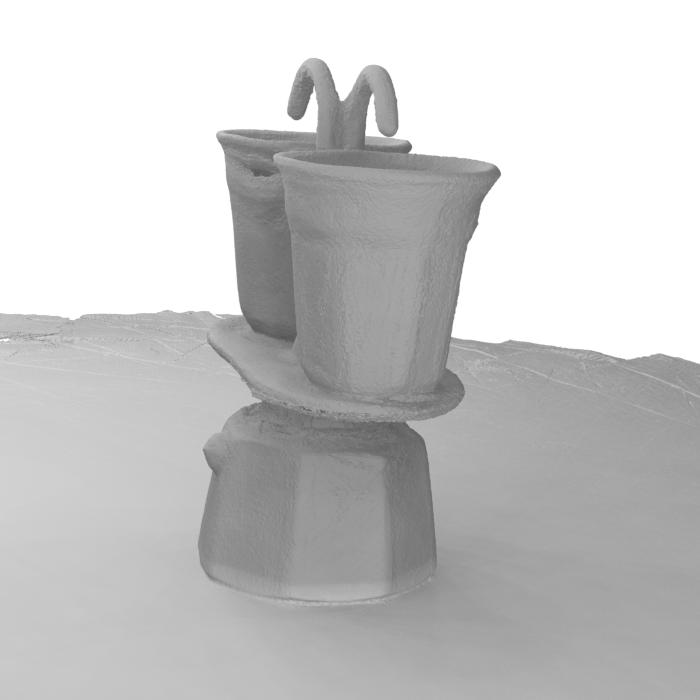} &
        \includegraphics[width=0.22\textwidth]{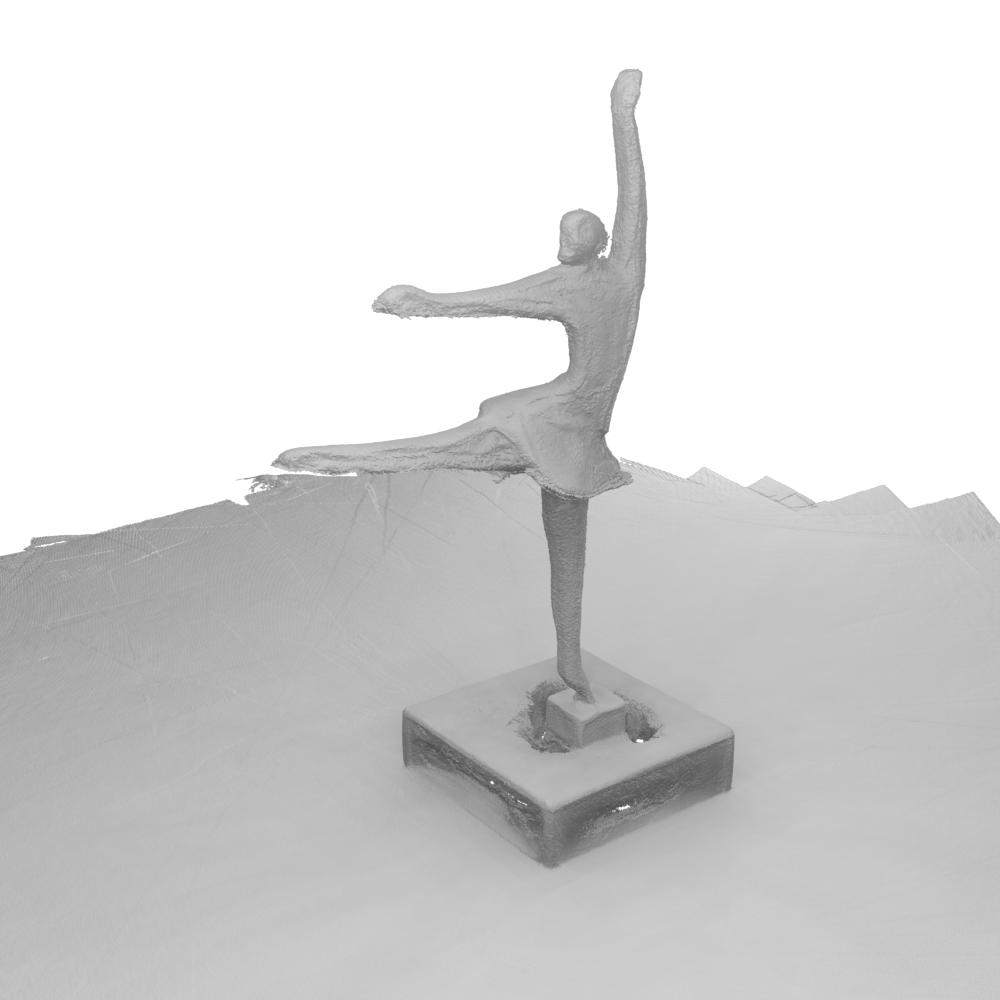} & 
        \includegraphics[width=0.22\textwidth]{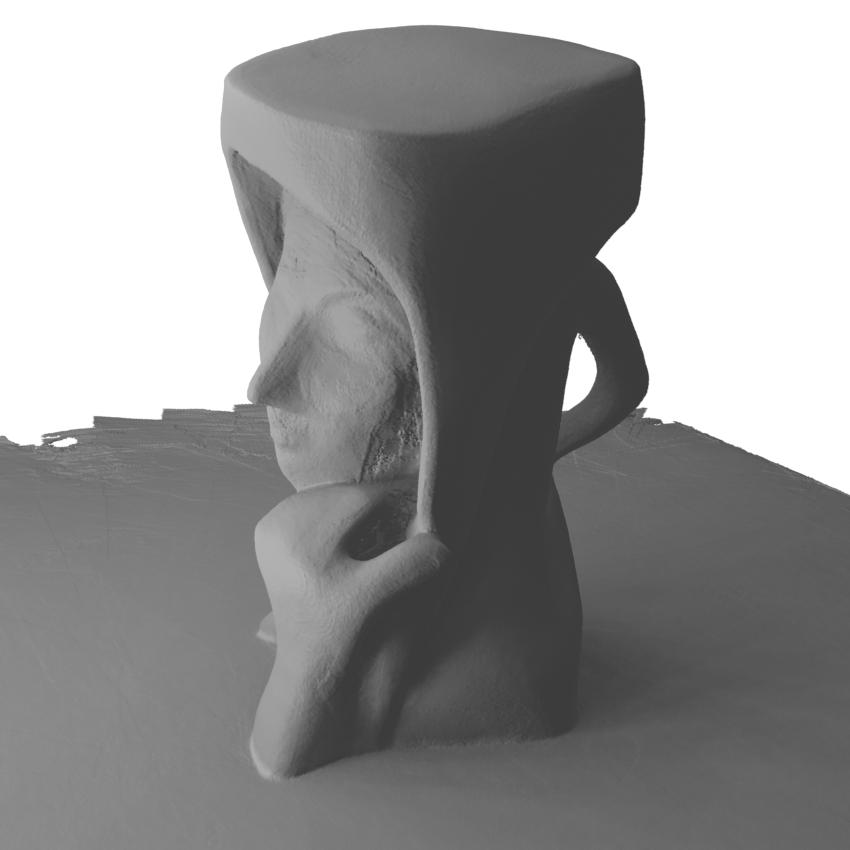} \\
    \end{tabular}
    \caption{Qualitative comparison of mesh reconstruction from in-the-wild videos between our method and SuGaR \cite{SuGaR}.}
    \label{fig:in-the-wild-comparison}
\end{figure}

%_______________________________________________

\subsection{Datasets}

\textbf{DTU \cite{DTU}.} 
This dataset is an MVS dataset, containing scans of small objects, as well as accurate camera poses and 3D point clouds.
We use the dataset and evaluation code from \cite{2DGS}, which calculates the Chamfer Distance (CD) between the reconstructed and ground-truth point clouds.

%_______________________________________________
\noindent\textbf{Tanks and Temples {(TnT)} \cite{tnt}}. 
This dataset contains videos of large objects such as vehicles, buildings and statues. 
These objects are scanned with a laser scanner for an accurate ground-truth 3D point cloud. 
As the videos and the laser scanned objects are difficult to align \cite{rotstein2022multimodal}, for evaluation we use the official TnT \cite{tnt} evaluation alignment method.
It first aligns the point clouds using ICP \cite{ICP}, and then calculates the precision, recall, and F1 score.

%_______________________________________________

\noindent\textbf{Mip-NeRF360 \cite{Mip-NeRF}.} 
This dataset contains scenes taken from a 360 degree view, with emphasis on minimizing photometric variations through controlled capture conditions. 
Since there is no ground-truth in terms of surface reconstruction, we leave this as a qualitative comparison only.

%_______________________________________________
\noindent\textbf{MobileBrick \cite{li2023mobilebrick}.} 
This dataset contains videos of LEGO models, with corresponding 3D ground-truth meshes created from the LEGO 3D model. 
The poses are manually refined and are more accurate than the COLMAP ones \cite{schoenberger2016mvs,schoenberger2016sfm}. 
This dataset is challenging since most of the videos in the test set are taken from a top view of the model, thus, creating occlusions and leaving areas in the model with little visibility.
We use the official evaluation code.

%_______________________________________________
\noindent\textbf{In-the-wild videos.} 
For reconstruction of in-the-wild objects, our method presents a favorable balance between accuracy and computation time.
To validate this claim, we captured scenes containing various objects such as plants, sculptures, figures and everyday items, with intricate geometries and textures, and reconstructed their surface. 
Each video contains one or two cycles of moving around the object, depending on the object's size, without any measures to maintain a persistent radius or pose of the camera, and without any control of the lighting in the environment.
Since these objects are filmed only with a smartphone camera, there is no ground-truth reconstruction for these objects.

%_______________________________________________
\subsection{Baselines}
\textbf{Gaussian Splatting-based methods.} For the DTU \cite{DTU} dataset, We compare our method with the vanilla 3DGS \cite{GS} and SuGaR \cite{SuGaR}, as well as additional state-of-the-art methods, namely 2DGS \cite{2DGS} and Gaussian Opacity Fields (GOF) \cite{GOF}. 
For the TNT \cite{tnt} and Mip-NeRF360 \cite{Mip-NeRF} datasets, as well as for in-the-wild scenes, we compare with SuGaR \cite{SuGaR}.

\noindent \textbf{Neural rendering methods.} 
For the DTU \cite{DTU} dataset, we compare our method with Neuralangelo \cite{Neuralangelo}, VolSDF \cite{VolSDF},  and NeuS \cite{Neus}. 
For the TnT \cite{tnt} dataset, we compare with Neuralangelo \cite{Neuralangelo}, NeuralWarp\cite{Neural_warping} and NeuS \cite{Neus}. 
For the Mip-NeRF360 \cite{Mip-NeRF} dataset, we compare with BakedSDF \cite{bakedsdf}.

\noindent \textbf{Deep MVS.} for in-the-wild scenes and the MobileBrick \cite{li2023mobilebrick} dataset, we compare with MVSformer\cite{mvsformer}, a state-of-the art deep MVS network.

%_______________________________________________
\subsection{Results}

\begin{table}[]
\caption{Quantitative results on the DTU\cite{DTU} dataset, comparing our method with state-of-the art neural and Gaussian Splatting-based methods. 
Chamfer distance - lower is better. \hlred{Red-$1^{st}$}, \hlorange{Orange-$2^{nd}$}, \hlyellow{Yellow-$3^{rd}$}. 
Table adapted from \cite{GOF}.}
\centering
\resizebox{\columnwidth}{!}{
\begin{tabular}{@{}llccccccccccccccc|c|c|cc@{}}
\hline
\multicolumn{2}{c}{} & 24 & 37 & 40 & 55 & 63 & 65 & 69 & 83 & 97 & 105 & 106 & 110 & 114 & 118 & 122 & Mean & Time \\ \cline{3-17} \cline{18-19}
\multirow{4}{*}{\rotatebox[origin=c]{90}{Neural}} & NeRF~\cite{Nerf} & 1.90 & 1.60 & 1.85 & 0.58 & 2.28 & 1.27 & 1.47 & 1.67 & 2.05 & 1.07 & 0.88 & 2.53 & 1.06 & 1.15 & 0.96 & 1.49 & $>$ 12h \\
 & VolSDF~\cite{VolSDF}  & 1.14 & 1.26 & 0.81 & 0.49 & 1.25 & \tbest 0.70 & 0.72 & 1.29 & 1.18 & 0.70 & 0.66 & 1.08 & 0.42 & 0.61 & 0.55 & 0.86 & $>$ 12h \\
 & NeuS~\cite{Neus} & 1.00 & 1.37 & 0.93 & 0.43 & 1.10 & \sbest 0.65 & \sbest 0.57 & 1.48 & 1.09 & 0.83 & \tbest 0.52 & 1.20 & \sbest 0.35 & \sbest 0.49 & 0.54 & 0.84 & $>$ 12h \\
 & Neuralangelo~\cite{Neuralangelo} & \best 0.37 & \best 0.72 & \best 0.35 & \best 0.35 & 0.87 & \best 0.54 & \best 0.53 & \tbest 1.29 & \tbest 0.97 & 0.73 & \best 0.47 & \tbest 0.74 & \best 0.32 & \best 0.41 & \best 0.43 & \best 0.61 & $>$ 12h \\
\cline{2-2} \cline{3-17} \cline{18-19}
\multirow{8}{*}{\rotatebox[origin=c]{90}{Gaussian Splatting}} & 3DGS~\cite{GS} & 2.14 & 1.53 & 2.08 & 1.68 & 3.49 & 2.21 & 1.43 & 2.07 & 2.22 & 1.75 & 1.79 & 2.55 & 1.53 & 1.52 & 1.50 & 1.96 & \best 11.2m \\
 & SuGaR~\cite{SuGaR} & 1.47 & 1.33 & 1.13 & 0.61 & 2.25 & 1.71 & 1.15 & 1.63 & 1.62 & 1.07 & 0.79 & 2.45 & 0.98 & 0.88 & 0.79 & 1.33 & $\sim$~1h \\
 & 2DGS~\cite{2DGS} & \sbest 0.48 & 0.91 & \tbest 0.39 & 0.39 & 1.01 & 0.83 & 0.81 & 1.36 & 1.27 & 0.76 & 0.70 & 1.40 & 0.40 & 0.76 & 0.52 & 0.80 & \sbest 18.8m \\
 & GOF~\cite{GOF} & \tbest 0.50 & 0.82 & \sbest 0.37 & \sbest 0.37 & 1.12 & 0.74 & 0.73 & \best 1.18 & 1.29 & 0.68 & 0.77 & 0.90 & 0.42 & 0.66 & 0.49 & 0.74 & 30m \\
\cline{2-2} \cline{3-17} \cline{18-19}
 & Ours - DLNR Baseline 3.5\% & 0.61 & 0.85 & 0.64 & 0.39 & 0.96 & 1.25 & 0.80 & 1.52 & 1.10 & 0.68 & 0.59 & 0.93 & 0.45 & 0.60 & 0.54 & 0.79 & \tbest $\sim$20m \\
 & Ours - DLNR Baseline 10.5\% & 0.69 & \tbest 0.81 & 0.95 & 0.51 & \sbest 0.82 & 1.06 & 0.72 & \best 1.18 & \best 0.93 & \sbest 0.61 & 0.54 & \best 0.66 & \tbest 0.37 & 0.54 & 0.50 & \tbest 0.73 & \tbest $\sim$20m \\
 & Ours - RAFT Baseline 7\% & 0.59 & \tbest 0.81 & 0.68 & 0.40 & \tbest 0.83 & 1.15 & 0.73 & 1.35 & 1.05 & \tbest 0.62 & 0.53 & 0.80 & 0390 & 0.55 & \tbest 0.49 & \tbest  0.73 & \tbest $\sim$20m \\
 \cline{2-2} \cline{3-17} \cline{18-19}
 & \textbf{Ours - DLNR Baseline 7\%} & \textbf{0.59} & \sbest \textbf{79} & \textbf{0.70} & \tbest \textbf{0.38} & \best \textbf{0.78} & \textbf{1.00} & \tbest \textbf{0.69} & \sbest \textbf{1.25} & \sbest \textbf{0.96} & \best \textbf{0.59} & \sbest \textbf{0.50} & \sbest \textbf{0.68} & \tbest \textbf{0.37} & \tbest \textbf{0.50} & \sbest \textbf{0.46} & \sbest \textbf{0.68} & \tbest  \textbf{$\sim$20m} \\
\hline
\end{tabular}
}
\label{tab:DTU}
\end{table}

\noindent{\textbf{DTU\cite{DTU}}.}
\cref{tab:DTU} presents quantitative results on the DTU \cite{DTU} dataset. 
We ran the 3DGS step of our method for 30000 iterations, but as we show in the supplementary material, we can achieve nearly identical results with only 7000 iterations, reaching a mean Chamfer Distance of $0.70$ with only $\sim12m$ of total runtime per scan. 
We used the same TSDF as in 2DGS \cite{2DGS} and GOF \cite{GOF}, which is based on the Open3D implementation \cite{TSDF}, for a fair comparison. 
We apply the mask supplied with the dataset before inputting the depths into the TSDF algorithm. 
The table is adapted from GOF \cite{GOF} for consistency. Within the splatting-based methods, our method achieves the best score, while maintaining a similar runtime. Additionally, when compared to the neural methods, which take more than 12 hours to reconstruct a single scene, our method surpasses some of the methods, and is comparable with Neuralangelo \cite{Neuralangelo}, the state-of-the-art method. 
Additionally, we test our method with RAFT\cite{RAFT} as the stereo model with the RVC weights \cite{raft_rvc} and with DLNR\cite{DLNR} as the stereo model with the Middlebury weights \cite{middlebury}, achieving better results with DLNR. 
Finally, we compare between three different horizontal baselines: 3.5\%, 7\% and 10.5\% of the scene radius. We achieve the best results with 7\%, noting that increasing or decreasing the horizontal baseline has a negative effect on the results. 
\cref{fig:dtu_example} shows an example of the intermediate representations of our method on one of DTU \cite{DTU} scan105, and the full set of reconstructed meshes is available in the supplementary material.

\begin{table}[]
\caption{Quantitative results on the Tanks and Temples \cite{tnt} benchmark. F1 score - higher is better.}
\centering
\resizebox{\columnwidth}{!}{
\begin{tabular}{@{}|llcccc|c|c|@{}}
\hline
\multicolumn{2}{|c}{} & Barn & Caterpillar & Ignatius & Truck & Mean F1 $\uparrow$ & Runtime \\ \cline{1-8}
\multirow{3}{*}{\rotatebox[origin=c]{90}{Neural}} & NeuralWarp~\cite{Neural_warping} & 0.22 & 0.18 & 0.02 & 0.35 & 0.19 &  \\
 & NeuS~\cite{Neus} & 0.29 & 0.29 & 0.83 & 0.45 & 0.47 & $\sim$16h-48h \\ 
 & Neuralangelo~\cite{Neuralangelo} & 0.70 & 0.36 & 0.89 & 0.48 & 0.61 &  \\ 
\cline{1-2} \cline{3-8}
\multirow{2}{*}{\rotatebox[origin=c]{90}{GS}} & SuGaR~\cite{SuGaR} & 0.01 (0.08) & 0.02 (0.09) & 0.06 (0.34) & 0.05 (0.17) & 0.04 (0.17) & $\sim$2h \\ 
 & \textbf{Ours} & 0.21 (0.22) & 0.17 (0.12) & 0.64 (0.68) & 0.46 (0.40) & 0.37 (0.36) & $\sim$1h \\ 
\hline
\end{tabular}
}
\label{table:tnt}
\end{table}

\begin{figure}[]
    \centering
    \begin{tabular}{ccc}
        % First image
        \begin{overpic}[width=0.25\textwidth]{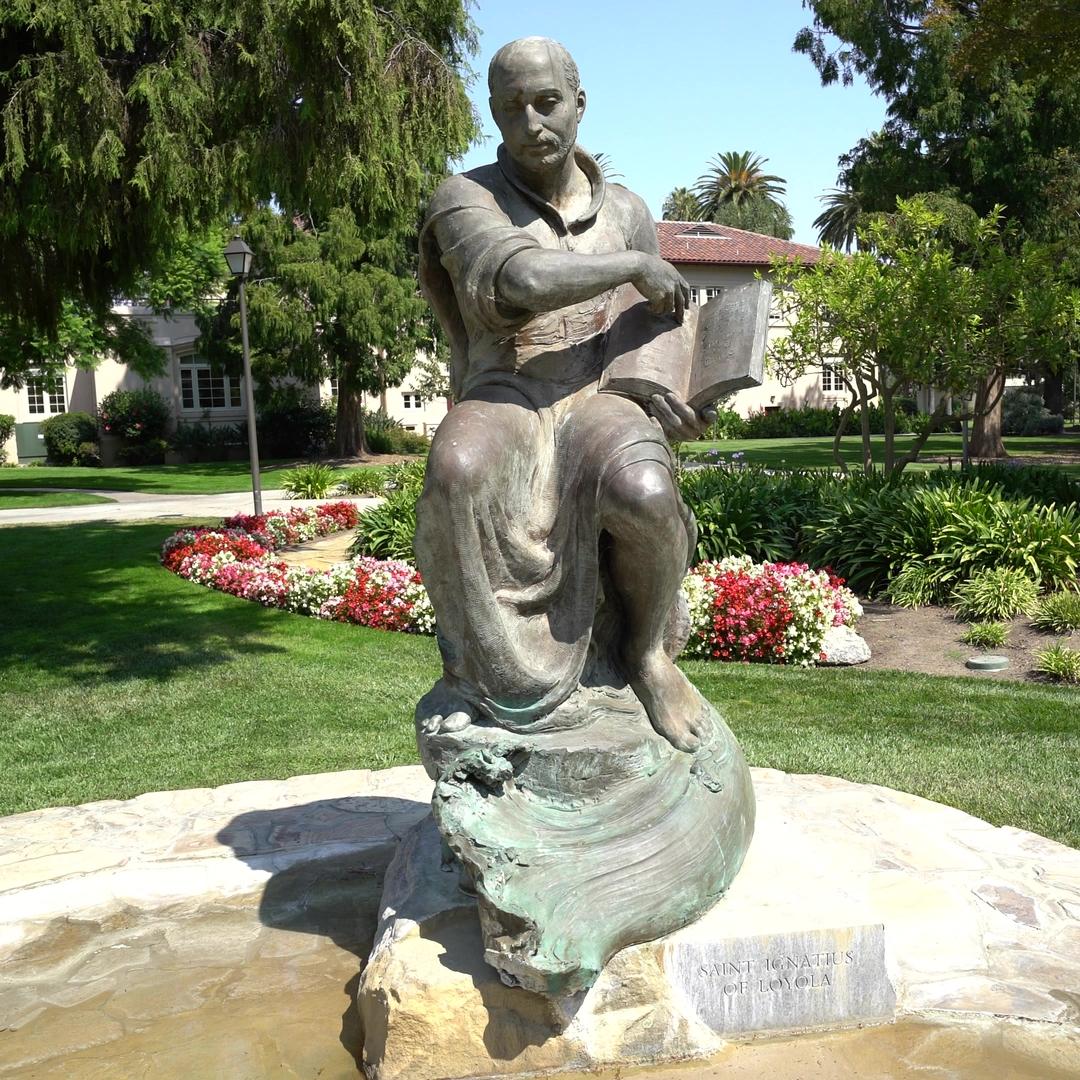}
            \put(5,5){\color{white}\textbf{Image}}
        \end{overpic} & 
        % Second image
        \begin{overpic}[width=0.25\textwidth]{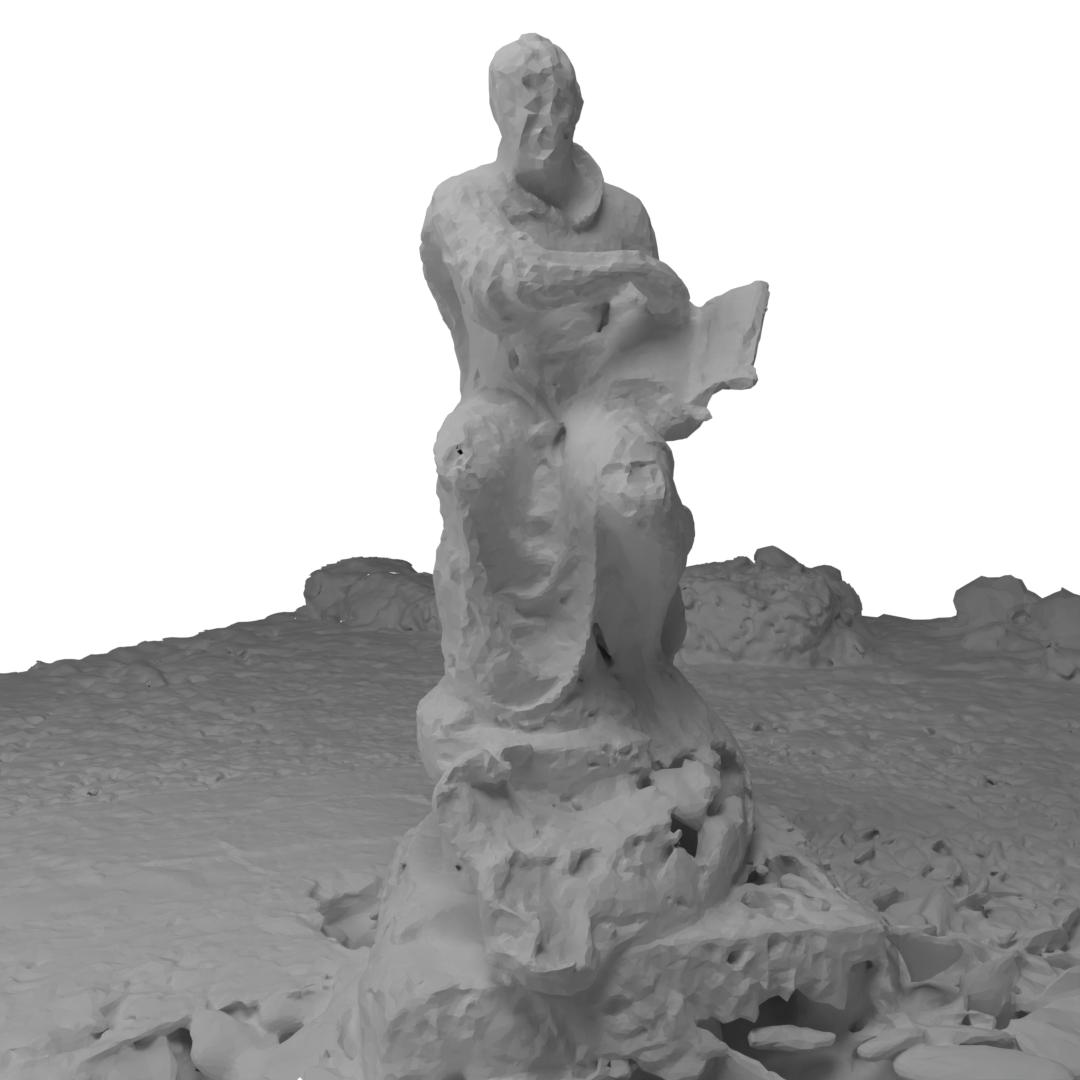}
            \put(5,5){\color{white}\textbf{SuGaR}}
        \end{overpic} &
        % Third image
        \begin{overpic}[width=0.25\textwidth]{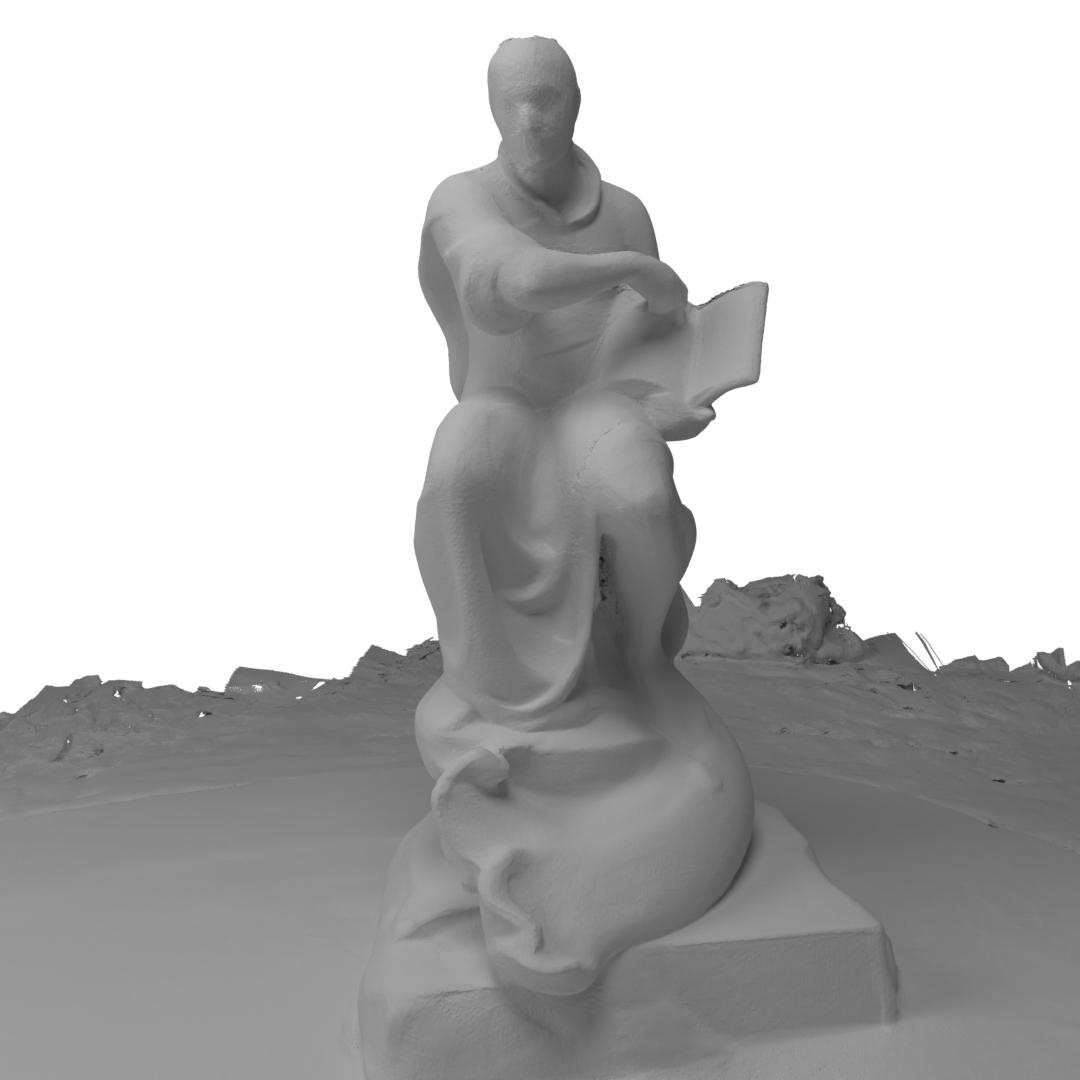}
            \put(5,5){\color{white}\textbf{Ours}}
        \end{overpic} \\
        % Fourth Image
        \multicolumn{3}{c}{
            \begin{overpic}[width=0.8\textwidth]{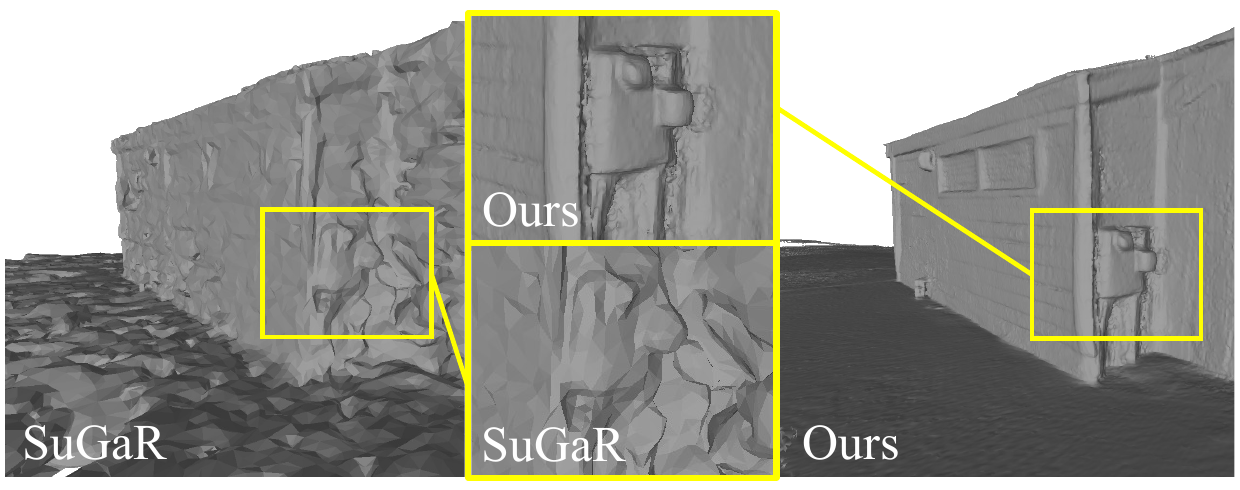}
            \end{overpic}
        } \\
    \end{tabular}
    \caption{Qualitative results on Tanks and Temples \cite{tnt}. 
    Top row: Ignatius scene, compared to SuGaR \cite{SuGaR}. Bottom row: 
    Barn scene, compared to SuGaR.}
    \label{fig:tnt}
\end{figure}

\noindent \textbf{Tanks and Temples\cite{tnt}.} \cref{table:tnt} presents a summary of the reconstruction results on the TnT \cite{tnt} benchmark. 
Since SuGaR \cite{SuGaR} yields a sparse mesh, and thus its recall drops significantly, we include a precision metric that is unaffected by mesh sparsity. However, it is important to note that this metric does not account for missing parts in the reconstruction.
The results show that our method outperforms SuGaR \cite{SuGaR} in both F1 and precision. 
Additionally, it is evident from \cref{fig:tnt} that our method is able to reconstruct fine details such as in the Barn scene.
Moreover, our method has a significant advantage in terms of processing time, requiring less than 60 minutes of total computation time per TnT\cite{tnt} scene, compared to the 16-48 hours needed by neural reconstruction methods. 
The reason for our relatively longer computation times for TnT \cite{tnt} is since each scene containing hundreds of frames, compared to a typical in-the-wild scene which contains less than 100 frames.
It is important to note that the TnT\cite{tnt} dataset predominantly features large scenes, whereas our method is based on 3DGS reconstruction that is designed for accurate reconstruction of smaller ones, and TSDF which is better suited for reconstruction of specific objects. 
This is particularly evident in the case of the Ignatius and Truck scenes, relatively small scenes, where our method performed on-par with the neural reconstruction methods.

\begin{figure}[]
    \centering
    \begin{tabular}{cc}
        % First image
        \begin{overpic}[width=0.35\textwidth]{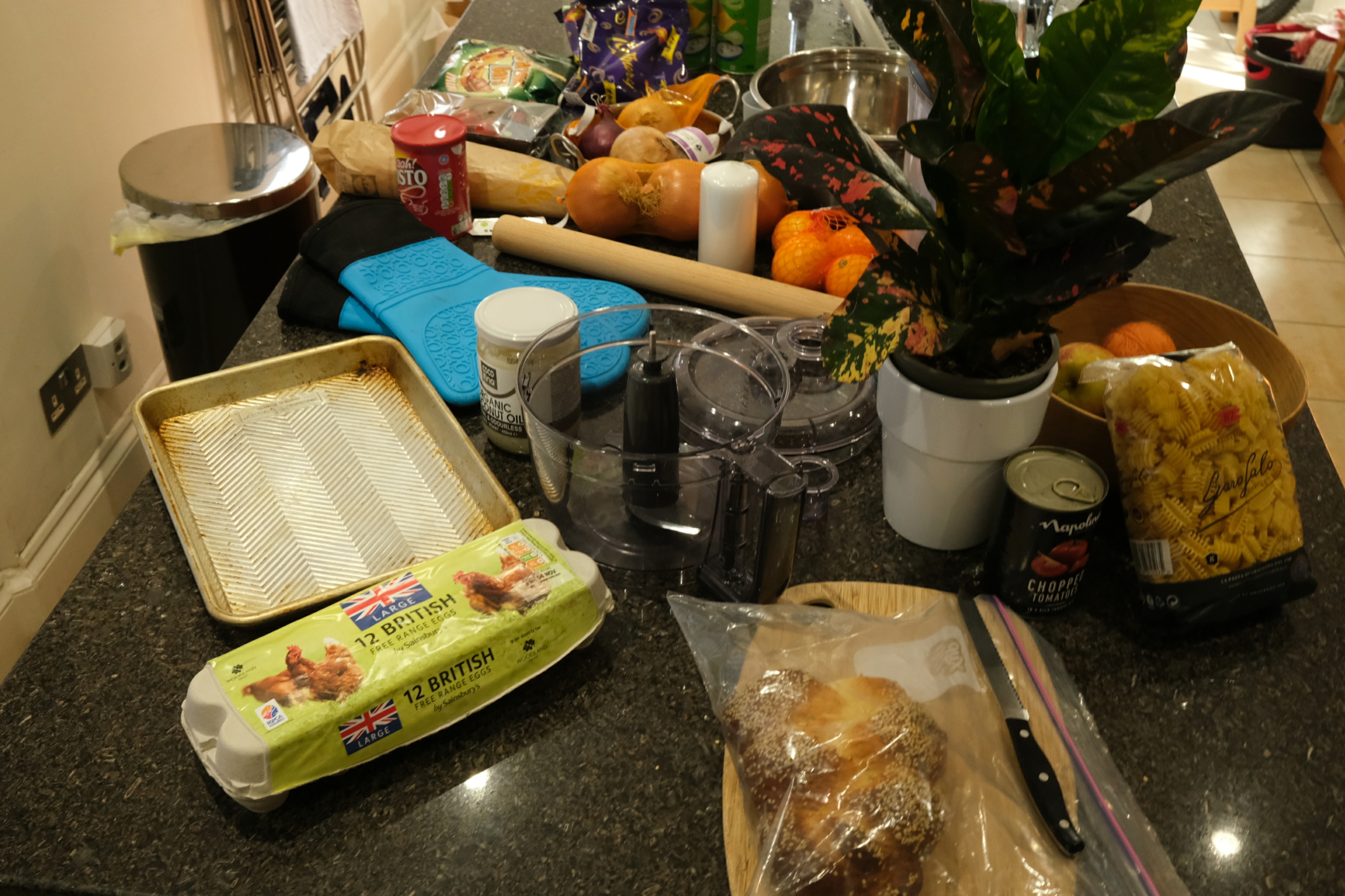}
            \put(5,5){\color{white}\textbf{Image}}
        \end{overpic} & 
        % Second image
        \begin{overpic}[width=0.35\textwidth]{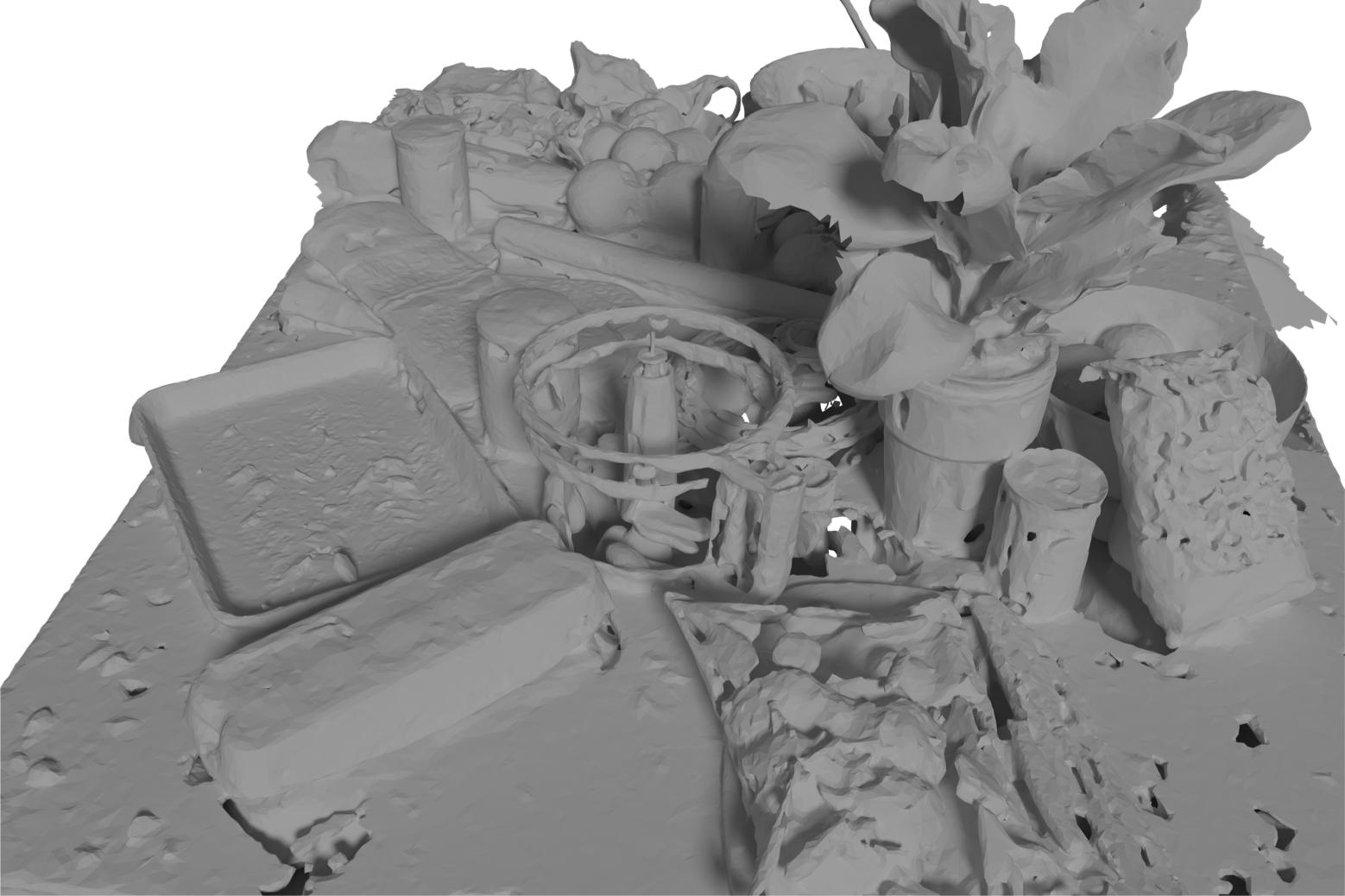}
            \put(5,5){\color{white}\textbf{SuGaR}}
        \end{overpic} \\
        % Third image
        \begin{overpic}[width=0.35\textwidth]{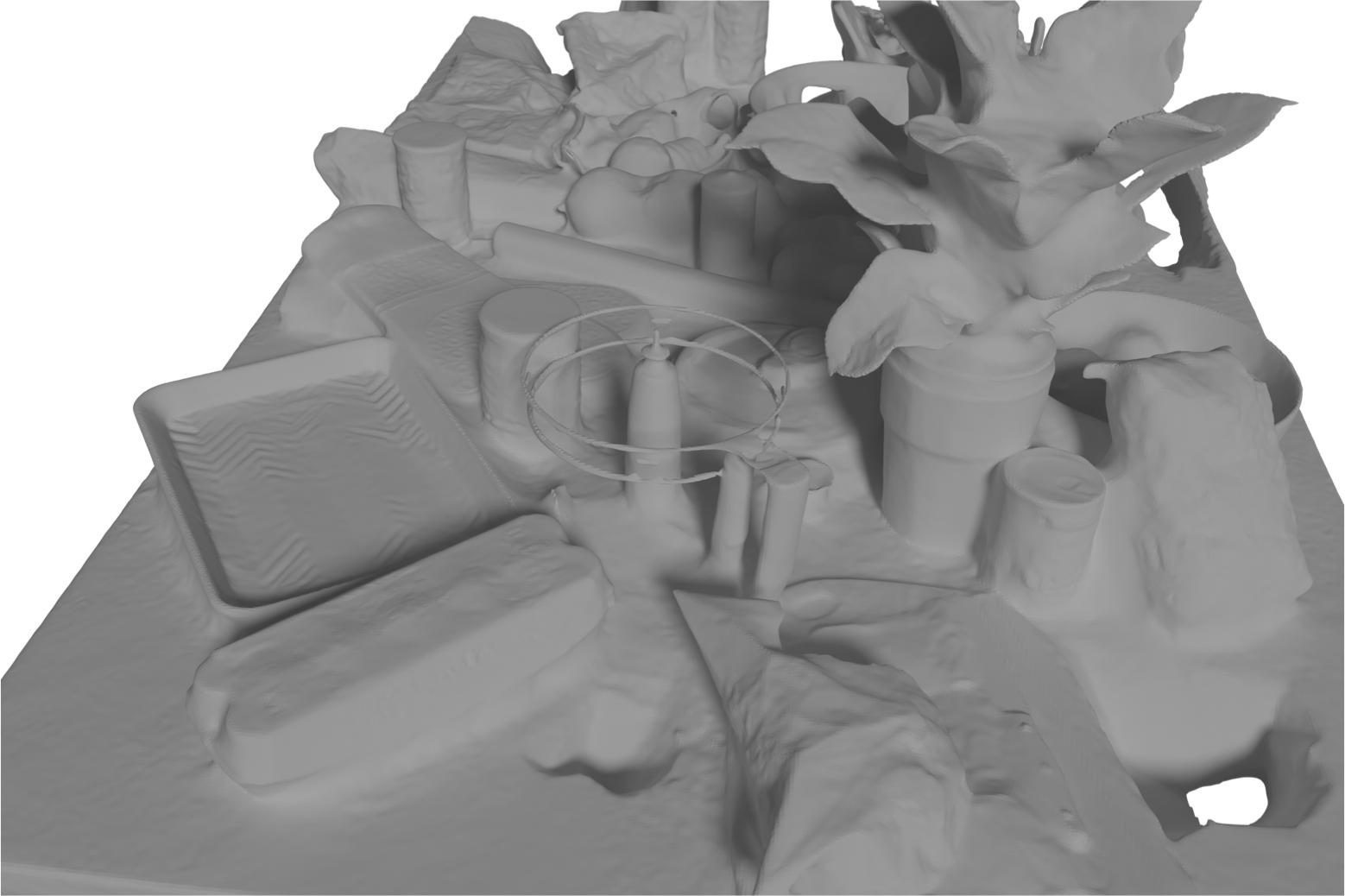}
            \put(5,5){\color{white}\textbf{BakedSDF}}
        \end{overpic} & 
        % Fourth image
        \begin{overpic}[width=0.35\textwidth]{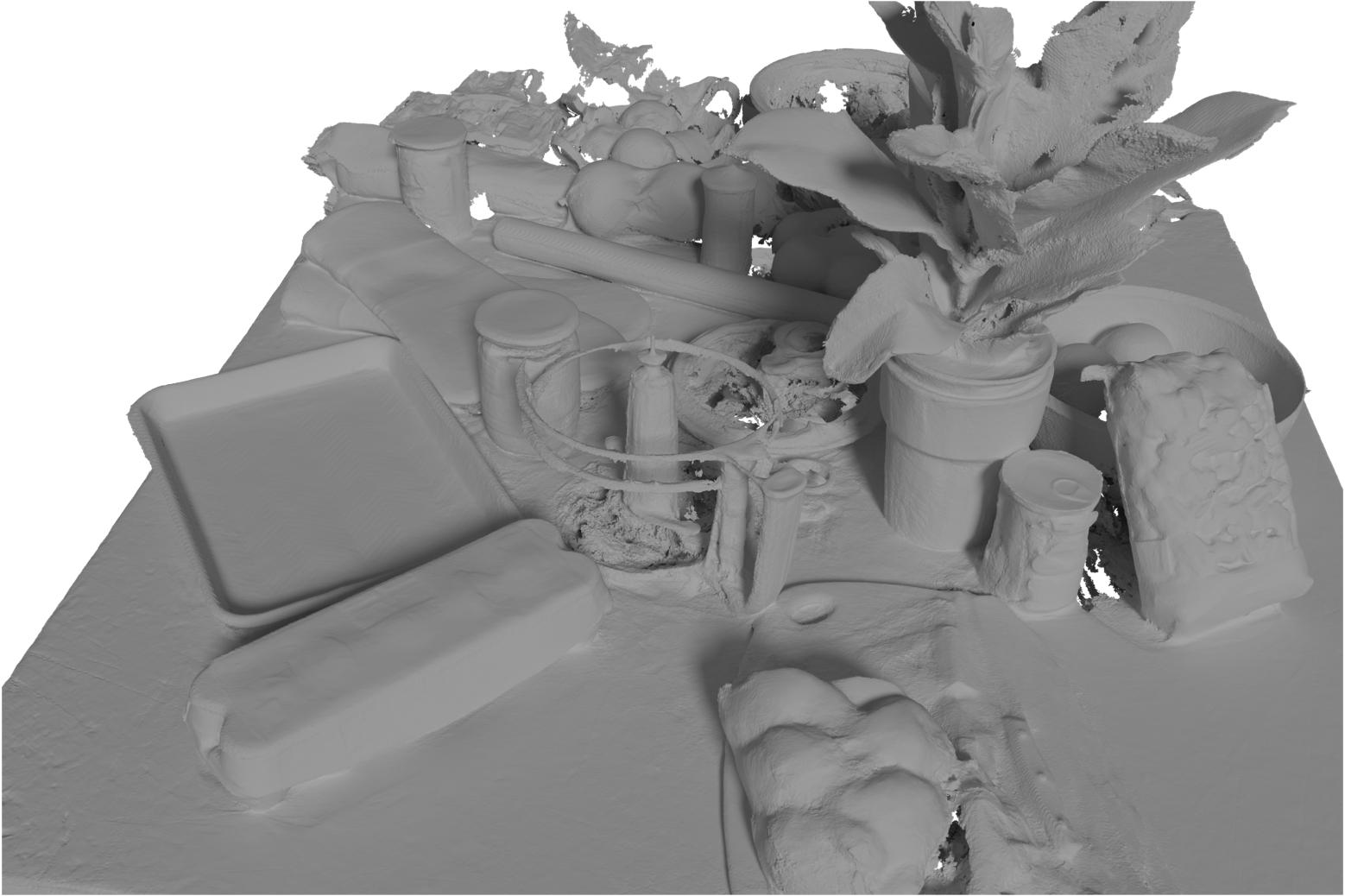}
            \put(5,5){\color{white}\textbf{Ours}}
        \end{overpic} \\
    \end{tabular}
    \caption{Qualitative comparison on Mip-NeRF360 \cite{Mip-NeRF} dataset with BakedSDF \cite{bakedsdf} and SuGaR \cite{SuGaR}.}
    \label{fig:Mip-NeRF_360}
\end{figure}

%___________________________
\noindent\textbf{Mip-NeRF360\cite{Mip-NeRF}.} As illustrated in \cref{fig:Mip2} and \cref{fig:Mip-NeRF_360}, we present a qualitative analysis of scenes from the Mip-NeRF360 \cite{Mip-NeRF} dataset. 
This comparison reveals that our approach surpasses SuGaR\cite{SuGaR} in terms of reconstruction quality and presents on-par results with BakedSDF \cite{bakedsdf}.
Notably, our method excels in reconstructing fine details; for instance, even the small groves in the garden scene's table are evident in the reconstruction, and there are intricate details in the objects on the countertop scene. 
Furthermore, while BakedSDF \cite{bakedsdf} requires 48 hours for training, our method achieves comparable results in less than an hour.
Compared to SuGaR\cite{SuGaR}, our method generates smoother and more realistic surfaces, especially in reflective areas; we note that our countertop is smooth and flat, while SuGaR's countertop has many bumps in areas with glare. 
This is likely due to our model's use of a small baseline for stereo matching, where the reconstruction distortion is relatively small, and additionally, due to our model integrating the reconstructed patches from various viewing directions, which further reduces potential distortions.

\noindent\textbf{In-the-wild comparison.} Our comprehensive in-the-wild comparisons demonstrate the superior performance of our method across various scenes, as illustrated in \cref{fig:in-the-wild-comparison}, with additional results provided in the supplementary material. 
Our method surpasses SuGaR \cite{SuGaR} in extracting accurate and noise-free meshes from 3DGS. 

%_______________________________________________
\subsection{Ablation Study}

 Our main contribution is the use of a pre-trained stereo model to extract depth from a 3DGS scene using novel stereo views. 
 To strengthen our claim of the benefit of using novel stereo views, we perform two ablations, which include replacing steps of our pipeline with deep MVS methods. 

\noindent\textbf{MVS on original images.} 
Our method extracts depth from each original pose, by creating novel stereo-aligned views from that pose and applying a pre-trained stereo matching model. 
One obvious comparison would be to take each original pose and extract the depth using a pre-trained deep MVS model which will take as input the original training set of the scene. 
In the first ablation, we run a pre-trained deep MVS model on the original images, and fuse the resulting depths using TSDF \cite{TSDF}.

\noindent\textbf{MVS on rendered images.} Applying 3DGS to the scene and re-rendering the images from the original poses can reduce distortion and camera noise, which may enhance the quality of the reconstruction regardles of the novel stereo views. 
In the second ablation, we run a pre-trained deep MVS model on the rendered images from the original poses, which are the left image of each stereo-aligned novel view, and fuse the resulting depths using TSDF \cite{tsdf_cuda}.

\noindent\textbf{Evaluation.} We evaluate on the MobileBrick \cite{li2023mobilebrick}  test set, and compare our method against MVSFormer \cite{mvsformer}, a state-of-the-art deep MVS model. 
To ensure a fair comparison, we use TSDF \cite{TSDF} as the fusion method both for our method and the deep MVS model. 

\begin{figure}[]
    \center
    \includegraphics[width=0.95\textwidth]{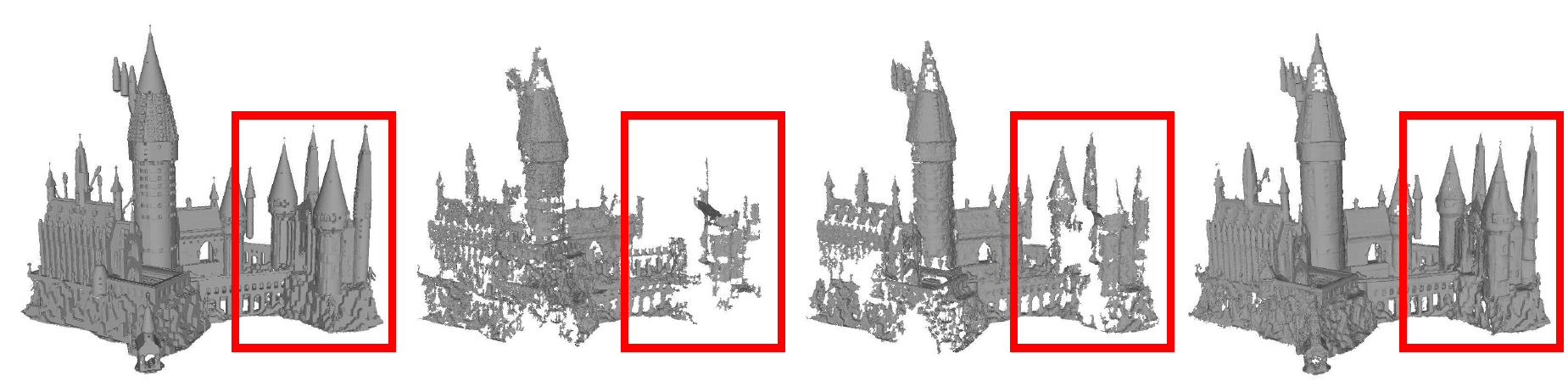}
    \caption{Example from MobileBrick \cite{li2023mobilebrick} dataset, on the castle scene. From left to right: the ground truth mesh, reconstruction of MVSFormer \cite{mvsformer} with original images, reconstruction of MVSFormer with rendered images, and our reconstruction.}
    \label{fig:mobile_brick_ablation}
\end{figure}

\noindent\textbf{Results.} 
\cref{tab:mobile_brick} shows the mean accuracy, recall, F1 and Chamfer Distance of Our method, compared to MVSFormer \cite{mvsformer} with the original and rendered images as input. 
\cref{fig:mobile_brick_ablation} shows a qualitative comparison on one of the scans in the MobileBrick \cite{li2023mobilebrick} dataset, with the rest of the scans, as well as additional examples from in-the-wild scenes, available in the supplementary material. 
Qualitative comparison shows that applying deep MVS directly on the original images results in a reconstruction filled with holes. 
Applying MVS on the rendered images slightly improves the quality of the reconstruction, however, our method still produces a significantly smoother reconstruction.
Quantitative comparison confirms that inputting rendered images to the same MVS model results in a smoother reconstructed surface, as evident by the higher recall, with a slight trade-off in accuracy. 
Overall, our method performs better, as evident by the higher recall and F1 and lower Chamfer distance, even though the manually refined poses given by the MobileBrick \cite{li2023mobilebrick} dataset should give an advantage to MVSFormer \cite{mvsformer}.

\begin{table}[]
    \caption{Ablation study results on MobileBrick \cite{li2023mobilebrick} dataset. 
    We compare our method against MVSFormer \cite{mvsformer} with two types of inputs: the original images with the original refined poses, and the rendered images with the same poses.}
    \centering
    \begin{tabular}{|l|ccc|ccc|c|}
    \hline
    \multicolumn{1}{|c|}{} & \multicolumn{3}{c|}{2.5mm Radius} & \multicolumn{3}{c|}{5mm Radius} & \multirow{2}{*}{\begin{tabular}[c]{@{}c@{}}Chamfer Distance \\ (mm) $\downarrow$\end{tabular}} \\
    \cline{2-7}
    \multicolumn{1}{|c|}{\multirow{-2}{*}{Method}} & Acc $\uparrow$ & Recall $\uparrow$ & F1 $\uparrow$ & Acc $\uparrow$ & Recall $\uparrow$ & F1 $\uparrow$ & \\
    \hline
    MVSFormer \cite{mvsformer} &  \textbf{80.77} &  55.02 &  64.60 &  96.33 &  71.32 &   81.14 &  9.11 \\
    MVSFormer + Rendered &  80.16 &  59.92 &  68.04 &  \textbf{96.84} &  77.50 &   85.59 &  7.10 \\
    \hline
    Ours &  68.77 &  \textbf{69.27} &  \textbf{68.94} &  89.46 &  \textbf{87.37} &  \textbf{88.28} &  \textbf{4.94}\\
    \hline
    \end{tabular}
    \label{tab:mobile_brick}
\end{table}

%_______________________________________________
\section{Limitations}

\begin{figure}[]
    \centering

    \mbox{
        \includegraphics[width=0.45\textwidth]{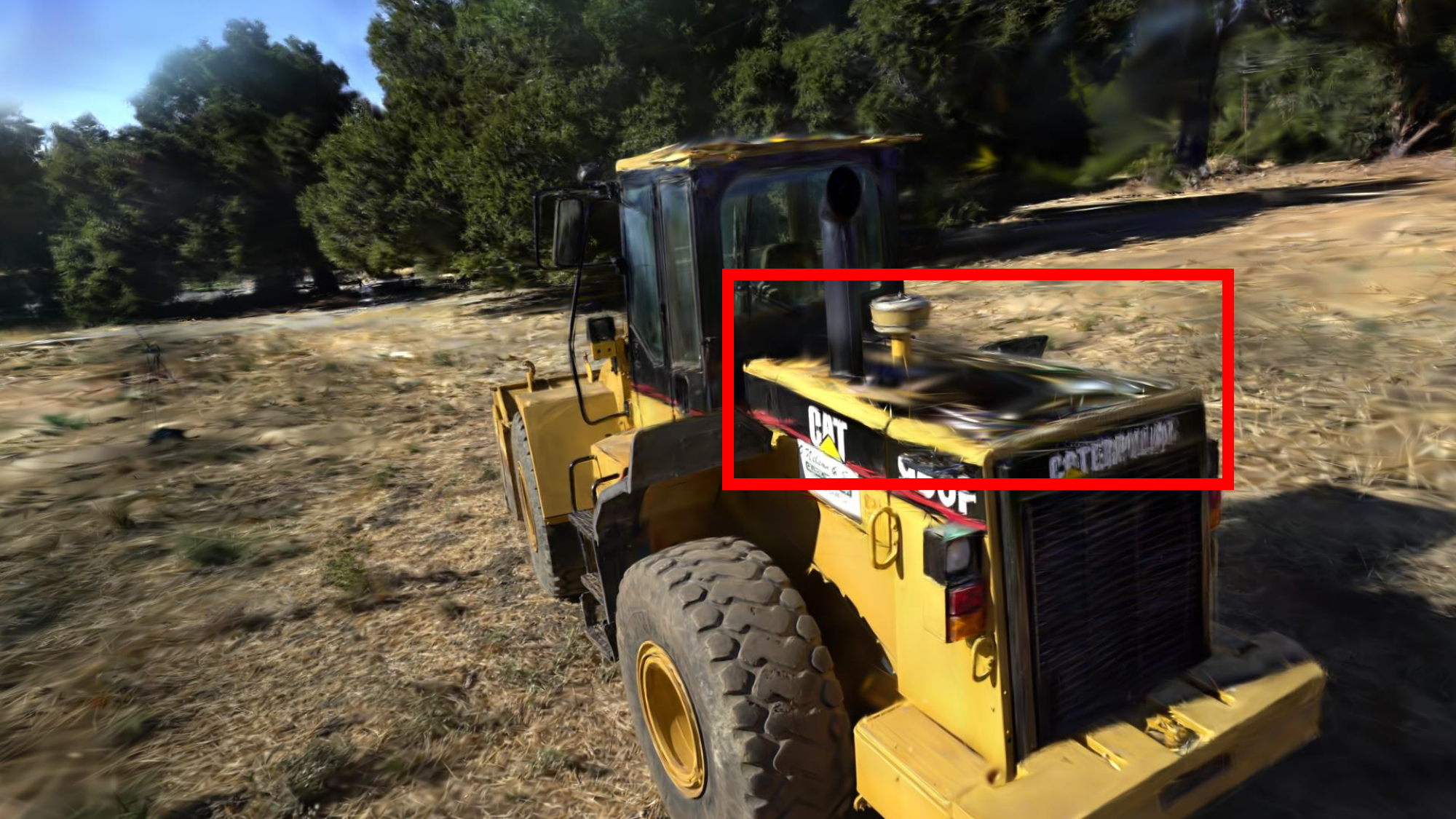}

        \includegraphics[width=0.45\textwidth]{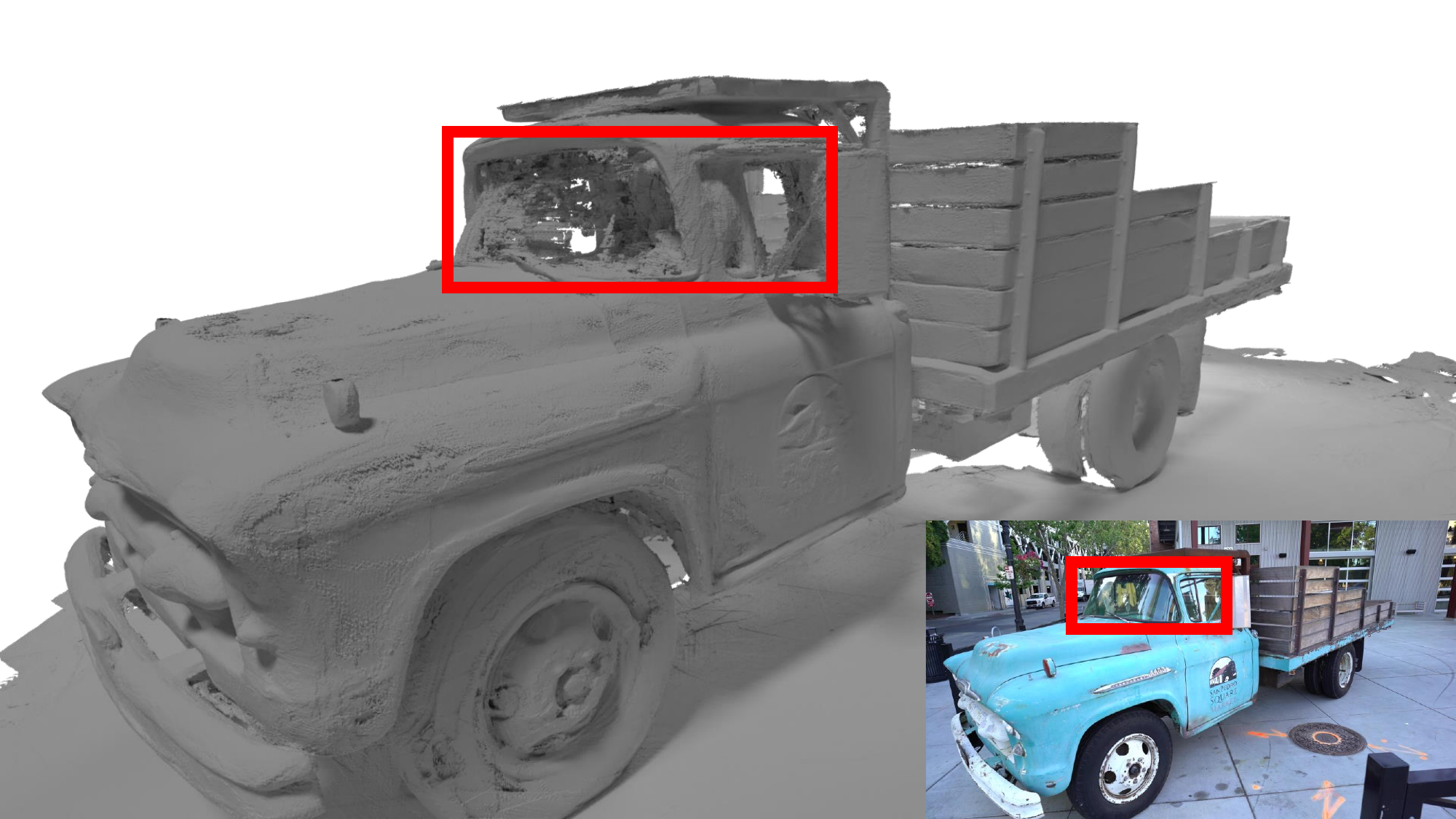}
        }
    \caption{Examples of limitations of our method. 
    On the left, we show a rendered image from the Caterpillar scene from TnT \cite{tnt} dataset, highlighting an area with ``floater''  Gaussians. 
    On the right, the Truck scene from TnT \cite{tnt} dataset, highlighting the missing windshield.}
    \label{fig:limitations}
\end{figure}

Our pipeline consists of 3DGS, depth extraction via stereo, and TSDF fusion. 
Each of these steps exhibits limitations that can impact the final reconstruction: 3DGS can produce noisy ``floater" Gaussians in areas which aren't sufficiently covered in the original training images, as can be seen in the right side of \cref{fig:limitations}. 
Additionally, stereo matching models are known to struggle with transparent surfaces, as can be seen in the left side of \cref{fig:limitations}. 
Finally, TSDF fusion does not scale well for larger scenes, such as the Meetingroom and Courthouse scenes from TnT \cite{tnt}. 
Swapping the 3DGS and stereo with future versions which will have improved accuracy and robustness, as well as adding fusion methods better suited for larger scenes, should help mitigate the effect of these limitations.

%_______________________________________________
\section{Conclusion}
We introduce a novel approach for bridging the gap between noisy Gaussian point clouds and smooth surfaces in 3D. 
Instead of applying geometric optimizations directly on the Gaussians and extracting the depth using their locations, we use a pre-trained stereo model as a geometric prior with real-world knowledge to extract the depth. 
While this approach preserves the inherent properties of the 3DGS representation, it also enhances the accuracy and fidelity of the reconstructed surfaces. 
Our experimental results on DTU \cite{DTU}, Tanks and Temples\cite{tnt}, Mip-NeRF360\cite{Mip-NeRF}, MobileBrick \cite{li2023mobilebrick} and real-world scenes captured using smartphones - demonstrate the superiority of our method over the current state-of-the-art methods for surface reconstruction from Gaussian splatting models, offering both improved accuracy and significantly shorter computation times compared to neural methods.

\bibliographystyle{splncs04}
\bibliography{main}

\clearpage\newpage
\clearpage

\appendix
\setcounter{page}{1}
\counterwithin{figure}{section}
\setcounter{figure}{1}
\counterwithin{table}{section}
\setcounter{table}{1}
{
\centering
\Large
\textbf{Surface Reconstruction from Gaussian Splatting via Novel Stereo Views}\\
Supplementary Material\\}
%_______________________________________________

%_______________________________________
\section{Additional Ablations}

\begin{table}[]
\caption{Additional ablations on the DTU\cite{DTU} dataset, regarding the number of 3DGS iterations and the use of occlusion masks. Chamfer distance - lower is better. \hlred{Red-$1^{st}$}, \hlorange{Orange-$2^{nd}$}, \hlyellow{Yellow-$3^{rd}$}.}
\centering
\resizebox{\columnwidth}{!}{
\begin{tabular}{@{}llccccccccccccccc|c|c|cc@{}}
\hline
\multicolumn{2}{c}{} & 24 & 37 & 40 & 55 & 63 & 65 & 69 & 83 & 97 & 105 & 106 & 110 & 114 & 118 & 122 & Mean & Time \\ 

\cline{3-17} \cline{18-19}

 & Ours - 7000 w/o occlusion & 0.75 & {0.81} & \sbest{0.67} & \tbest{0.40} & 0.91 & \sbest{0.98} & 0.82 & {1.46} & 1.03 & 0.76 & 0.79 & {0.89} & \tbest{0.47} & 0.58 & \sbest{0.46} & 0.78 & \best $\sim$12m \\
 
 & Ours - 7000 w/ occlusion & \tbest 0.62 & \tbest{0.78} & \tbest{0.70} & \best{0.37} & \tbest 0.80 & {1.01} & \tbest 0.71 & \sbest{1.28} & \tbest0.98 & \sbest 0.60 & \tbest 0.55 & \sbest{0.76} & \best{0.35} & \sbest 0.53 & \sbest{0.46} & \tbest 0.70 & \best $\sim$12m \\
 
 & Ours - 30000 w/o occlusion & {0.71} & 0.82 & \best{0.66} & {0.42} & 0.88 & \sbest{0.98} & 0.78 & {1.46} & 1.03 & \tbest{0.75} & {0.72} & \tbest{0.82} & {0.48} & {0.57} & \sbest{0.46} & 0.77 & \sbest $\sim$20m \\
 
 \cline{2-2} \cline{3-17} \cline{18-19}
 
 & \textbf{Ours - 30000 w/ occlusion} & \best{0.59} & \best{0.79} & \tbest{0.70} & \sbest{0.38} & \sbest{0.78} & \tbest{1.00} & \sbest{0.69} & \best{1.25} & \best{0.96} & \best{0.59} & \best{0.50} & \best{0.68} & \sbest{0.37} & \best{0.50} & \sbest{0.46} & \best{0.68} & \sbest $\sim$20m \\
 
 \cline{2-2} \cline{3-17} \cline{18-19}
 
 & Ours - 60000 w/o occlusion & 0.72 & 0.82 & \best{0.66} & 0.43 & {0.86} & \best{0.96} & {0.77} & \tbest{1.42} & {1.02} & \tbest{0.75} & 0.74 & \tbest{0.82} & 0.49 & \tbest{0.56} & \best{0.45} & {0.76} & \tbest $\sim$30m \\
 
 & Ours - 60000 w/ occlusion & \sbest{0.60} & \sbest{0.80} & 0.71 & \sbest{0.38} & \best{0.75} & \tbest{1.00} & \best{0.68} & \best{1.25} & \sbest{0.97} & \sbest{0.60} & \sbest{0.53} & \best{0.68} & \sbest{0.37} & \best{0.50} & \sbest{0.46} & \sbest{0.69} & \tbest $\sim$30m \\
 
\hline
\end{tabular}
}
\label{tab:DTU_ablations}
\end{table}

\begin{figure}[]
    \centering
    \begin{tabular}{cccc}
    \textbf{Image} & \textbf{7000 it.} & \textbf{30000 it.} & \textbf{60000 it.} \\
    \includegraphics[width=0.22\textwidth]{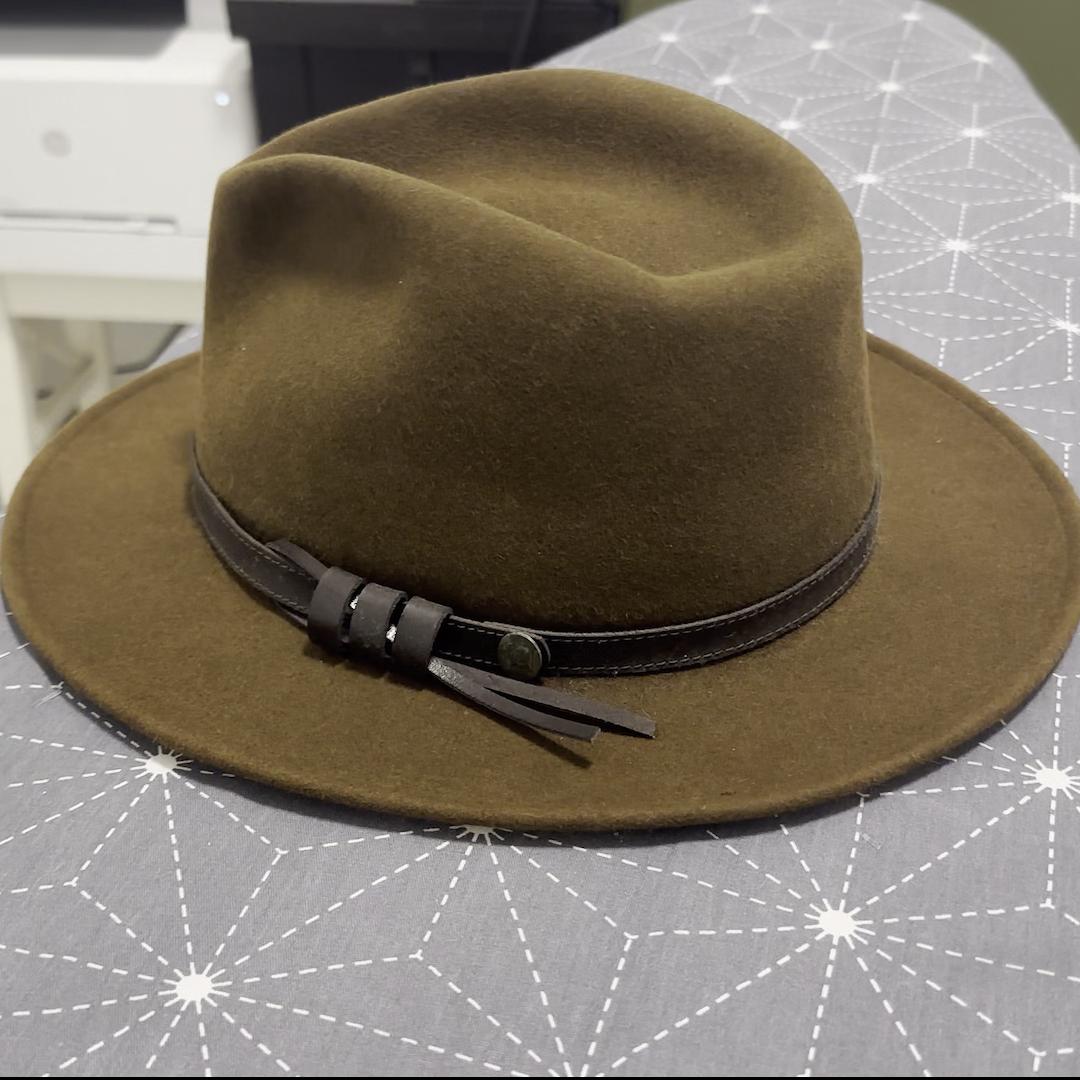} & 
    \includegraphics[width=0.22\textwidth]{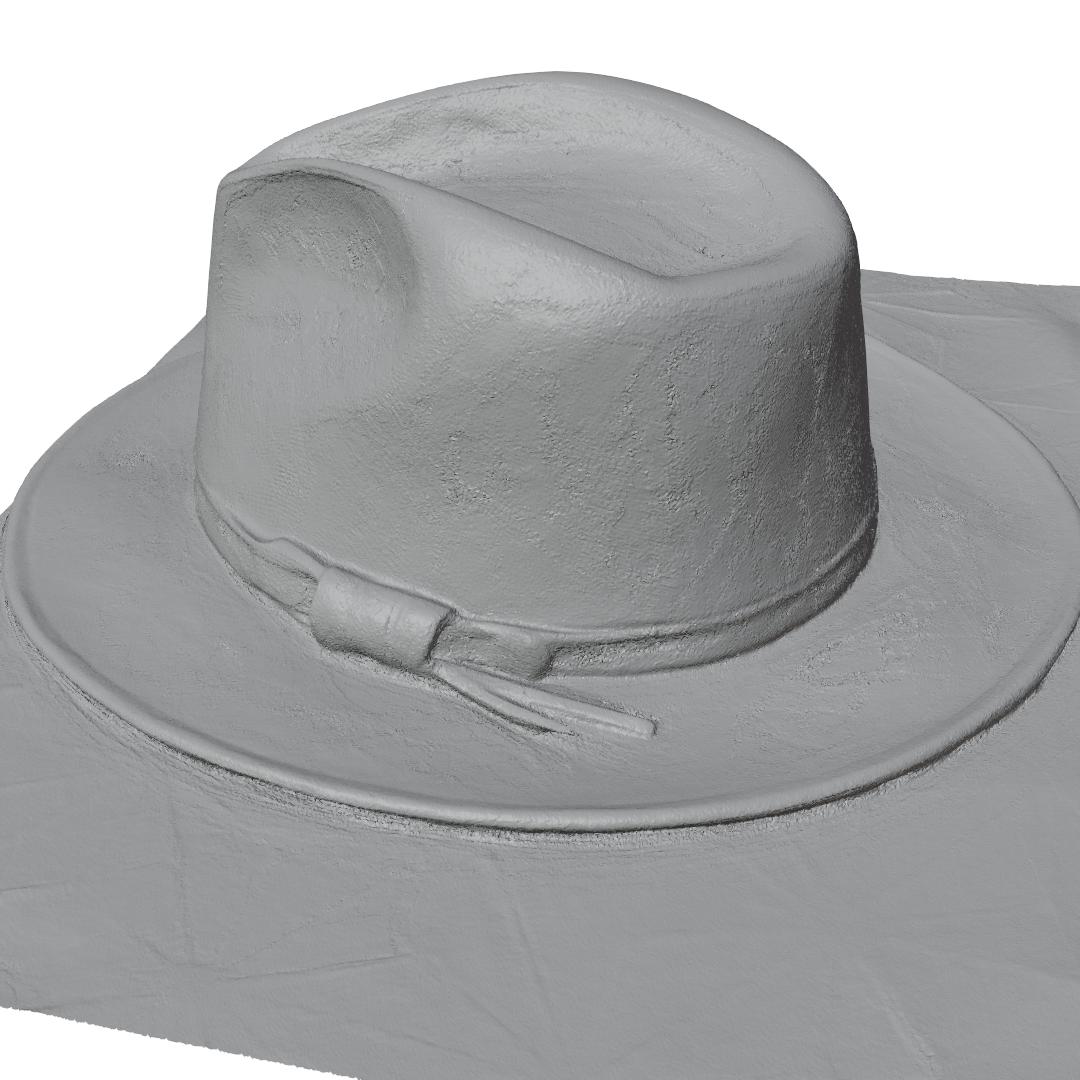} &
    \includegraphics[width=0.22\textwidth]{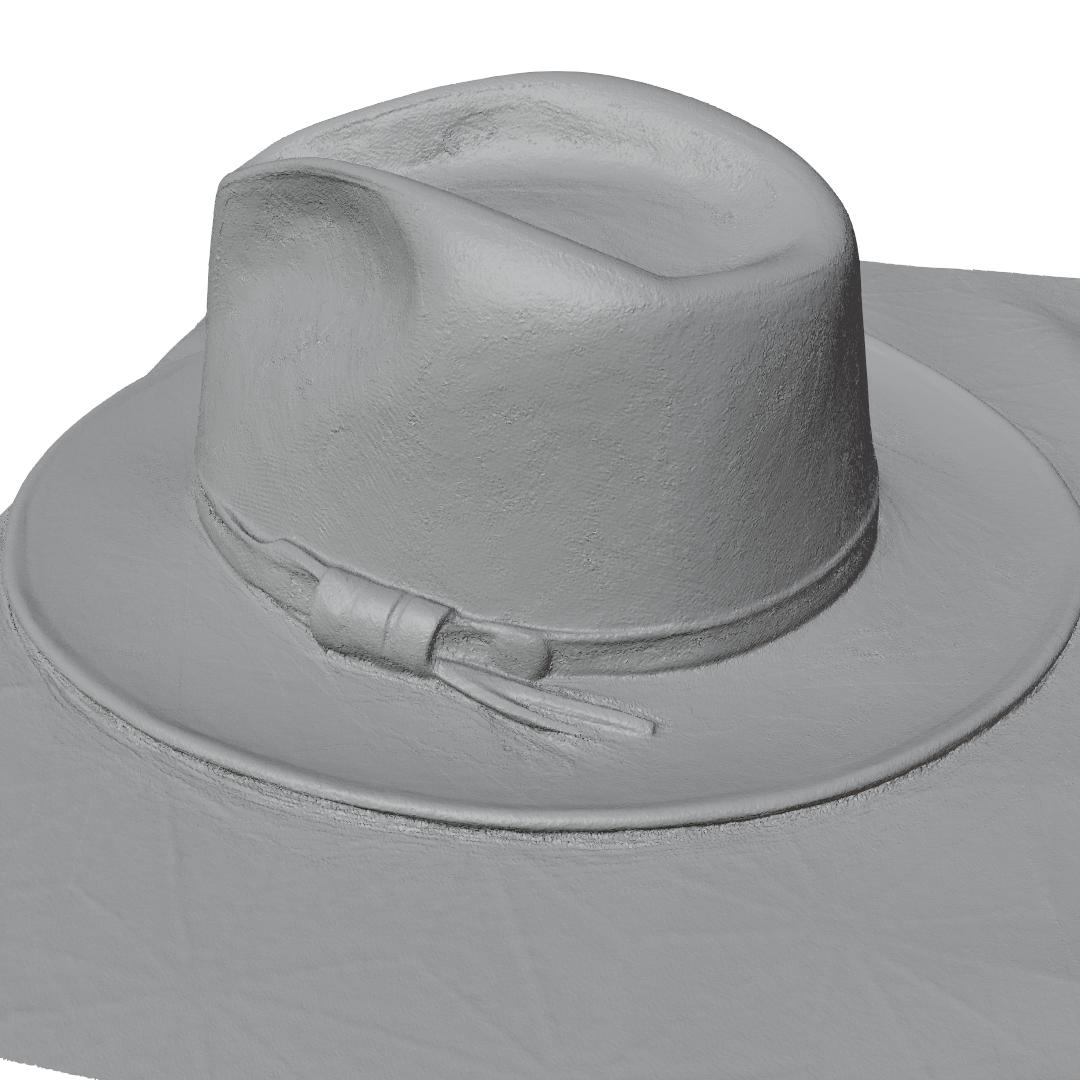} &
    \includegraphics[width=0.22\textwidth]{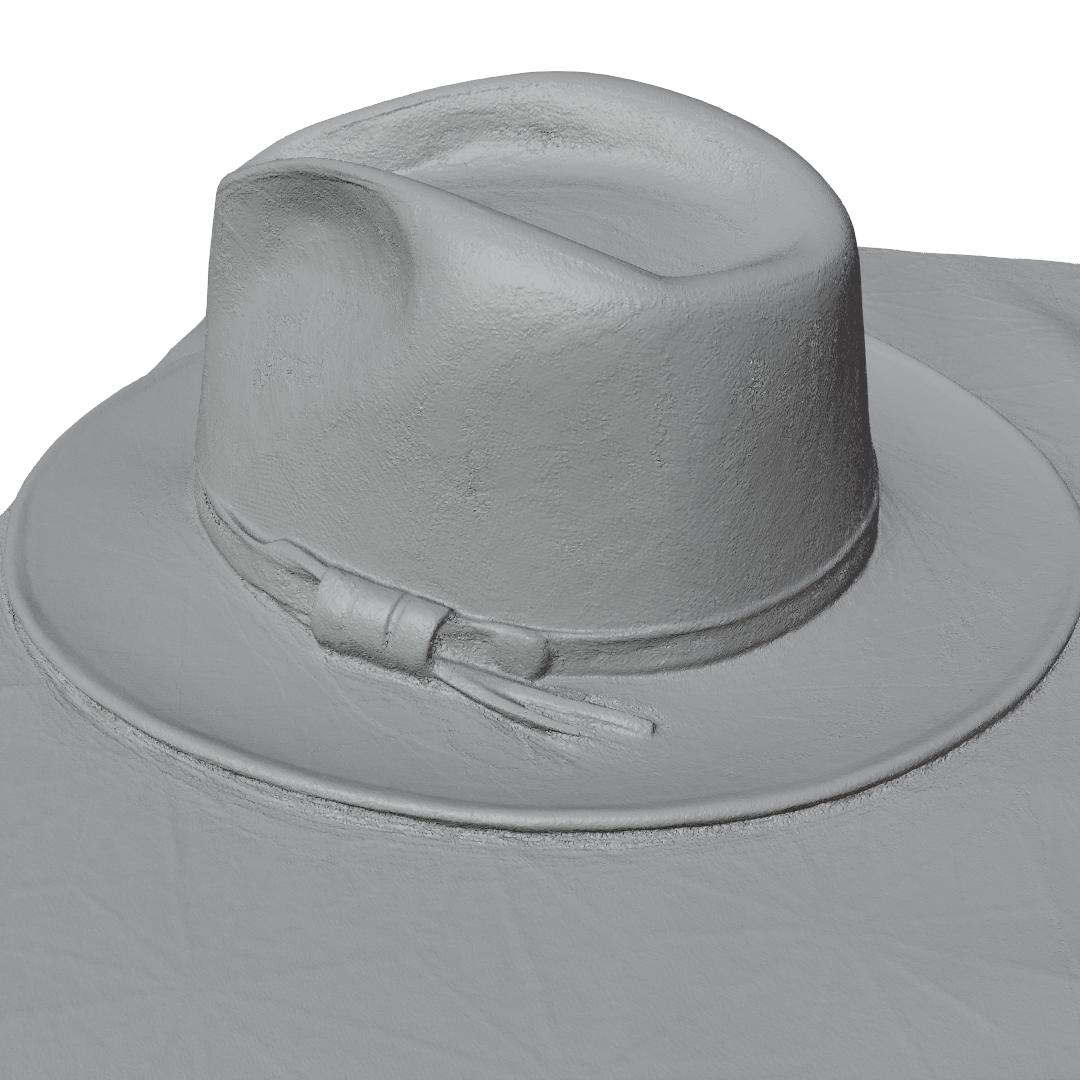} \\
    \includegraphics[width=0.22\textwidth]{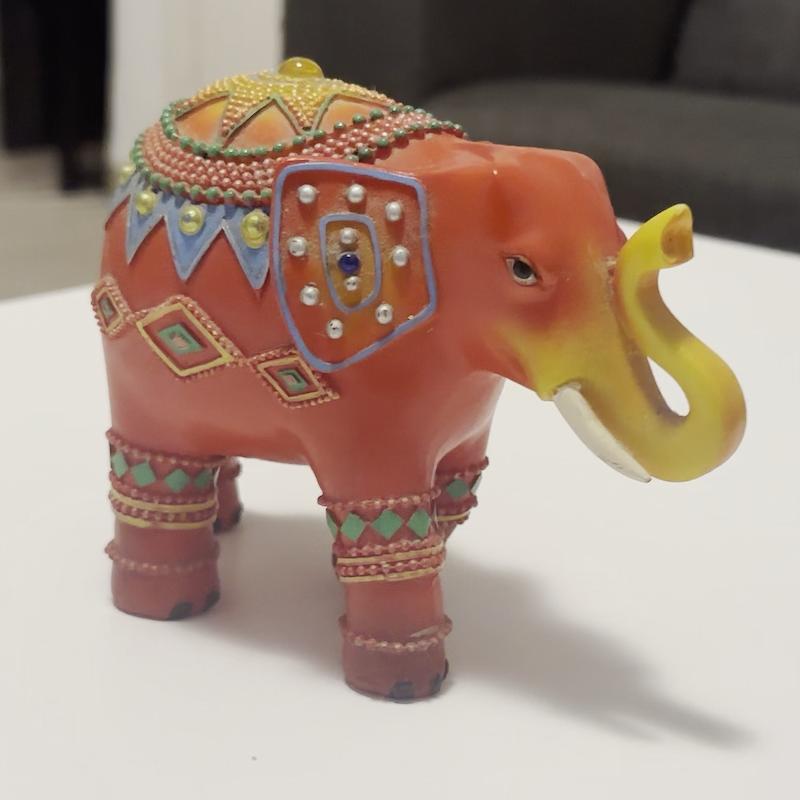} & 
    \includegraphics[width=0.22\textwidth]{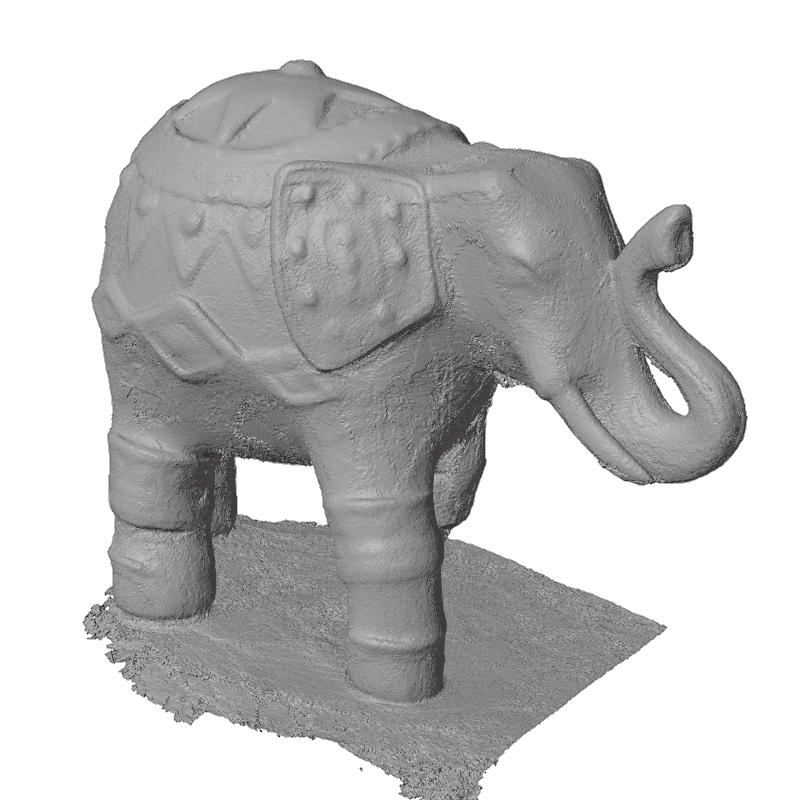} &
    \includegraphics[width=0.22\textwidth]{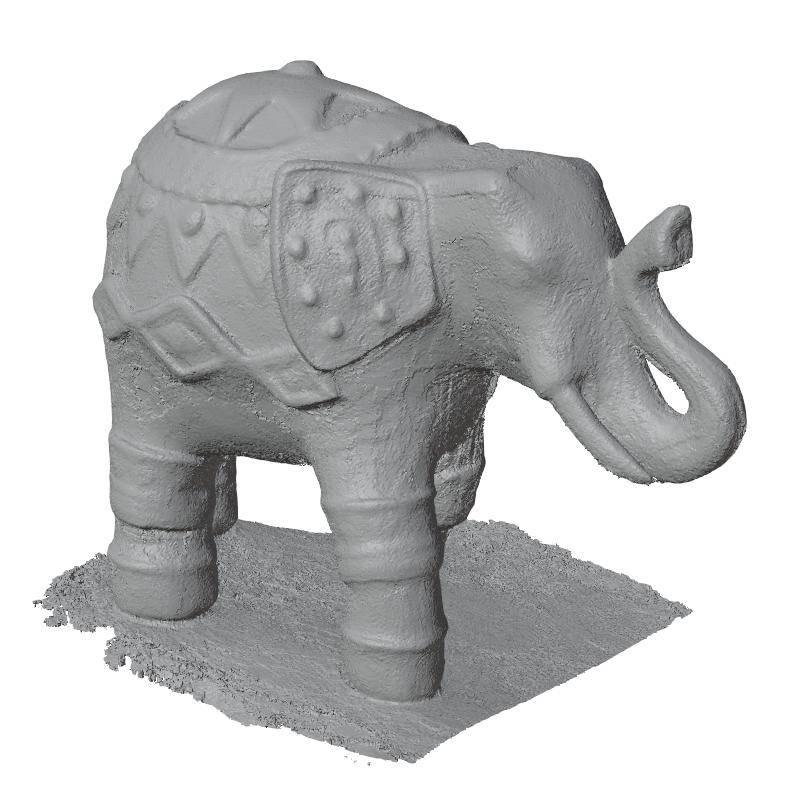} &
    \includegraphics[width=0.22\textwidth]{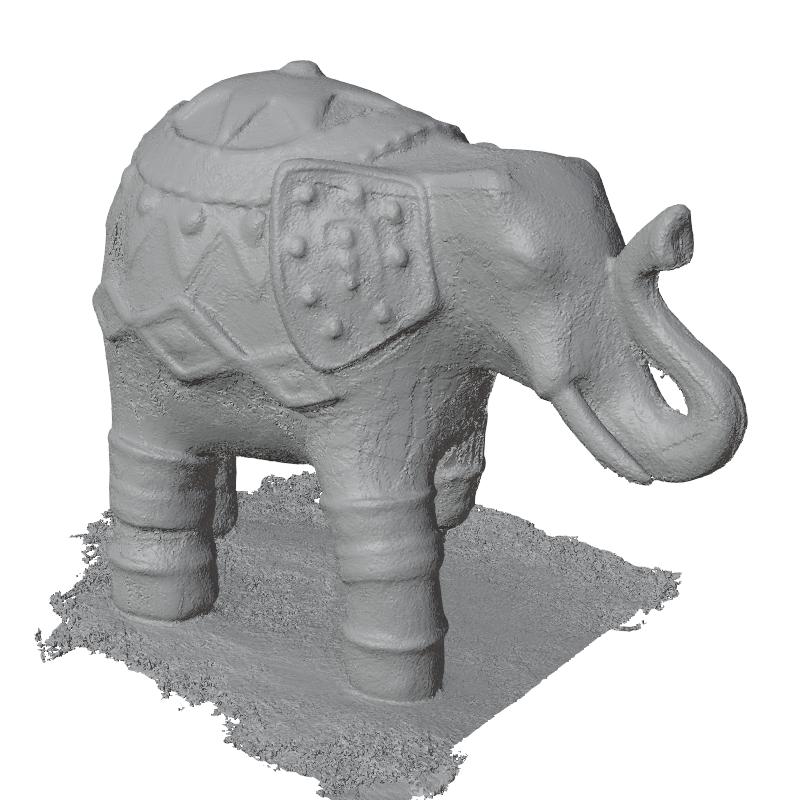} \\
    \end{tabular}
    \caption{From left to right: One of the input images, and the reconstructed mesh using our method, after 7000, 30000 and 60000 3DGS iterations.}
    \label{fig:3DGS_iterations}
\end{figure}

\subsection{Number of GS Iterations}
We claim that our method runs significantly faster compared to other reconstruction methods, and that the bottleneck of our runtime is the 3DGS optimization time. For a typical in-the-wild scene containing a central object, with roughly 80 images, 3DGS optimization takes 5-30 minutes on an Nvidia L40 GPU, depending on the number of iterations: 5-10 minutes for 7000 iterations, 10-20 minutes for 30000 iterations, and 20-30 minutes for 60000 iterations. In \cref{fig:3DGS_iterations}, we show that there are no noticeable differences in reconstructions of in-the-wild objects after 7000, 30000 and 60000 iterations. Additionally, in \cref{tab:DTU_ablations}, we evaluate our method on the DTU \cite{DTU} dataset after 7000, 30000 and 60000 iterations, showing that even with only 7000 iterations we achieve nearly identical results, demonstrating the power of our real-world regularization using a pre-trained stereo model.

\subsection{Necessity of Occlusion Mask}
\cref{tab:DTU_ablations} additionally validates the contribution of the occlusion mask that avoids using the depth information in areas which weren't visible in both the left and right views, showing an improvement in metrics regardless of the number of 3DGS iterations.

\section{Object Segmentation using Depth and SAM}
\begin{figure}[]
    \centering
    \begin{tabular}{cccc}
        \textbf{Image} & \textbf{Ours} & \textbf{Ours Masked} & \\  % Row headers
        \includegraphics[width=0.22\textwidth]{figures/itw_us_vs_sugar/cropped_apples_image.jpeg} & 
        \includegraphics[width=0.22\textwidth]{figures/itw_us_vs_sugar/cropped_apples_ours.jpeg} & 
        \includegraphics[width=0.22\textwidth]{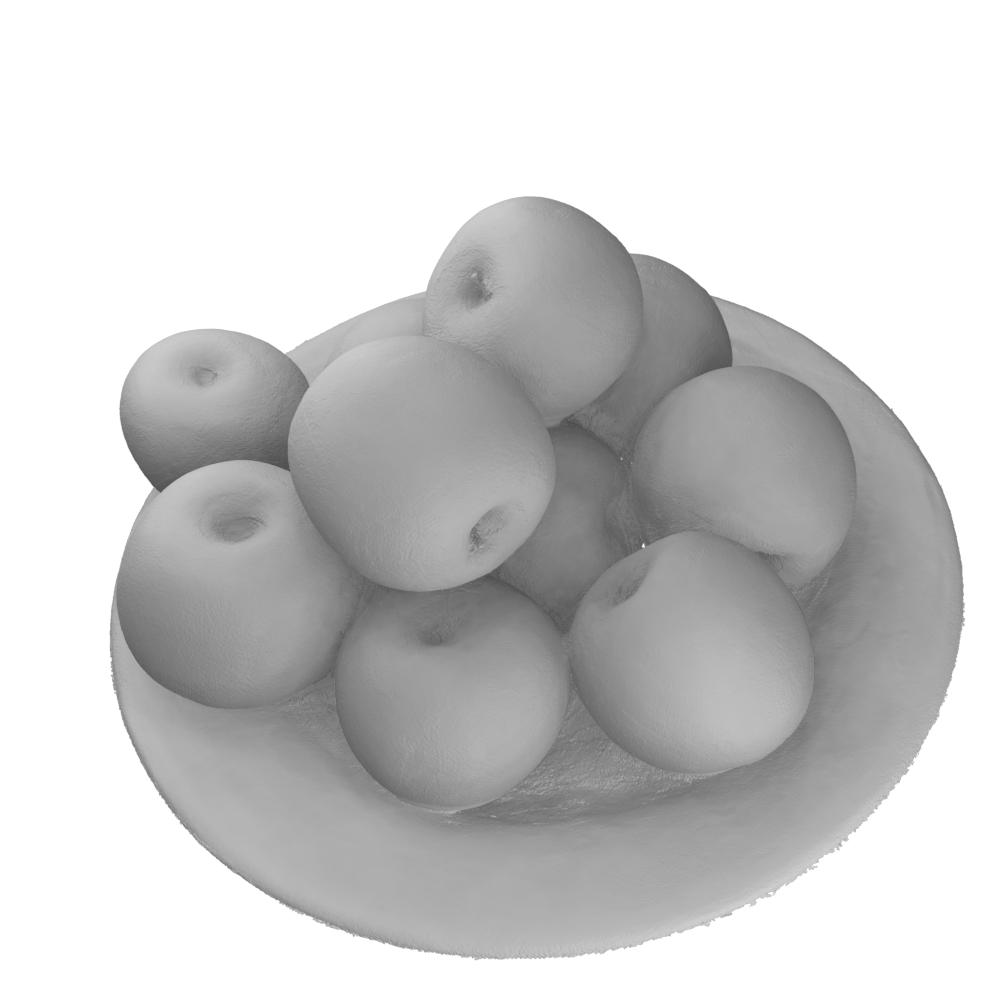} & \\

        \includegraphics[width=0.22\textwidth]{figures/itw_us_vs_sugar/cropped_coffee_image.jpeg} & 
        \includegraphics[width=0.22\textwidth]{figures/itw_us_vs_sugar/cropped_coffee_ours.jpeg} & 
        \includegraphics[width=0.22\textwidth]{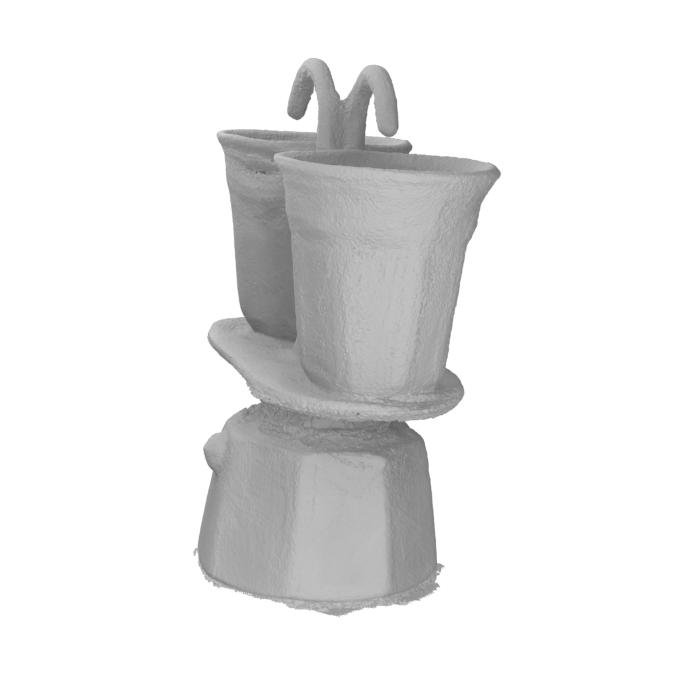} & \\

        \includegraphics[width=0.22\textwidth]{figures/itw_us_vs_sugar/cropped_dancer_image.jpeg} & 
        \includegraphics[width=0.22\textwidth]{figures/itw_us_vs_sugar/cropped_dancer_ours.jpeg} & 
        \includegraphics[width=0.22\textwidth]{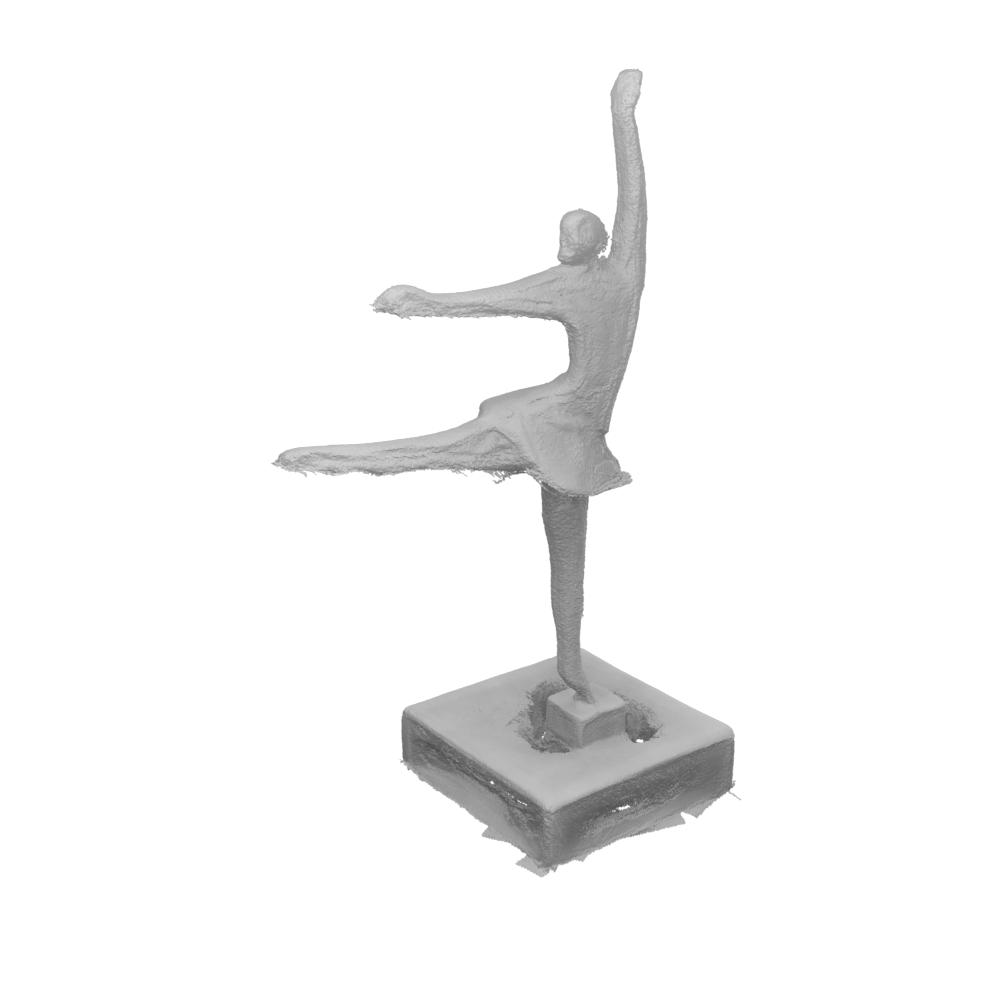} & \\

        \includegraphics[width=0.22\textwidth]{figures/itw_us_vs_sugar/cropped_sculpture_image.jpeg} & 
        \includegraphics[width=0.22\textwidth]{figures/itw_us_vs_sugar/cropped_sculpture_ours.jpeg} & 
        \includegraphics[width=0.22\textwidth]{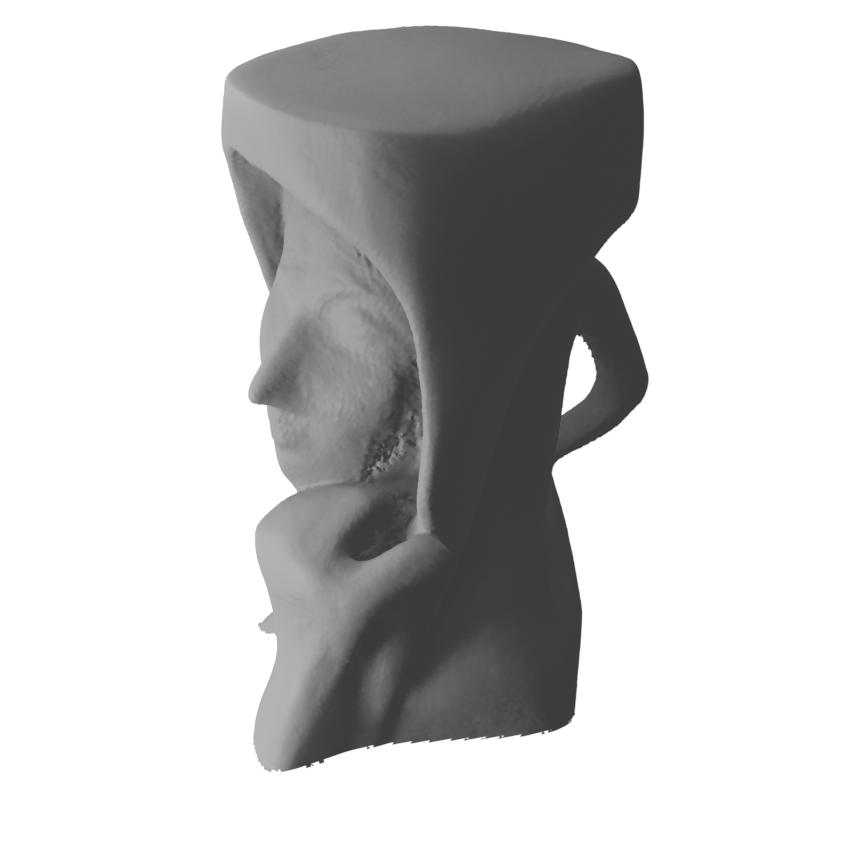} & \\
    \end{tabular}
    \caption{Examples of our method's ability to segment objects using SAM \cite{SAM} with depth projections, on in-the-wild videos taken by a smartphone.}
    \label{fig:SAM}
\end{figure}

In some cases we want to extract only the surface of a given object in the scene.
Other methods such as Segment-Any-Gaussians require additional training after the 3DGS, and are more suited for scenes with multiple objects rather than 360 object-centric scenes \cite{cen2023saga}. Thus, we choose to segment each image.
The naive approach to object segmentation in a scene would involve independently segmenting every image, a method that can be labor-intensive. 
Instead, we employ a technique that leverages the power of Segment Anything Model (SAM) \cite{SAM} in conjunction with depth information and geometric transformations. 
Initially, we annotate the first image of the scene using SAM to obtain a precise object mask. 
This initial segmentation acts as a foundation for tracking and segmenting the object across subsequent images. 
By performing this process after obtaining the depth of the identified object, we can project its mask onto the next image in the sequence using the extrinsic camera parameters.
To accommodate for potential errors in SAM, we dilate the projected mask in the new image. 
Then, utilizing farthest point sampling, we select points that represent the extremities of the object within this dilated mask. 
These points are used as the seed for a new SAM annotation for the next image. 
This process is applied iteratively to each subsequent image in the series, allowing for dynamic and precise object segmentation throughout the scene. \cref{fig:SAM} shows examples of our method's segmentation abilities on in-the-wild videos taken by a smartphone.

\newpage 
\section{Additional Examples}
\subsection{Additional In-the-Wild Examples}
\begin{figure}[]
    \centering
    \begin{tabular}{ccc}
    \textbf{Image} & \textbf{SuGaR} & \textbf{Ours}\\
    \includegraphics[width=0.22\textwidth]{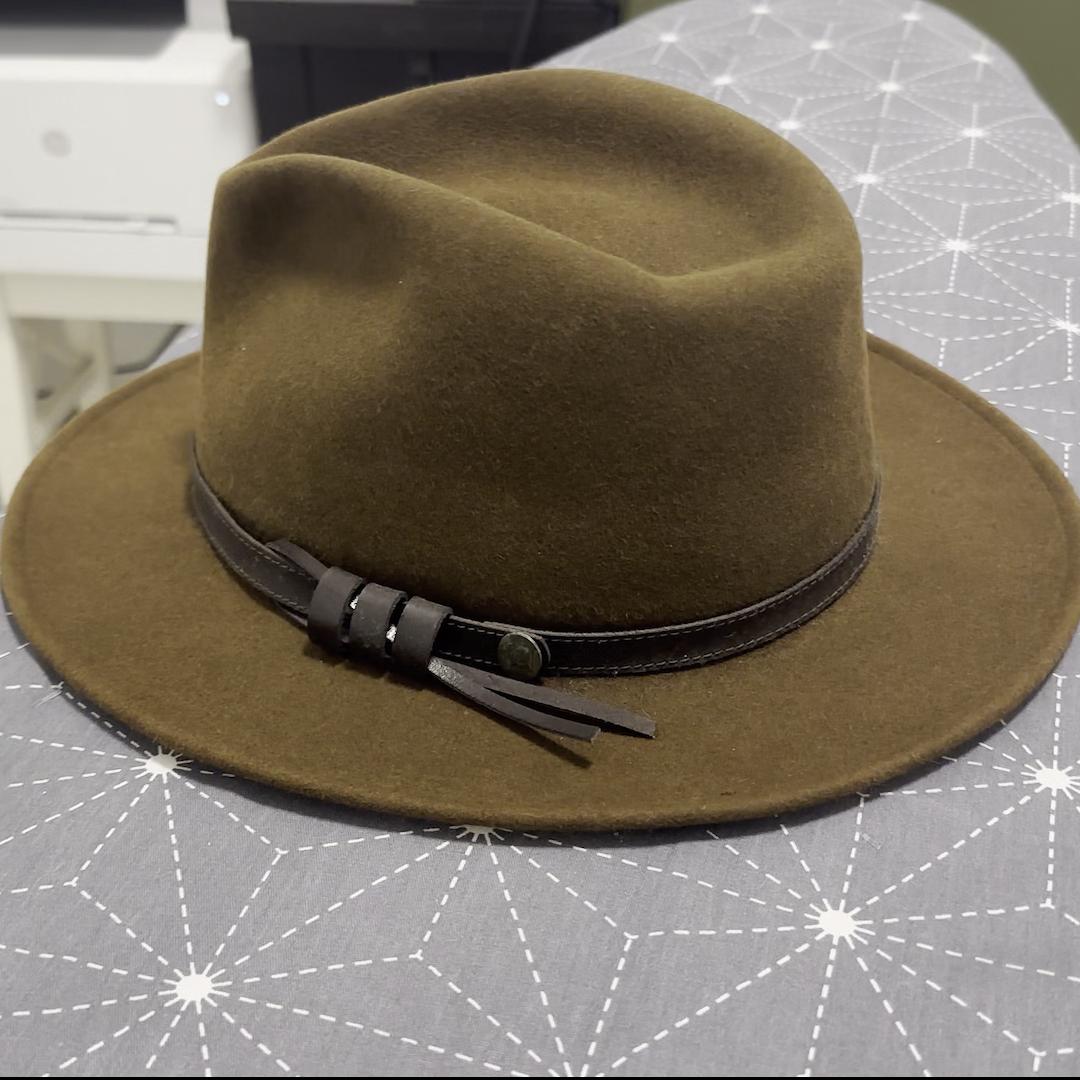} & 
    \includegraphics[width=0.22\textwidth]{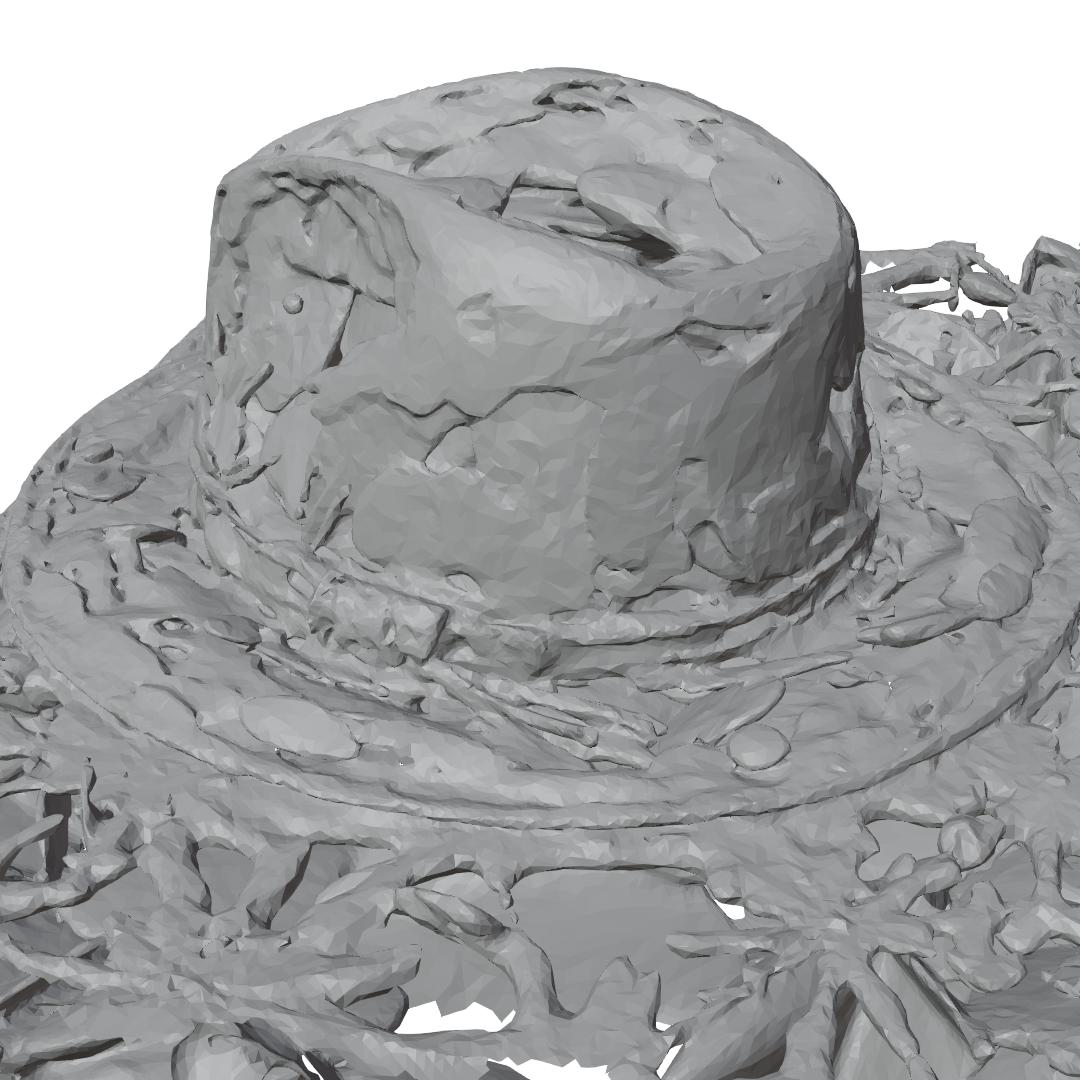} &
    \includegraphics[width=0.22\textwidth]{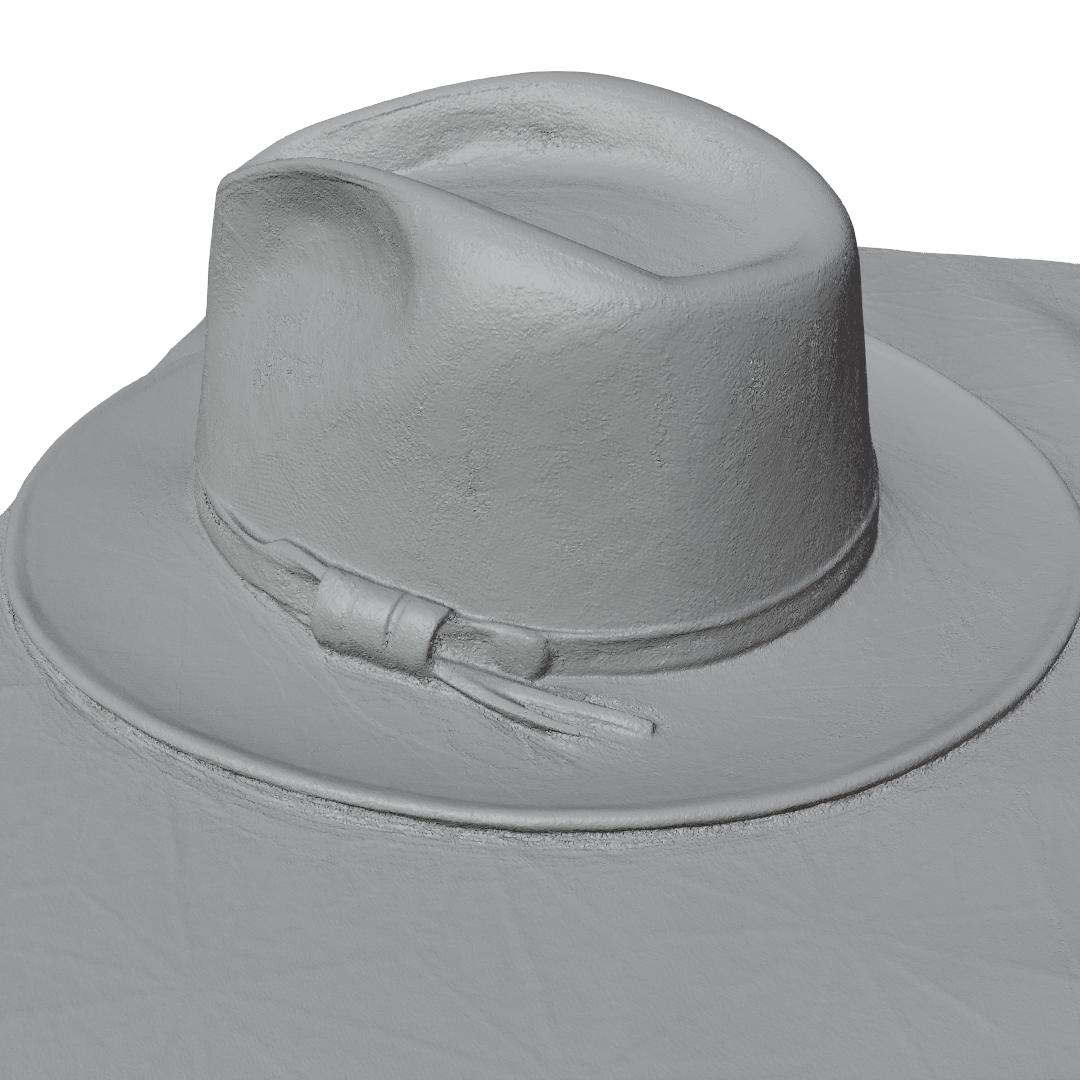} \\
    \includegraphics[width=0.22\textwidth]{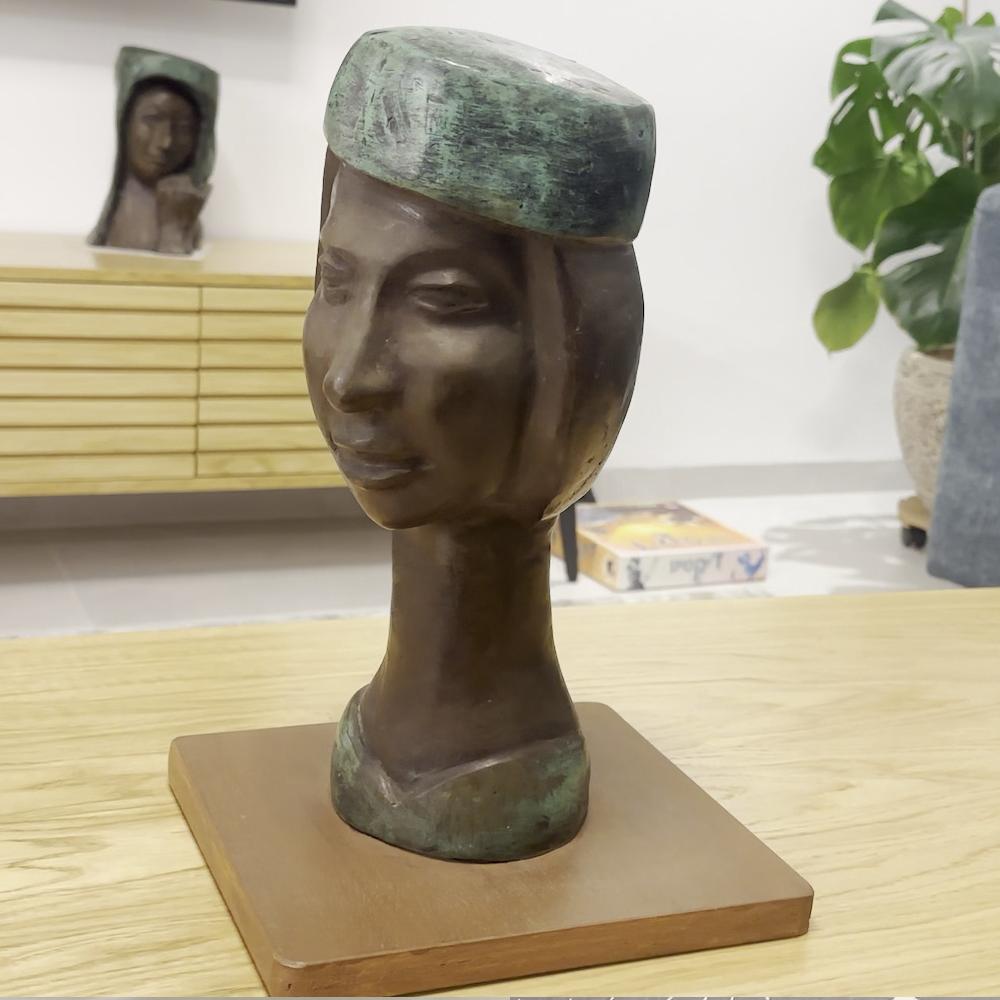} & 
    \includegraphics[width=0.22\textwidth]{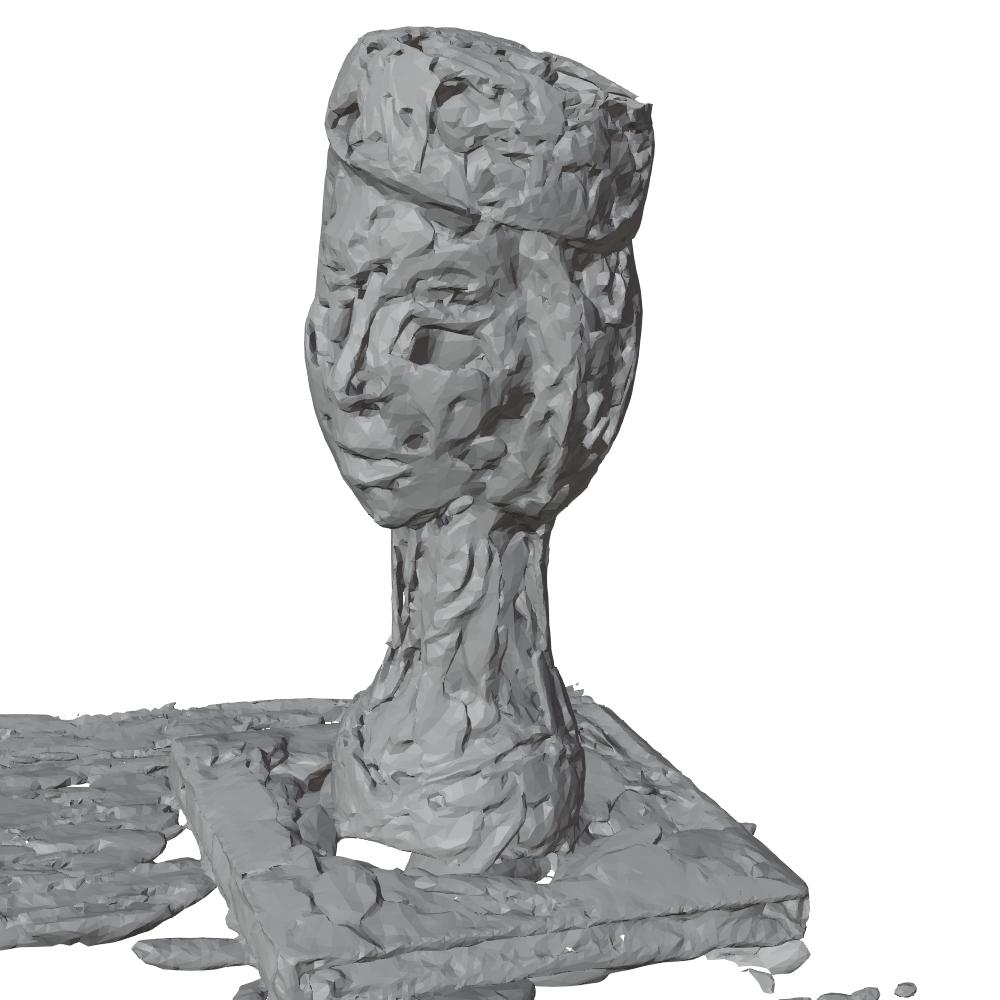} &
    \includegraphics[width=0.22\textwidth]{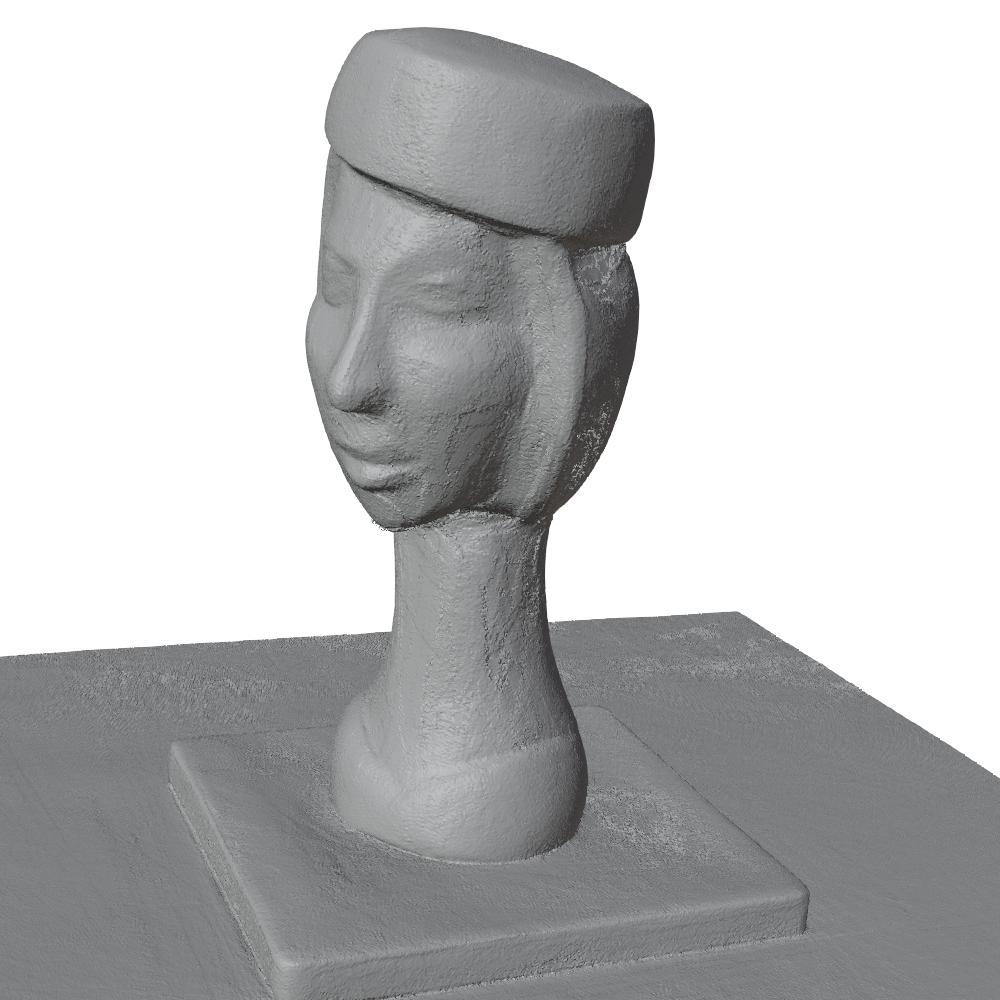} \\
    \includegraphics[width=0.22\textwidth]{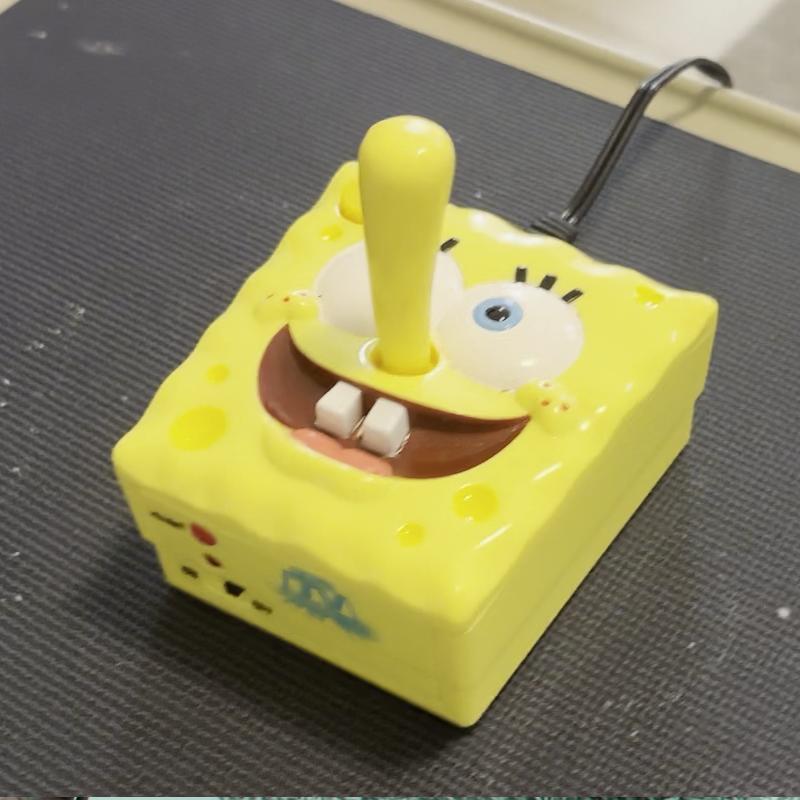} & 
    \includegraphics[width=0.22\textwidth]{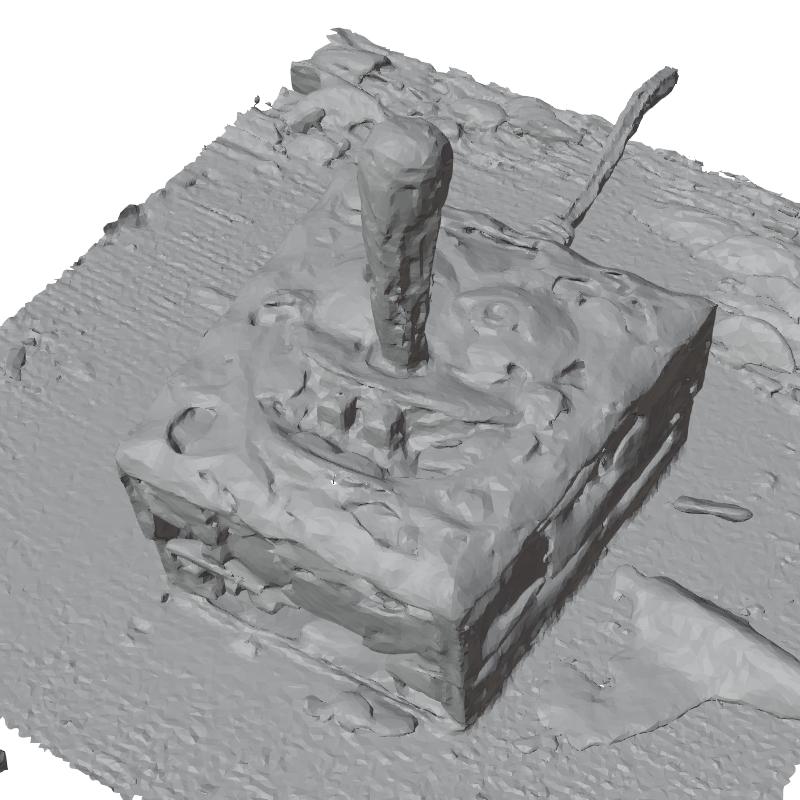} &
    \includegraphics[width=0.22\textwidth]{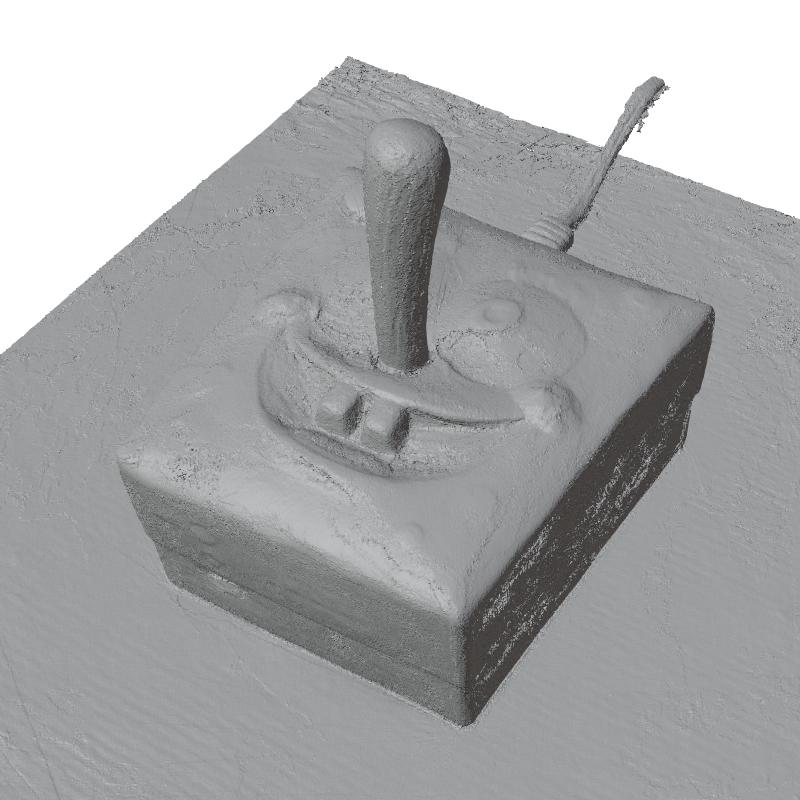} \\
    \includegraphics[width=0.22\textwidth]{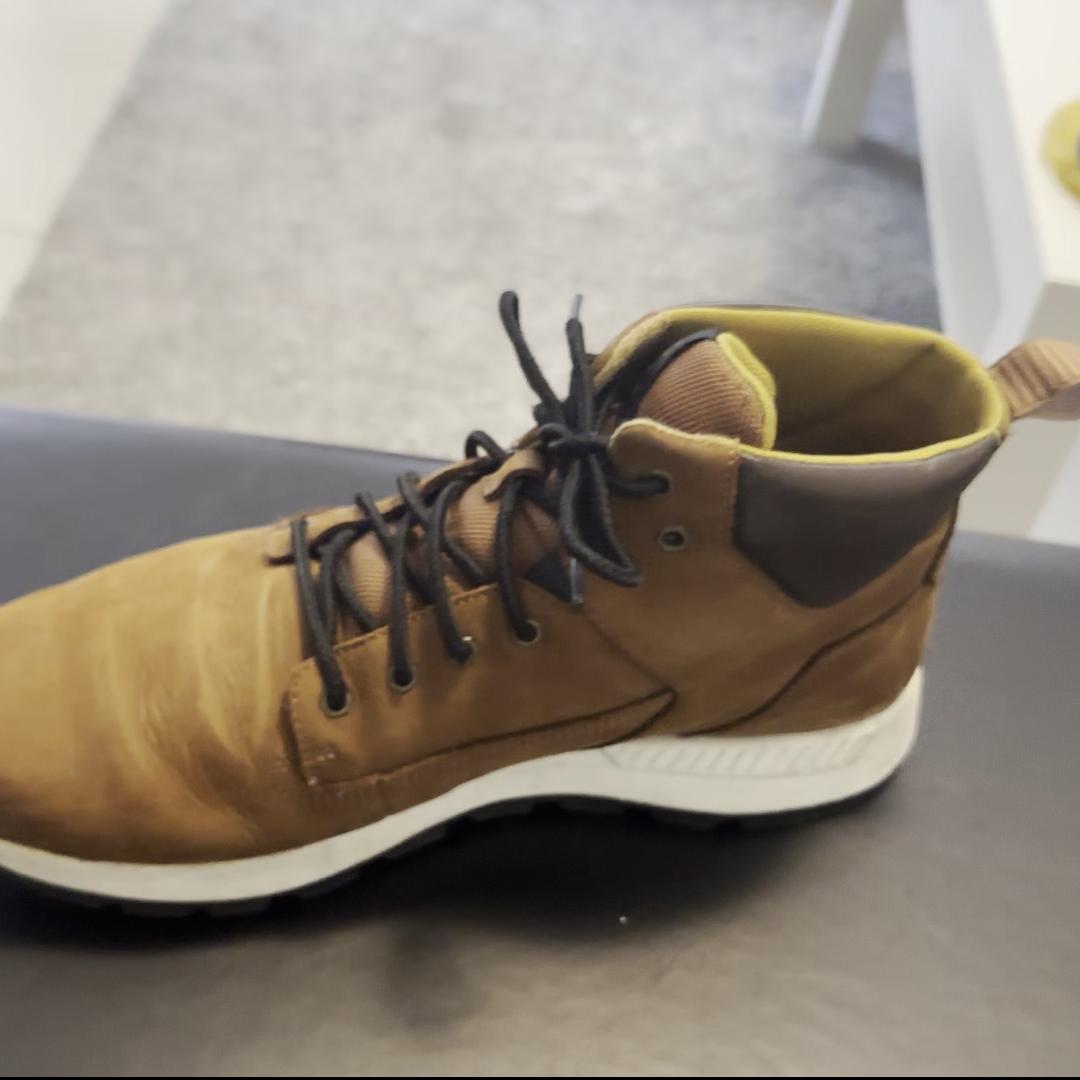} & 
    \includegraphics[width=0.22\textwidth]{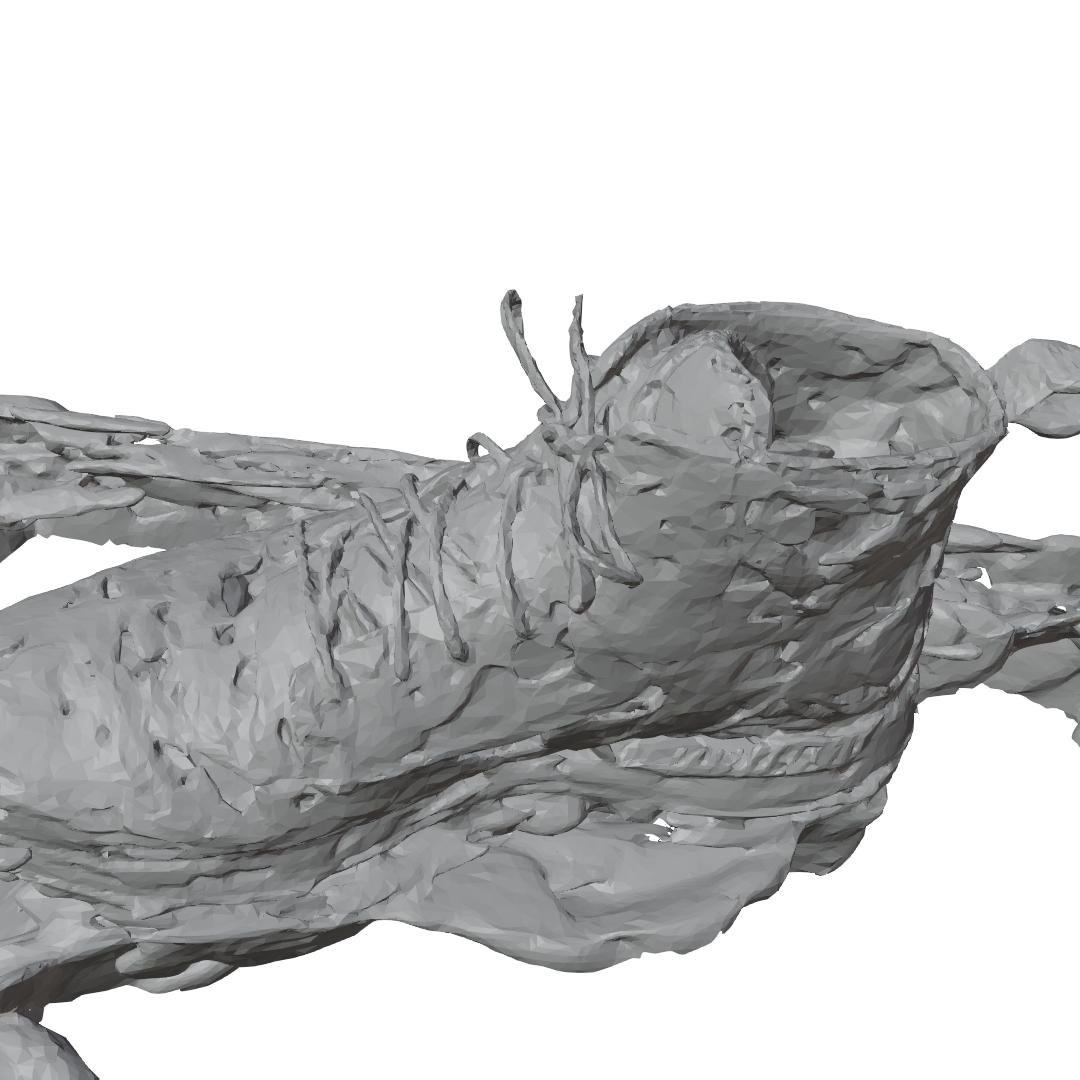} &
    \includegraphics[width=0.22\textwidth]{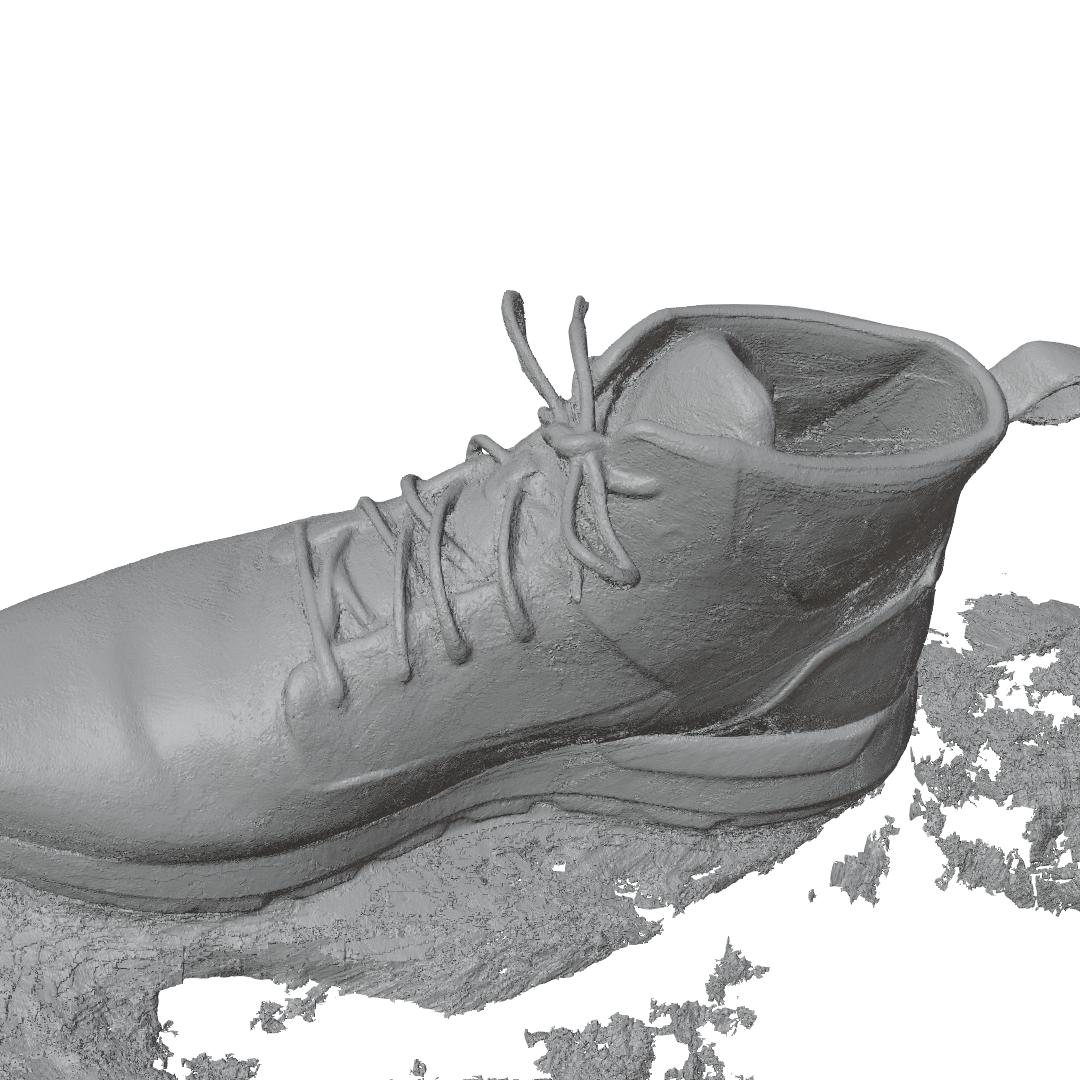} \\
    \includegraphics[width=0.22\textwidth]{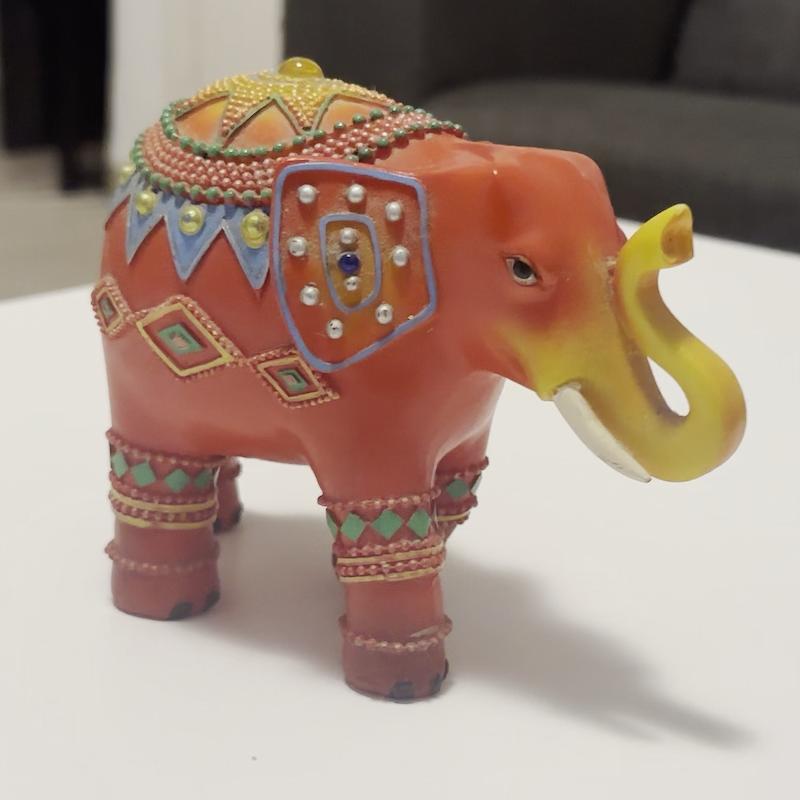} & 
    \includegraphics[width=0.22\textwidth]{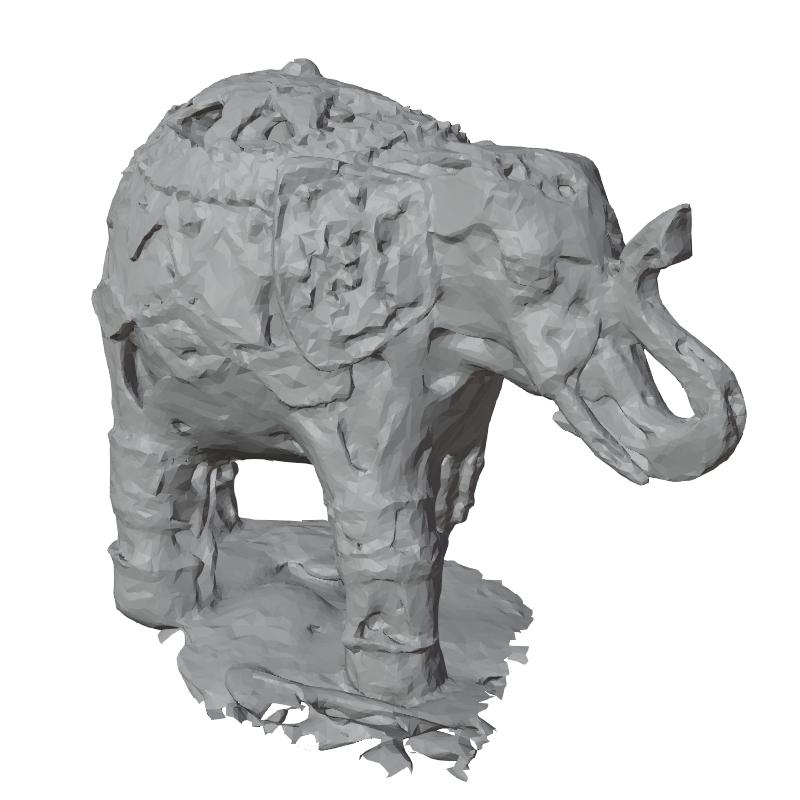} &
    \includegraphics[width=0.22\textwidth]{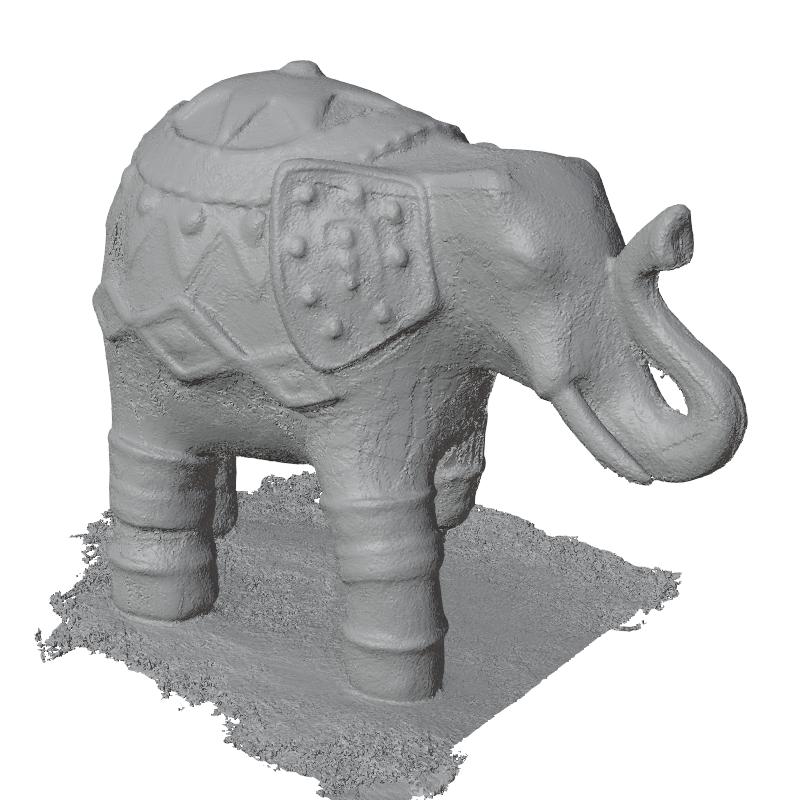} \\
    \includegraphics[width=0.22\textwidth]{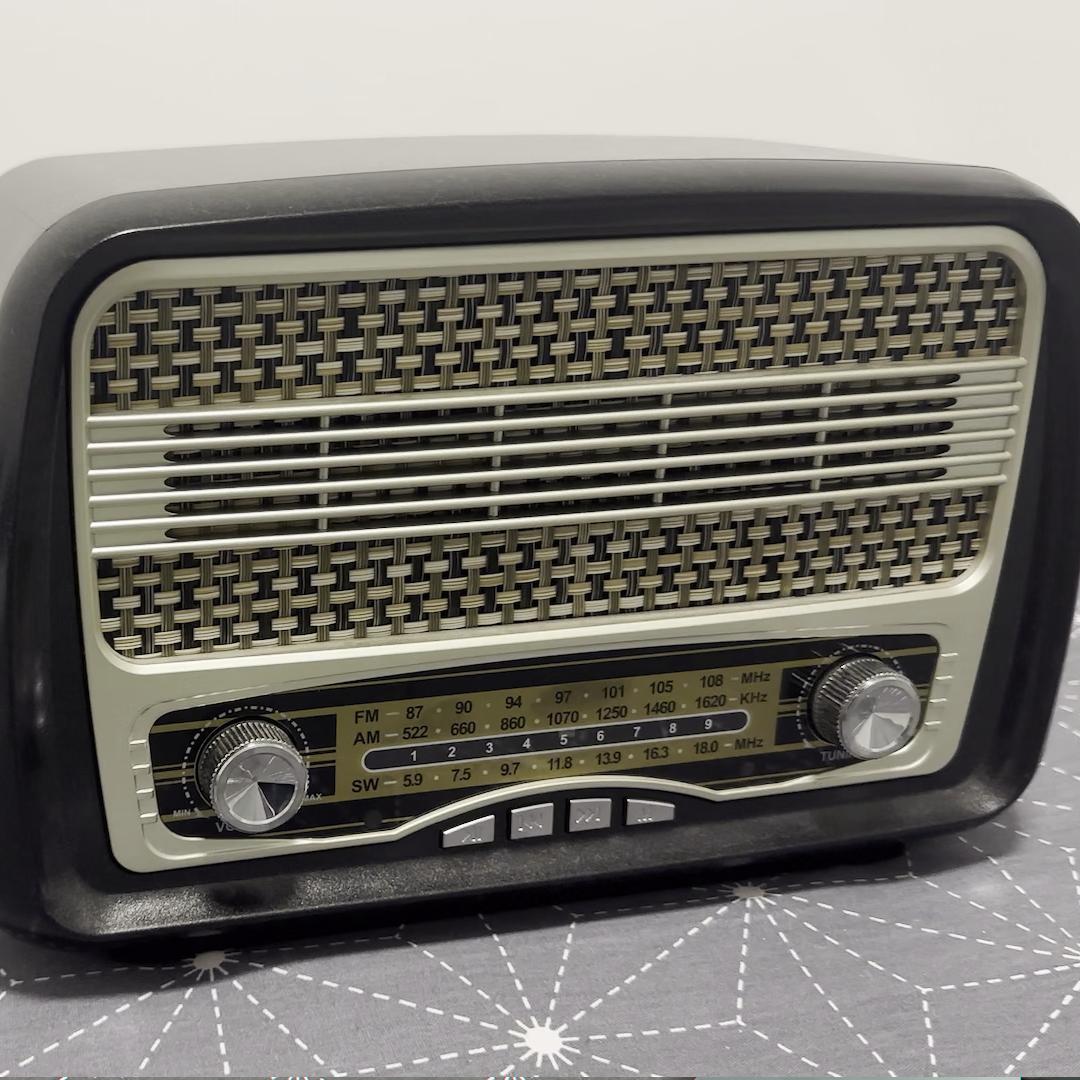} & 
    \includegraphics[width=0.22\textwidth]{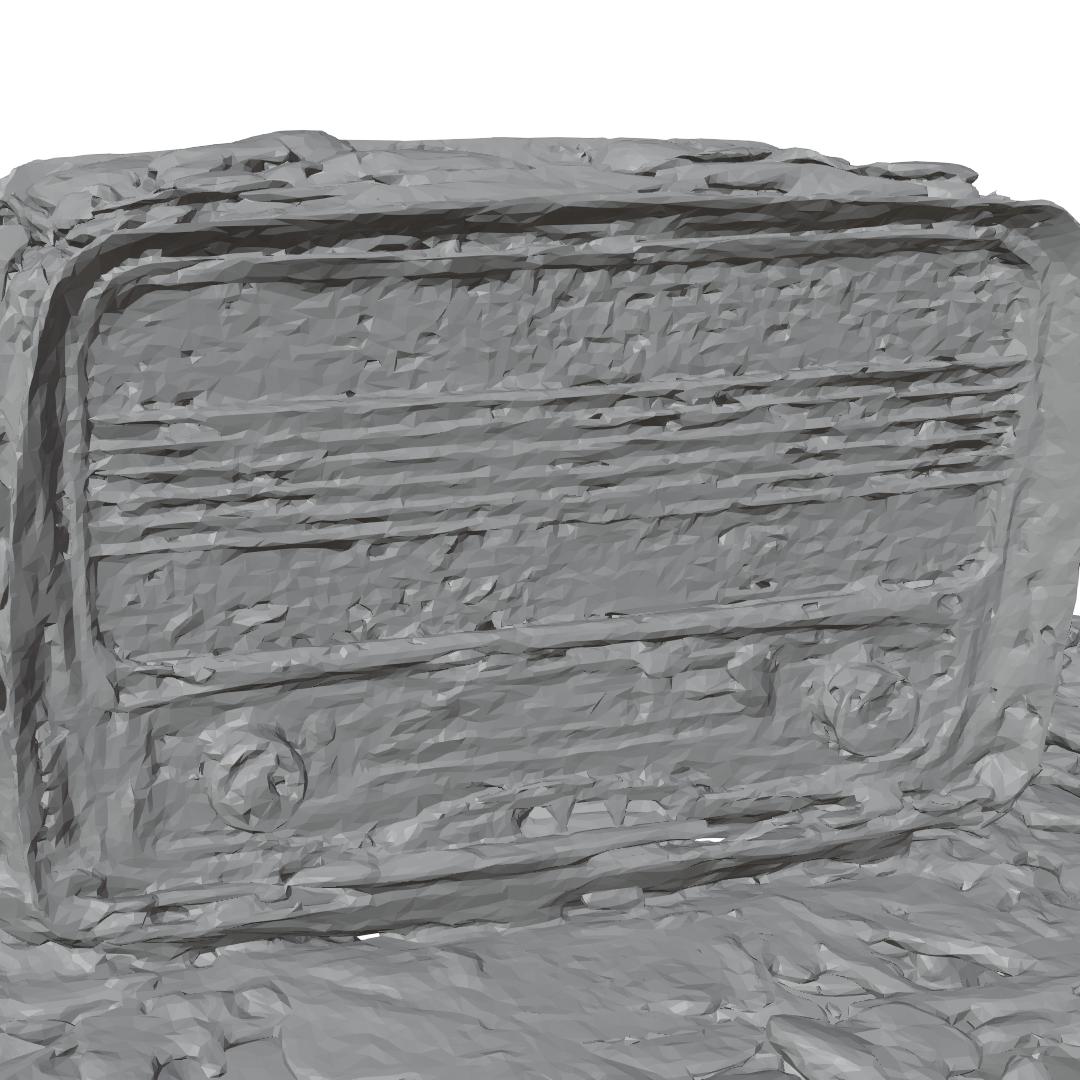} &
    \includegraphics[width=0.22\textwidth]{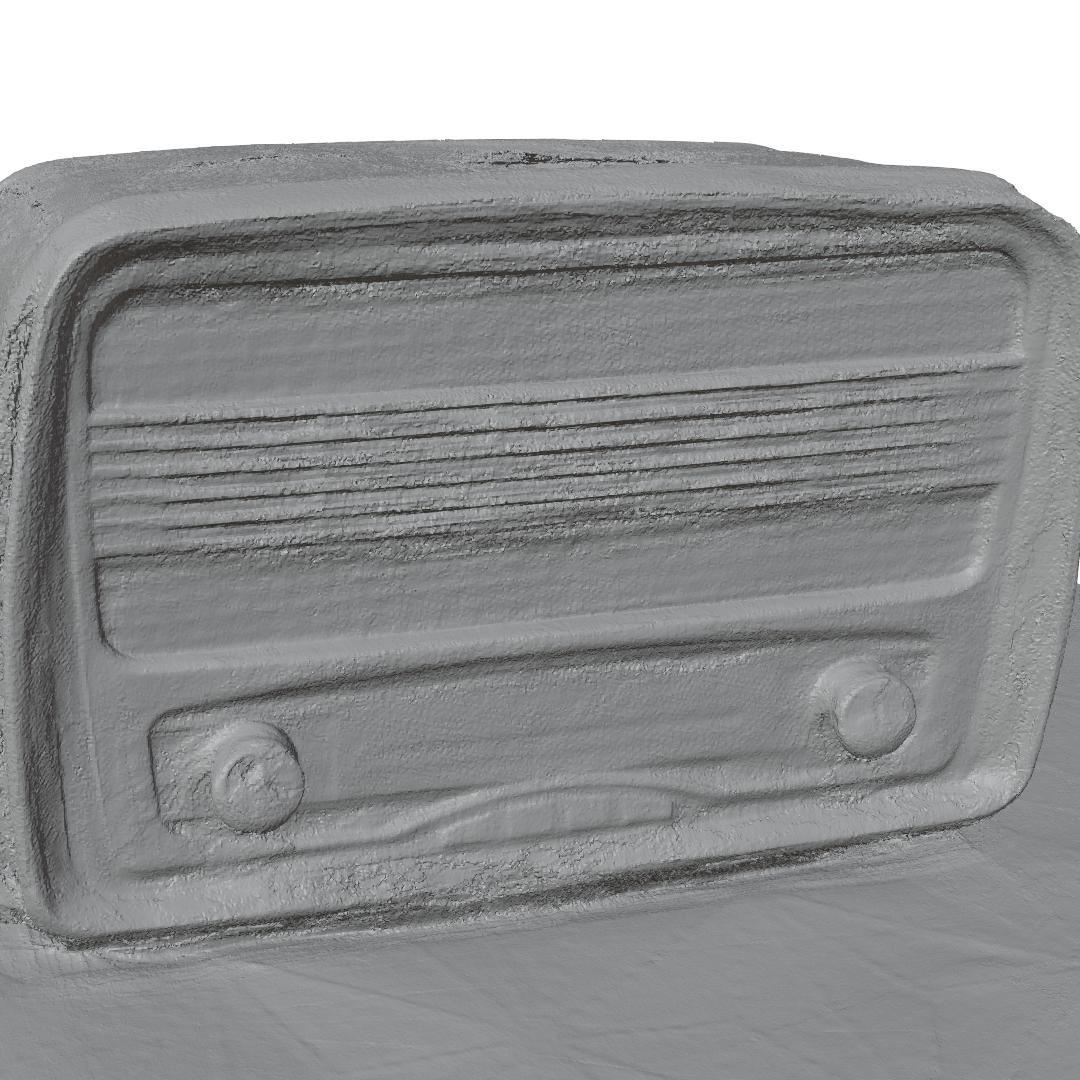} \\
    \includegraphics[width=0.22\textwidth]{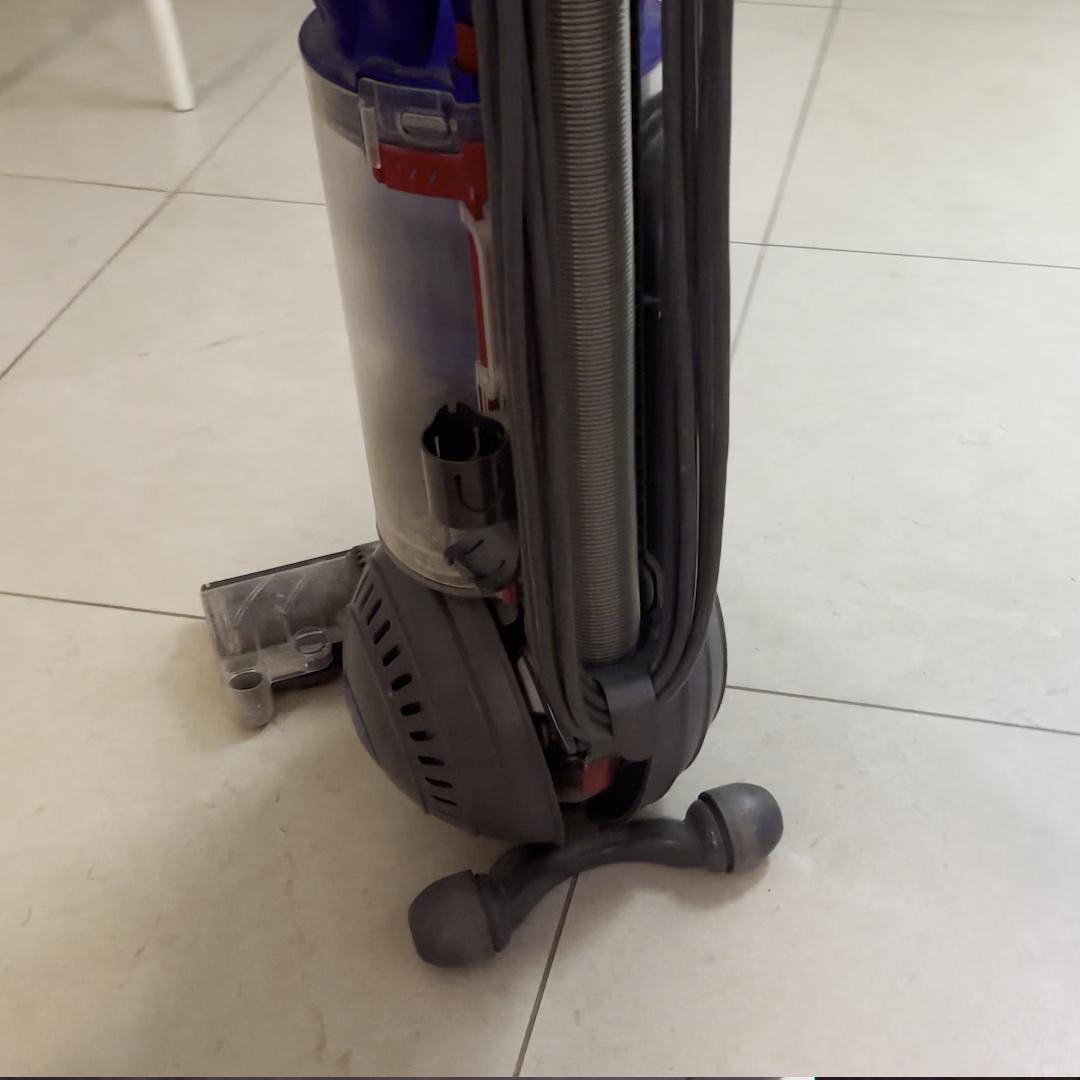} & 
    \includegraphics[width=0.22\textwidth]{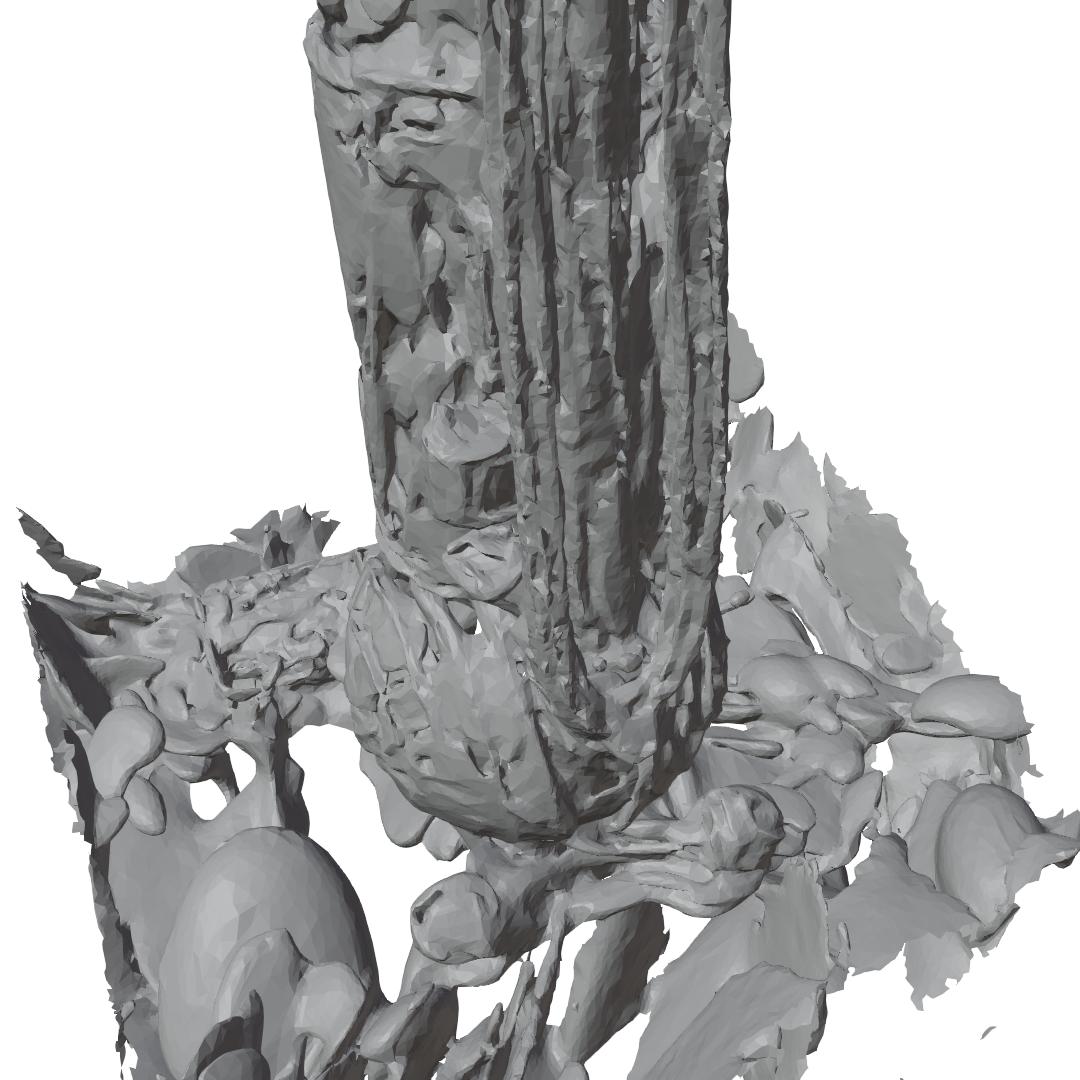} &
    \includegraphics[width=0.22\textwidth]{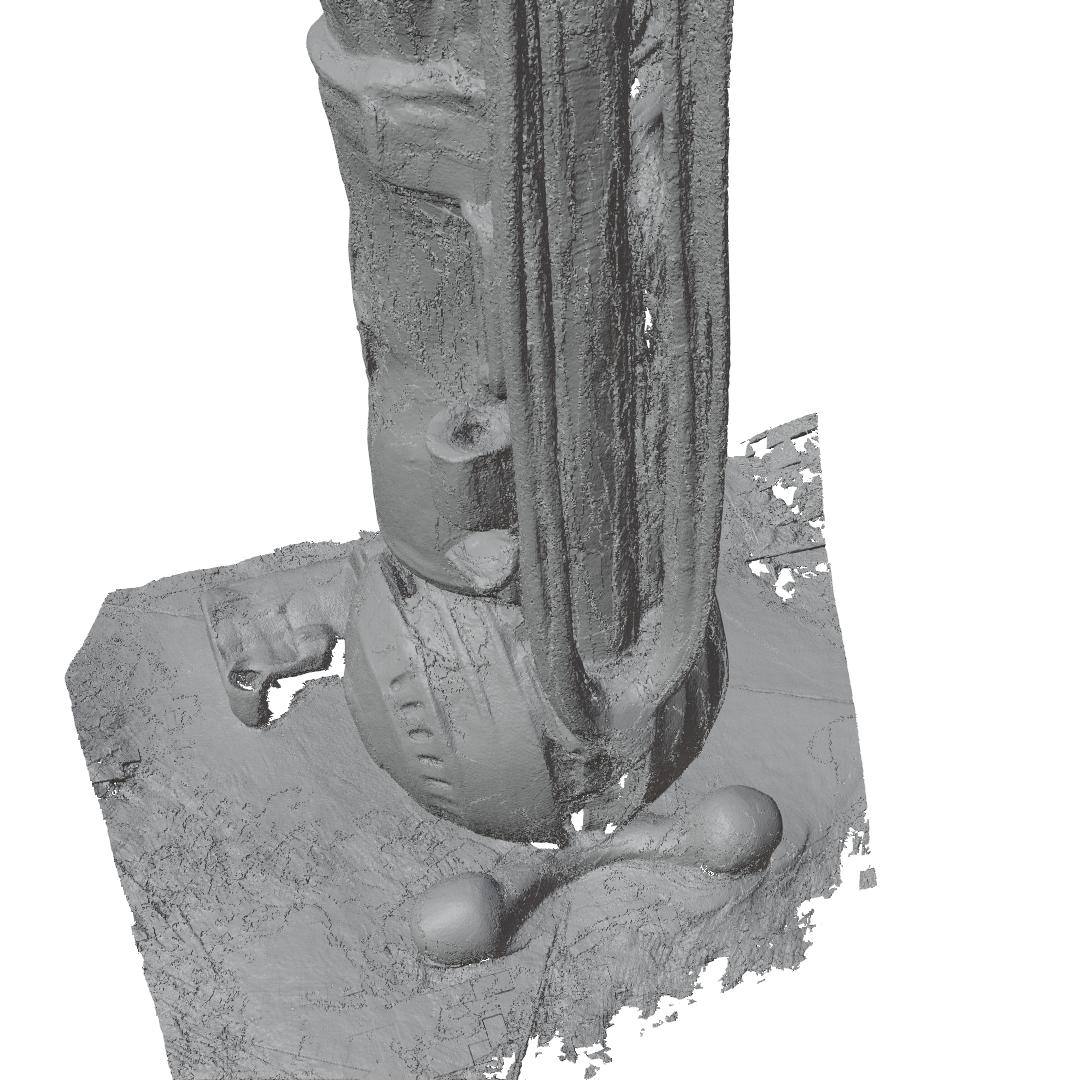} \\
    \end{tabular}
    \caption{Additional qualitative comparisons between our method and SuGaR \cite{SuGaR} on surface reconstruction from in-the-wild videos.}
    \label{fig:in-the-wild-comparison_sup}
\end{figure}

In \cref{fig:in-the-wild-comparison_sup} We present additional qualitative comparisons between reconstructions from our method and reconstructions from SuGaR \cite{SuGaR}, on in-the-wild videos taken by a smartphone in an uncontrolled environment. We run SuGaR according to the instructions from their official repository, with density regularization, no train/eval split, and 15000 refinement iterations.

\subsection{Additional Examples from the DTU dataset}
\begin{figure}[]
    \centering
    \begin{tabular}{ccccc}
        \includegraphics[width=0.3\textwidth]{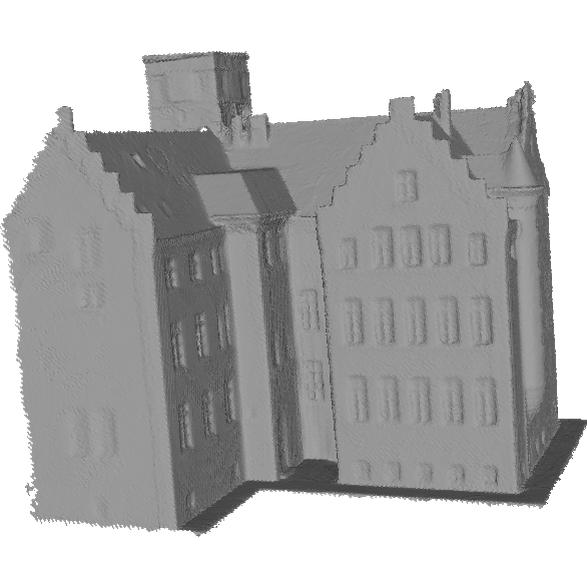} & 
        \includegraphics[width=0.3\textwidth]{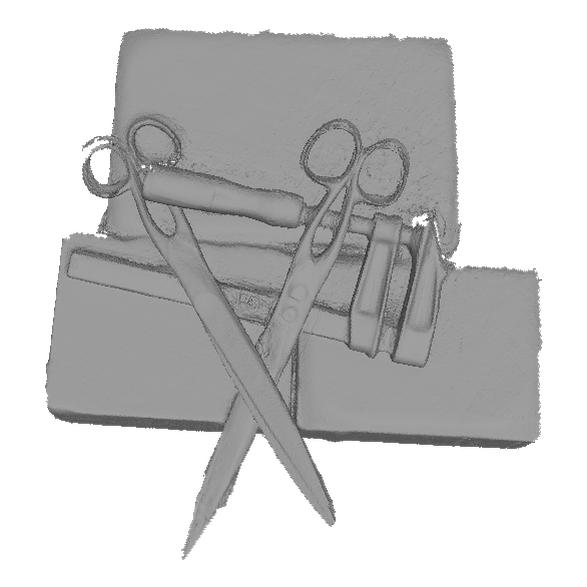} &
        \includegraphics[width=0.3\textwidth]{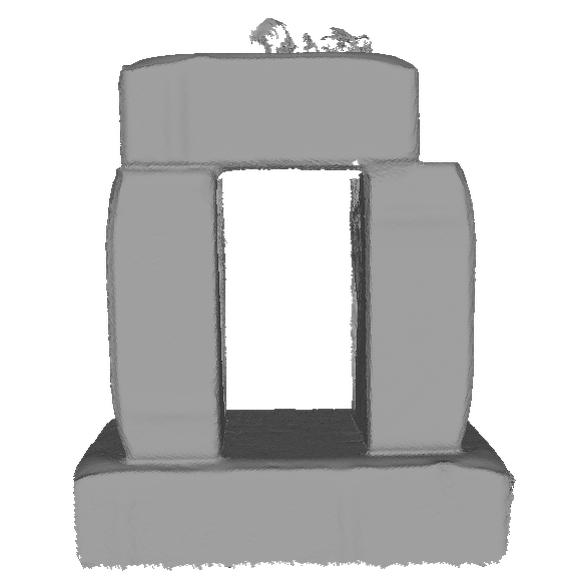} \\
        scan24 & scan37 & scan40 \\

        \includegraphics[width=0.3\textwidth]{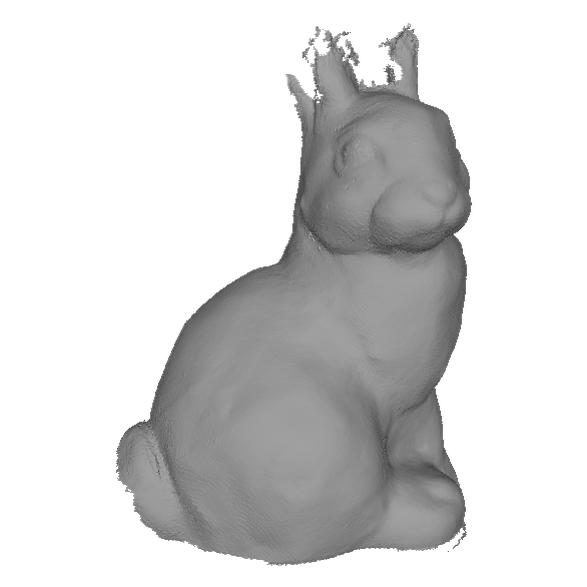} &
        \includegraphics[width=0.3\textwidth]{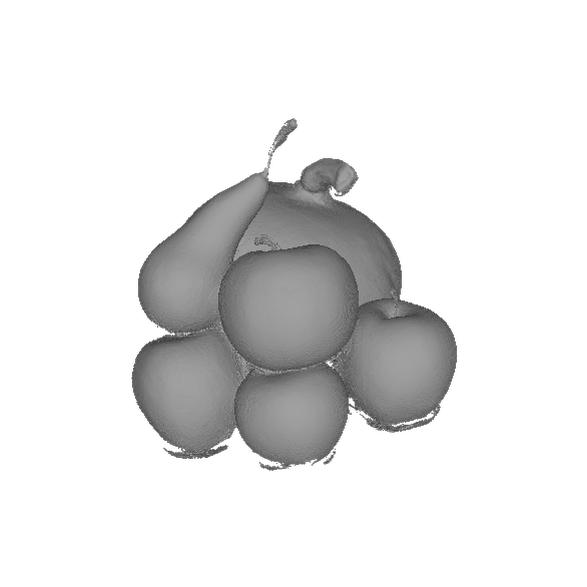} &
        \includegraphics[width=0.3\textwidth]{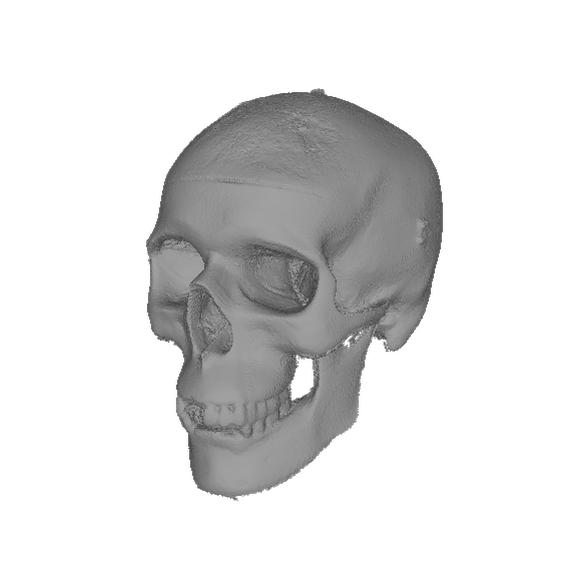} \\
        scan55 & scan63 & scan65 \\

        \includegraphics[width=0.3\textwidth]{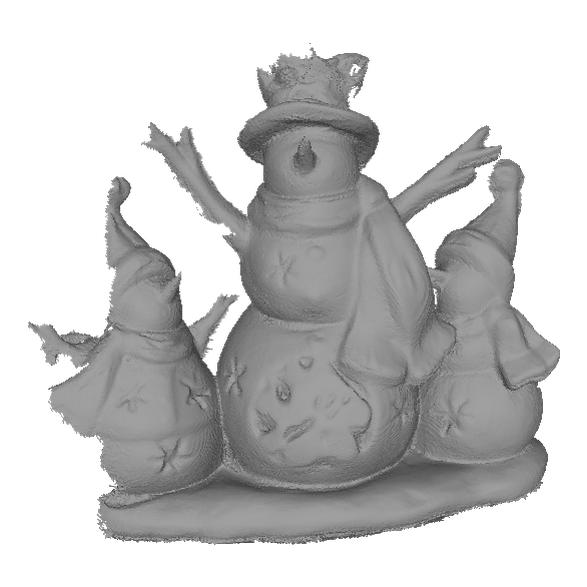} &
        \includegraphics[width=0.3\textwidth]{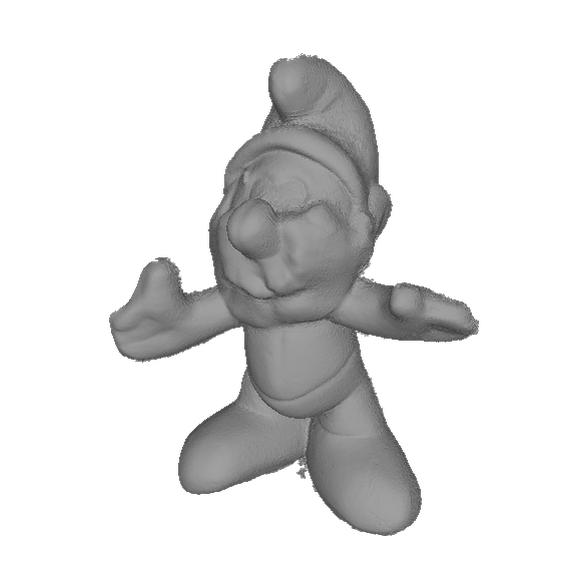} &
        \includegraphics[width=0.3\textwidth]{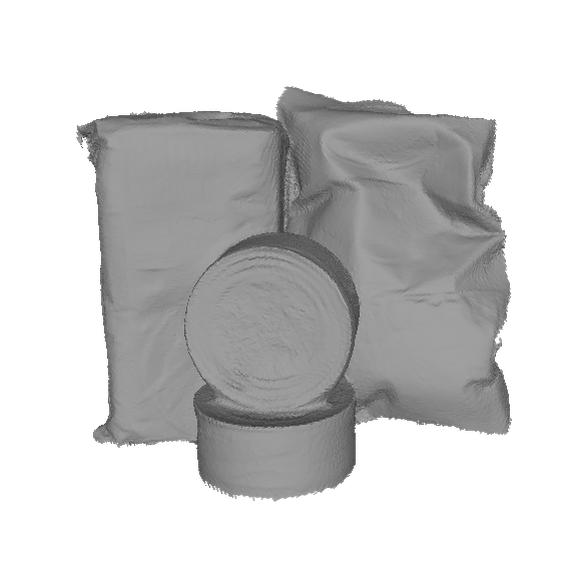} \\
        scan69 & scan83 & scan97 \\

        \includegraphics[width=0.3\textwidth]{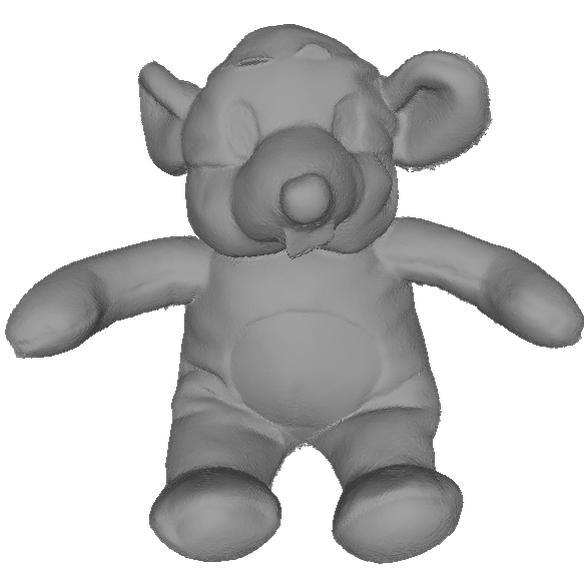} &
        \includegraphics[width=0.3\textwidth]{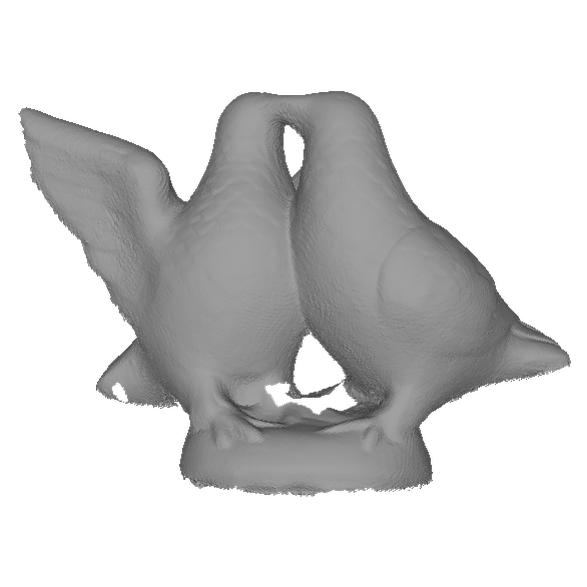} &
        \includegraphics[width=0.3\textwidth]{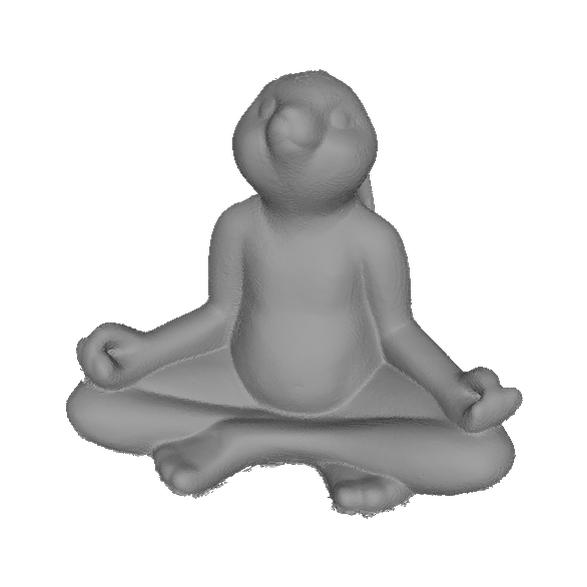} \\
        scan105 & scan106 & scan110 \\

        \includegraphics[width=0.3\textwidth]{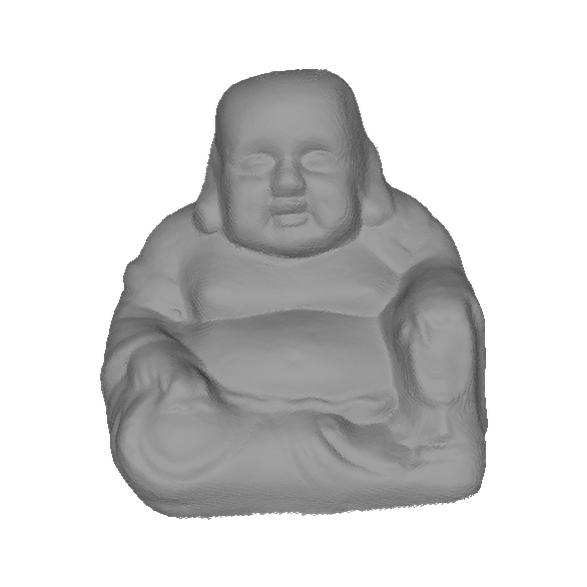} &
        \includegraphics[width=0.3\textwidth]{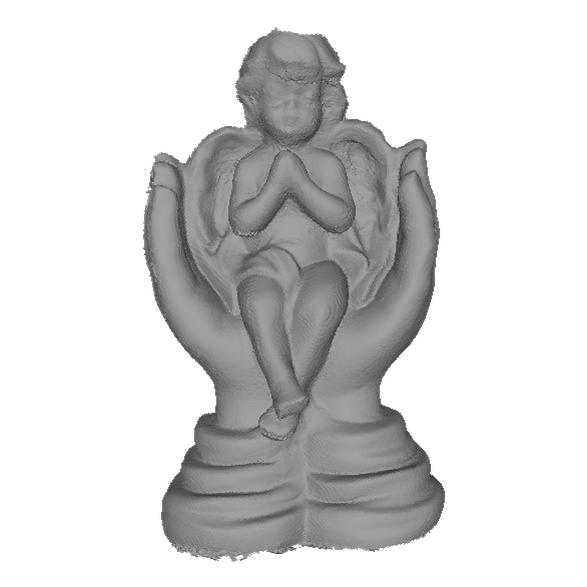} &
        \includegraphics[width=0.3\textwidth]{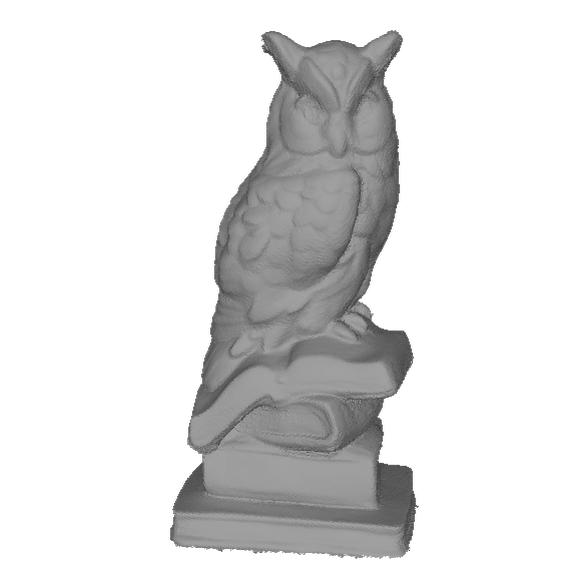} \\
        scan114 & scan118 & scan122 \\
    \end{tabular}
    \caption{Our method's reconstructions of the DTU \cite{DTU} dataset.}
    \label{fig:DTU}
\end{figure}

\cref{fig:DTU} shows our method's reconstruction of all of the DTU \cite{DTU} dataset.

\subsection{Additional Examples from The Ablation Study}

\begin{figure}[]
    \centering
    \begin{tabular}{cccc}
    \textbf{Ground} & \textbf{MVSFormer} & \textbf{MVSFormer} & \textbf{Ours} \\
    \textbf{Truth} & & \textbf{+ Rendered} & \\
    \includegraphics[width=0.22\textwidth]{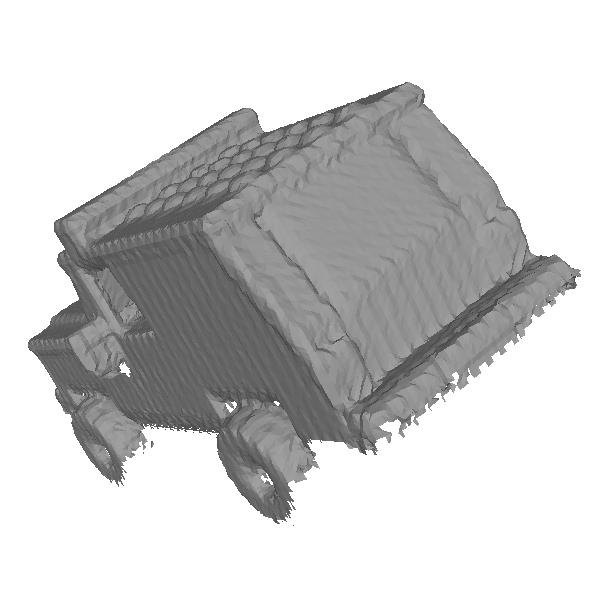} & 
    \includegraphics[width=0.22\textwidth]{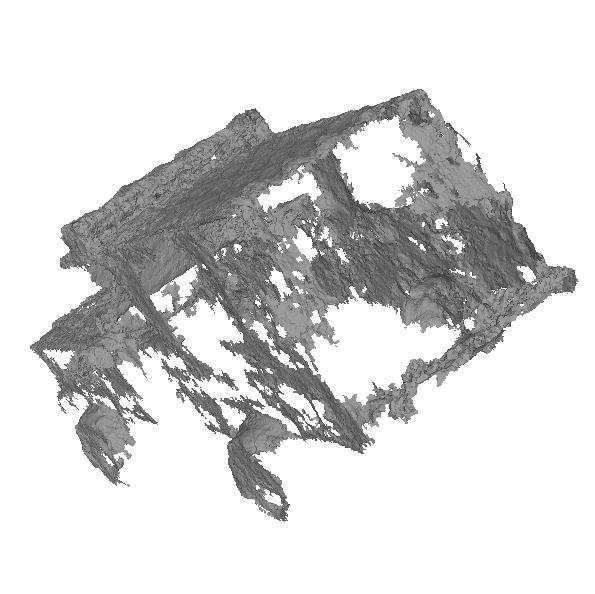} &
    \includegraphics[width=0.22\textwidth]{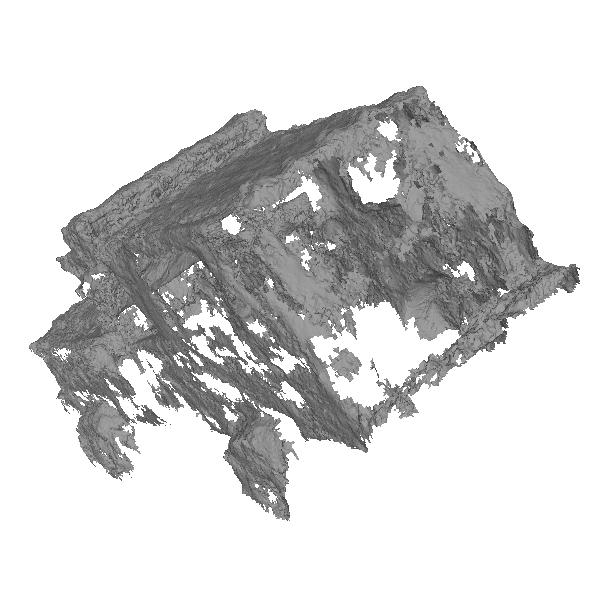} &
    \includegraphics[width=0.22\textwidth]{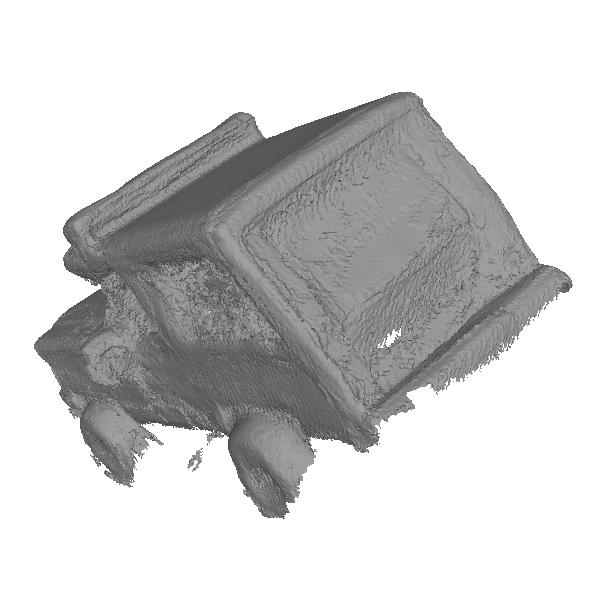} \\
    \multicolumn{4}{c}{Jeep} \\
    \includegraphics[width=0.22\textwidth]{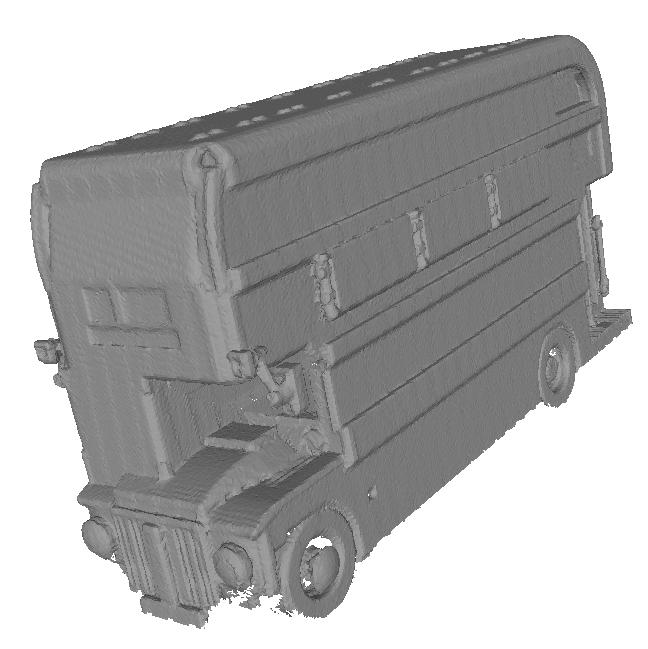} & 
    \includegraphics[width=0.22\textwidth]{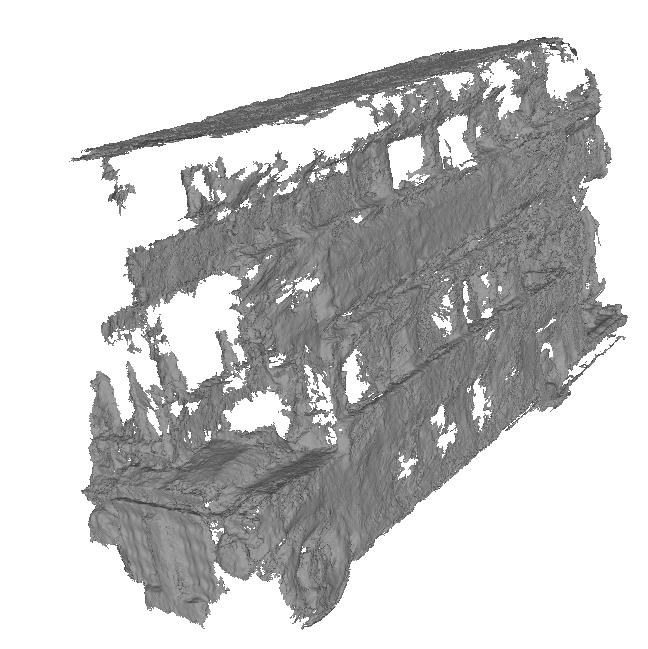} &
    \includegraphics[width=0.22\textwidth]{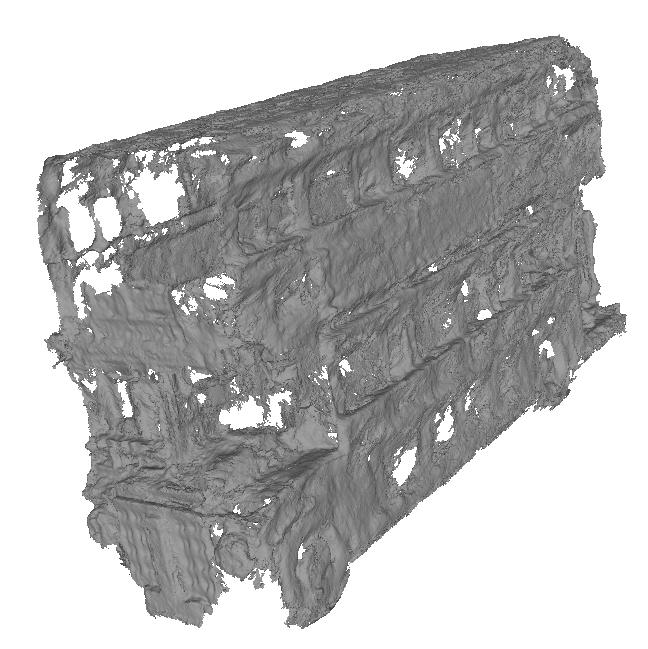} &
    \includegraphics[width=0.22\textwidth]{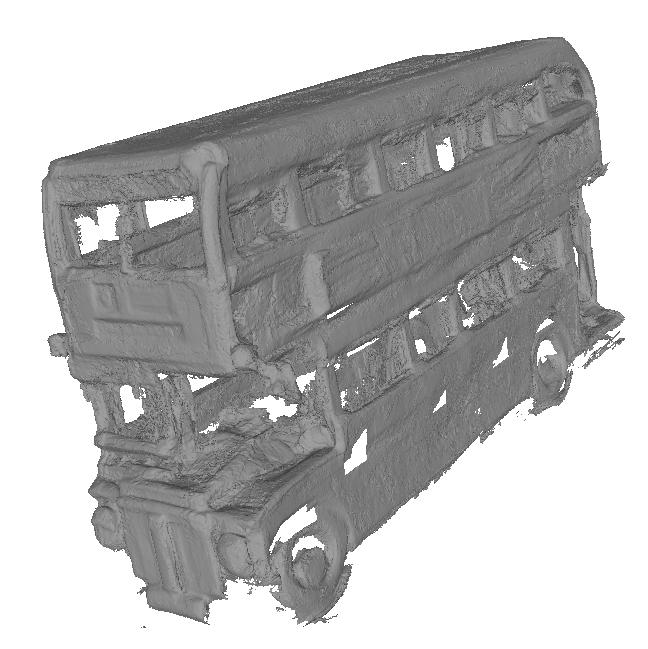} \\
    \multicolumn{4}{c}{London Bus} \\
    \includegraphics[width=0.22\textwidth]{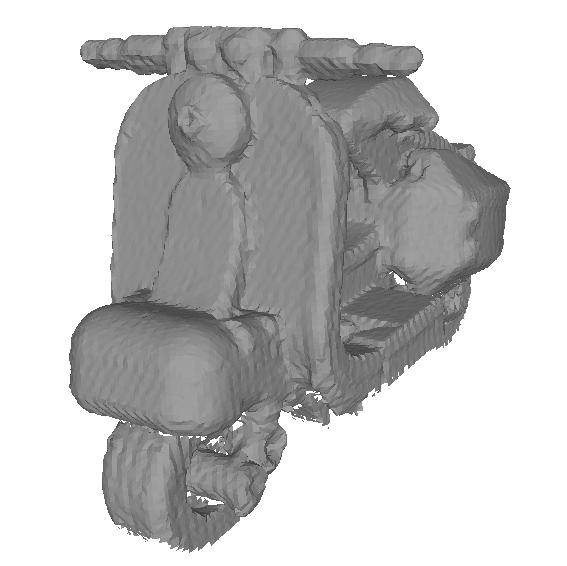} & 
    \includegraphics[width=0.22\textwidth]{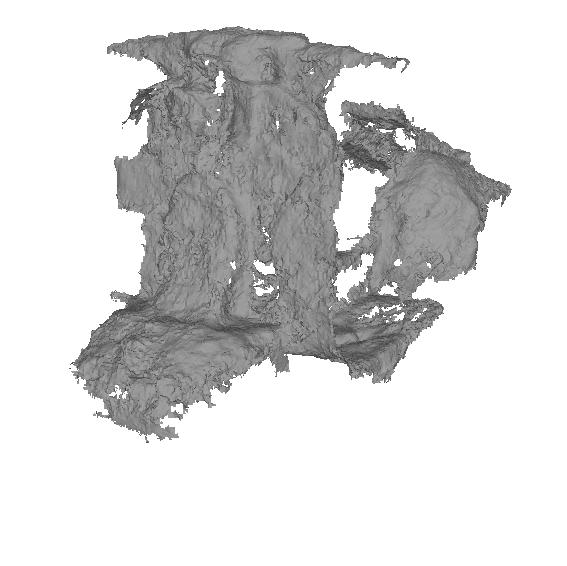} &
    \includegraphics[width=0.22\textwidth]{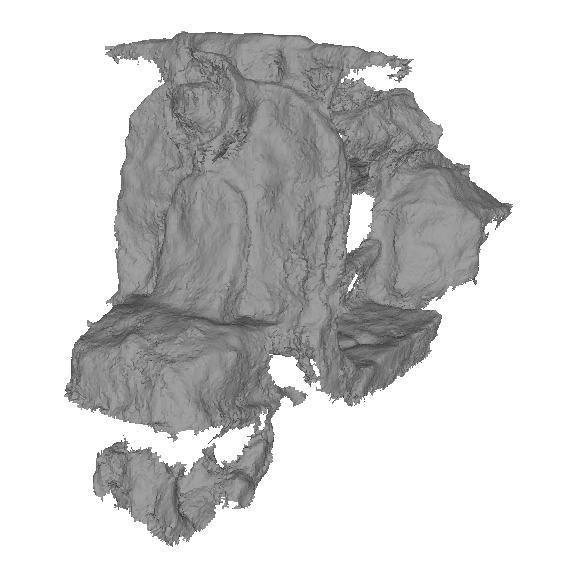} &
    \includegraphics[width=0.22\textwidth]{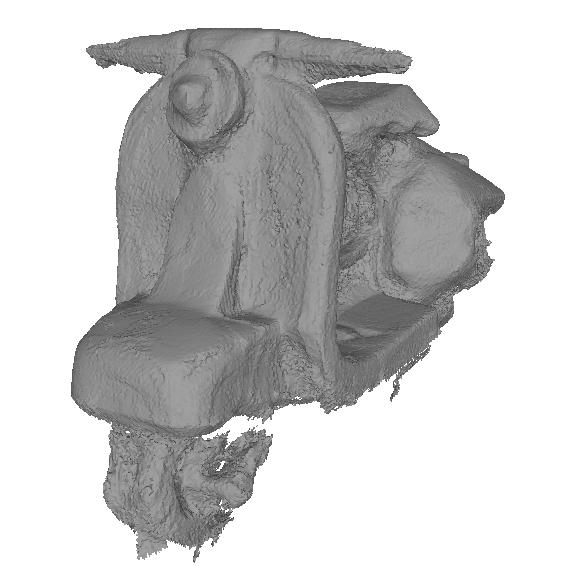} \\
    \multicolumn{4}{c}{Motorcycle} \\
    \includegraphics[width=0.22\textwidth]{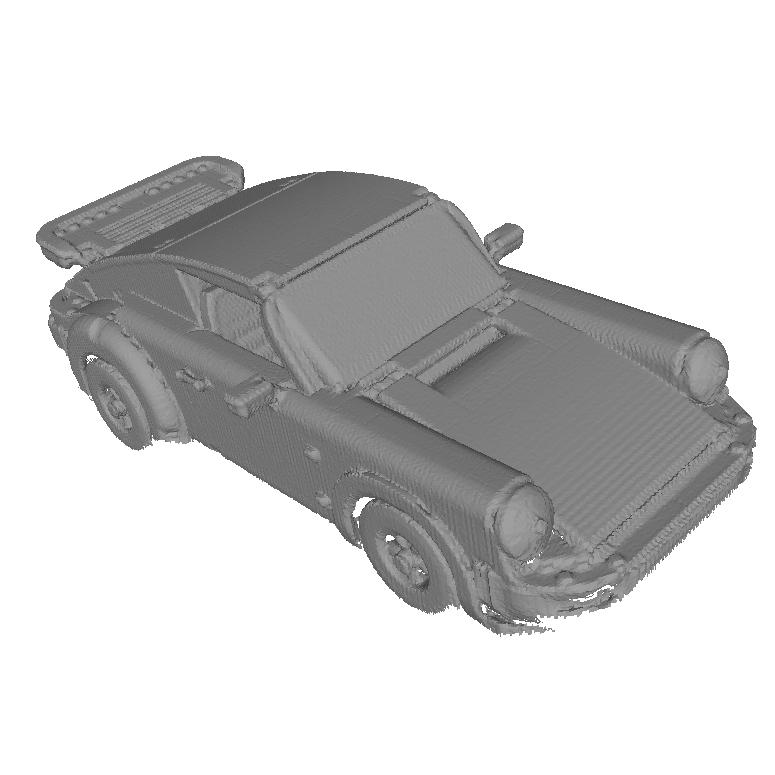} & 
    \includegraphics[width=0.22\textwidth]{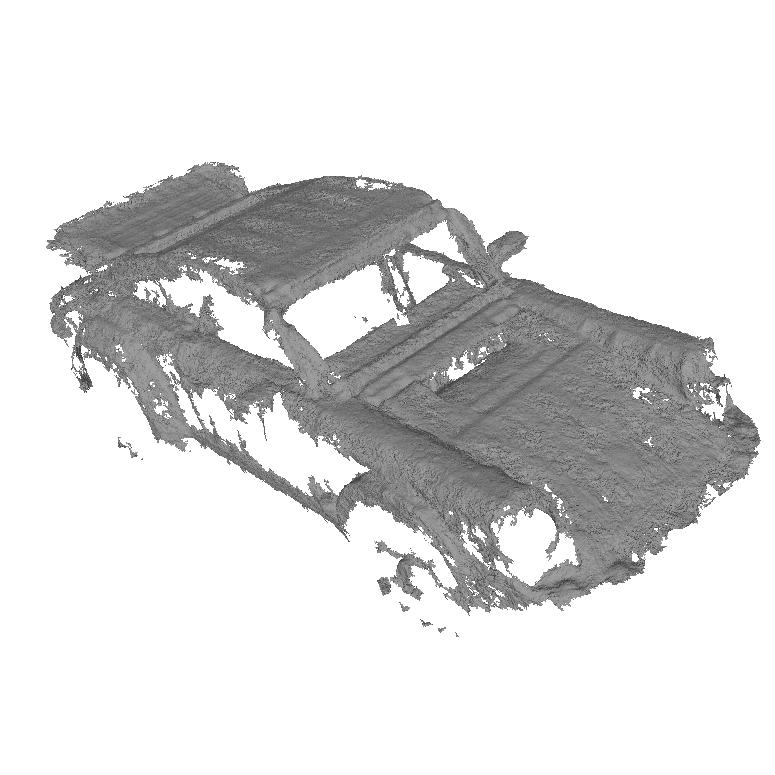} &
    \includegraphics[width=0.22\textwidth]{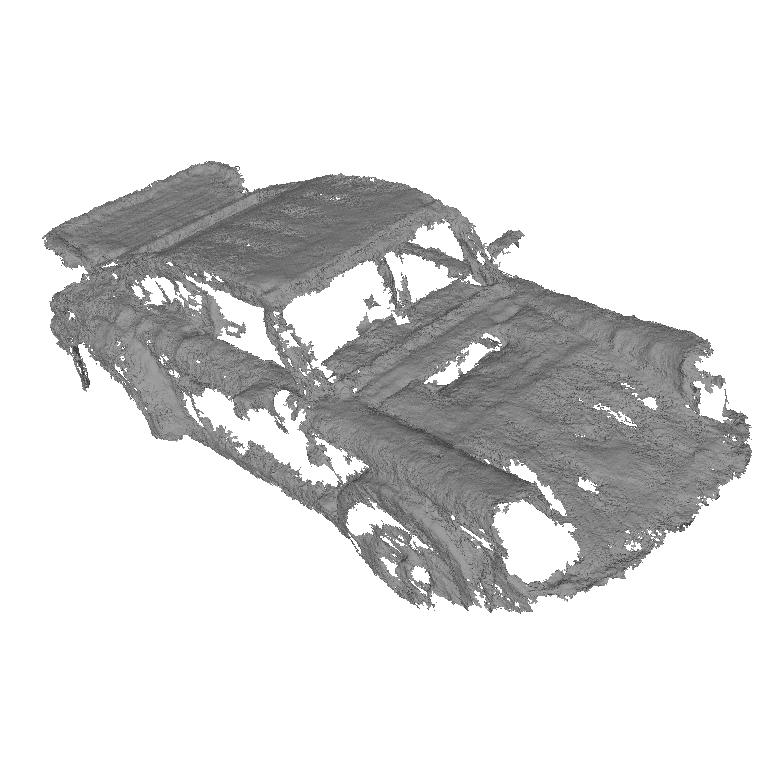} &
    \includegraphics[width=0.22\textwidth]{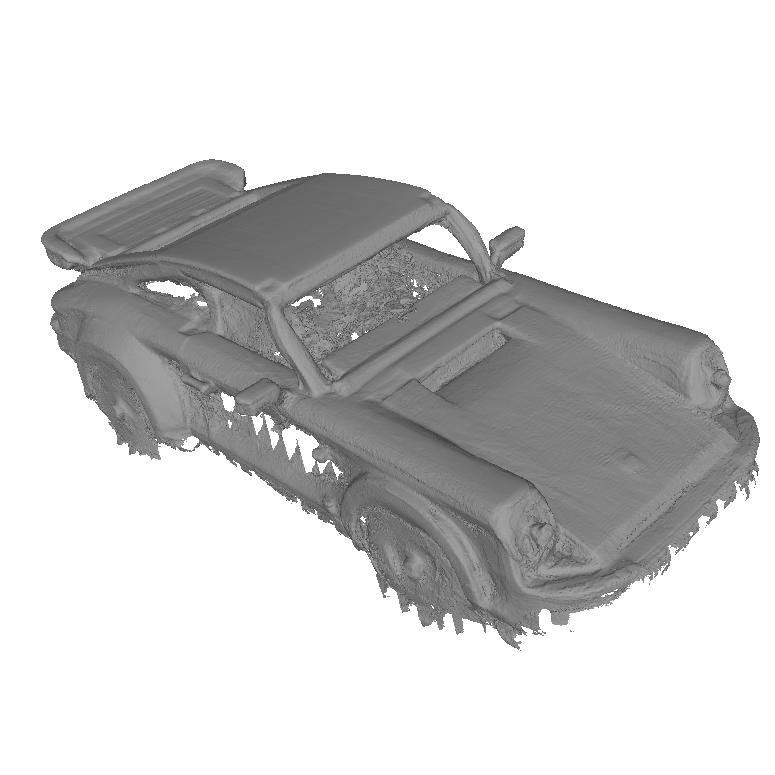} \\
    \multicolumn{4}{c}{Porsche} \\
    \includegraphics[width=0.22\textwidth]{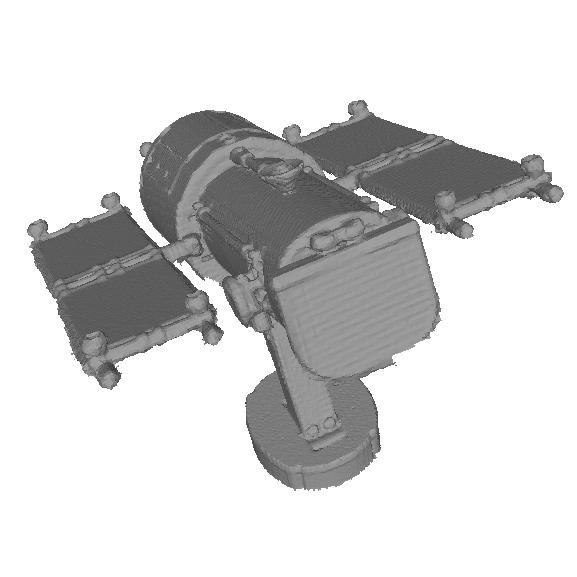} & 
    \includegraphics[width=0.22\textwidth]{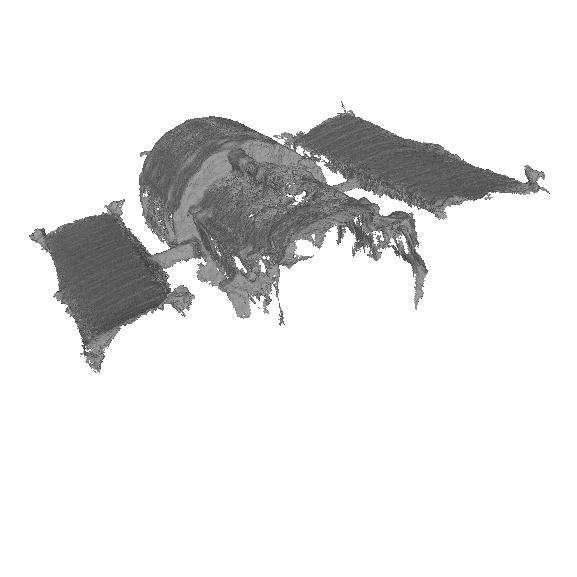} &
    \includegraphics[width=0.22\textwidth]{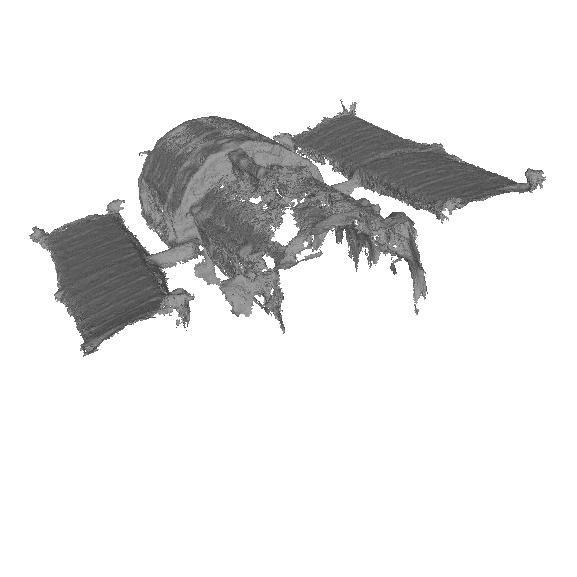} &
    \includegraphics[width=0.22\textwidth]{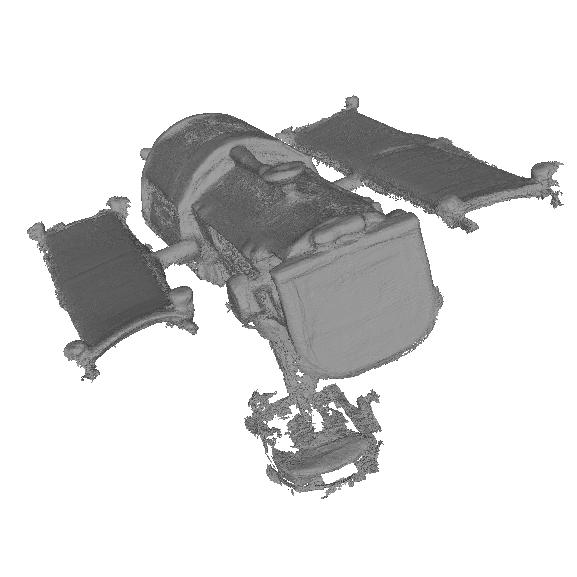} \\
    \multicolumn{4}{c}{Satellite} \\
    \includegraphics[width=0.22\textwidth]{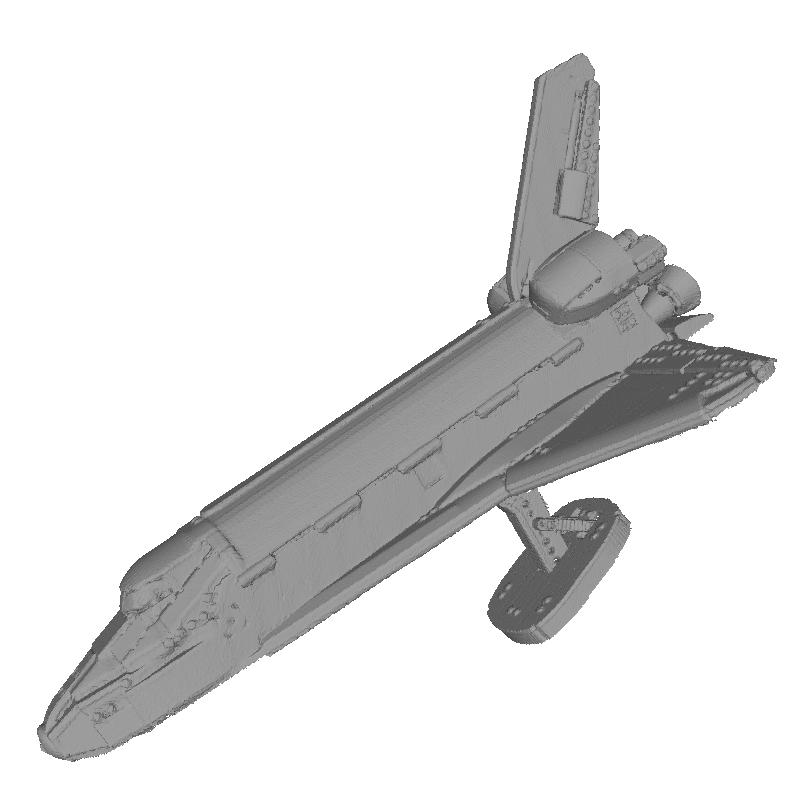} & 
    \includegraphics[width=0.22\textwidth]{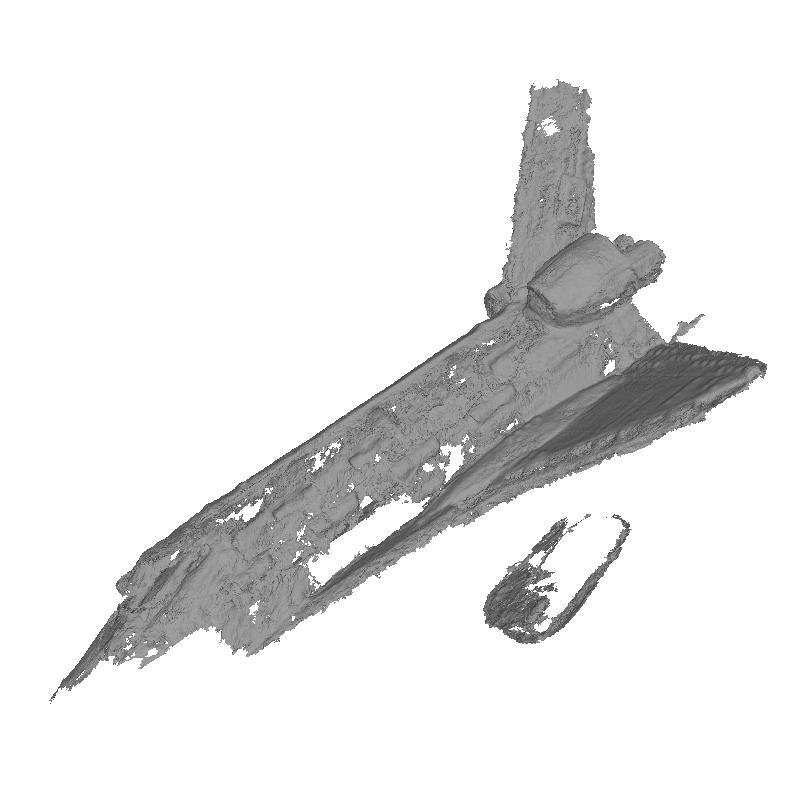} &
    \includegraphics[width=0.22\textwidth]{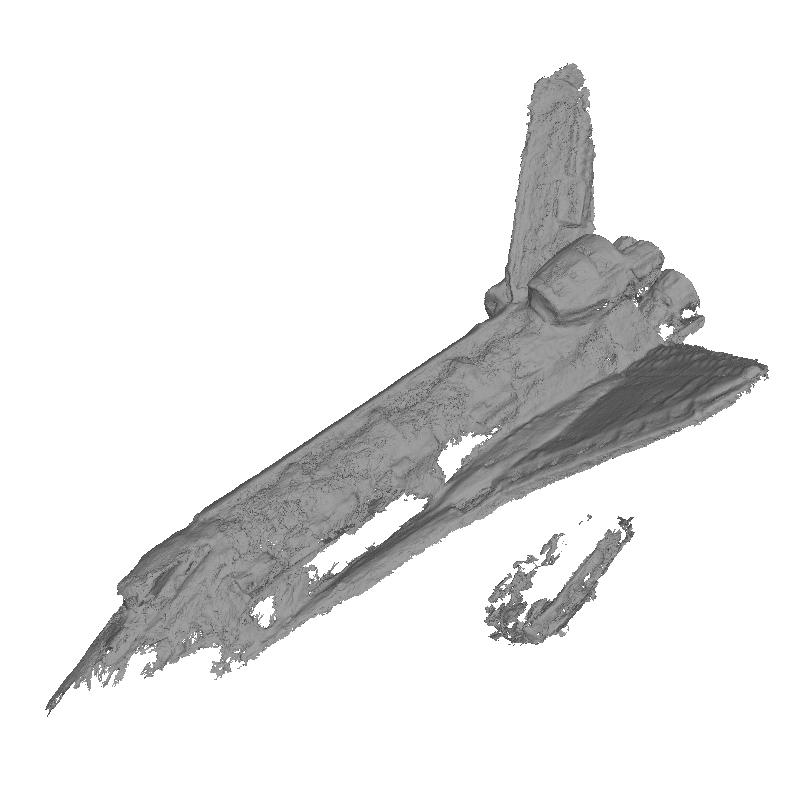} &
    \includegraphics[width=0.22\textwidth]{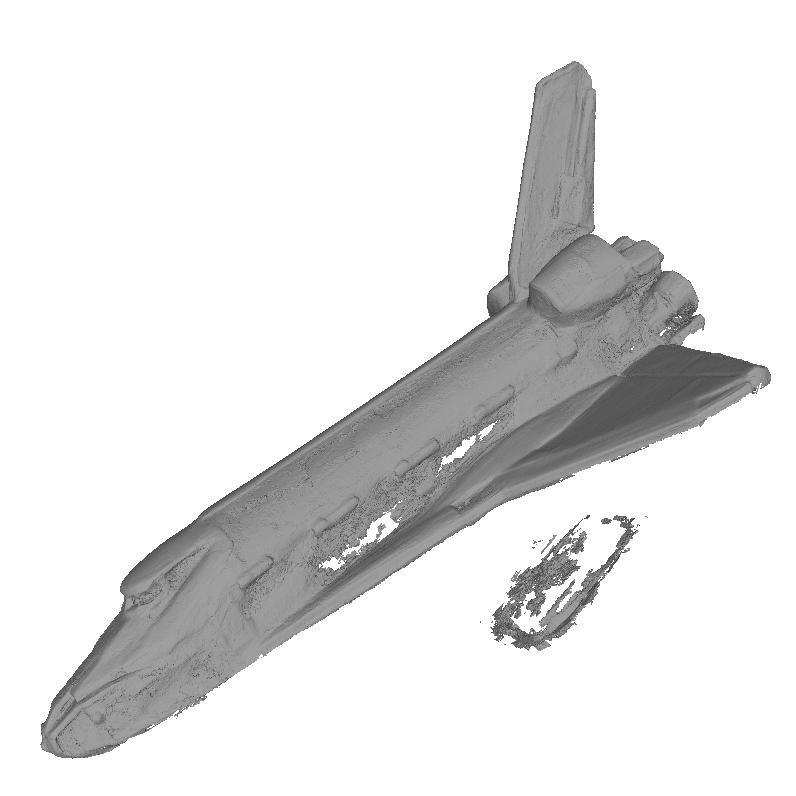} \\
    \multicolumn{4}{c}{Space Shuttle} \\
    \end{tabular}
    \caption{Left to right: ground truths of MobileBrick \cite{li2023mobilebrick} dataset, reconstructions from MVSFormer \cite{mvsformer} with original and rendered images, as well as our reconstruction.}
    \label{fig:mobile_brick_sup}
\end{figure}

\begin{figure}[]
    \centering
    \begin{tabular}{cccc}
    \textbf{Image} & \textbf{MVSFormer} & \textbf{MVSFormer} & \textbf{Ours} \\
    & & \textbf{+ Rendered} & \\
    \includegraphics[width=0.22\textwidth]{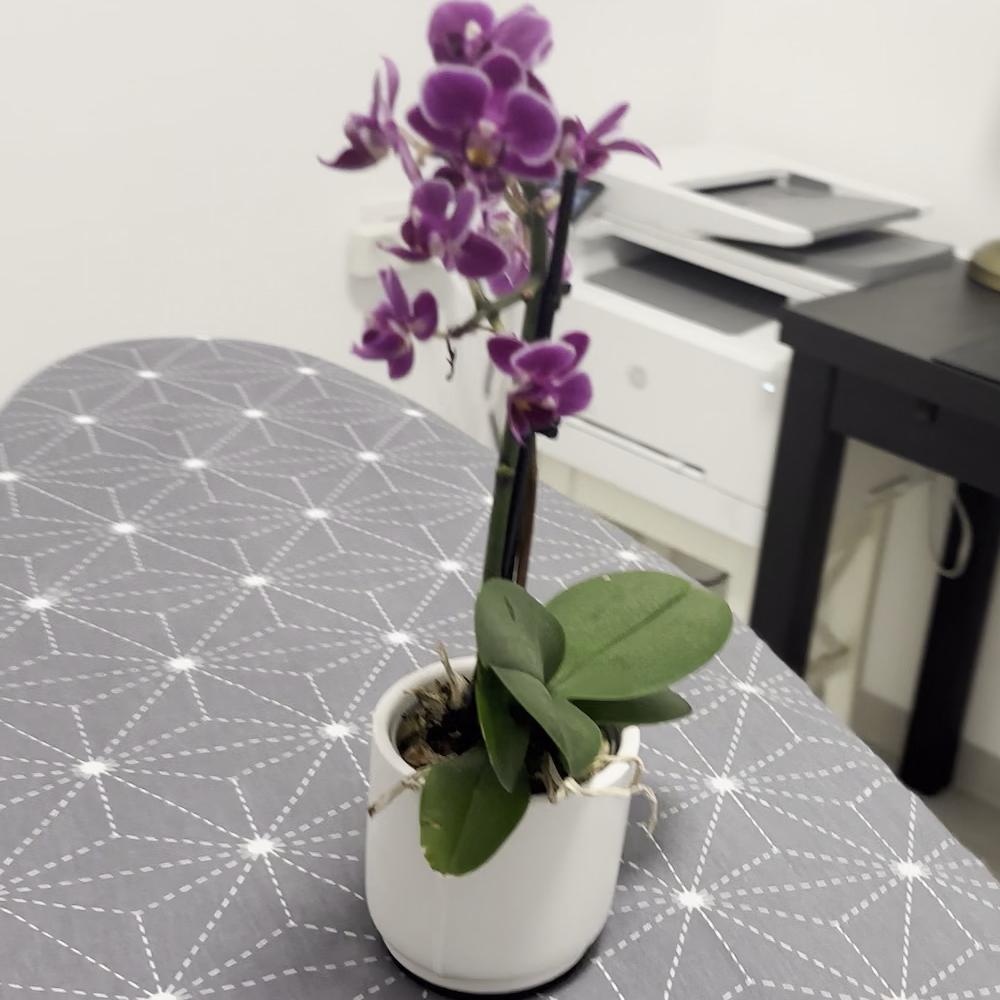} & 
    \includegraphics[width=0.22\textwidth]{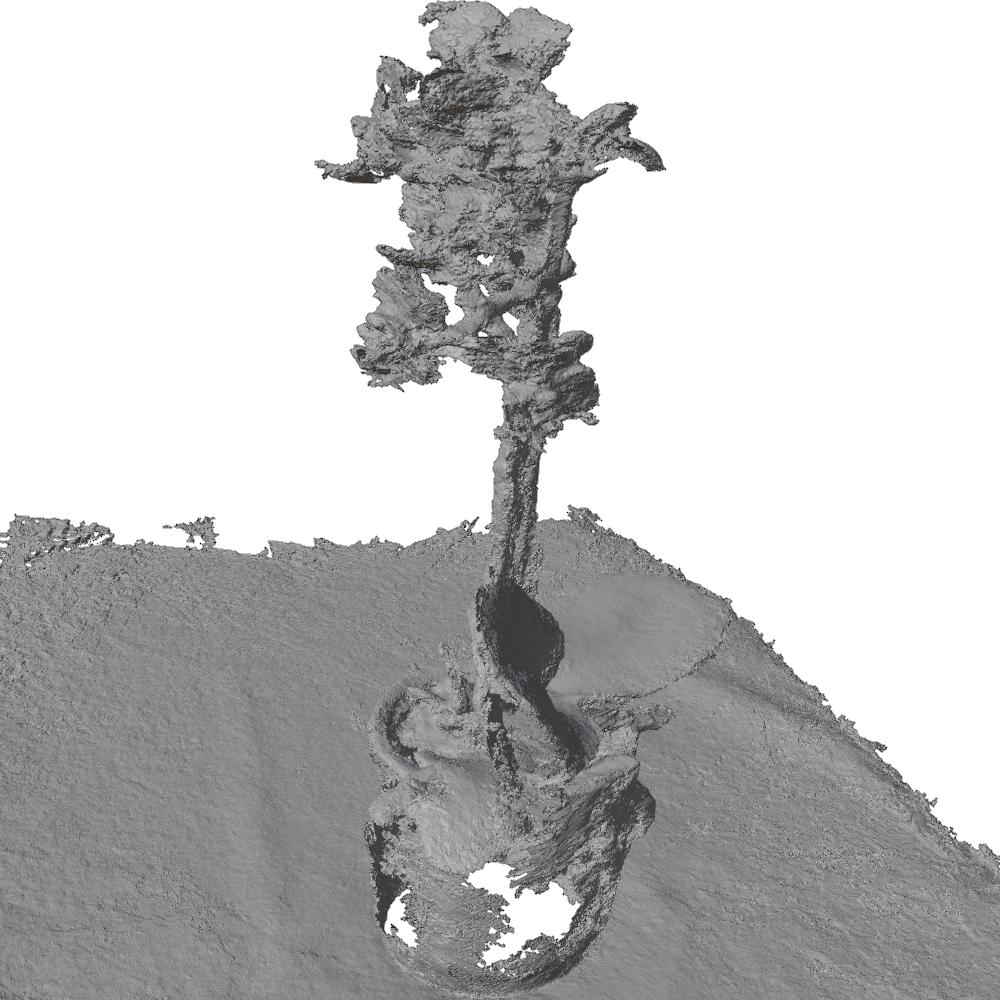} &
    \includegraphics[width=0.22\textwidth]{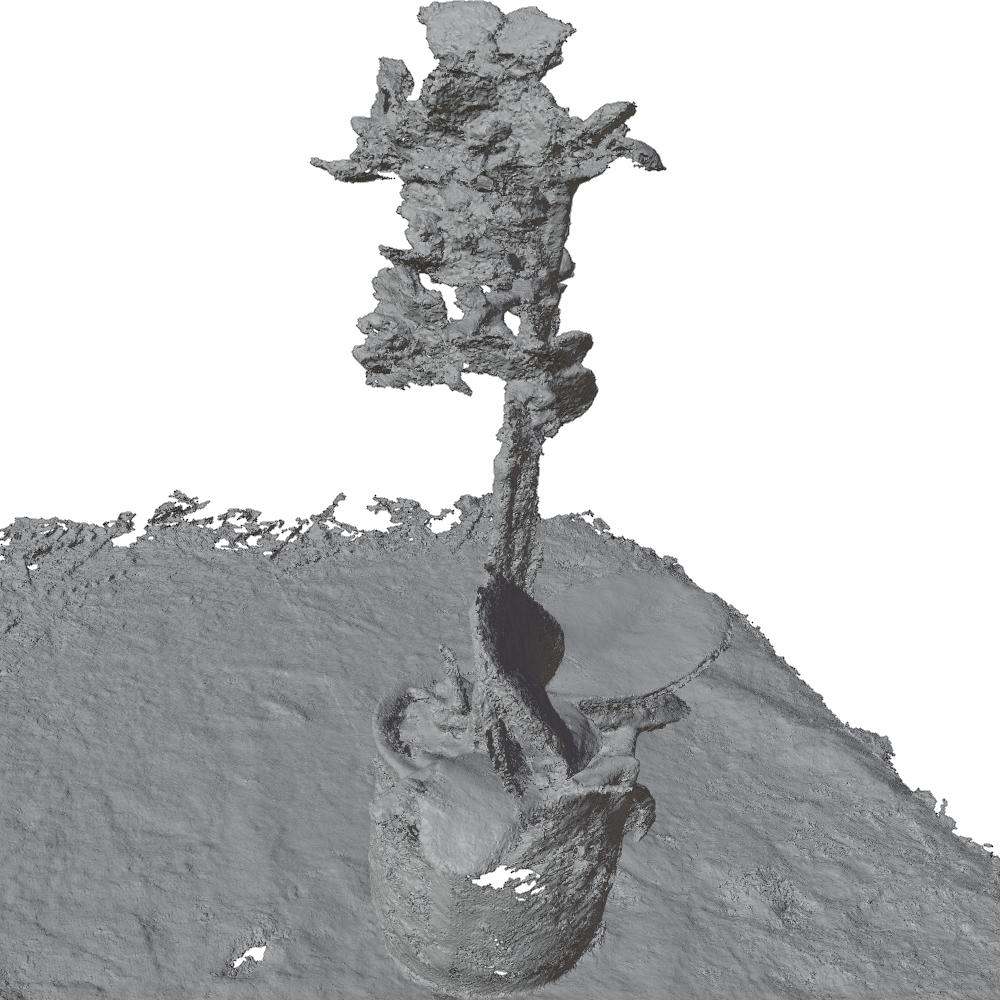} &
    \includegraphics[width=0.22\textwidth]{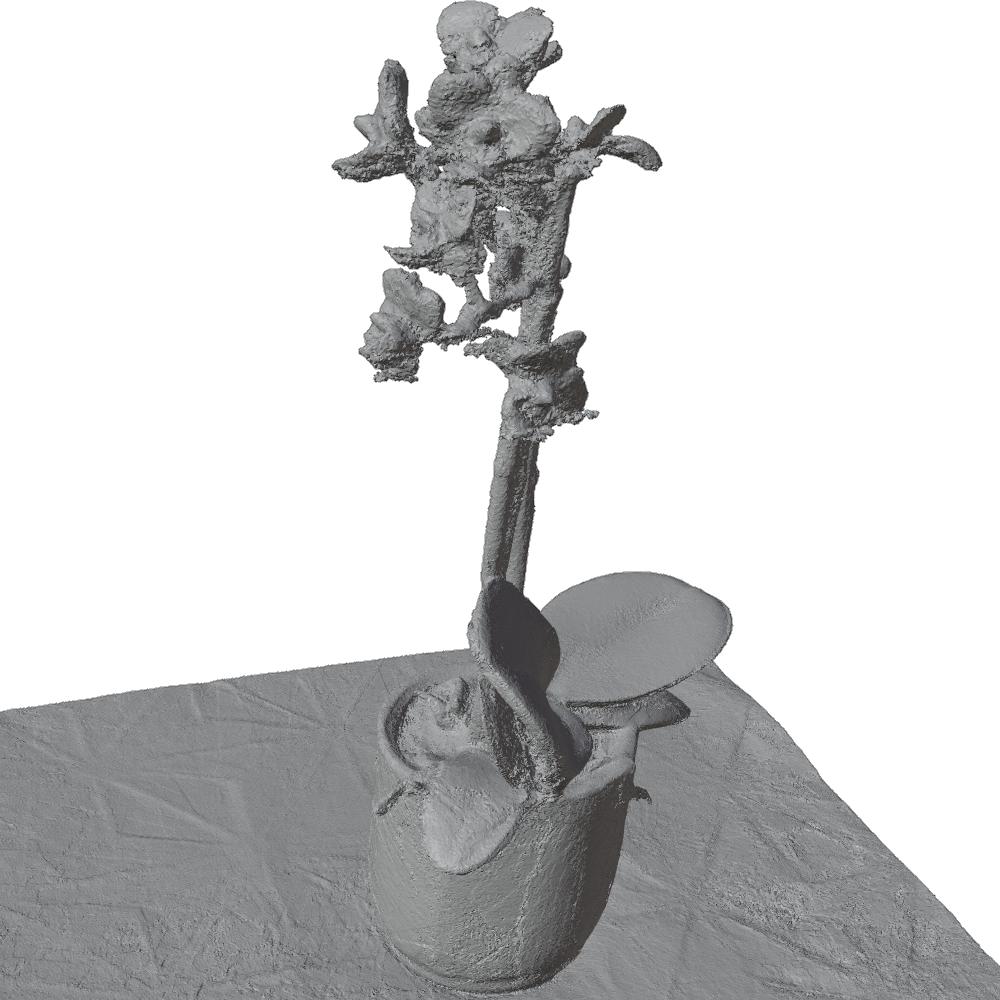} \\
    \includegraphics[width=0.22\textwidth]{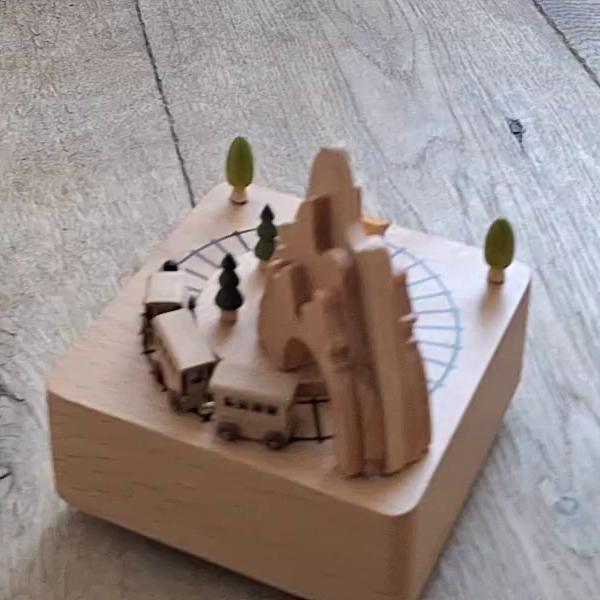} & 
    \includegraphics[width=0.22\textwidth]{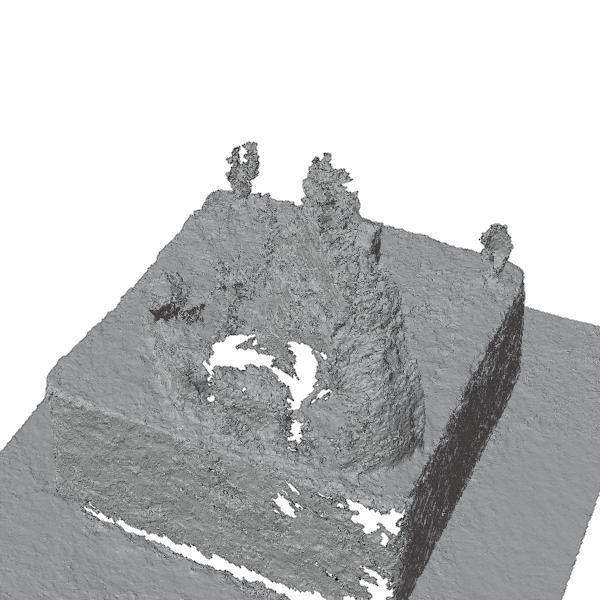} &
    \includegraphics[width=0.22\textwidth]{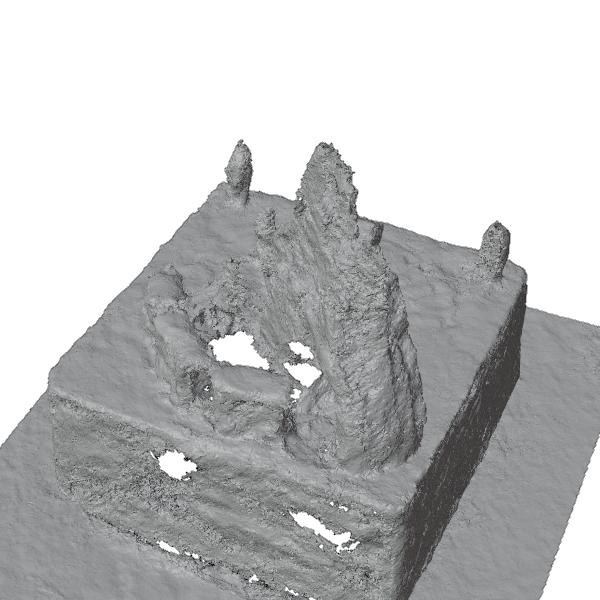} &
    \includegraphics[width=0.22\textwidth]{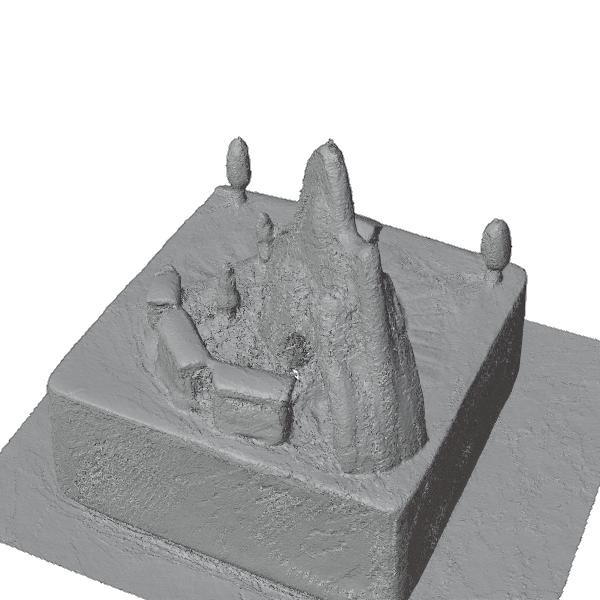} \\
    \includegraphics[width=0.22\textwidth]{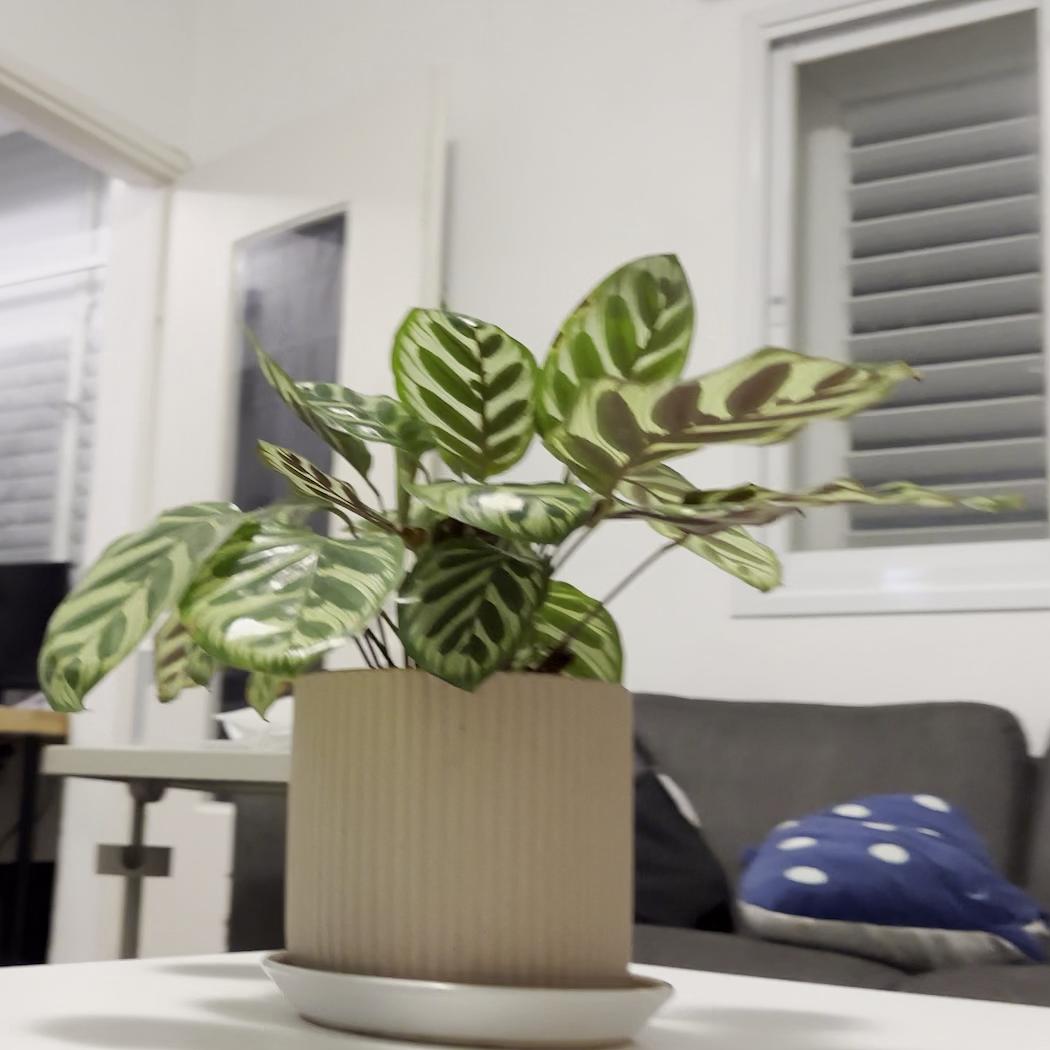} & 
    \includegraphics[width=0.22\textwidth]{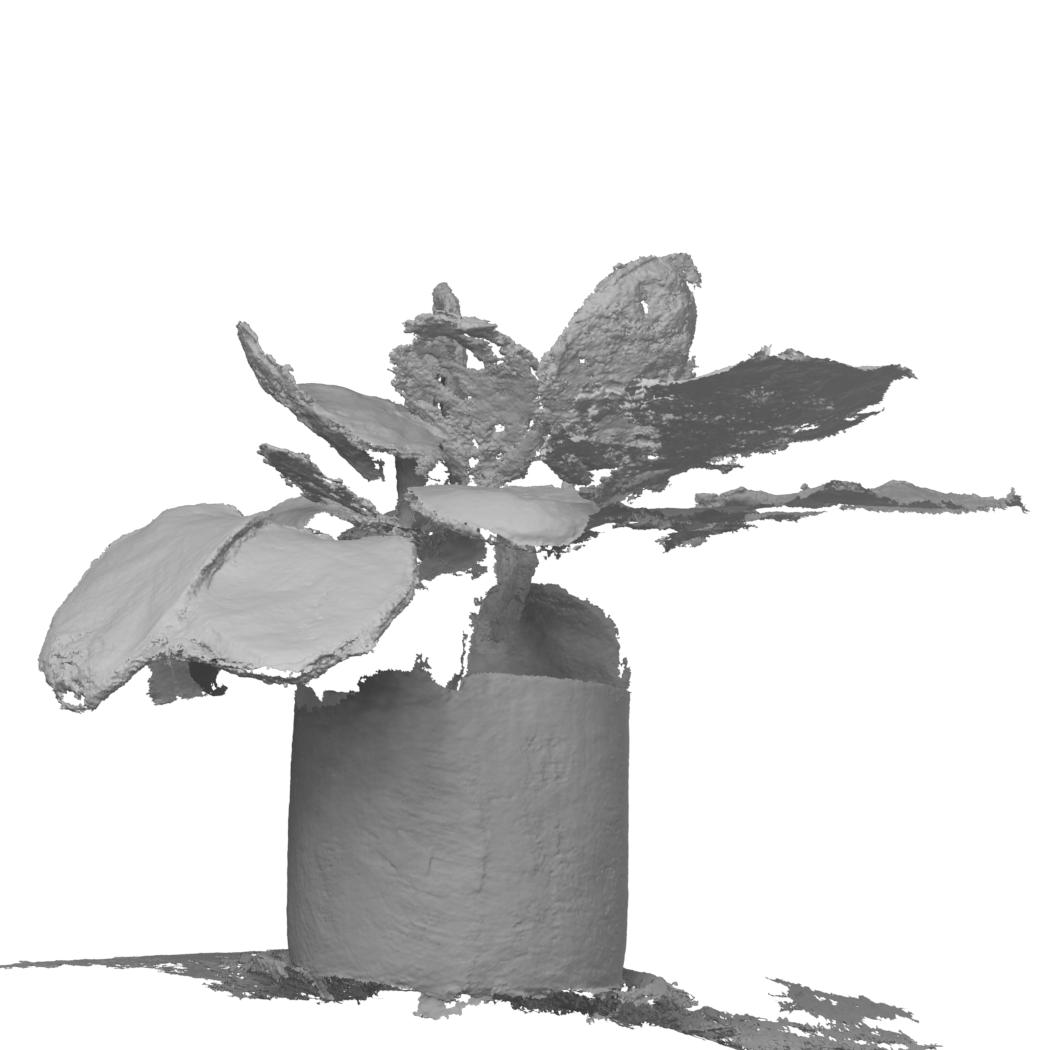} &
    \includegraphics[width=0.22\textwidth]{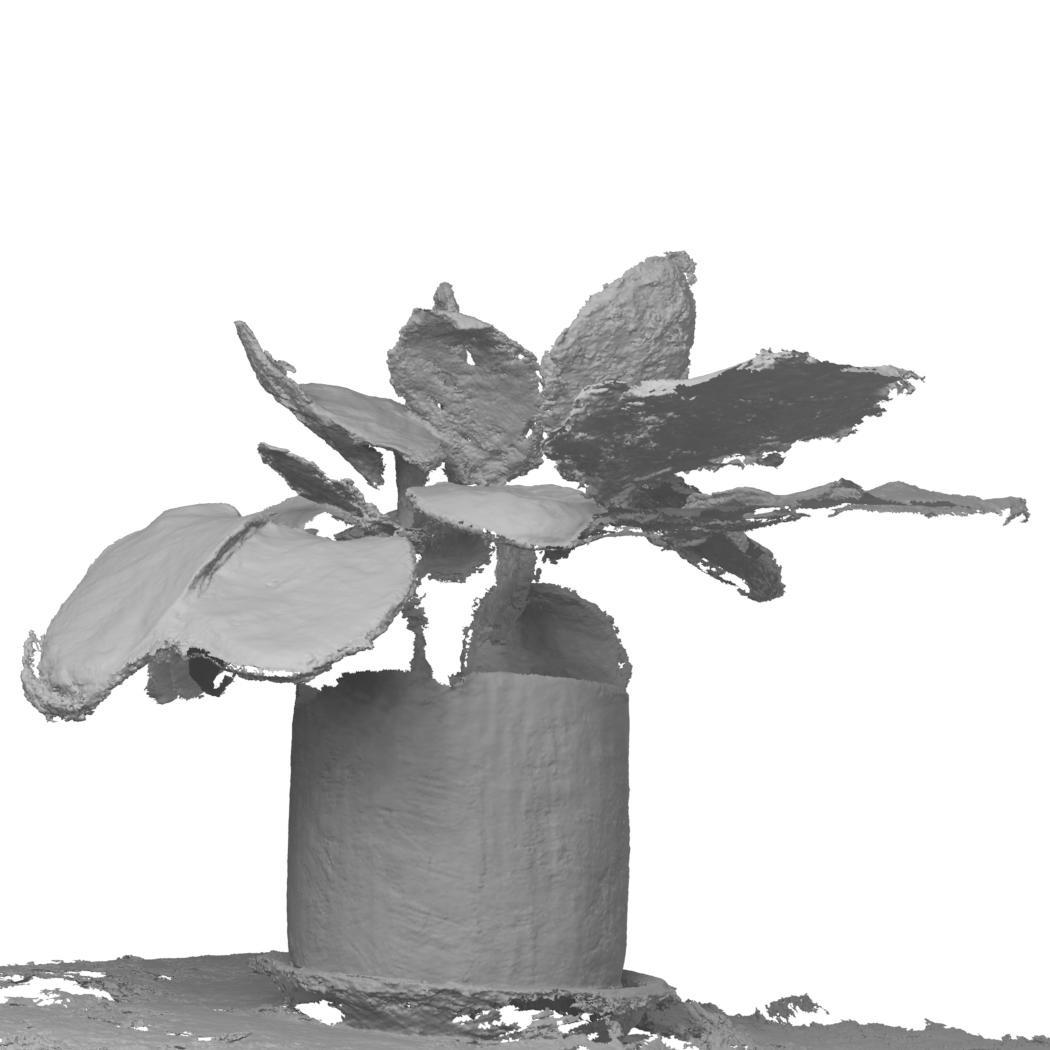} &
    \includegraphics[width=0.22\textwidth]{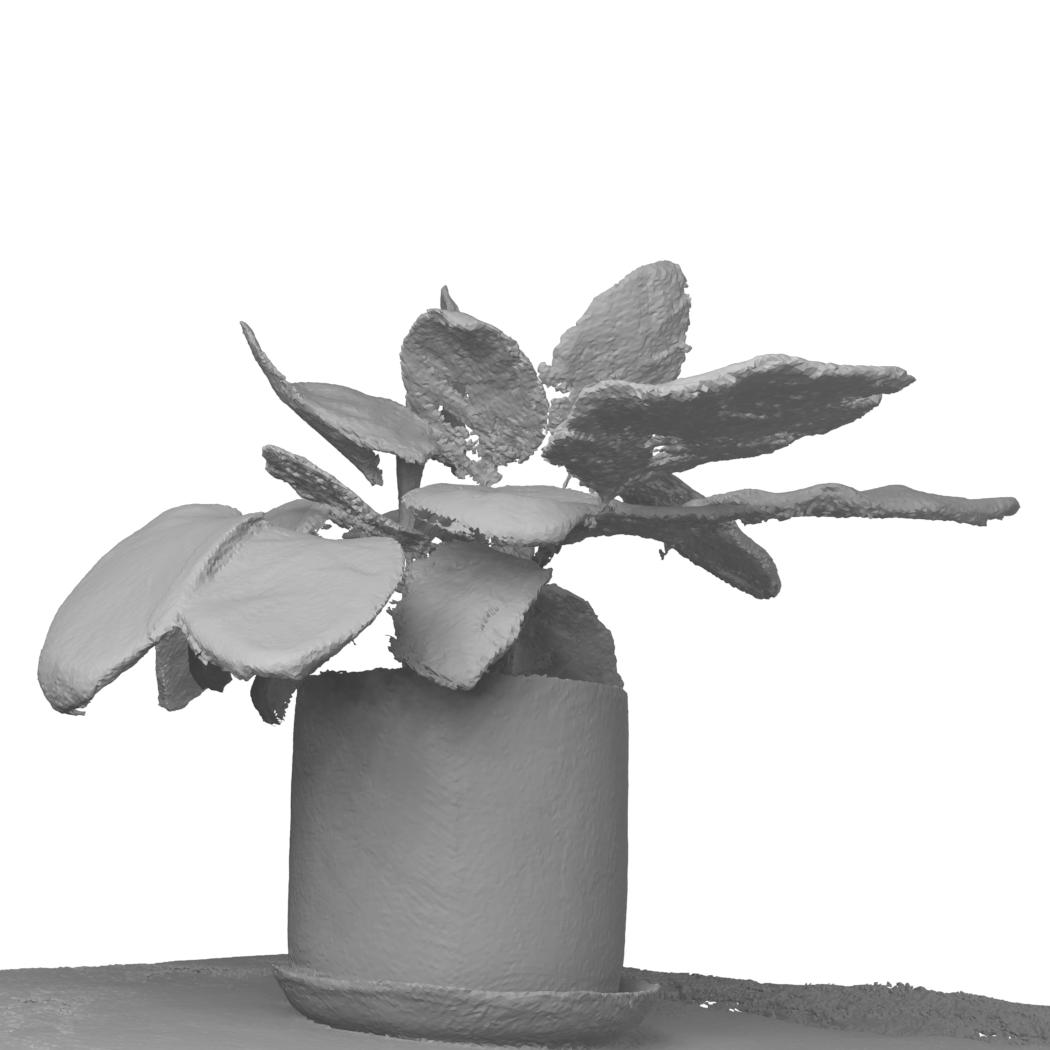} \\
    \includegraphics[width=0.22\textwidth]{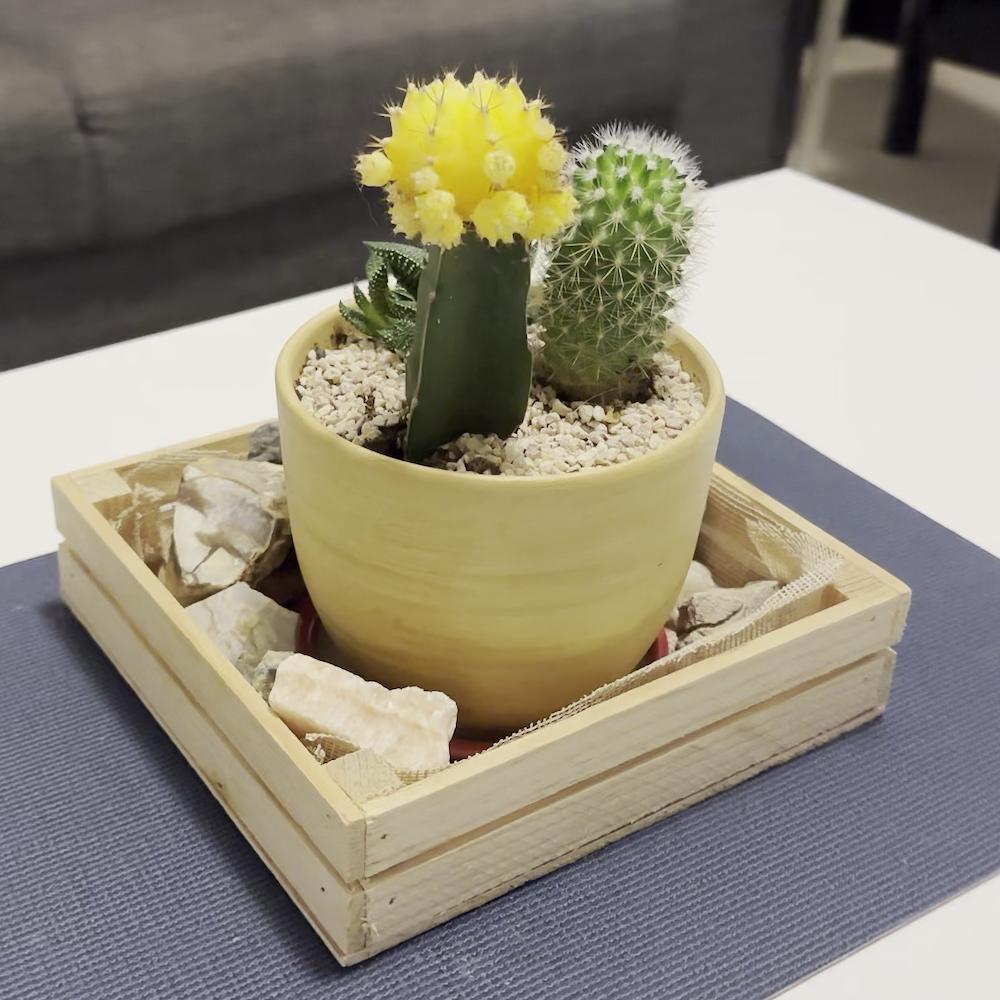} & 
    \includegraphics[width=0.22\textwidth]{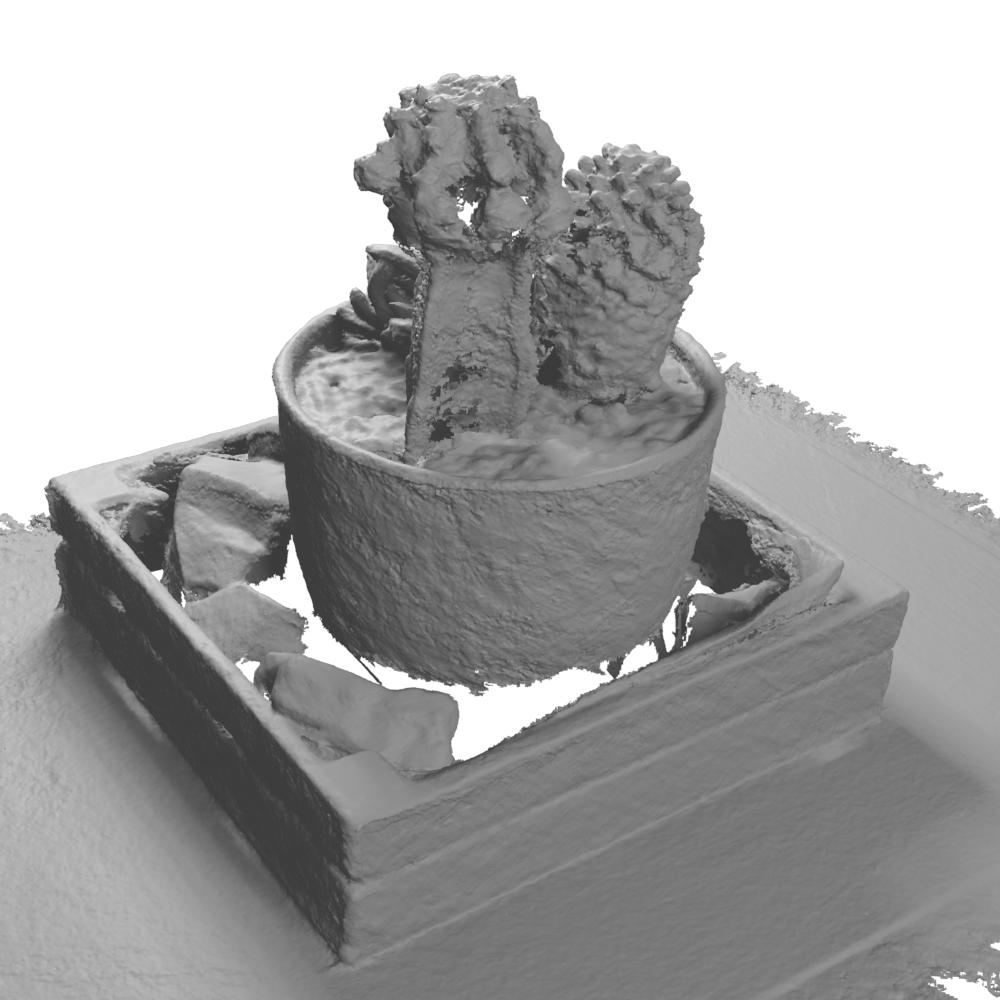} &
    \includegraphics[width=0.22\textwidth]{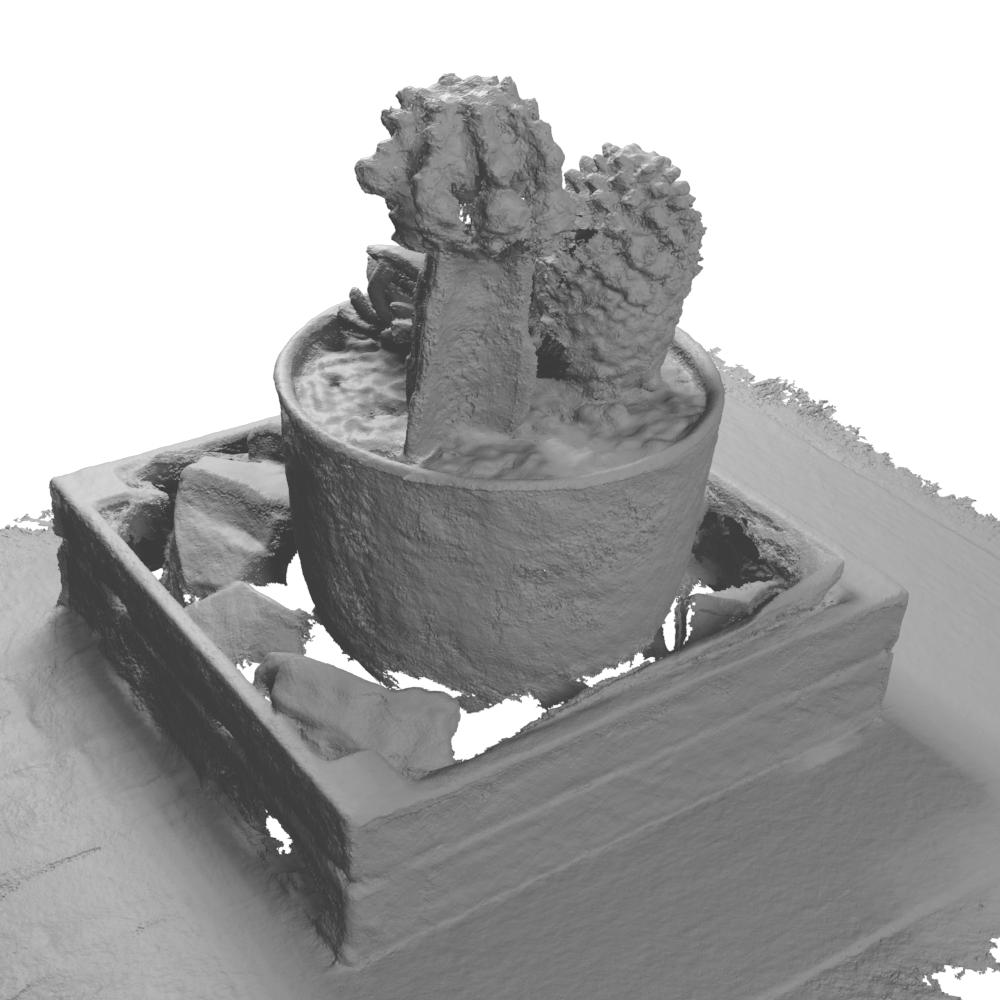} &
    \includegraphics[width=0.22\textwidth]{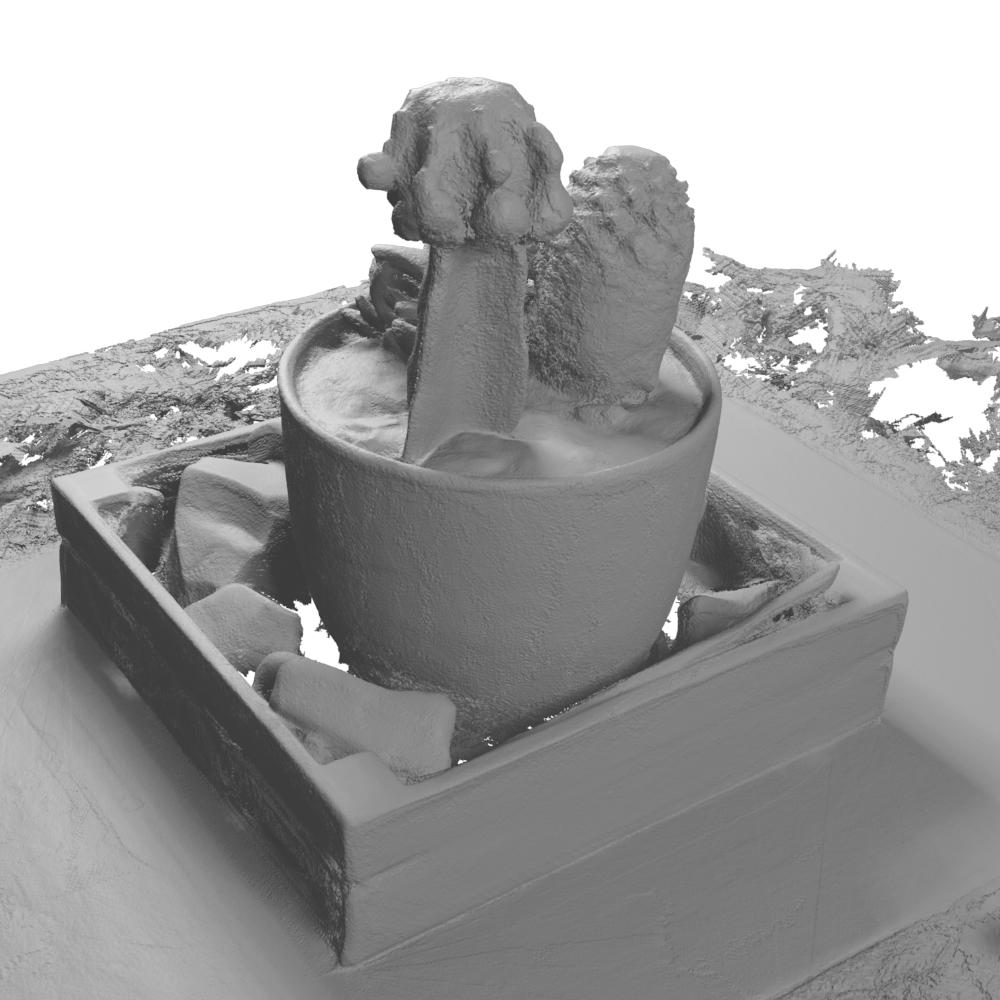} \\
    \includegraphics[width=0.22\textwidth]{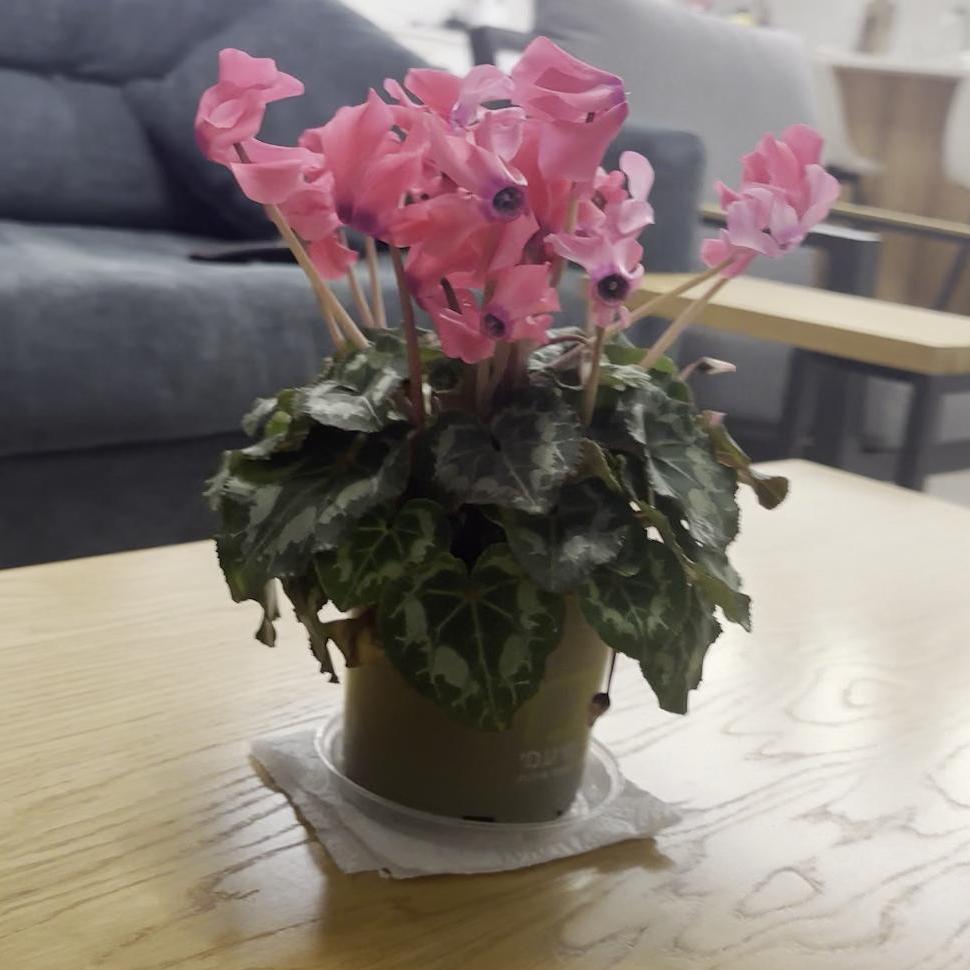} & 
    \includegraphics[width=0.22\textwidth]{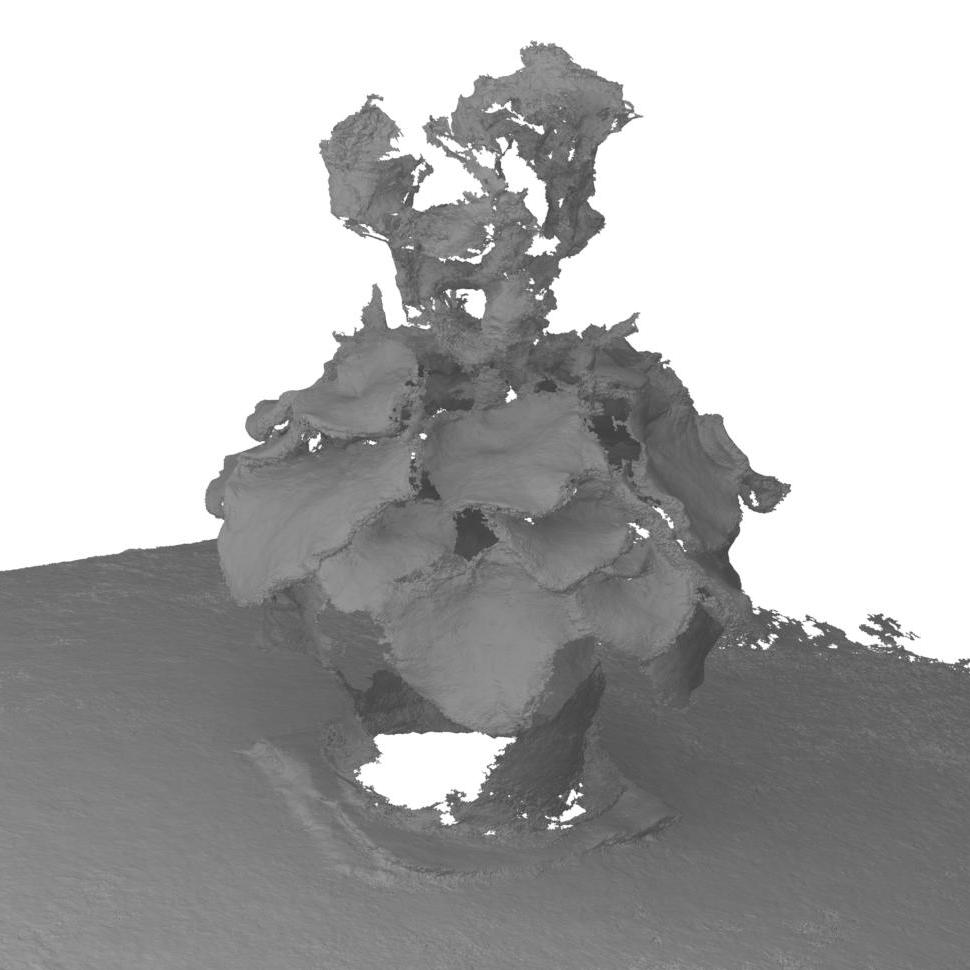} &
    \includegraphics[width=0.22\textwidth]{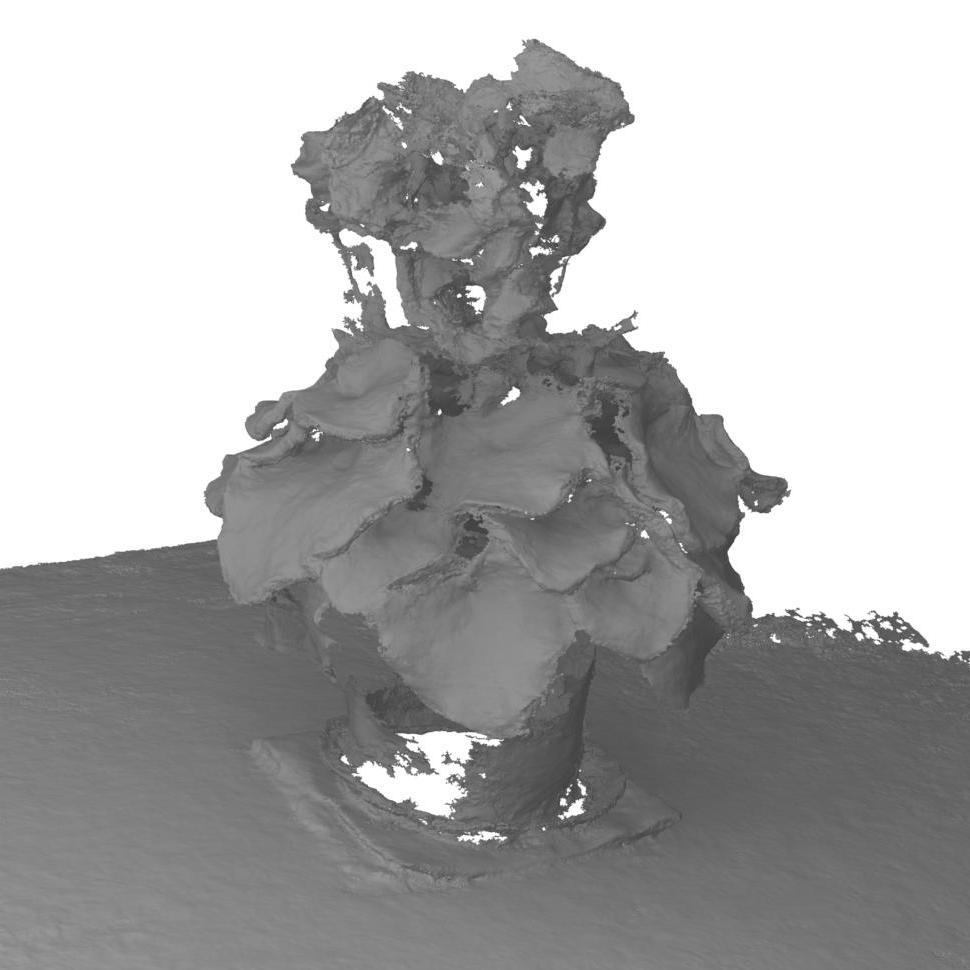} &
    \includegraphics[width=0.22\textwidth]{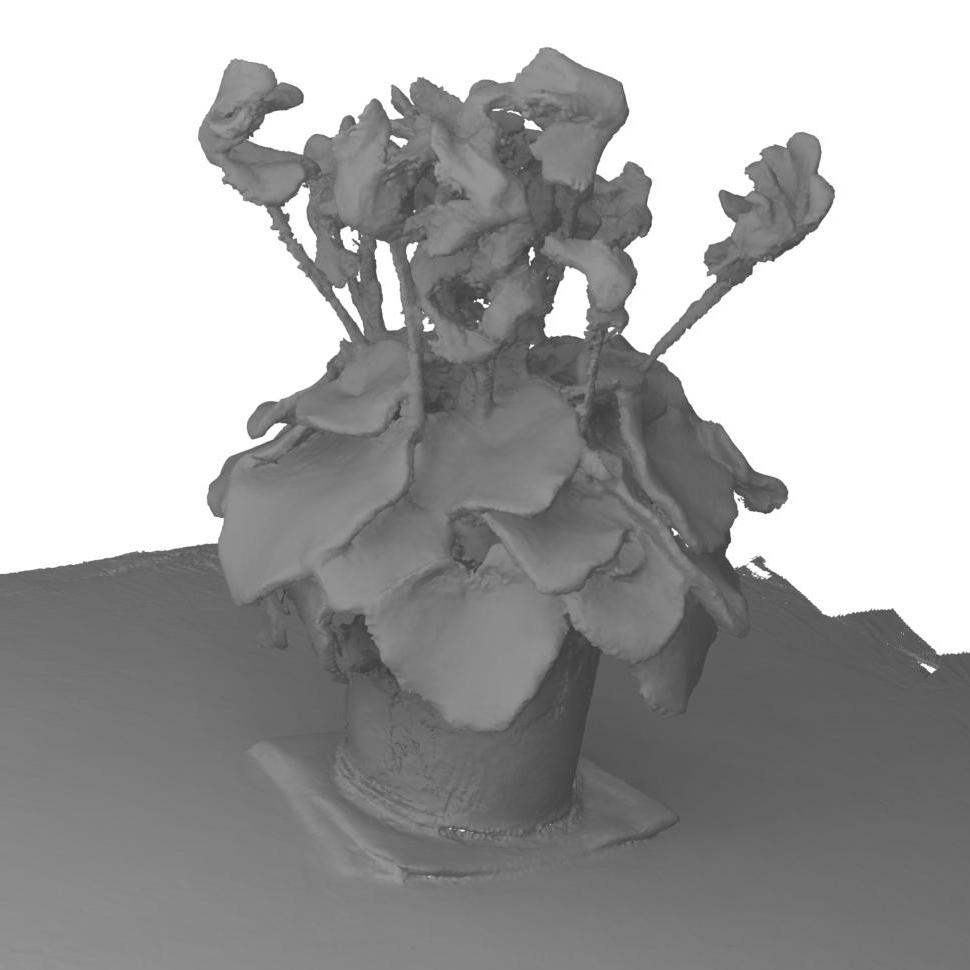} \\
    \end{tabular}
    \caption{Additional examples from the ablation study, showing reconstructions done by MVSFormer \cite{mvsformer} on the original images and on the rendered images from the same poses, as well as our method's reconstruction.}
    \label{fig:Stereo matching_vs._MVS_sup}
\end{figure}

\cref{fig:mobile_brick_sup} presents results from the ablation study on the MobileBrick \cite{li2023mobilebrick} test set, showing MVSFormer\cite{mvsformer}, MVSFormer with rendered images as input, and our method's reconstructions, as well as the ground truth. Additional results on in-the-wild scenes taken by a smartphone are shown in \cref{fig:Stereo matching_vs._MVS_sup}.

\begin{figure}[]
    \centering
    \begin{tabular}{cccc}
    \textbf{Ground} & \textbf{MVSFormer} & \textbf{MVSFormer} & \textbf{Ours} \\
    \textbf{Truth} & & \textbf{+ Rendered} & \\
    \includegraphics[width=0.22\textwidth]{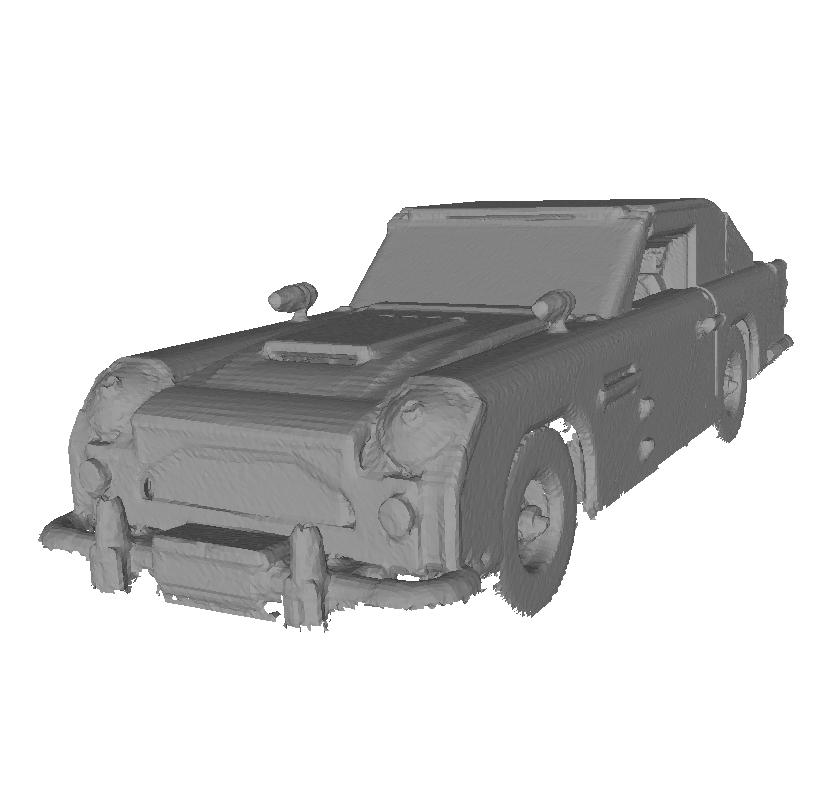} & 
    \includegraphics[width=0.22\textwidth]{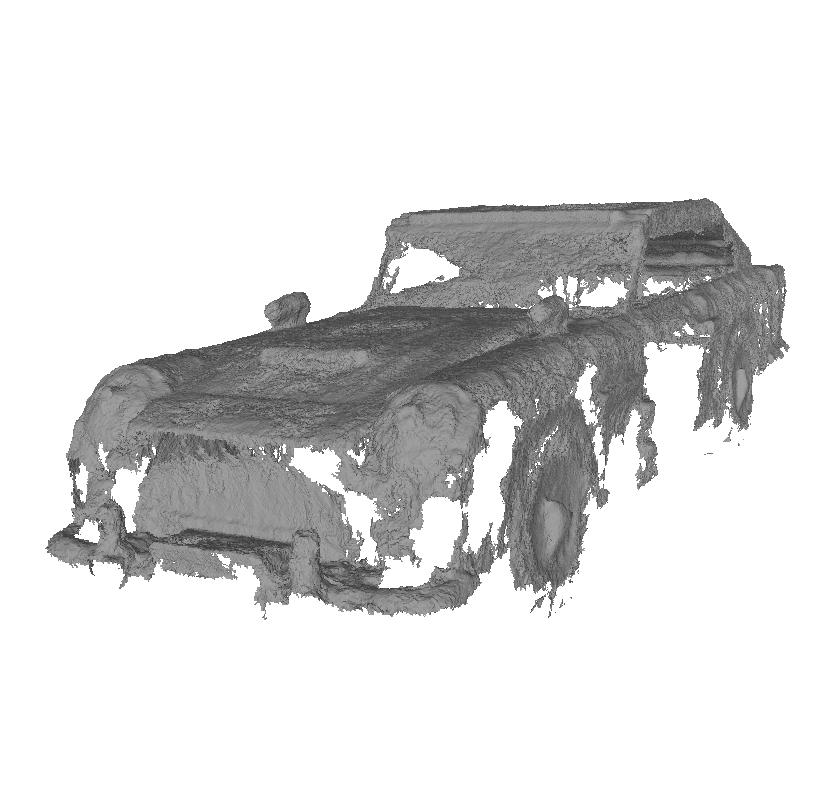} &
    \includegraphics[width=0.22\textwidth]{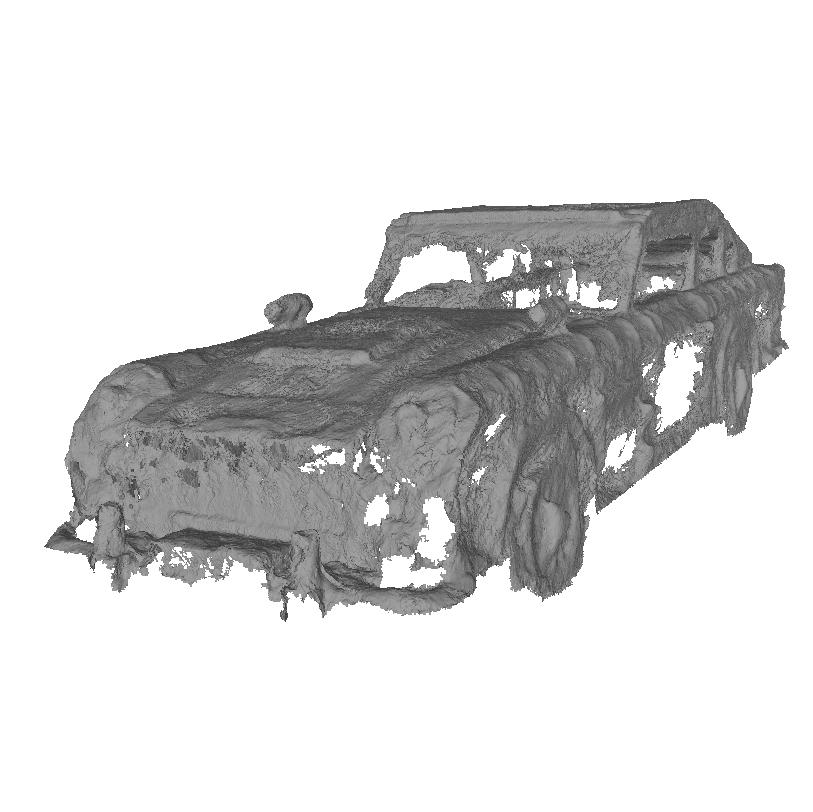} &
    \includegraphics[width=0.22\textwidth]{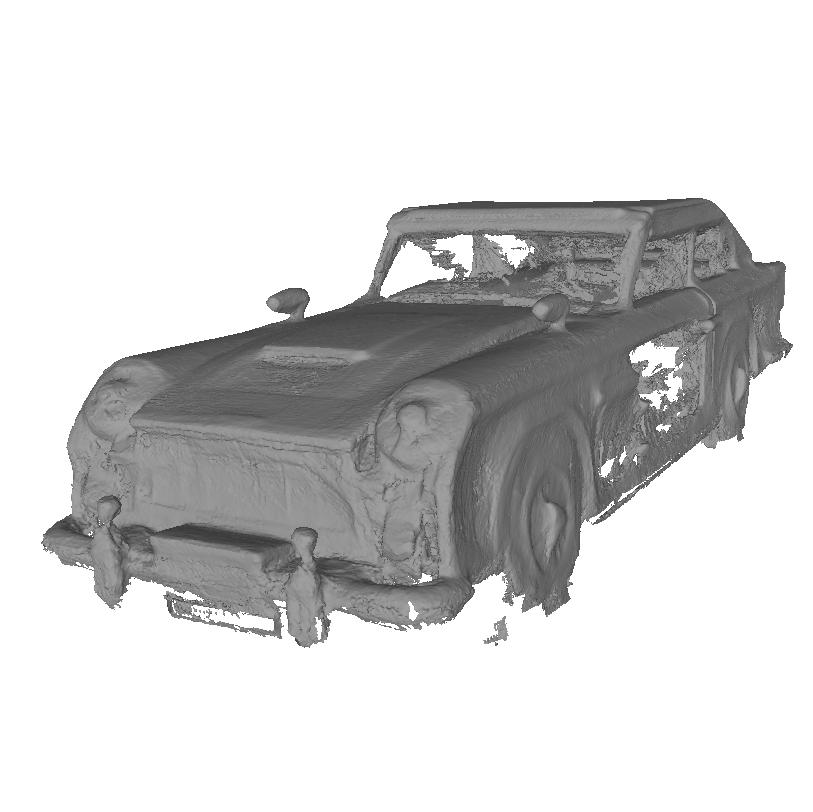} \\    
    \multicolumn{4}{c}{Aston} \\
    \includegraphics[width=0.22\textwidth]{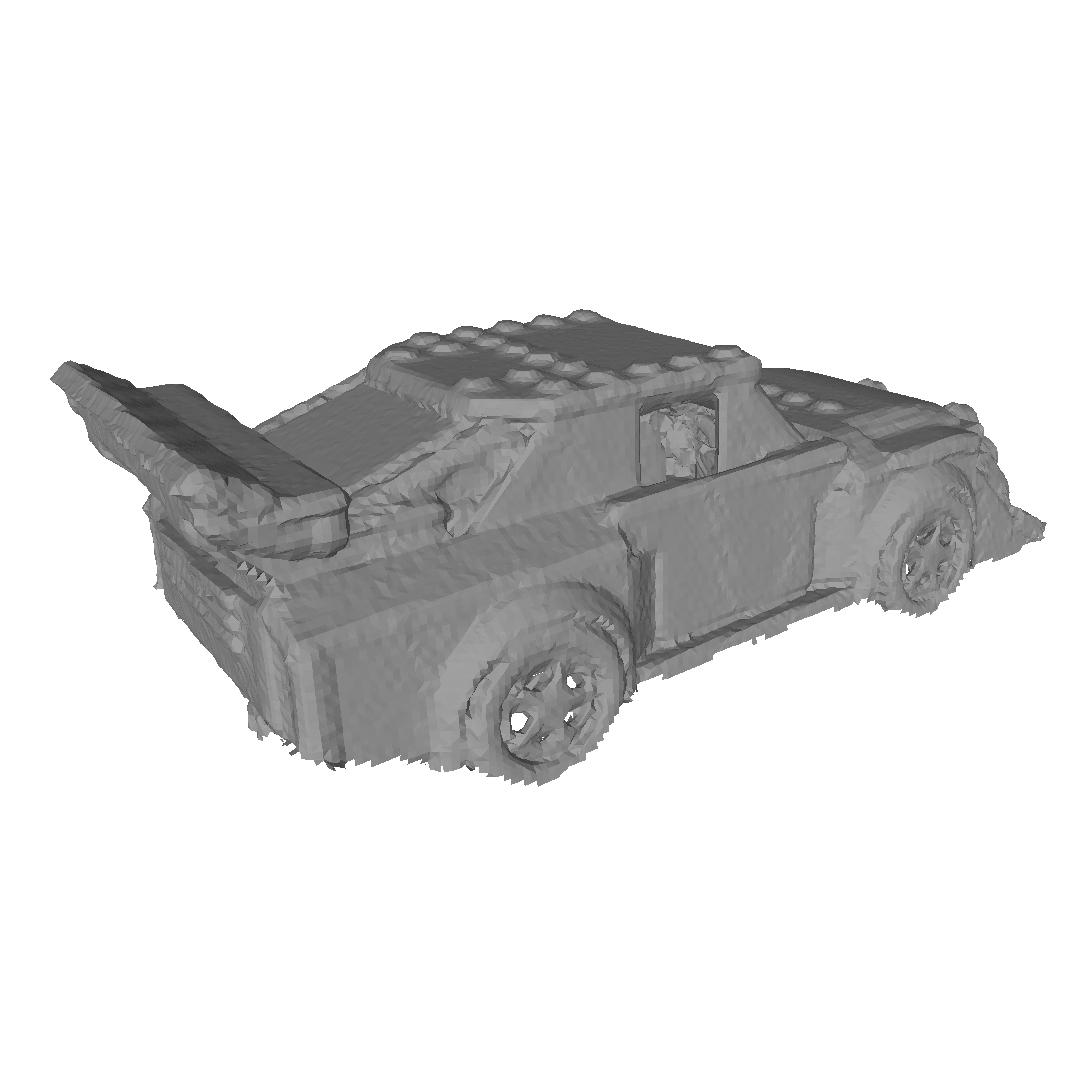} & 
    \includegraphics[width=0.22\textwidth]{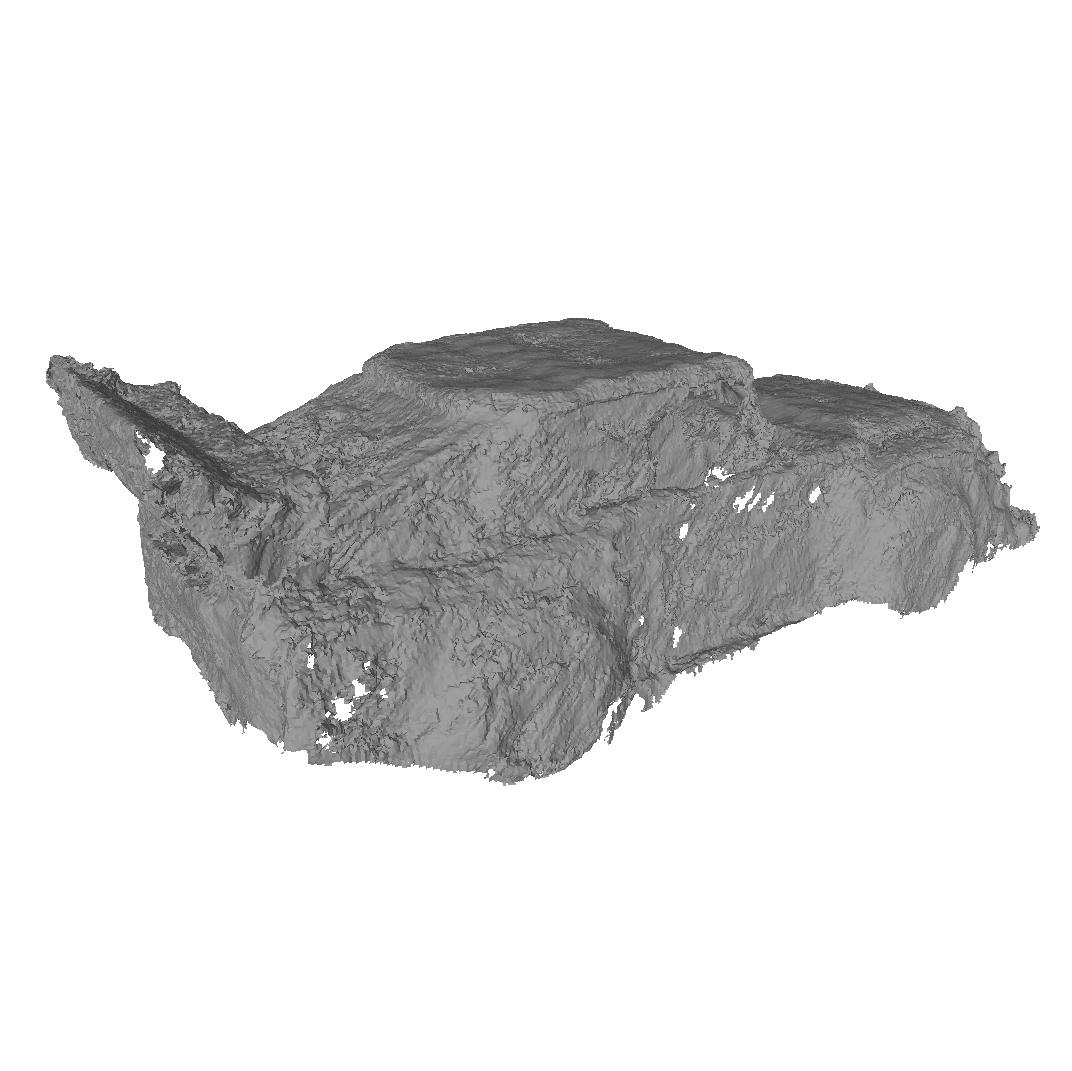} &
    \includegraphics[width=0.22\textwidth]{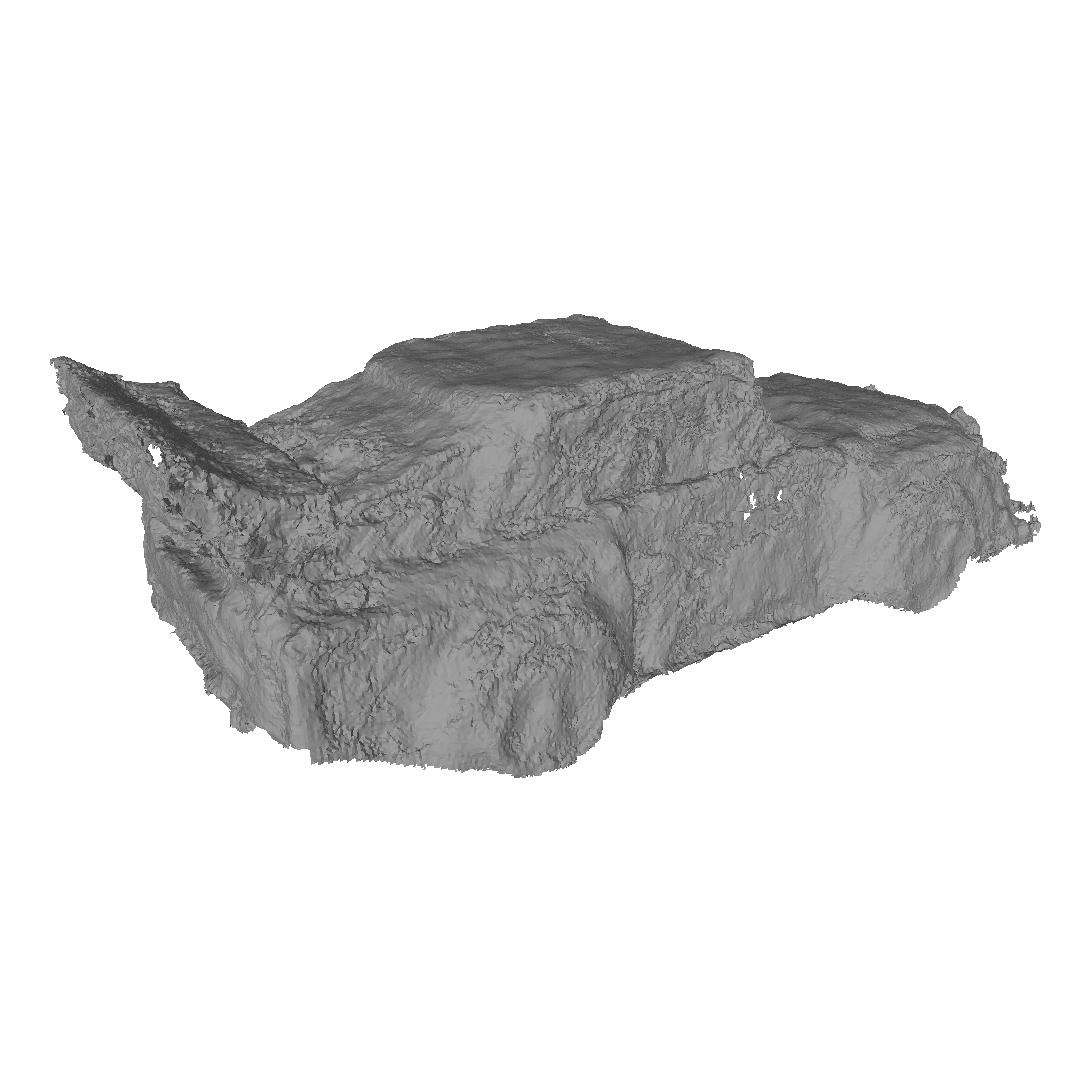} &
    \includegraphics[width=0.22\textwidth]{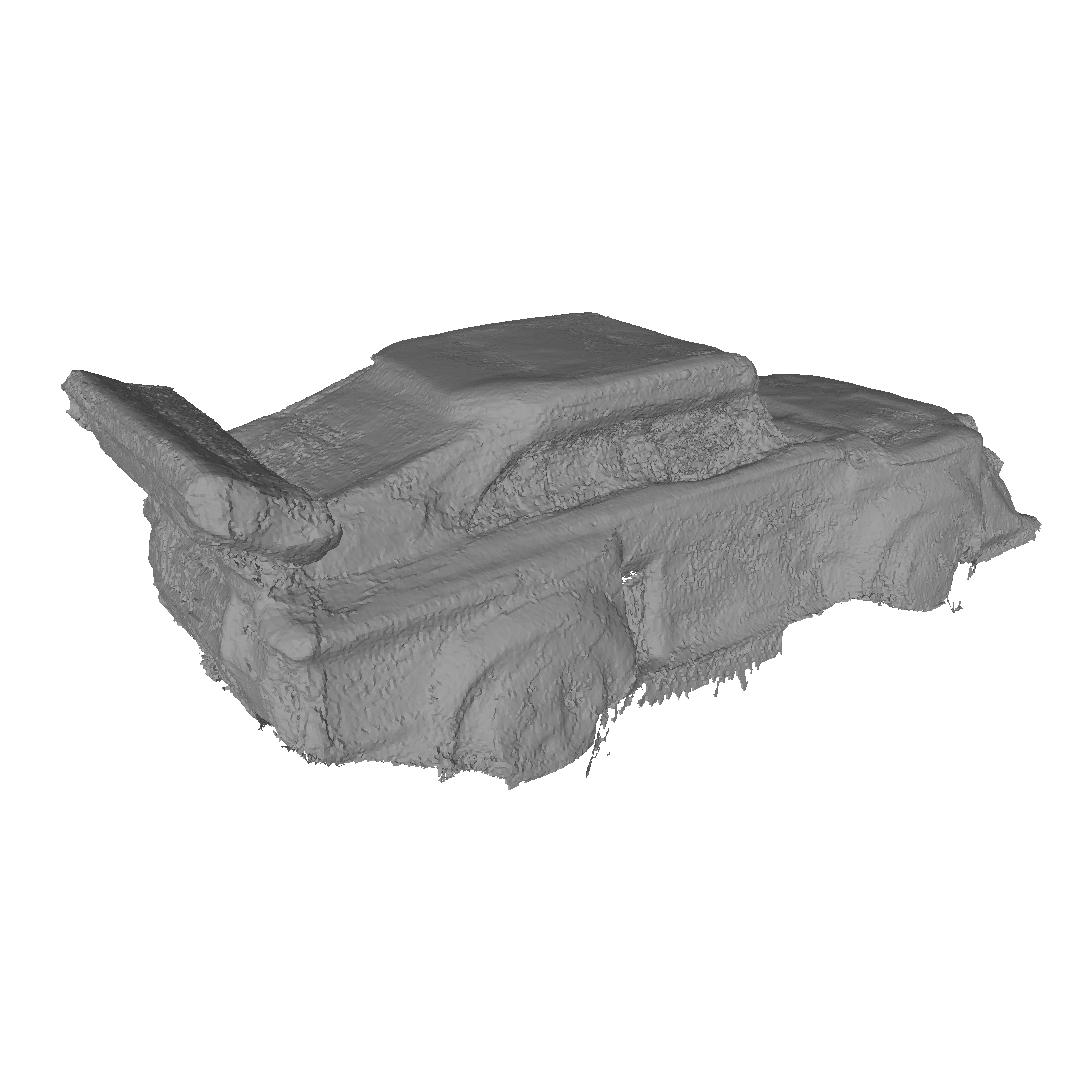} \\
    \multicolumn{4}{c}{Audi} \\
    \includegraphics[width=0.22\textwidth]{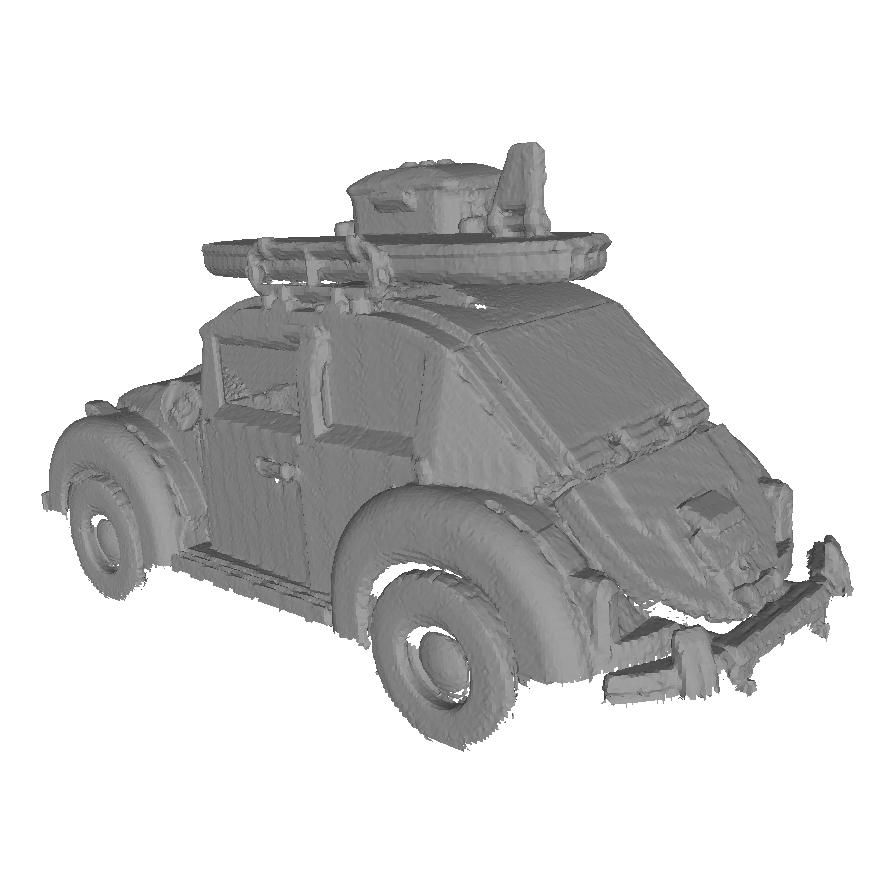} & 
    \includegraphics[width=0.22\textwidth]{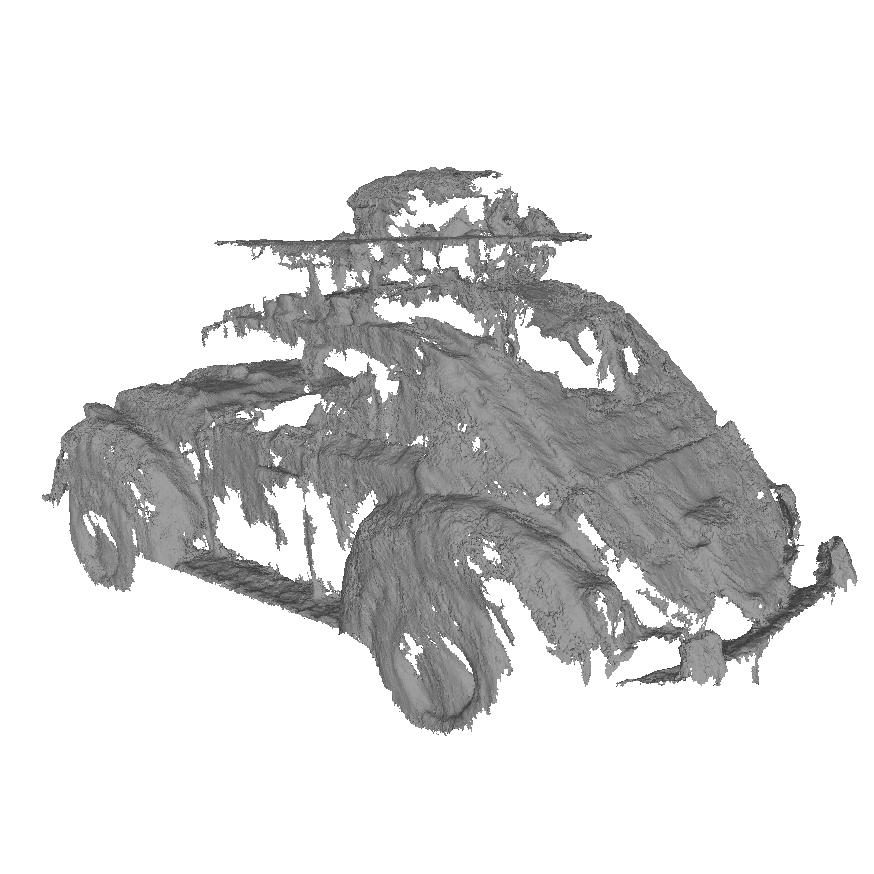} &
    \includegraphics[width=0.22\textwidth]{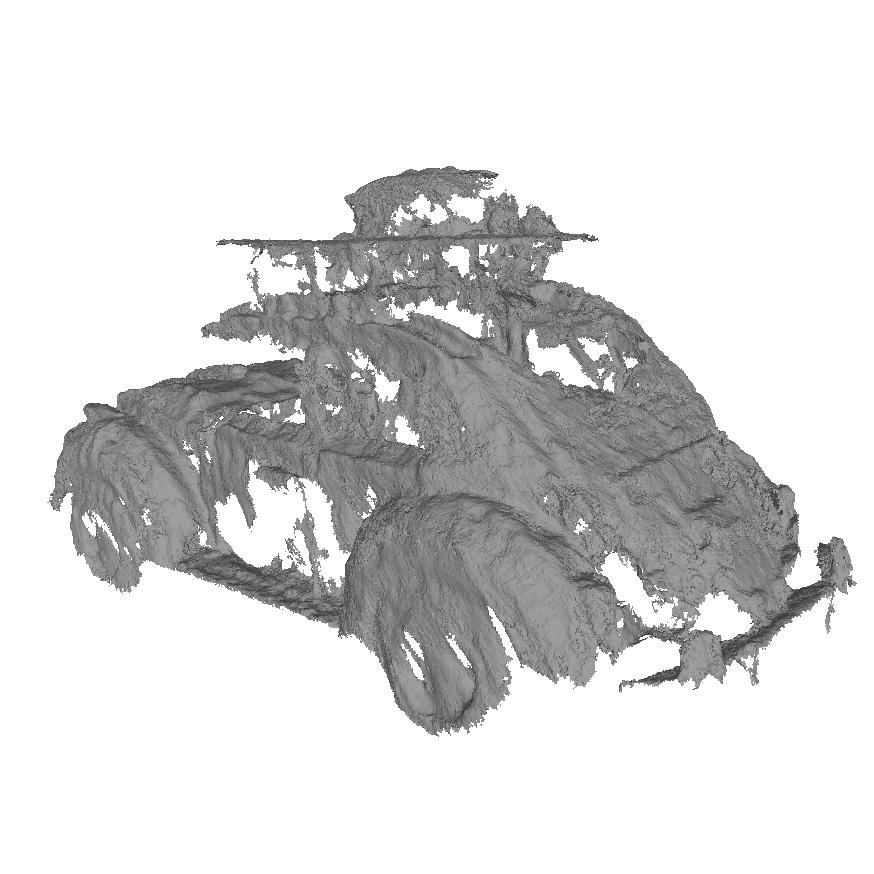} &
    \includegraphics[width=0.22\textwidth]{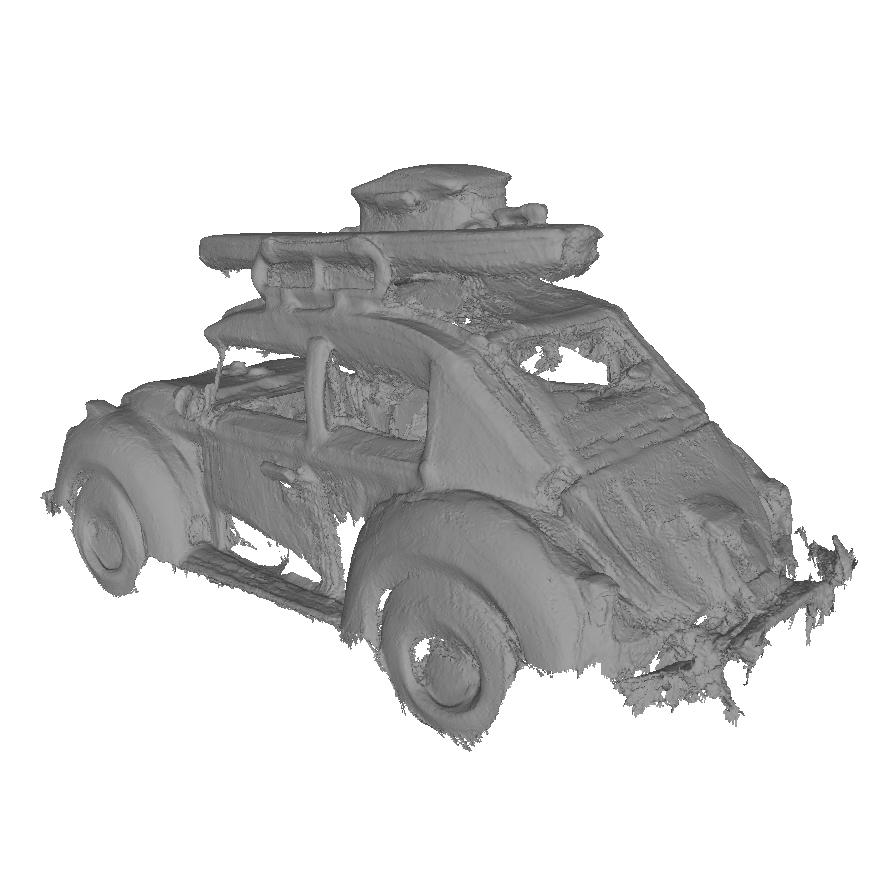} \\
    \multicolumn{4}{c}{Beetles} \\
    \includegraphics[width=0.22\textwidth]{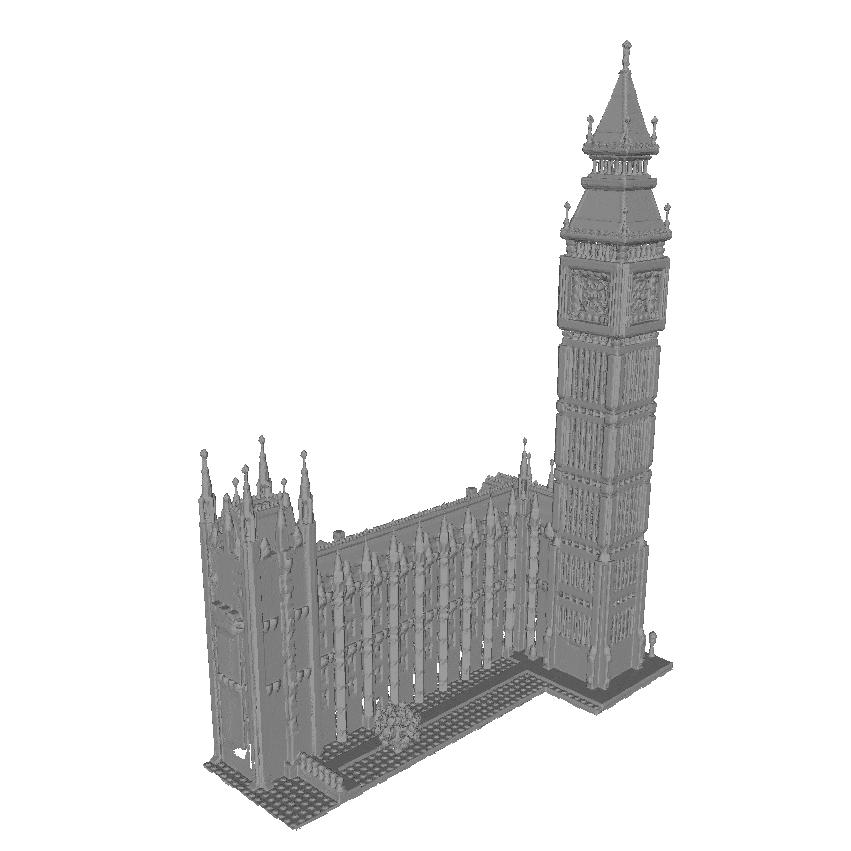} & 
    \includegraphics[width=0.22\textwidth]{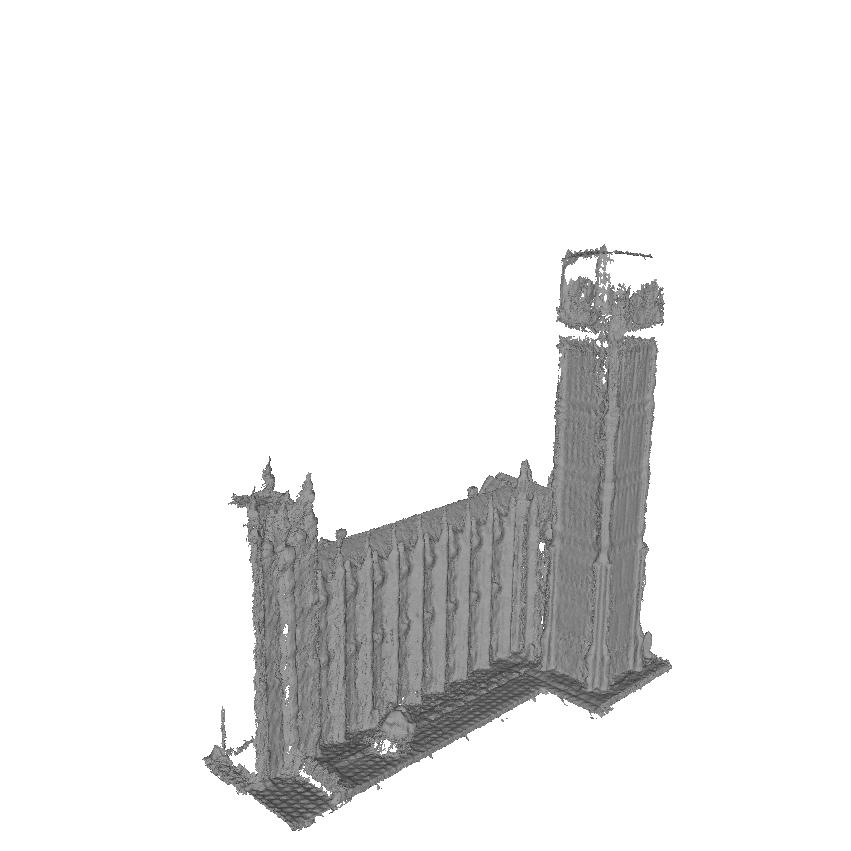} &
    \includegraphics[width=0.22\textwidth]{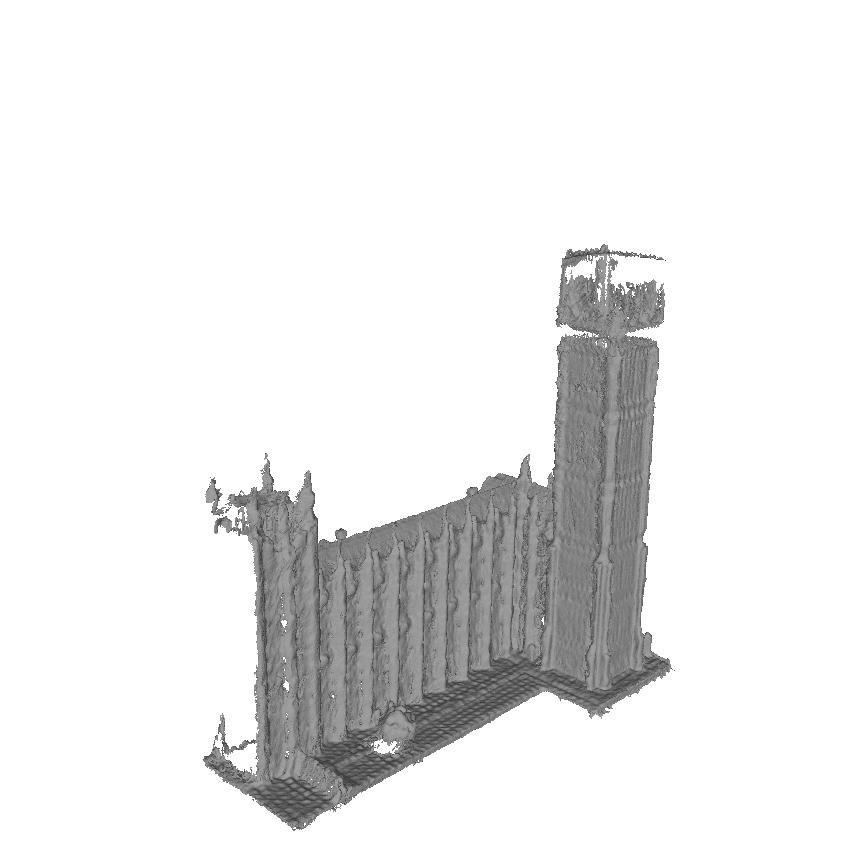} &
    \includegraphics[width=0.22\textwidth]{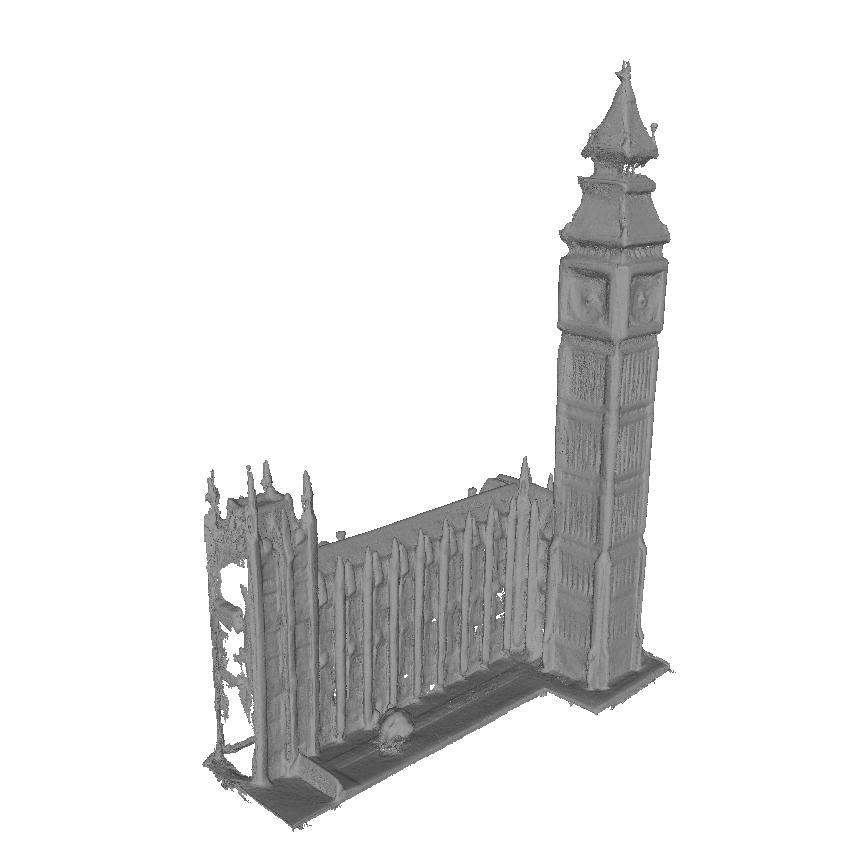} \\    
    \multicolumn{4}{c}{Big Ben} \\
    \includegraphics[width=0.22\textwidth]{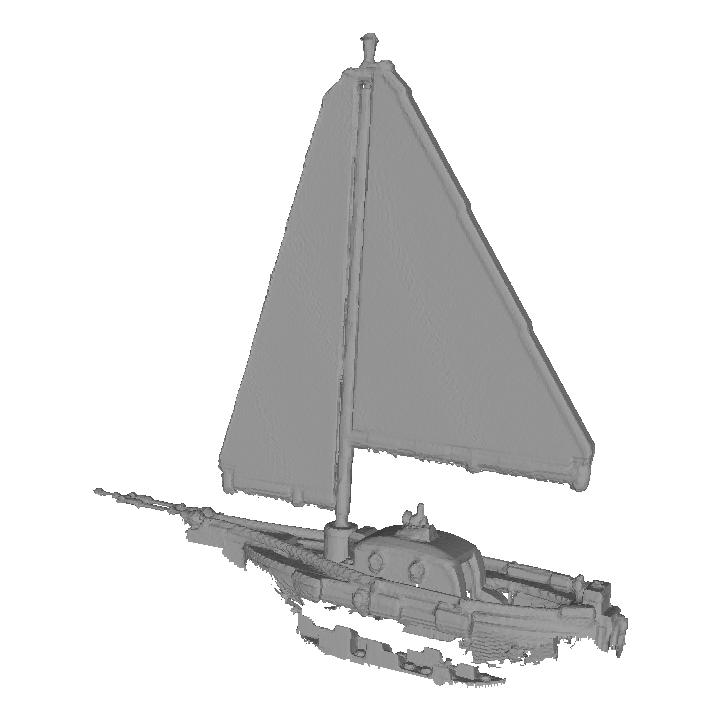} & 
    \includegraphics[width=0.22\textwidth]{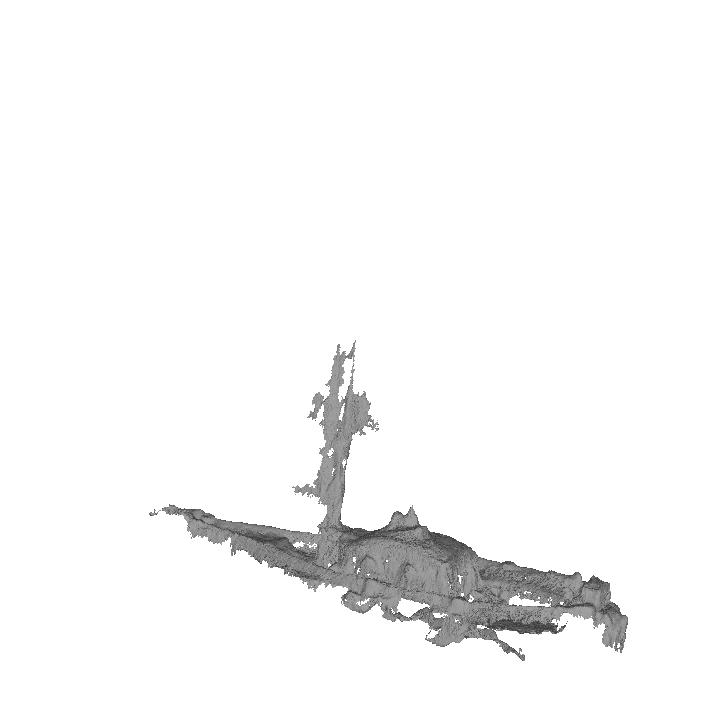} &
    \includegraphics[width=0.22\textwidth]{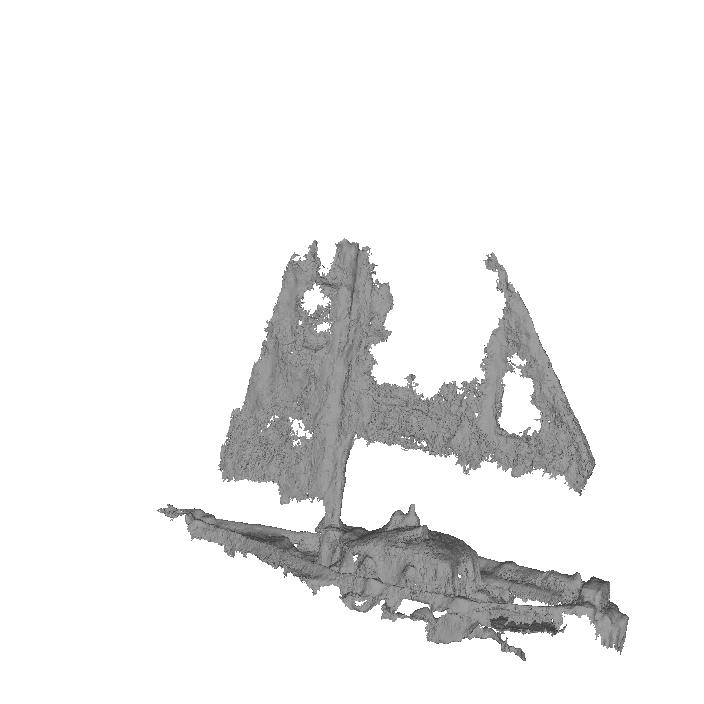} &
    \includegraphics[width=0.22\textwidth]{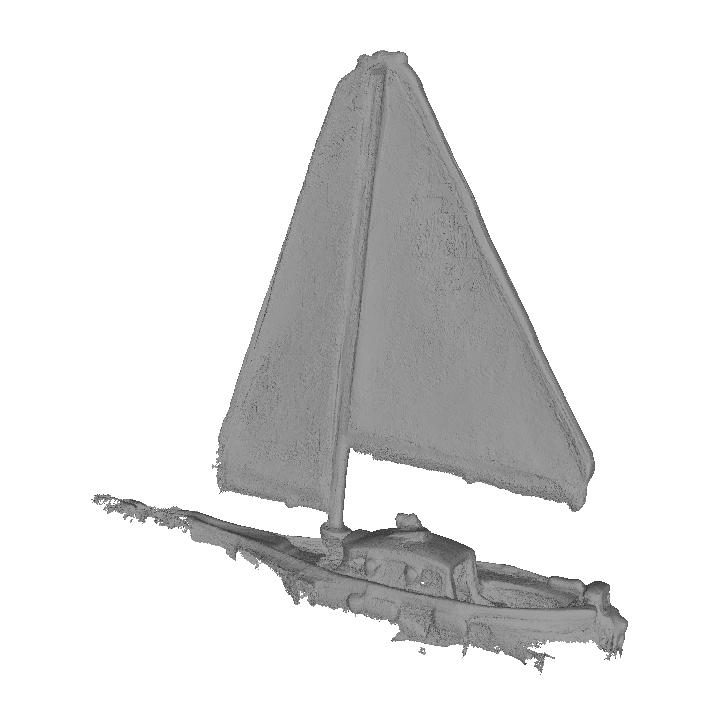} \\
    \multicolumn{4}{c}{Boat} \\
    \includegraphics[width=0.22\textwidth]{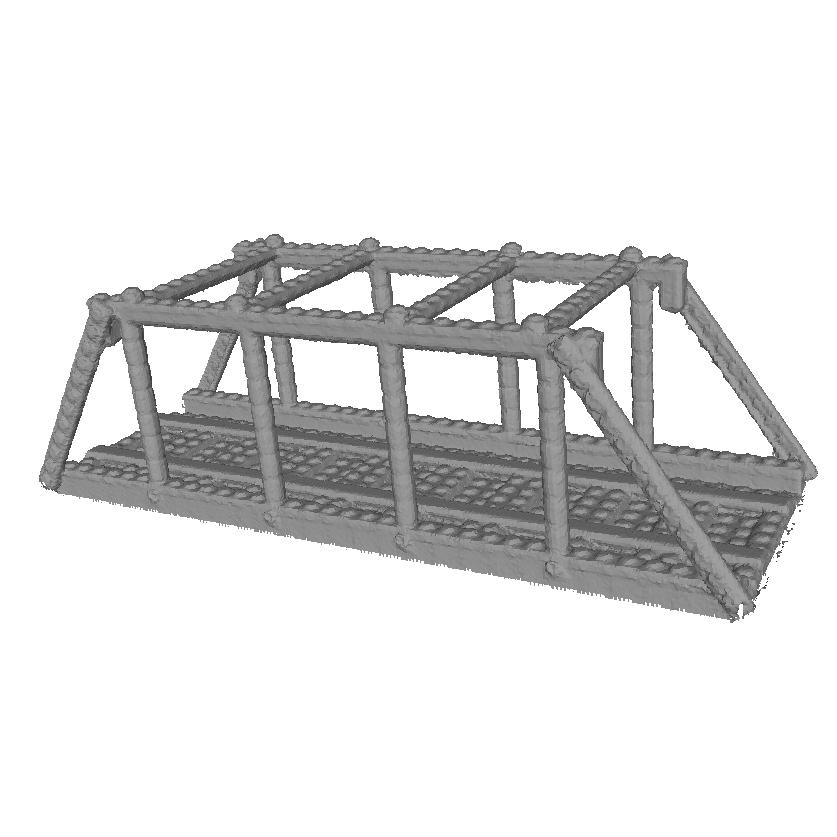} & 
    \includegraphics[width=0.22\textwidth]{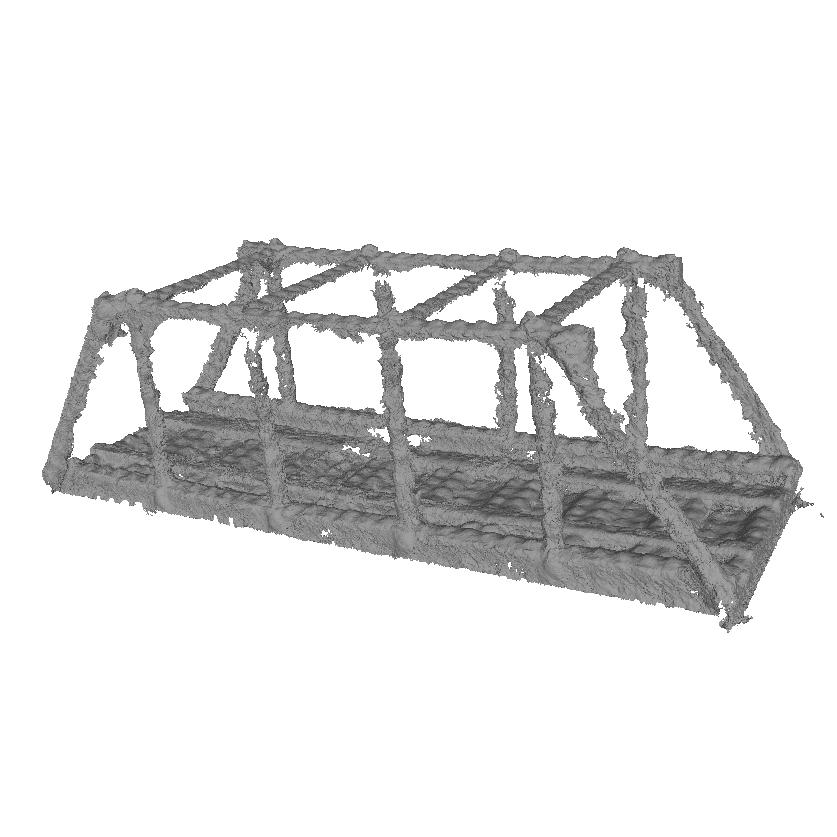} &
    \includegraphics[width=0.22\textwidth]{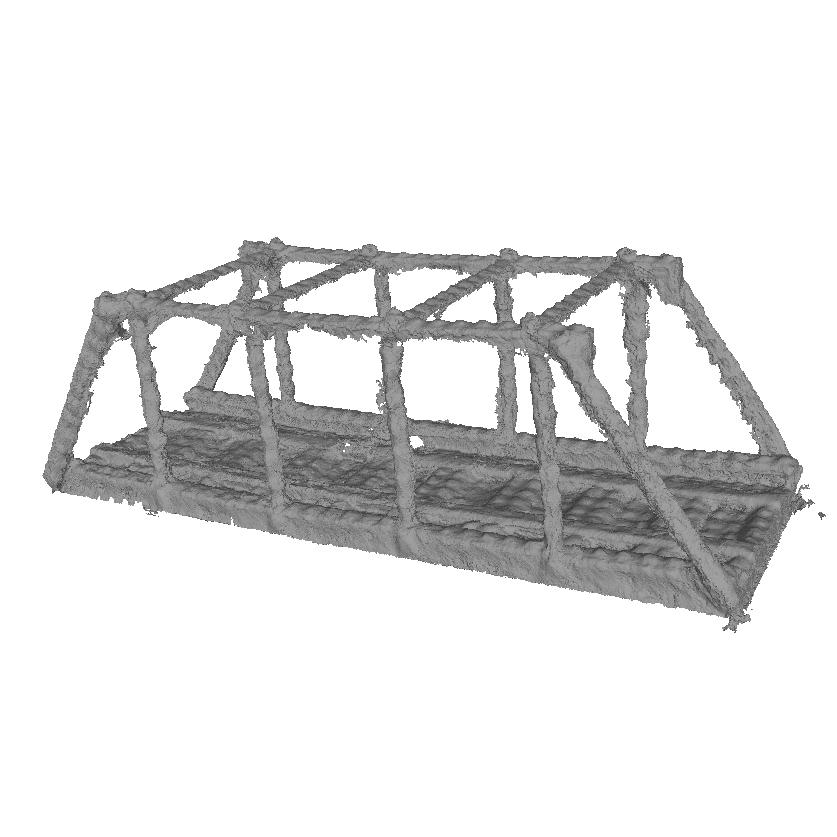} &
    \includegraphics[width=0.22\textwidth]{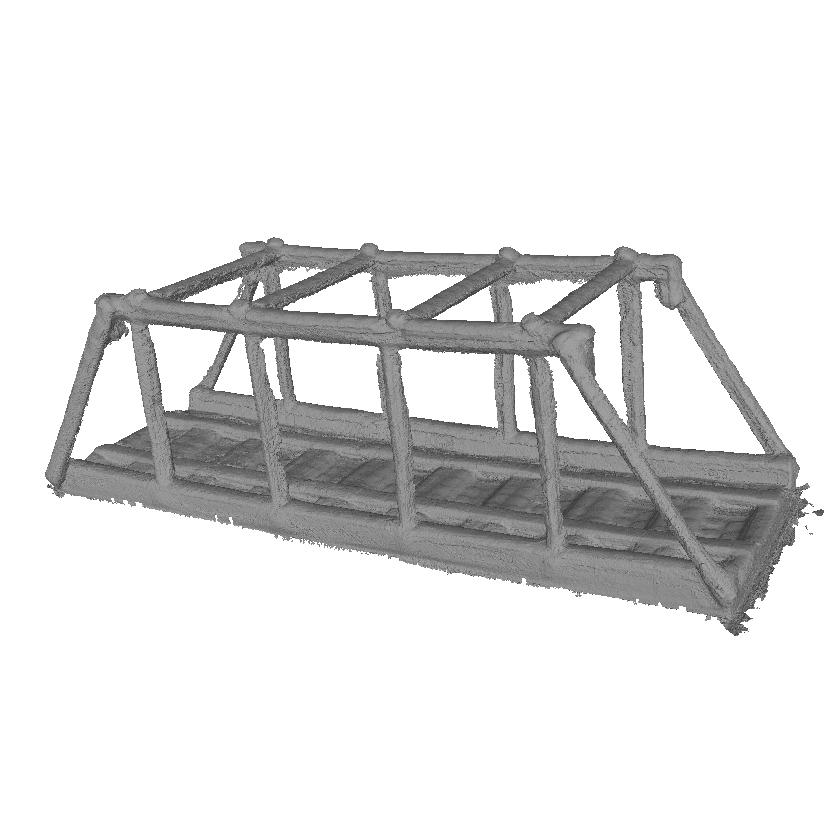} \\
    \multicolumn{4}{c}{Bridge} \\
    \end{tabular}
\end{figure}
\begin{figure}[]
    \centering
    \begin{tabular}{cccc}
    \textbf{Ground} & \textbf{MVSFormer} & \textbf{MVSFormer} & \textbf{Ours} \\
    \textbf{Truth} & & \textbf{+ Rendered} & \\
    \includegraphics[width=0.22\textwidth]{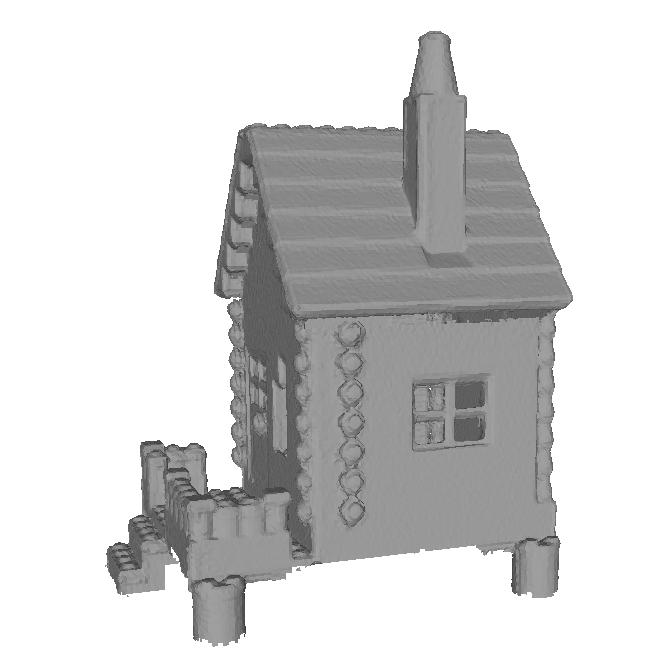} & 
    \includegraphics[width=0.22\textwidth]{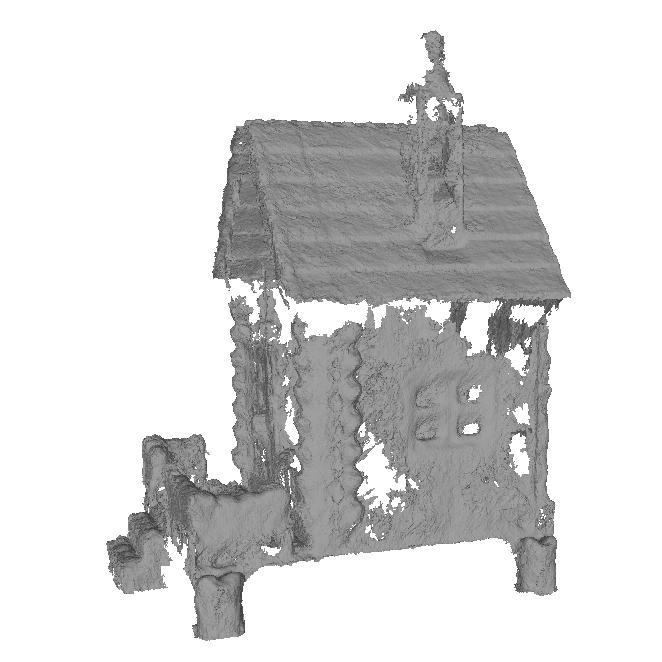} &
    \includegraphics[width=0.22\textwidth]{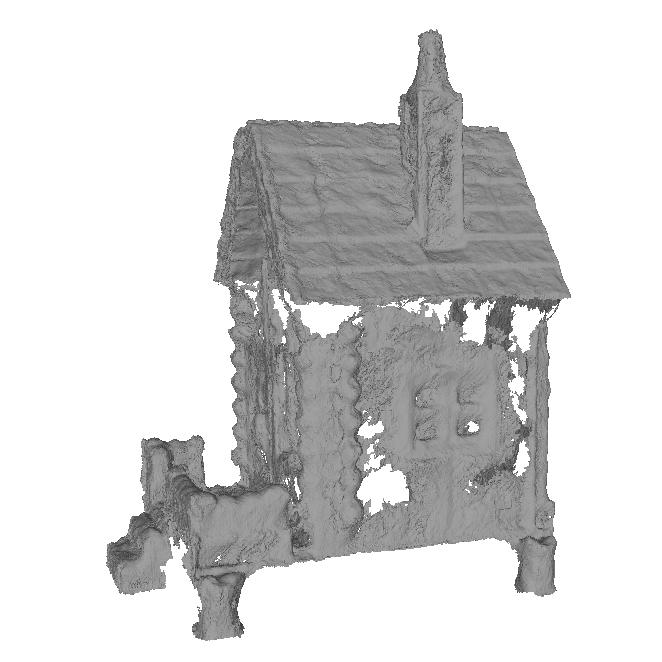} &
    \includegraphics[width=0.22\textwidth]{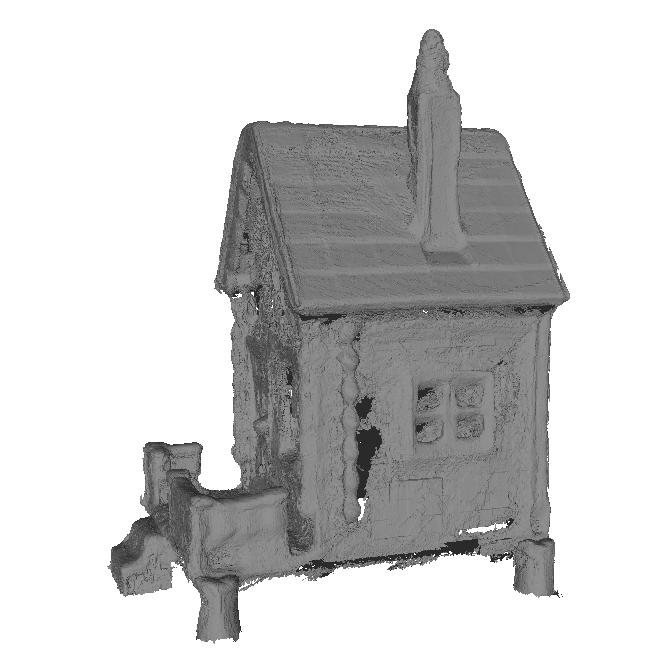} \\
    \multicolumn{4}{c}{Cabin} \\
    \includegraphics[width=0.22\textwidth]{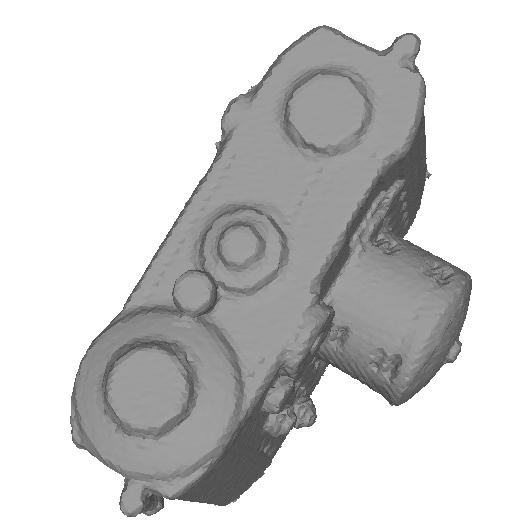} & 
    \includegraphics[width=0.22\textwidth]{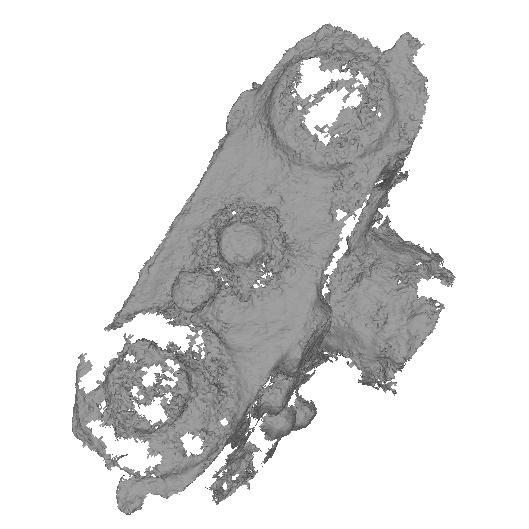} &
    \includegraphics[width=0.22\textwidth]{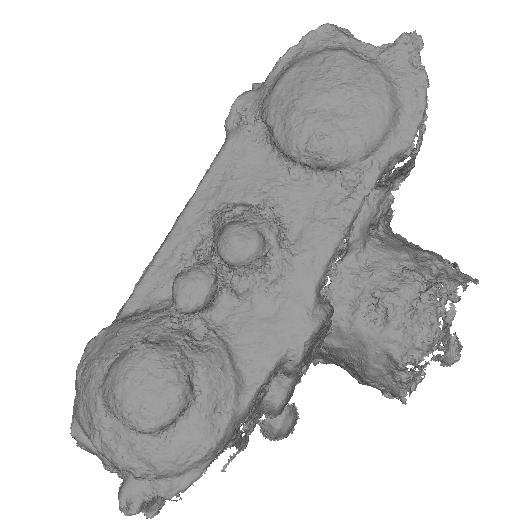} &
    \includegraphics[width=0.22\textwidth]{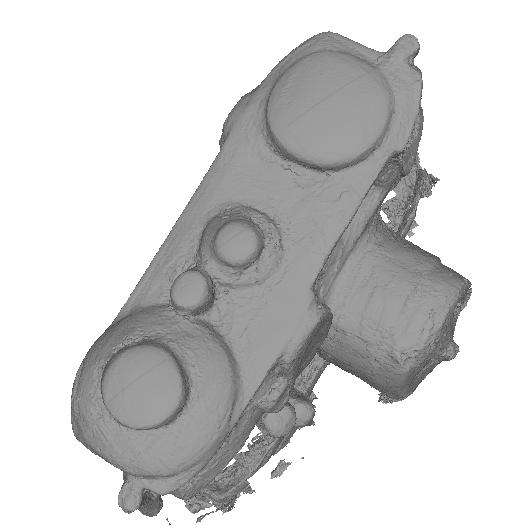} \\
    \multicolumn{4}{c}{Camera} \\
    \includegraphics[width=0.22\textwidth]{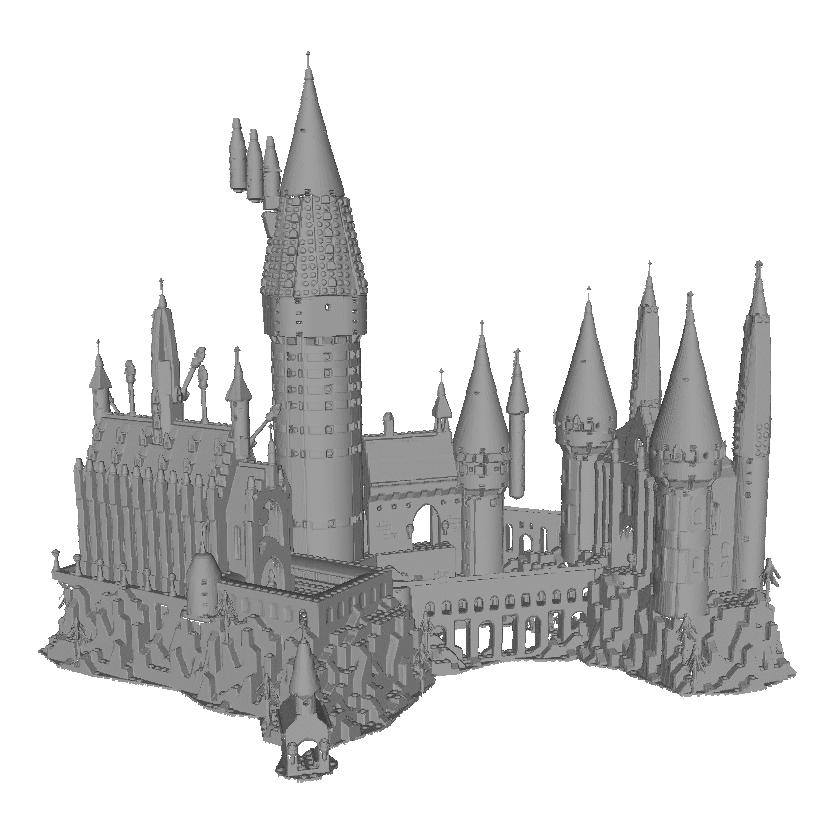} & 
    \includegraphics[width=0.22\textwidth]{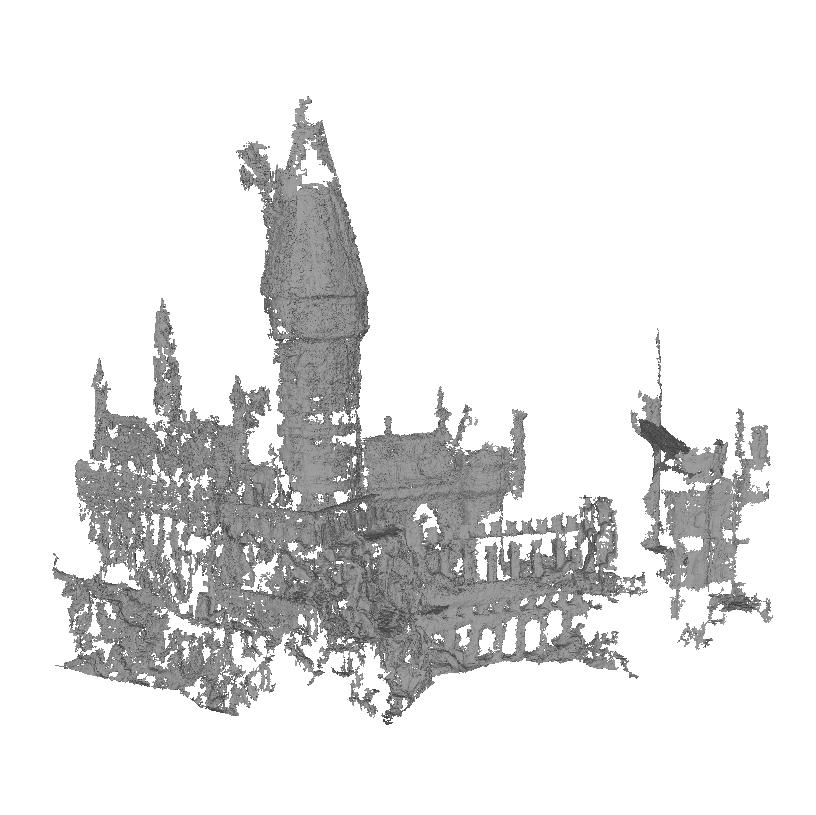} &
    \includegraphics[width=0.22\textwidth]{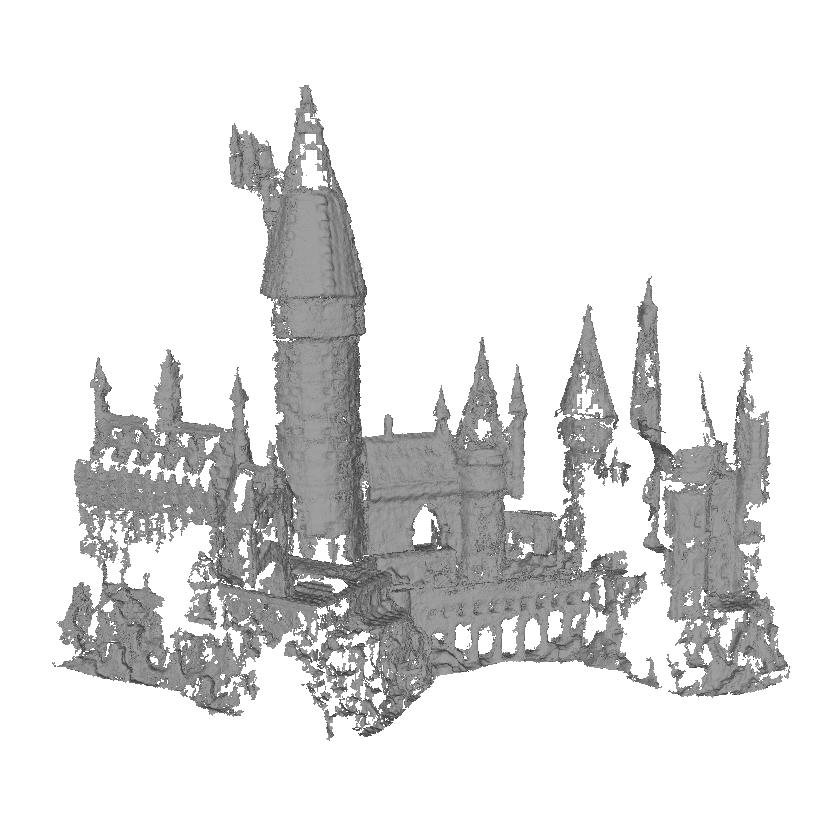} &
    \includegraphics[width=0.22\textwidth]{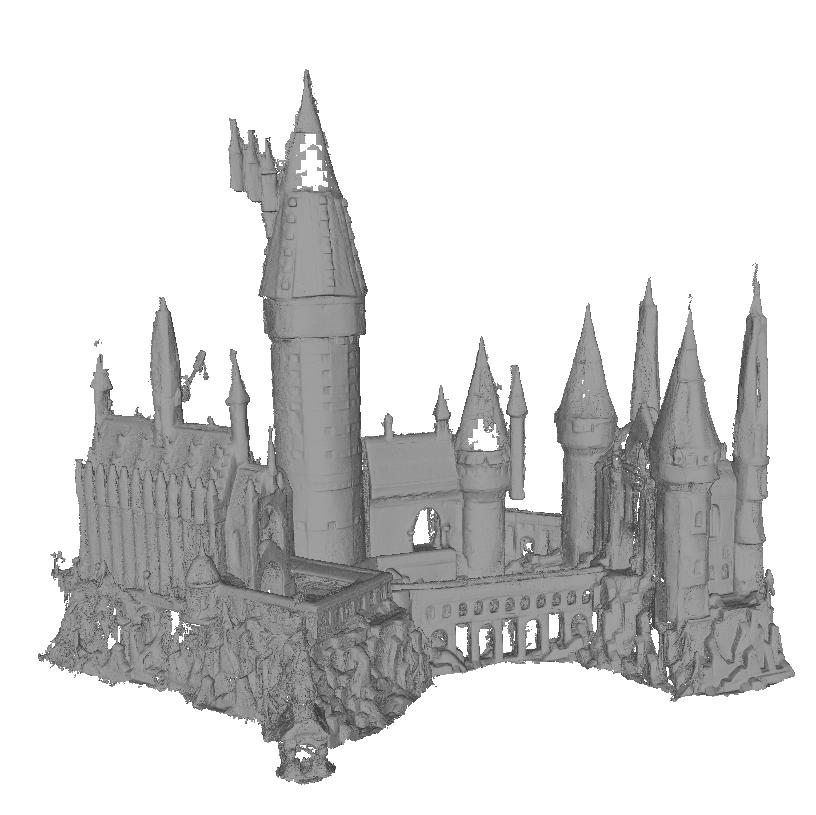} \\
    \multicolumn{4}{c}{Castle} \\
    \includegraphics[width=0.22\textwidth]{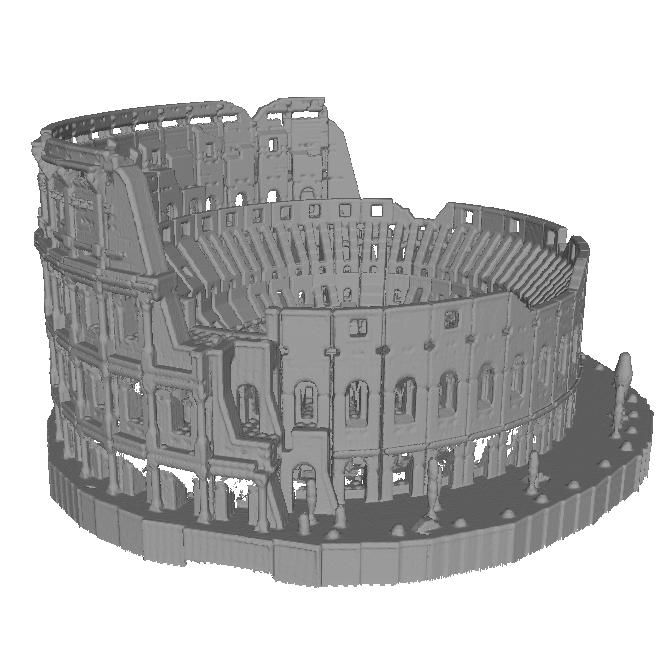} & 
    \includegraphics[width=0.22\textwidth]{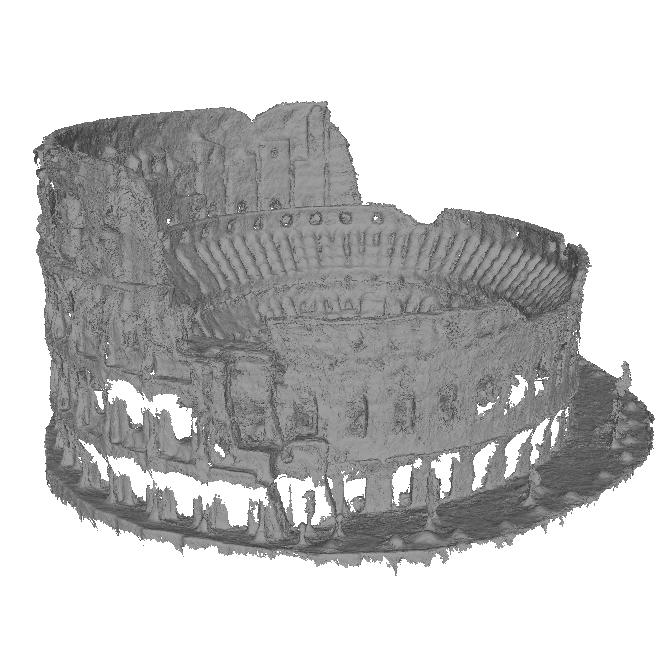} &
    \includegraphics[width=0.22\textwidth]{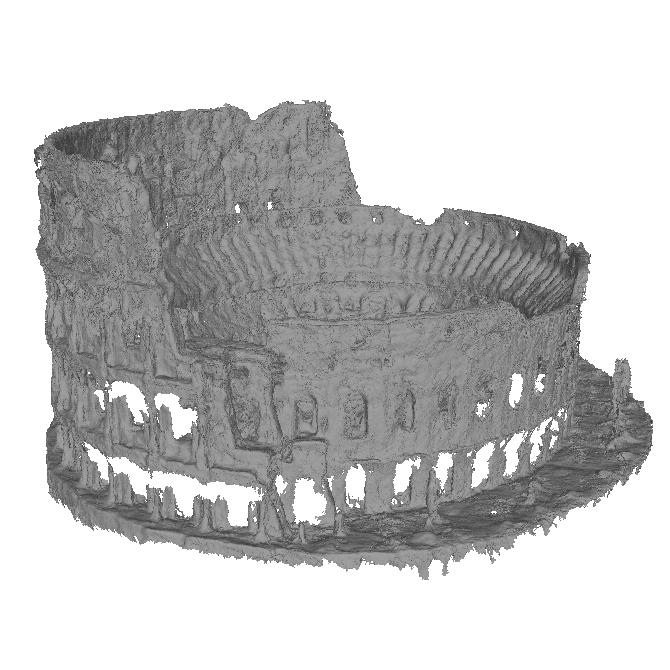} &
    \includegraphics[width=0.22\textwidth]{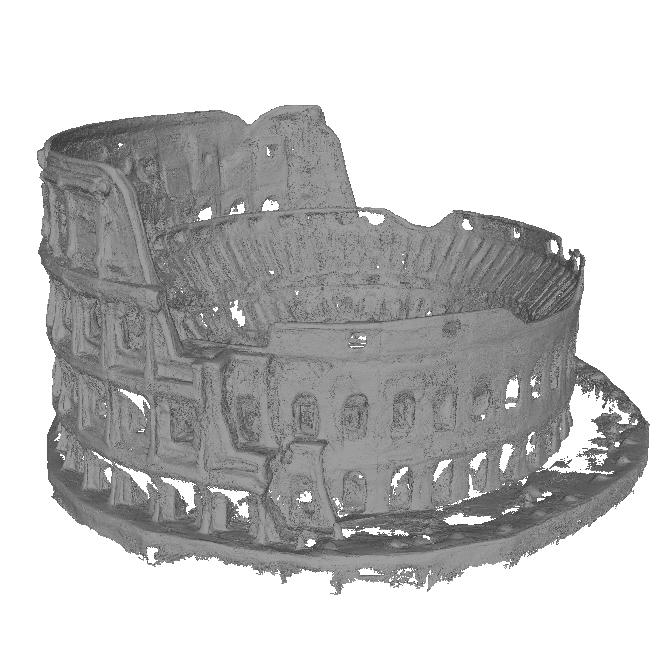} \\
    \multicolumn{4}{c}{Colosseum} \\
    \includegraphics[width=0.22\textwidth]{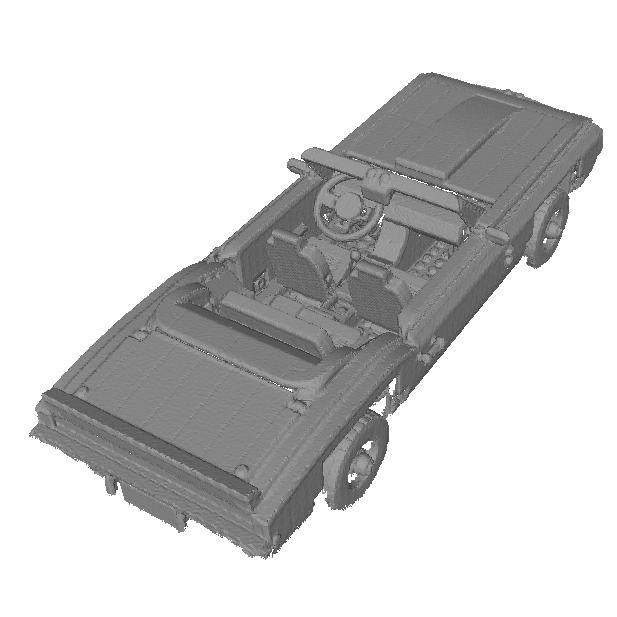} & 
    \includegraphics[width=0.22\textwidth]{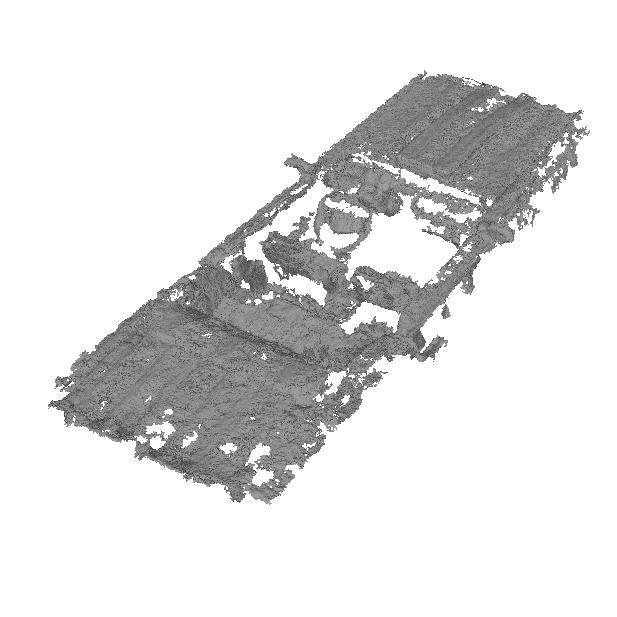} &
    \includegraphics[width=0.22\textwidth]{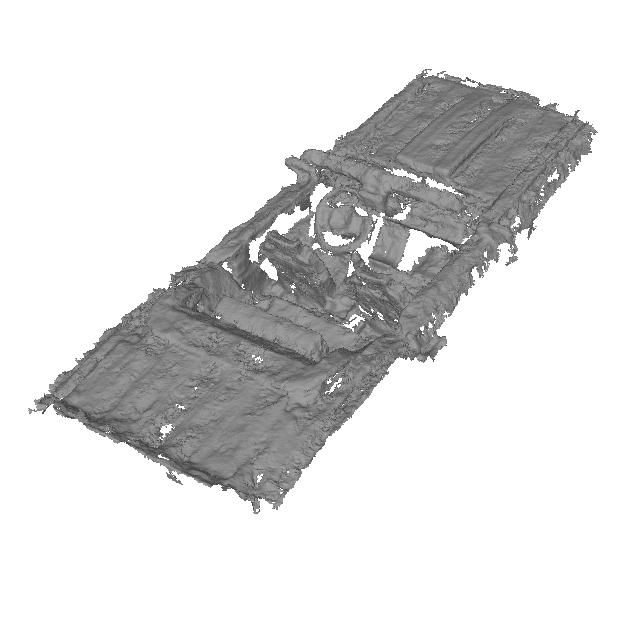} &
    \includegraphics[width=0.22\textwidth]{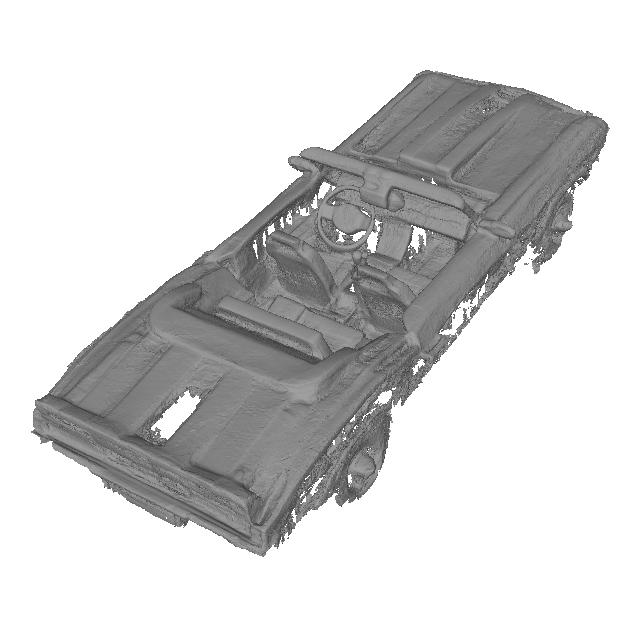} \\
    \multicolumn{4}{c}{Convertible} \\
    \includegraphics[width=0.22\textwidth]{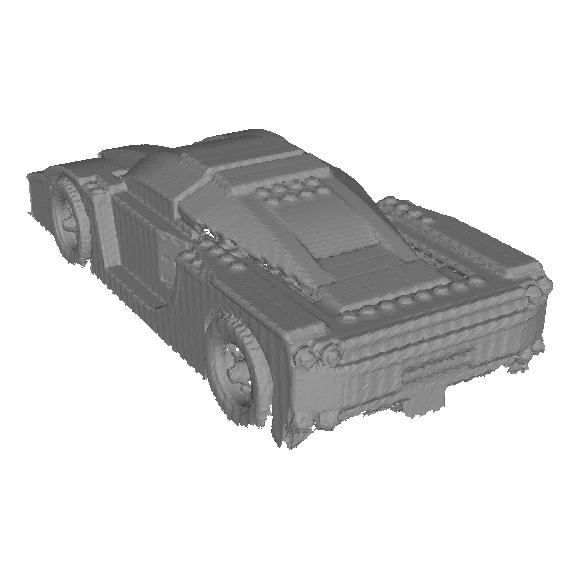} & 
    \includegraphics[width=0.22\textwidth]{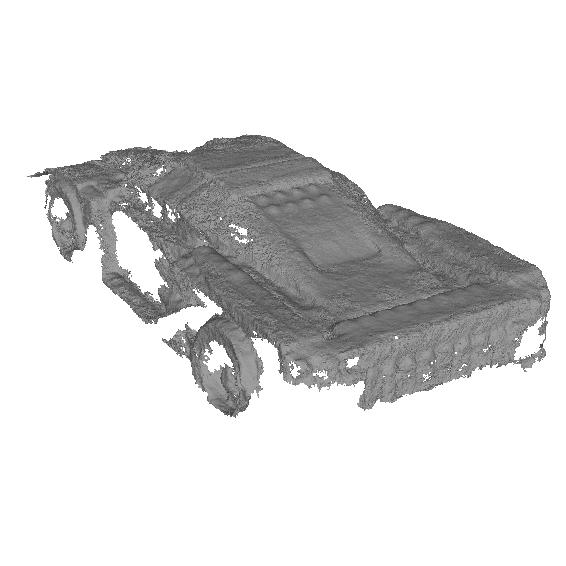} &
    \includegraphics[width=0.22\textwidth]{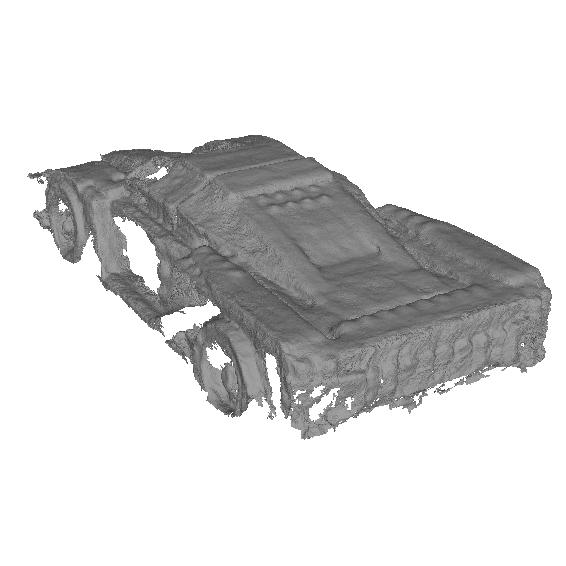} &
    \includegraphics[width=0.22\textwidth]{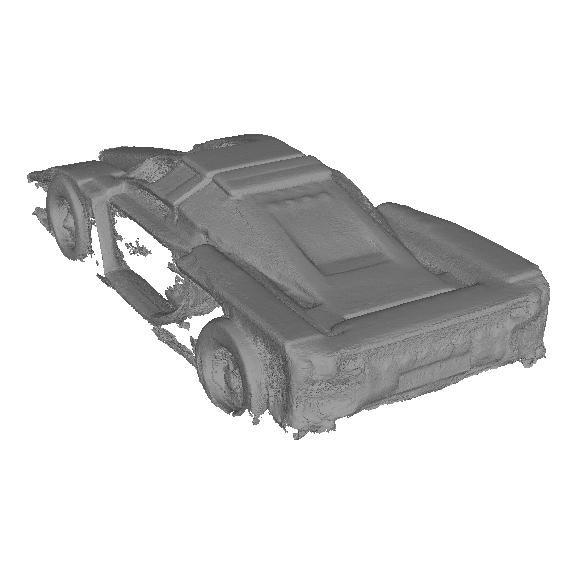} \\
    \multicolumn{4}{c}{Ferrari} \\
    \end{tabular}
\end{figure}

\end{document}